%% file: main.tex
\documentclass{article}

\usepackage{silence}
\usepackage[table,dvipsnames]{xcolor}
\definecolor{mydarkblue}{rgb}{0,0.08,0.45}
\definecolor{note_fontcolor}{rgb}{0.800781, 0.800781, 0.800781}

    \PassOptionsToPackage{numbers, compress, sort}{natbib}

     \usepackage[final]{neuripsx}

\usepackage[utf8]{inputenc} %
\usepackage[T1]{fontenc}    %
\usepackage[hyphens]{url}            %
\usepackage[
    colorlinks,
    bookmarks=true,
    breaklinks=true,
    linkcolor=mydarkblue,
    citecolor=mydarkblue,
    filecolor=mydarkblue,
    urlcolor=mydarkblue,]{hyperref}       %
\usepackage{booktabs}       %
\usepackage{amsfonts}       %
\usepackage{nicefrac}       %
\usepackage{microtype}      %
\usepackage{multicol}
\usepackage{subcaption}
\usepackage{amssymb, amsmath, amsthm}
\usepackage{thmtools}
\usepackage{mathtools}
\usepackage{tcolorbox}
\usepackage{wrapfig}

\usepackage{subfiles}

\usepackage[inline]{enumitem}
\usepackage{breakurl}
\usepackage[capitalize,nameinlink]{cleveref}
\usepackage{mymacros}

\newcommand{\Zz}{Z}
\newcommand{\Zhat}{\hat Z}
\newcommand{\Zdot}{\dot Z}
\newcommand{\Gg}{\mathcal G}

\newtheorem{thm}{Theorem}[section]
\newtheorem{cor}[thm]{Corollary}
\newtheorem{fact}[thm]{Fact}

\newtheorem{conj}[thm]{Conjecture}
\newtheorem{claim}{Claim}[section]
\newtheorem{assm}[thm]{Assumption}

\newtheorem{setup}[thm]{Setup}
\newtheorem{lemma}[thm]{Lemma}
\newtheorem{prop}[thm]{Proposition}
\theoremstyle{definition}
\newtheorem{defn}[thm]{Definition}
\theoremstyle{remark}
\newtheorem{rem}[thm]{Remark}
\newtheorem{exmp}[thm]{Example}

\newtheorem{cond}{Condition}

\newtcolorbox[auto counter,crefname={Box}{Boxes}]{pabox}[2][]{%
title=Box~\thetcbcounter $\quad$ #2, label={#1}}

\usepackage{bbm}
\crefname{thm}{\text{Theorem}}{\text{Theorems}}
\crefname{assm}{\text{Assumption}}{\text{Assumptions}}
\crefname{defn}{\text{Definition}}{\text{Definitions}}
\crefname{prop}{\text{Proposition}}{\text{Propositions}}
\crefname{cor}{\text{Corollary}}{\text{Corollaries}}
\crefname{lemma}{\text{Lemma}}{\text{Lemmas}}
\crefname{algorithm}{\text{Program}}{\text{Programs}}

\newcommand{\bigtheta}{\Theta}

\newcommand{\trsp}{\top}
\newcommand{\distto}{\xrar{\mathrm{d}}}
\newcommand{\asto}{\xrar{\mathrm{a.s.}}}
\newcommand{\probto}{\xrar{\mathrm{p}}}
\newcommand{\disteq}{\overset{\mathrm{d}}{=}}
\newcommand{\aseq}{\overset{\mathrm{a.s.}}{=}}
\newcommand{\meanto}{\xrar{L^1}}

\newcommand{\cdc}{\mathfrak{c}}

\newcommand{\Ss}{\mathcal{S}}

\newcommand{\odefeq}{\mathbin{\overset{\text{\tiny{def}}}{=}}}

\newcommand{\defeq}{\mathrel{\raisebox{-0.3ex}{$\odefeq$}}}

\newcommand{\Jac}[2]{\f{\dd #1}{\dd #2}}

\renewcommand{\cite}{\citep}

\renewcommand{\cite}{\citep}

\newcommand{\netsor}{\texorpdfstring{{$\textsc{Netsor}$}}{Netsor}}
\newcommand{\netsort}{\texorpdfstring{{$\textsc{Netsor}\trsp$}}{NetsorT}}
\newcommand{\netsorplus}{\texorpdfstring{{$\textsc{Netsor}^+$}}{Netsor+}}
\newcommand{\netsortplus}{\texorpdfstring{{$\textsc{Netsor}\trsp^+$}}{NetsorT+}}

\global\long\def\musc{\mu_{\mathrm{sc}}}%

\newcommand{\xme}{x}
\newcommand{\yme}{y}
\newcommand{\ume}{u}
\newcommand{\vme}{v}
\newcommand{\Xme}{\mathbf{X}}
\newcommand{\Yme}{\mathbf{Y}}
\newcommand{\Ume}{\mathbf{U}}
\newcommand{\Vme}{\mathbf{V}}
\newcommand{\gammame}{\gamma}
\newcommand{\Upsilonme}{\Upsilon}
\newcommand{\deltame}{\delta}
\newcommand{\Lambdame}{\Lambda}
\newcommand{\dme}{d}
\newcommand{\eme}{e}
\newcommand{\varepsilonhat}{\hat\varepsilon}
\newcommand{\varepsiloncheck}{\check\varepsilon}
\newcommand{\PP}{\boldsymbol{\Pi}}

\newcommand{\refZInput}{\textbf{\ref{Z:Initial}}}
\newcommand{\refZMatMul}{\textbf{\ref{Z:MatMul}}}
\newcommand{\refZNonlin}{\textbf{\ref{Z:Nonlin}}}
\newcommand{\refZhat}{\textbf{\ref{Z:Zhat}}}
\newcommand{\refZdot}{\textbf{\ref{Z:Zdot}}}
\newcommand{\refMatMul}{\textbf{\ref{instr:matmul}}}
\newcommand{\refNonlin}{\textbf{\ref{instr:nonlin}}}
\newcommand{\refMatMulPlus}{\textbf{\ref{instr:matmul+}}}
\newcommand{\refNonlinPlus}{\textbf{\ref{instr:nonlin+}}}
\newcommand{\refMoment}{\textbf{\ref{instr:moment}}}

\newcommand{\refMatMulVarDim}{\textbf{\ref{instr:matmulVarDim}}}
\newcommand{\refNonlinVarDim}{\textbf{\ref{instr:nonlinVarDim}}}

\newcommand{\refNonlinPlusVarDim}{\textbf{\ref{instr:nonlin+VarDim}}}

\makeatletter
\let\orgdescriptionlabel\descriptionlabel
\newcommand*{\@restrictlabeltext}[1]{#1\protected@edef\@currentlabel{#1}}
\newcommand*{\nolabel}[1]{#1}%
\renewcommand*{\descriptionlabel}[1]{%
  \let\orglabel\label
  \let\label\@gobble
  \let\orig@hfil\hfil
  \def\hfil{}%
  \let\nolabel\@gobble
  \let\restrictlabeltext\@firstofone
  \phantomsection
  \protected@edef\@currentlabel{#1}%
  \let\hfil\orig@hfil
  \let\label\orglabel
  \let\restrictlabeltext\@restrictlabeltext
  \orgdescriptionlabel{#1}%
}
\makeatother

\title{Tensor Programs III:
Neural Matrix Laws}

\author{%
  Greg Yang\\
  Microsoft Research AI\\
  \texttt{gregyang@microsoft.com} \\
}

\begin{document}

\maketitle

\begin{abstract}
In a neural network (NN), \emph{weight matrices} linearly transform inputs into \emph{preactivations} that are then transformed nonlinearly into \emph{activations}.
A typical NN interleaves multitudes of such linear and nonlinear transforms to express complex functions.
Thus, the (pre-)activations depend on the weights in an intricate manner.
We show that, surprisingly, (pre-)activations of a randomly initialized NN become \emph{independent} from the weights as the NN's widths tend to infinity, in the sense of \emph{asymptotic freeness} in random matrix theory.
We call this the \emph{Free Independence Principle (FIP)}, which has these consequences:
1) It rigorously justifies the calculation of asymptotic Jacobian singular value distribution of an NN in \citet{pennington_resurrecting_2017,pennington_emergence_2018}, essential for training ultra-deep NNs \citep{xiao_dynamical_2018}.
2) It gives a new justification of \emph{gradient independence assumption} used for calculating the \emph{Neural Tangent Kernel} of a neural network.
FIP and these results hold for any neural architecture.
We show FIP by proving a Master Theorem for any Tensor Program, as introduced in \citet{yangTP1,yangTP2}, generalizing the Master Theorems proved in those works.
As warmup demonstrations of this new Master Theorem, we give new proofs of the semicircle and Marchenko-Pastur laws, which benchmarks our framework against these fundamental mathematical results.
\end{abstract}

\section{Introduction}

A neural network (NN), at a high level, is an interleaved composition of linear and nonlinear transformations. Its \emph{weights} are the matrices involved in the linear transformations. The image (resp.\ input) of nonlinear transformations are called \emph{activations} (resp.\ \emph{preactivations}). Thus, \emph{a priori}, (pre-)activations depend in complicated ways on the weights. We show that, surprisingly, the (pre-)activations become roughly \emph{independent} from the weights in the sense of random matrix theory, when the NN is randomly initialized and as its width tends to infinity. More formally, we prove that the weights are \emph{asymptotically free} from the diagonal matrices whose diagonals are the images of preactivation vectors under any bounded coordinatewise function.\footnote{Note, however, that we make no claim about trained weights, just random weights.} This result holds for any neural architecture (i.e. architectural universality)\footnote{In this work, \emph{architecture} refers to the network topology along with the random initialization of parameters, as specified by a \netsort{} program (\cref{defn:netsort}).} . We call this result the \emph{Free Independence Principle} (FIP). A major application of FIP is in rigorously justifying a prevalent free independence assumption in the calculation of NN Jacobian singular value distribution \citep{pennington_resurrecting_2017,pennington_emergence_2018,tarnowski_dynamical_2018,ling_spectrum_2019,xiao_dynamical_2018}, as we overview below. 

Besides FIP, we also prove several other interesting results. In the next few subsections, we discuss the context required to state them. A summary of all of our contributions appears at the end of this introduction, and the reader in a hurry may feel free to skip directly to it.

\subsection{Random Matrix Theory in Deep Learning}

Random matrix theory (RMT) has a successful history of being applied in deep learning \citep{pennington_resurrecting_2017,pennington_emergence_2018,tarnowski_dynamical_2018,ling_spectrum_2019,xiao_dynamical_2018}. For example, RMT has been used to calculate the Jacobian%
\footnote{Here we are concerned with the input-output Jacobian of a neural network on a fix input. Some works consider the Jacobian with respect to the parameters, where the Jacobian has dimension \#parameters $\times$ \#data points (e.g. \citep{pennington_spectrum_2018}). This is a related but distinct case from the input-output Jacobian, which is what we focus on in this work.}
singular value distribution of a wide neural network, which is an important indicator of its architectural soundness: If this distribution becomes very diffuse as the network gets deeper (i.e.\ more layers), then an error signal (i.e.\ the gradient) will be badly distorted as it is backpropagated. On the other hand, if the distribution concentrates around 1 even when the network gets deeper, then the error signal is largely preserved, and all layers of the networks receive adequate signal to improve. This idea is known as \emph{dynamical isometry} \citep{pennington_resurrecting_2017} and has been successfully applied in practice to train ultra-deep networks (as deep as 10,000 layers! \citep{xiao_dynamical_2018}).

This calculation of the NN Jacobian singular values distribution can be seen as a vast nonlinear generalization of the classical problem of calculating the singular value distribution for an iid random matrix. The solution of this latter problem is famously given by (the square root of) the Marchenko-Pastur distribution. A related classical random matrix law is the semicircle law, which says that the eigenvalue distribution of the sum $A+A^{\trsp}$ of a random real iid matrix $A$ and its transpose $A^{\trsp}$ is asympotically shaped like a semicircle. Like the utility of NN Jacobian singular values, Marchenko-Pastur and the semicircle law have both been impactful in science and engineering and are considered two fundamental laws of random matrix theory. In fact, the semicircle law was first motivated by a study of the distribution of energy levels of an atom \citep{lane_giant_1955}.

Up until recently, the study of the NN Jacobian singular value distribution lacked a rigorous foundation. The first calculation of it assumed (free) independence (between the weights and the preactivations) that was only empirically checked but not proved. It was unknown how widespread this phenomenon actually was. One purpose of this work is filling in that hole: FIP reveals the architectural universality of this free independence (i.e. the neural network can be as complex as any modern manifestations and the principle still holds).

\subsection{New Approach to Random Matrix Theory}
\label{sec:newRMT}
Another purpose of this work is to illustrate a new way to go about random matrix theory, one that allows us to methodically attack the nonlinear problems of deep neural networks. In contrast, classical methods of random matrix theory typically heavily rely on the linearity of classical random matrix ensembles.
The \emph{expansion technique} is one such method.
We review how it's used in linear problems and show it is ineffective for (nonlinear) neural networks problems.

\paragraph{The expansion technique of classical RMT}
For example, a typical quantity to calculate is the expectation of the trace of some matrix power $\tr(A^{\trsp}A)^{k}$ for some iid random matrix $A$. One common way to proceed is expanding this expression in terms of monomials of the entries of $A$, and then counting the different kinds of monomials that arise: such as monomials multilinear in the entries of $A$, or monomials where every entry appears quadratically, etc. 
Given this expansion, the trace expectation then follows easily.

\paragraph{The expansion technique is ineffective in neural network problems}
Now, in the (nonlinear) case of neural networks, the matrix $A$ is usually something more complicated.
Take, for example, $A=D_{2}BD_{1}B$ for some iid matrix $B$ and diagonal matrices $D_{i}$ that depend nonlinearly on $B$.
This dependence can, for example, take the following form: for some iid vector $v$ and $\phi=\tanh$ (to be applied coordinatewise), we define $u_{1}=\phi(Bv),u_{2}=\phi(Bu_{1})$ and set $D_{i}$ to have diagonal $u_{i}$.
Such ensemble $A$ commonly appears as the Jacobian of a neural network.
Inspired by the classical expansion technique described above, one may attack this problem by expanding $\tr(A^{\trsp}A)^{k}$ in terms of entries of $A$ or in terms of entries of $B$. Because the entries of $A$ are correlated in complicated ways, it will be hard to use the expansion in $A$. On the other hand, we cannot even expand $\tr(A^{\trsp}A)^{k}$ cleanly in $B$ because of the nonlinear dependence of $D_{i}$ on $B$ (particularly because of $\phi$)%
\footnote{We can Taylor expand $D_{i}$ in terms of entries of $B$ but that gets ugly really fast.}.

Thus the usual expansion technique runs into a wall very quickly.

\paragraph*{Our proposed method}
We express the trace $\tr(A^{\trsp}A)^{k}$ as an expectation $\EV_{v}v^{\trsp}(A^{\trsp}A)^{k}v$ where $v$ is a standard Gaussian vector%
\footnote{but this identity holds for any $v$ with iid, zero-mean, unit-variance entries}. Then we inductively \emph{analyze} the vectors $Av,A^{\trsp}Av,AA^{\trsp}Av,(A^{\trsp}A)^{2}v,\ldots,(A^{\trsp}A)^{k}v$, in a way we will soon describe below. Finally, this analysis of $(A^{\trsp}A)^{k}v$ will allow us to calculate $\EV_{v}v^{\trsp}(A^{\trsp}A)^{k}v$.

It turns out that the vectors $v,Av,A^{\trsp}Av,AA^{\trsp}Av,(A^{\trsp}A)^{2}v,\ldots,(A^{\trsp}A)^{k}v$ will 
\begin{enumerate}
  \item all have \emph{approximately iid} coordinates in the large dimension limit, in a suitable sense, i.e. $\{(Av)_{\alpha}\}_{\alpha}$ are approximately iid, and similarly for $\{(A^{\trsp}Av)_{\alpha}\}_{\alpha}$, $\{((A^{\trsp}A)^{2}v)_{\alpha}\}_{\alpha}$, etc, where $\alpha$ denotes coordinate index; \label{item:_approxiid}
  \item but the sequence of coordinates $v_{\alpha},(Av)_{\alpha},(A^{\trsp}Av)_{\alpha},\ldots,((A^{\trsp}A)^{k}v)_{\alpha}$ are correlated for each fixed index $\alpha$, in the same way across all $\alpha$. \label{item:_cor}
\end{enumerate}
By \emph{analyze}, we mean to calculate such correlations.
Then $\EV_{v}v^{\trsp}(A^{\trsp}A)^{k}v$ is given by the correlation between $v_\alpha$ and $((A^{\trsp}A)^{k}v)_{\alpha}$.

\paragraph*{Example}
Let us use the classical example of iid $A$ to make these two points more concrete.

If $v\in\R^{n}$ is a standard Gaussian vector, and $A\in\R^{n\times n},A_{\alpha\beta}\sim\Gaus(0,1/n)$ is iid and independent from $v$, then it's not hard to see $Av$ has coordinates which tend to iid Gaussians via some kind of central limit argument (illustraing point (1)). Next, we will soon see intuitively that $A^{\trsp}Av$ is asymptotically the sum of $v$ and a random Gaussian vector with iid coordinates, independent from $v$ (illustrating point (1)). Thus $A^{\trsp}Av$ is \emph{coordinatewise }correlated with $v$, illustrating point (2). 

To get this intuition, write each coordinate 
\begin{equation}(A^{\trsp}Av)_{\alpha}=\sum_{\beta,\gamma}A_{\beta\alpha}A_{\beta\gamma}v_{\gamma}=\sum_{\beta}A_{\beta\alpha}^{2}v_{\alpha}+\sum_{\beta;\gamma\ne\alpha}A_{\beta\alpha}A_{\beta\gamma}v_{\gamma},\quad \beta,\gamma\in[n].
\label{eqn:ATAv}
\end{equation}
Because each $A_{\beta\alpha}$ has variance $1/n$, the first sum will converge via law of large numbers to $v_{\alpha}$. In the second sum, one can note that each summand is uncorrelated with others (and higher order correlations drop off rapidly with $n$), so one may expect the second sum to converge to a Gaussian through some central limit behavior. We can calculate that the second sum $\sum_{\beta;\gamma\ne\alpha}A_{\beta\alpha}A_{\beta\gamma}v_{\gamma}$ for different $\alpha$s will be uncorrelated, so their limits for different $\alpha$ should be independent Gaussians. Likewise, it should be asymptotically independent from $v$ for the same reason.

This finishes our example and illustrates the overarching philosophy of our proposed method. But it is still tedious if we have to manually derive the correlation between $v_{\alpha},(Av)_{\alpha},(A^{\trsp}Av)_{\alpha},\ldots,((A^{\trsp}A)^{k}v)_{\alpha}$. In addition, it's not entirely clear at this point that this method can handle the nonlinear ensemble example $A=D_{2}BD_{1}B$ given above. This is where the Tensor Programs framework is crucial, which we overview now.

\subsection{The Tensor Programs Framework}

The Tensor Programs framework can be thought of as a clean, scalable, and rigorous packaging of the intuition explained in the last section.
We first demonstrate how one can apply this framework to compute the correlations between $v_{\alpha},(Av)_{\alpha},(A^{\trsp}Av)_{\alpha},\ldots,((A^{\trsp}A)^{k}v)_{\alpha}$.
Then we describe in more general terms what the framework is about.

\paragraph{Calculating $\EV_v v^\trsp (A^\trsp A)^k v$ with Tensor Programs}
The framework associates a random variable $Z^u \in \R$ to each $u \in \{v,Av,A^{\trsp}Av,\ldots,(A^{\trsp}A)^{k}v\}$, representing the distribution of coordinates of $u$ in the large $n$ limit.
It also comes with a set of symbolic rules to derive $Z^{Av}$ given $Z^v$, $Z^{A^\trsp Av}$ given $Z^{Av}$ and $Z^v$, and so on.
For example, consistent with the intuition laid out in \cref{sec:newRMT}, $Z^v$ and $Z^{Av}$ are defined as independent standard Gaussians (since $v$ and $Av$ are asymptotically standard Gaussian vectors, independent from each other), and $Z^{A^\trsp Av} = Z^v + S$ for a standard Gaussian $S \in \R$ independent from $Z^v$ (and $Z^{Av}$).
Finally, we can calculate $\lim_{n\to\infty}\EV_v v^\trsp (A^\trsp A)^k v = \EV Z^v Z^{(A^{\trsp}A)^{k}v}$.

This set of $Z$-rules handle nonlinearities.
For example, we have $Z^{\phi(v)} \defeq \phi(Z^{v})$ for any $\phi: \R \to \R$ (where $\phi$ on the LHS is applied coordinatewise to the vector $v$, but is applied in the usual sense to the scalar random variable $Z^v$ on the RHS.)
This reflects the intuition that $Z^{\phi(v)}$ is the distribution of $\phi(v)$'s coordinates.
As such, it is not hard to see our proposed method in \cref{sec:newRMT} can deal with nonlinear random matrix ensembles like the example $A=D_{2}BD_{1}B$ given there.
The full description of these rules can be found in \cref{defn:Z}, but we will not dive into more details here.

\paragraph{Tensor Programs in general}
While a major focus of this paper is on RMT, the original motivation of Tensor Programs is in understanding wide neural networks.
There were 3 lines of research that molded the framework, which we briefly overview so we can later state some other contributions of this paper:
\emph{(Research Line 1)} In 1994, Radford Neal \citep{neal_bayesian_1995} discovered that shallow neural networks with random weights converge to Gaussian processes as their widths tend to infinity. This Neural Network-Gaussian Process correspondence has become a fascinating topic ever since, with a flurry of activities in recent years extending it \citep{{williams_computing_1997,le_roux_continuous_2007,hazan_steps_2015,daniely_toward_2016},lee_deep_2018,matthews_gaussian_2018_arxiv,novak_bayesian_2018}. \emph{(Research Line 2)} Relatedly, a line of work, closely connected to the study of Jacobian singular values mentioned above, researched how signal propagates inside a neural network with random weights \citep{poole_exponential_2016,schoenholz_deep_2017,yang_mean_2017,yangVarianceVariation,hanin_which_2018,hanin_how_2018,chen_dynamical_2018,yang_mean_2019,pennington_resurrecting_2017,hayou_selection_2018,philipp_nonlinearity_2018}, and use such insights to improve the initialization scheme or architecture of neural networks. \emph{(Research Line 3)} Recently, Jacot et al.\ and others \citep{jacot_neural_2018,allen-zhu_convergence_2018,allen-zhu_convergence_2018-1,allen-zhu_learning_2018,du_gradient_2018,zou_stochastic_2018,lee_wide_2019,arora_exact_2019} found that, when trained with gradient descent, wide, fully connected neural networks evolve like linear models. This resolved many open questions regarding the training and generalization of neural networks, and was followed by intense activity generalizing these results to more modern neural architectures \citep{{yangScalingLimitsWide2019arXiv.org},arora_exact_2019,du2019graph,littwin2020optimization,alemohammad2020recurrent,hron2020infinite}.

In all three lines of research, each new architecture required a new paper. For example, going from fully connected networks to convolutional neural networks required careful thinking about how the newly added dimension of pixel position in the latter changes the argument for the former. It was the expectation that every architecture is special in some way, and extending the theory to catch up with modern practice would require a paper covering every major architectural advancement starting from convolutional neural networks (e.g. normalization layers, attention, etc). Considering the breakneck pace of empirical deep learning, this ``catch-up'' might never happen.

It was against this backdrop that the Tensor Programs framework \citep{yangScalingLimitsWide2019arXiv.org,yangTP1,yangTP2} surprisingly showed all three lines of research can be unified into one and simultaneously be generalized to all practically relevant neural architectures, now or in the future. There are two major insights that make this possible: 
\begin{description}
  \item[Insight 1] Every computation done in deep learning can be written as a program (i.e. a composition) of matrix multiplication and coordinatewise nonlinearities%
  \footnote{A coordinatewise nonlinearity is, in the simplest case, a function that is applied to each coordinate of an input vector independently. More generally, see \refNonlin{}.} (for example, a simple neural network can be written this way as $h=Wx,y=\phi(h)$ for some activation function $\phi$, like $\tanh$, applied coordinatewise; the example of $A=D_{2}BD_{1}B$ given above can be written likewise).
  \item[Insight 2] In every such computation, if the matrices are randomized, then every vector computed over the course of such a program has roughly iid coordinates when the matrices' sizes are large. However, for each $\alpha$, the $\alpha$th coordinates of different vectors will in general be correlated, in the same way across $\alpha$, and in a way that can be inductively calculated from the program structure. The machinery of Tensor Programs allows one to mechanically perform this calculation.
\end{description}
Note that insight 2 is a generalization of (\ref{item:_approxiid}) and (\ref{item:_cor}) in \cref{sec:newRMT}, and is reflected in the $\EV_v v^\trsp (A^\trsp A)^k v$ example above.
Given these two insights, it then becomes rather straightforward to generalize the three lines of research to any relevant architecture.

\subsection{Tensor Programs Master Theorems}
\label{sec:IntroTPMasterThm}
The Tensor Programs framework was first laid out in (the unpublished manuscript) \citet{yangScalingLimitsWide2019arXiv.org} but was written densely. Subsequently, \citep{yangTP1} and \citep{yangTP2} rewrote, in an accessible, pedagogical way, the parts of \citet{yangScalingLimitsWide2019arXiv.org} that deal with research lines 1 and 3 above (resp.\ generalizing Neal and Jacot et al.) (and research line 2 is essentially covered by the combination of \citep{yangTP1} and \citep{yangTP2}). The main technical result of each of \citep{yangTP1} and \citep{yangTP2} is a \emph{Master Theorem} that tells one how to mechanically track the correlation between vectors computed in a program. The difference between their versions of the Master Theorems is in the generality of programs considered: In \citep{yangTP1}, the Master Theorem applies to programs that allow one to re-use matrices, but one cannot use both a matrix and its transpose at the same time. For example, it applies to the computation $AAv$ but not to $A^{\trsp}Av$ (both computations reuse the matrix $A$ but the latter do so through its transpose $A^{\trsp}$). The Master Theorem of \citep{yangTP2} allows $A^{\trsp}$, but only under some restrictive condition. For example, it 
1) allows $DA^{\trsp}u$ where $D$ is a diagonal matrix with diagonal given by $Av$, and $u$ and $v$ are independent standard Gaussian vectors, but
2) disallows $A^{\trsp}Av$.

Nevertheless, the restrictions of \citep{yangTP1} and \citep{yangTP2}'s Master Theorems are natural for their respective settings. Thus, \citep{yangTP1} and \citep{yangTP2} are able to generalize research lines 1 and 3 to their most practically relevant scenarios (which is more than enough for the purpose of deep learning), but not all practically \emph{conceivable} scenarios (for example, when a weight matrix and its transpose are both used in the forward pass of a neural network). Furthermore, their Master Theorems are \emph{not enough} for proving the random matrix results of this paper, which requires machinery that handles programs like $A^{\trsp}Av$.
A major contribution of this paper is \emph{to prove a Master Theorem for any Tensor Program}, without the restrictions above%
\footnote{A version of this general Master Theorem was already presented in \citet{yangScalingLimitsWide2019arXiv.org}. Here we focus on giving a more organized, pedagogical proof of it as well as a succinct outline.}.
From this would naturally follow the aforementioned generalizations of \citep{yangTP1,yangTP2} as well as all of the RMT results in this paper.

\paragraph*{The \emph{Tensor Programs} Series}

This paper will complete the rewriting of \citet{yangScalingLimitsWide2019arXiv.org}. Between \citep{yangTP1}, \citep{yangTP2}, and this paper, most results of \citet{yangScalingLimitsWide2019arXiv.org} are now re-presented in an accessible way. This paper also finishes the foundation for the current version of the Tensor Programs machinery, which we will rely on crucially for several future papers. We intend this paper to be an authoritative reference for the technical details and the formulation of this foundation, going forward.

\paragraph*{Summary of Our Contributions}
\begin{enumerate*}
\item We prove the unrestricted Master Theorem, as explained above (\cref{sec:netsort}).
\item Give new proofs of the semicircle and Marchenko-Pastur laws, benchmarking our theoretical framework against these fundamental mathematical results (\cref{sec:semicircle,sec:MP}).
\item Prove the Free Independence Principle (FIP) (\cref{sec:FIPMain}).
\item Apply FIP to rigorously compute the Jacobian singular value distribution of a randomly initialized NN (\cref{sec:jac}).
\end{enumerate*}

For readers familiar with \citep{yangTP1,yangTP2}, we also: 5.\ Generalize the GP (research line 1) and NTK (research line 3) results of \citep{yangTP1,yangTP2} by allowing the transpose of any weight matrix in the NN forward computation (\cref{sec:generalizedNTK4A,sec:NNGP}).
6.\ Use FIP to give a new proof of how \emph{Simple GIA Check} allows one to assume GIA when computing NTK rigorously (\cref{{sec:GIAFIP}}).

Taken together with \citep{yangTP1,yangTP2}, this paper shows the versatility and power of the \emph{Tensor Programs technique}: To calculate (or show the existence of) some limit, one can just express the quantity of concern in a Tensor Program and apply the Master Theorem mechanically.

\subfile{NetsorT2.tex}

\subfile{semicircleLaw.tex}

\subfile{FreeIndependence.tex}

\subfile{NNjacobian.tex}

\subfile{proofsketch.tex}

\section{Related Works}

\paragraph{Jacobian Singular Value Distribution of a Neural Network}
\citet{pennington_resurrecting_2017,pennington_emergence_2018} originally studied the multilayer-perceptron (MLP)'s Jacobian singular value distribution, in the limit of large width.
\citet{tarnowski_dynamical_2018,ling_spectrum_2019} generalized this analysis to residual MLPs and \citet{xiao_dynamical_2018} to convolutional networks.
These works assumed certain asymptotic freeness as in \cref{{stmt:AsymptoticFreenessAssumption}}, which was first proven rigorously by \citet{yangScalingLimitsWide2019arXiv.org} and presented in a more accessible way here.
Recently, \citet{pastur2020random} gave a new, direct proof of this asymptotic distribution for MLP by induction in its depth and standard random matrix machinery.
In comparison, our technique here is more general, both in the type of architectures allowed (any that is expressible in \netsort{}, which by \citet{yangTP1,yangTP2} includes practically all architectures) and in the nonlinearities involved (\citet{pastur2020random} assumes $\phi, \phi'$ are both bounded but we only require $\phi'$ is bounded).

In this paper, we are concerned with the input-output Jacobian of a neural network on a fix input. 
Other works have considered the Jacobian with respect to the parameters, where the Jacobian has dimension \#parameters $\times$ \#data points (e.g. \citep{pennington_spectrum_2018}). This is a related but distinct case from the input-output Jacobian.

\paragraph*{$\Zdot^{Wx}$ as the \emph{Onsager Correction Term}}

This $\Zdot^{Wx}$ has appeared before, in a limited setting, in the literature of asymmetric message passing as the \emph{Onsager correction term}. Asymmetric message passing \citep{donoho_message_2009} is an algorithm that tries to recover a ground truth signal $v$ from a noisy measurement $Wv$ of it obtained by a matrix $W$. This algorithm repeatedly multiplies the measurement by $W$ and its transpose $W^{\trsp}$, interleaving with coordinatewise nonlinearities. In between matrix multiplications, this Onsager correction term is subtracted explicitly. \citet{bayati_dynamics_2011} famously proved the recovery properties of this algorithm as the matrix size tends to infinity. This algorithm can be written down in a \netsort{} program, and (a version of) \citet{bayati_dynamics_2011}'s results can be realized as a corollary of the \netsort{} Master Theorem (see \citet{yangScalingLimitsWide2019arXiv.org}). In contrast, the Master Theorem keeps track of the Onsager correction term throughout arbitrary computation. This allows us to prove powerful theorems like the Semicircle Law and the Free Independence Principle, as shown in this paper.

\paragraph*{The \emph{Tensor Programs} Series}

This paper is the third in the \emph{Tensor Programs} series, following \citet{yangTP1,yangTP2}. The unrestricted Master Theorem, the new proofs of semicircle and Marchenko-Pastur laws, and the singular value distribution result for multilayer-perceptron originally appeared in \citet{yangScalingLimitsWide2019arXiv.org}, but here we present a clear, pedagogical presentation of them, with some technical improvements as well.
The Free Independence Principle is new and unique to this paper. Compared to the Master Theorem of \citet{yangTP2}, our Master Theorem here does not assume the BP-like condition (which implies that $W^\trsp$ can be assumed independent from $W$, i.e. GIA), but rather works for \emph{any} \netsort{} program.

\section{Conclusion}

In this work, we proved a Master Theorem for any \netsort{} program and applied this new theorem to give new proofs of the semicircle and Marchenko-Pastur laws, as well as to propose and prove the \emph{Free Independence Principle (FIP)}.
FIP then allows us to calculate the asymptotic Jacobian singular value distribution of a neural network.
These results suggest a new way to approach nonlinear random matrix theory pertaining to deep learning.
More generally, in combination with \citet{yangTP1,yangTP2}, they demonstrate the versatility of the \emph{Tensor Programs} technique.

\section*{Acknowledgements}

We thank Boris Hanin, Sam Schoenholz, Zhiyuan Li, Jeffrey Pennington, Etai Littwin, Ilya Razenshteyn, Bobby He, Ryan O'Donnell, Edward Hu, Michael Santacroce, Jason Lee, Judy Shen, Jascha Sohl-Dickstein for feedback on working copies of this manuscript.

\bibliography{references}
\bibliographystyle{plainnat}
\newpage

\appendix

\subfile{extraTheorems.tex}

\subfile{Generalized_GP4A_NTK4A.tex}

\subfile{advancedNetsorT.tex}

\subfile{RMTBackground.tex}

\subfile{MarchenkoPastur.tex}

\subfile{FIPProof.tex}

\subfile{proofs.tex}

\end{document}

%% file: NetsorT2.tex
\begin{figure}[h]
    \centering
    \includegraphics[width=0.9\textwidth]{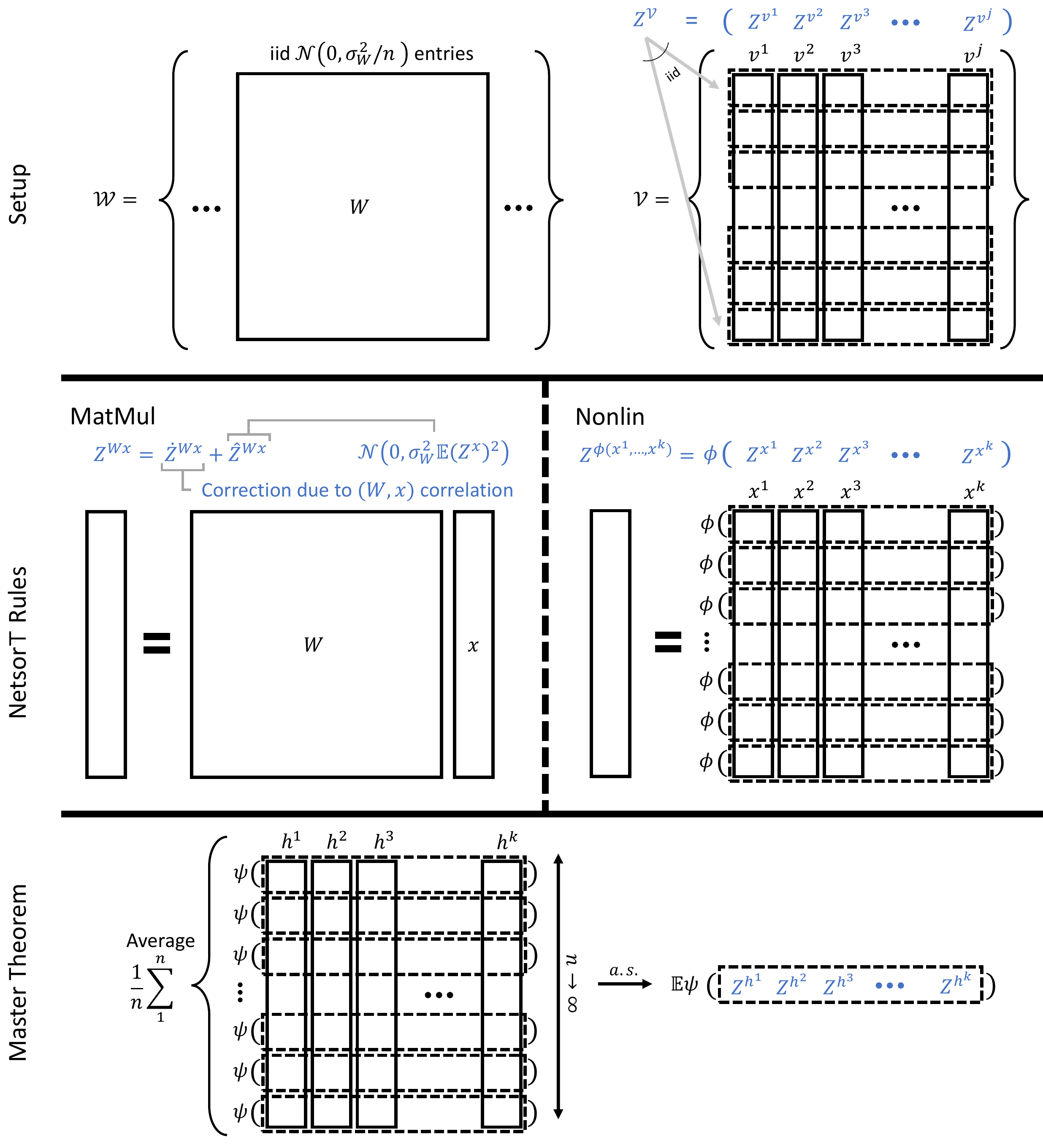}
    \caption{Graphical summary of \netsort{} and its Master Theorem.
    Vectors $v^i$ are initial vectors of the program, but $x^i$ and $h^i$ can be any vector in the program.}
    \label{fig:summary}
\end{figure}

\section{\texorpdfstring{\netsort{}}{NetsorT}: Language for Neural Computation}
\label{sec:netsort}

There are many versions of languages for Tensor Programs, but we shall focus on one version called \netsort{} here.
This entire section is summarized in \cref{fig:summary}, which we encourage the reader to regularly consult over the course of this section.
Originally, \netsort{}%
\footnote{pronounced \emph{netsert} or \emph{netser-T}. The \emph{ts} is pronounced like in \emph{tsar}. The \emph{or} is pronounced as in tens\emph{or}. Another way of thinking is \netsor{} is just \emph{tensor} with \emph{ten} reversed.}
was motivated by a desire to understand the behavior of large (wide) neural networks.
For example, a simple $L$-hidden-layer perceptron can be described by alternating applications of nonlinearities and matrix multiplication: For weight matrices $W^1 \in \R^{n\times d}$ and
$W^2, \ldots, W^L \in \R^{n\times n}$, and nonlinearity $\phi:\R \to \R$, such a neural network on input $\xi\in \R^d$ is given by $h^{1}(\xi)=W^{1}\xi\in\R^{n}$, and%
\footnote{
    Following neural network convention, $\phi(x) = (\phi(x_1), \ldots, \phi(x_n))$.
}
\begin{align}
    x^{l}(\xi)=\phi(h^{l}(\xi)) \in \R^n,
    \quad
    h^{l+1}(\xi)=W^{l+1}x^{l}(\xi)\in\R^{n},
    \quad\text{for $l=1,\ldots, L-1$,}
    \label{eqn:MLP}
\end{align}
and the network output is $f(\xi) = v^\trsp x^L(\xi)$ for some weight vector $v \in \R^n$.%
\footnote{
    For simplicity, we omit biases and set equal the widths of all layers, but these two simplifications can easily be removed; see \cref{appendix:VarDim}.
}
Thus, intuitively, a language of solely \emph{nonlinearity application} and \emph{matrix multiplication} seems to strike a good balance between 1) simplicity (and ease of analysis), and 2) generality.
Indeed, as shown in \citet{yangTP1,yangTP2}, such a language can express \emph{practically all} of modern deep learning, beyond the toy example here.
\citet{yangTP2} formalized a version of this language, called \netsort{}, which we recall here.   
\begin{defn}\label{defn:netsort}
A \netsort{} program is a sequence of $\R^{n}$ vectors (which we will refer to as \emph{vectors in the program}) inductively generated via one of the following ways from an initial set $\mathcal{V}$ of random $\R^{n}$ vectors and a set $\mathcal{W}$ of random $n\times n$ matrices 
\begin{description}
\item [\texttt{Nonlin\label{instr:nonlin}}] Given $\phi:\R^{k}\to\R$ and $x^{1},\ldots,x^{k}\in\R^{n}$, we can generate%
\footnote{Again, $\phi$ is applied coordinatewise.
Here $\phi$ and $k$ should be thought of as fixed while $n \to \infty$.}
$\phi(x^{1},\ldots,x^{k})\in\R^{n}$, where
\[\phi(x^{1},\ldots,x^{k})_\alpha \defeq\phi(x^{1}_\alpha,\ldots,x^{k}_\alpha),\quad\text{for each $\alpha \in [n]$.}\] 
\item [\texttt{MatMul\label{instr:matmul}}] Given $W\in\R^{n\times n}$ and $x\in\R^{n}$, we can generate $Wx\in\R^{n}$ or $W^{\trsp}x\in\R^{n}$.
\end{description}
\end{defn}
For example, in \cref{eqn:MLP}, $h^{l}(\xi)=W^{l}x^{l-1}(\xi)$ is an instance of \refMatMul{}, while $x^{l}(\xi)=\phi(h^{l}(\xi))$ is an instance of \refNonlin{}.
We are interested in understanding the behavior of \netsort{} programs when $n$ is large, and when $\mathcal{V}$ and $\mathcal{W}$ are sampled as follows:
\begin{setup}[\netsort{}]\label{setup:netsort}
1) For each initial $W\in\mathcal{W}$, we sample iid $W_{\alpha\beta}\sim\Gaus(0,\sigma_{W}^{2}/n)$ for some variance $\sigma_{W}^{2}$ associated to $W$, independent of other $W' \in \mathcal {W}$; 2) for some multivariate Gaussian $Z^{\mathcal{V}}=\left\{ Z^{h}:h\in\mathcal{V}\right\} \in\R^{\mathcal{V}}$, we sample the initial set of vectors $\mathcal{V}$ like $\left\{ h_{\alpha}:h\in\mathcal{V}\right\} \sim Z^{\mathcal{V}}$ iid for each $\alpha\in[n]$.
\end{setup}
\begin{exmp}\label{exmp:MLPsetup}
For example, we will often be interested in understanding the behavior of \cref{eqn:MLP} when $W^1,\ldots, W^L$ are sampled randomly.
In this case, $\mathcal W$ consists of $W^2, \ldots, W^L \in \R^{n\times n}$, each of which is sampled like $W^l_{\alpha\beta}\sim\Gaus(0, 1/n)$.
On the other hand, $\mathcal V$ consists of $h^1(\xi)=W^1 \xi$, which is distributed like $h^1(\xi)_\alpha \disteq \Gaus(0, \|\xi\|^2/d)$ if $W^1 \in \R^{n\times d}$ is sampled like $W^1_{\alpha\beta} \sim \Gaus(0,1/d)$.%
\footnote{In neural network settings, we are interested in the limit where the intermediate dimensions $n\to\infty$ but the input dimension $d$ stays fixed.
Thus, we treat the input-to-hidden matrix $W^1$ differently from other $W^l$.}
If we compute the MLP in \cref{eqn:MLP} on other inputs $\xi_1, \ldots, \xi_k$ as well, then $\mathcal V = \{W^1 \xi, W^1 \xi_1, \ldots, W^1 \xi_k\}$ with $Z^{\mathcal V}$ being the multivariate Gaussian with covariance $\Cov(Z^{W^1 \xi_i}, Z^{W^1 \xi_j}) = \xi_i^\trsp \xi_j / m$.
\end{exmp}

As discussed in \cref{sec:newRMT}, to understand a random matrix, it suffices to understand random vectors calculated from it.
How should we understand the random vectors in a \netsort{} program as $n\to\infty$? As hinted in \citet{yangTP1,yangTP2}, it turns out that every vector will have roughly iid coordinates in this limit, even though matrix multiplication by $W$ or $W^{\trsp}$ will introduce correlation between coordinates for any finite $n$.
Following \citet{yangTP2}, we shall define in \cref{defn:Z} a random variable $Z^h$ that describes this asymptotic coordinate distribution of each vector $h$; but first, let's use a few examples to motivate the rules governing $Z^h$.

\begin{exmp}\label{exmp:MLPZs}
In \cref{exmp:MLPsetup}, by \cref{setup:netsort}, we already have $Z^{h^1(\xi)} = \Gaus(0, \|\xi\|^2/n)$, reflecting the fact that each coordinate $h^1(\xi)_\alpha$ is an iid sample of $\Gaus(0, \|\xi\|^2/n)$.
For brevity, we drop the explicit dependence of $h$ and $x$ on $\xi$ below.
Next, since ${x^1} = {\phi(h^1)}$, we intuitively have $Z^{x^1} = \phi(Z^{h^1})$, as $\phi$ is applied to each coordinate separately.

Now, ${h^2} = W^2 x^1$ has approximately iid Gaussian coordinates, due to $W^2$ being sampled independent of $x^1$ and a central limit argument.
By some simple back-of-the-envelope calculation, the Gaussian coordinates should asymptotically have zero mean and variance $\EV(Z^{x^1})^2$, and they are uncorrelated with $x^1$.
Thus, it makes sense to set $Z^{h^2} = \Gaus(0, \EV(Z^{x^1})^2)$, independent from $Z^{x^1}$ and $Z^{h^1}$.

Our reasoning above can be repeated, and we derive, recursively, that $Z^{h^l} =  \Gaus(0, \EV (Z^{x^{l-1}})^2), Z^{x^l} = \phi(Z^{h^l})$, with $Z^{h^l}, Z^{x^l}$ independent from $Z^{h^r}, Z^{x^r}$ for all $r < l$.
\end{exmp}

In this example, the derivation of the $Z$s seems like a fairly simple repackaging of the central limit theorem.
However, when both a matrix $A$ and its transpose get involved, the derivation can become complex and subtle quickly, as we show below.

\begin{exmp}\label{exmp:ATAvZ}
This is already apparent in the $A^\trsp A v$ example in the introduction, where $v\in\R^{n}$ is a standard Gaussian vector, and $A\in\R^{n\times n},A_{\alpha\beta}\sim\Gaus(0,1/n)$ is iid and independent from $v$.
By \cref{eqn:ATAv}, $A^\trsp A v$ is roughly $v + g$ for some standard Gaussian vector $g$ independent from $v$.
Therefore, we should set $Z^{A^\trsp A v} = Z^v + G$ for a standard Gaussian $G\in \R$ independent from $Z^v$.

Compare this with $\tilde A A v$ for some independent copy $\tilde A$ of $A^\trsp$.
Then the calculation of \cref{exmp:MLPZs} would have $Z^{\tilde A A v}$ be a standard Gaussian independent from $Z^v$.
This shows the derivation of the $Z$s can require nuance, depending on the interaction between a matrix and its transpose.

It will turn out that, for any vector $u$ that may depend on $A$ and $A^\trsp$, $A^\trsp u$ is always a sum of an asymptotically Gaussian part and a correction term.
The Gaussian part comes from assuming $A$ to be independent from $A^\trsp$ and applying a central limit argument as in \cref{exmp:MLPZs}.
The correction term captures the interaction between $A$ and $A^\trsp$.
For instance, if we take $u = Av$ in the $A^\trsp A v$ example above, then the Gaussian part is $g$ and the correction term is $v$.
In contrast, $\tilde A A v$ doesn't have the correction term because $\tilde A$ and $A$ are independent.

In the formal definition (\cref{defn:Z}) of $Z$s, this decomposition of $A^\trsp u$ is reflected in the definition of $Z^{A^\trsp u}$ as a sum $\Zdot^{A^\trsp u} + \Zhat^{A^\trsp u}$.
Here $\Zhat^{A^\trsp u}$ represents the Gaussian part and $\Zdot^{A^\trsp u}$ represents the correction.
\end{exmp}

Finally, we state the formal definition of $Z$.
Here, the \refZNonlin{} and \refZhat{} rules are probably intuitive given the examples above, but the \refZdot{} rule may appear cryptic at first.
We digest the \refZdot{} rule more slowly in \cref{rem:ZIntuition} after \cref{defn:Z}.

\begin{pabox}[defn:Z]{Key Takeaways for Understanding a \netsort{} Program}
    Each vector $h$ will have coordinates roughly distributed as some random variable $Z^{h} \in \R$ (in a sense to be formalized in \cref{thm:NetsorTMasterTheorem}), which are \emph{symbolically defined} recursively as:
    \begin{description}
    \item [\texttt{ZInit}\label{Z:Initial}] If $h\in\mathcal{V}$, then $Z^{h}$ is defined as the random variable given in \cref{setup:netsort}. We also set $\hat{Z}^{h}\defeq Z^{h}$ and $\Zdot^{h}\defeq 0$.
    \item [\texttt{ZNonlin}\label{Z:Nonlin}] For any fixed (i.e. constant as $n\to\infty$) $k$ and $\phi:\R^{k}\to\R$, we have
    \[
    Z^{\phi(x^{1},\ldots,x^{k})}\defeq \phi(Z^{x^{1}},\ldots,Z^{x^{k}}).
    \]
    \item [\texttt{ZMatMul}\label{Z:MatMul}] $Z^{Wx}\defeq \hat{Z}^{Wx}+\Zdot^{Wx}$
    for every $W_{\alpha\beta}\sim\Gaus(0,\sigma_{W}^{2}/n)$ and vector $x$, where
    \begin{description}
    \item [\texttt{ZHat}\label{Z:Zhat}] {$\hat{Z}^{Wx}$} is a Gaussian variable with zero mean. Let $\mathcal V_W$ denote the set of all vectors in the program of the form $W y$ for some $y$.
    Then $\{\hat Z^{W y}: W y \in \mathcal V_W\}$ is defined to be jointly Gaussian with zero mean and covariance
    \[
    \Cov\left(\hat{Z}^{Wx},\hat{Z}^{W y}\right)\defeq \sigma_{W}^{2}\EV Z^{x}Z^{y},\quad\text{for any \ensuremath{Wx,W y\in\mathcal{V}_W}.}
    \]
    Furthermore, $\{\hat Z^{W y}: W y \in \mathcal V_W\}$ is mutually independent from $\{\hat Z^{v }: v \in \mathcal V \cup \bigcup_{\bar W \ne W} \mathcal V_{\bar W}\}$, where $\bar{W}$ ranges over $\mathcal{W}\cup\{A^{\trsp}:A\in\mathcal{W}\}$.
    \item [\texttt{ZDot}\label{Z:Zdot}]
    By the definition in this box, $Z^x$ is always a deterministic function of a set of $\Zhat^{\bullet}$ random variables.
    Then the partial derivative $\partial Z^x / \partial \Zhat^\bullet$ can be defined symbolically and is another random variable.
    Then we set
    \[\Zdot^{Wx}\defeq \sigma_{W}^{2}\sum Z^{y}\EV\frac{\partial Z^{x}}{\partial\hat{Z}^{W^{\trsp}y}},\]
    summing over $\hat Z^{W^\trsp y}, W^\trsp y \in \mathcal V_{W^\trsp},$ that $Z^x$ is a function of (where $\mathcal V_{W^\trsp}$ is defined in \refZhat{}).
    There is some nuiance in this definition, so see \cref{rem:PartialDer} and \ref{rem:ExpectationPartialDer}.
    \end{description}
    \end{description}
\end{pabox}

The rules above are largely the same as in \cite{yangTP2}, except the definition of $\hat{Z}^{Wx}$ and $\Zdot^{Wx}$. In the restricted \netsort{} programs of \citep{yangTP2}, $\Zdot^{Wx}$ turns out to be 0 (see \cref{thm:GIA}), so $\hat{Z}^{Wx}$ was implicitly identified with $Z^{Wx}$. However, a general \netsort{} program will have $\Zdot^{Wx}\ne0$. 

\begin{rem}
Note that $Z^{x},\hat{Z}^{x},\Zdot^{x}$ only depend on how $x$ is computed in the \netsort{} program (i.e. the program structure), not the specific (random) value of $x$.
\end{rem}

\begin{exmp}
In \cref{exmp:MLPZs}, $\Zdot^{h^l}$ as defined in \refZdot{} is always zero (because we never use $W^{l\trsp}$ in \cref{eqn:MLP}), and by \refZMatMul{}, we simply have $Z^{h^1} = \hat Z^{h^1} = \Gaus(0, \|\xi\|^2/d)$, and recursively, $Z^{h^l} = \hat Z^{h^l} = \Gaus(0, \EV (Z^{x^{l-1}})^2), Z^{x^l} = \phi(Z^{h^l})$, exactly as derived in \cref{exmp:MLPZs}.
\end{exmp}

\begin{exmp}\label{exmp:ATAvZdot}
    Write $x = A^\trsp A v$ in \cref{exmp:ATAvZ} as an explicit \netsort{} program $y = A v, x = A^\trsp y$.
    Then by \refZhat{}, $\Zhat^x = \Gaus(0, 1)$, independent from $Z^y$ and $Z^v$.
    By \refZdot{}, $\Zdot^x = Z^v \EV \f{\partial Z^y}{\partial \Zhat^{A v}} = Z^v \EV \f{\partial (\Zhat^{Av} + \Zdot^{Av})}{\partial \Zhat^{A v}} = Z^v \EV 1 = Z^v$.
    This verifies the decomposition $A^\trsp A v \approx v + g$ in \cref{exmp:ATAvZ}.
\end{exmp}

\begin{rem}[{Intuition for definition of $Z^h$}]\label{rem:ZIntuition}

    The rules \refZInput{} and \refZNonlin{} aptly follow the stated intuition of ``$Z^h$ as coordinate distribution of $h$.''
    The \refZMatMul{} rule is more complex, so let's digest it a bit.
    
    First suppose $W^\trsp$ is never used in the program.
    Then $\Zdot^{Wx}$ vanishes, and $Z^{Wx} = \hat Z^{Wx}$.
    The definition of $\hat Z^{Wx}$ then roughly says $Wx$ is an isotropic Gaussian vector, which is correlated with other vectors of the form $W\bar x$ in a natural way:
    $\la Wx, W\bar x \ra \approx \sigma_W^2 \la x, \bar x \ra$, where $\la,\ra$ denotes dot product.
    
    However, if $W^\trsp$ is used previously, then this ``Gaussian description'' of $Wx$ needs a correction.
    The definition of $\Zdot^{Wx}$ says this correction is a linear combination of previous vectors $y^i$ such that $W^\trsp y^i$ has been used to compute $x$.

    Let us illustrate the meaning of the coefficients of this linear combination through some simple calculations.
    If $W\in\R^{n\times n}$, $W_{\alpha\beta}\sim\Gaus(0,1/n)$, and $x\in\R^{n}$, then $Wx$ should be correlated with $W$.
    If $x$ depends on $W^\trsp y^i$ for a collection of vectors $\{y^i\}_i$, then we can detect this correlation as follows:
    with $\la,\ra$ denoting dot product, for each $i$,
    \[
    \langle y^{i},Wx\rangle=\langle W^{\trsp}y^{i},x\rangle.
    \]
    Thus, if $x$ is correlated with $W^{\trsp}y^{i}$, then $Wx$ should also be correlated with $y^{i}$. The \refZdot{} rule, remarkably, says that such correlations are the \emph{only} correction needed to the ``Gaussian description'' of $Wx$: ${Wx}$ splits into the sum of 1) a component with coordinates $\approx \Zdot^{Wx}$ that resides in the linear span of $\{y^{i}\}_{i}$, which is entirely determined by the inner products $\langle y^{i},Wx\rangle=\langle W^{\trsp}y^{i},x\rangle$ for all $i$,%
    \footnote{
        The set of coefficients $\{\f 1 n \langle W^{\trsp}y^{i},x\rangle \approx \EV Z^{W^{\trsp}y^{i}} Z^x\}_i$ is linearly related to the coefficients $\{\EV \hat Z^{W^{\trsp}y^{i}} Z^x\}_i$, as can be seen from an easy inductive argument.
        The latter is then linearly related to the partial derivative expectations of \refZdot{} by \cref{rem:ExpectationPartialDer}.
    } and 2) another orthogonal component with coordinates $\approx\hat{Z}^{Wx}$ that comes from naively assuming $W^{\trsp}$ to be independent of $W$.

\end{rem}

The following \emph{Master Theorem} rigorously relates the \emph{symbolically constructed} random variables $Z^h$ to the vectors $h$ in the program and their \emph{analytic limits}.
It is one of our main results and will also be our main workhorse in this paper.

\begin{thm}[\netsort{} Master Theorem]
\label{thm:NetsorTMasterTheorem} Fix a \netsort{} program. Suppose the initial matrices $\mathcal{W}$ and vectors $\mathcal{V}$ are sampled in the fashion of \cref{setup:netsort}.
Assume all nonlinearities $\phi$ used in \refNonlin{} are polynomially bounded. Then for any fixed $k$ and any \emph{polynomially bounded}%
\footnote{
    $\phi: \R^k \to \R$ is \emph{polynomially-bounded} if $|\phi(x)| \le C\|x\|^p + c$ for some $p, C, c > 0$, for all $x \in \R^k$.
}
$\psi:\R^{k}\to\R$, as $n\to\infty$,
\begin{equation}
\f 1n\sum_{\alpha=1}^{n}\psi(h_{\alpha}^{1},\ldots,h_{\alpha}^{k})\asto\EV\psi(Z^{h^{1}},\ldots,Z^{h^{k}}),
\label{eqn:NetsorTMasterTheorem}
\end{equation}
for any collection of vectors $h^{1},\ldots,h^{k}$ in the program, where $Z^{h^{i}}$ are defined in \cref{defn:Z}.%
\footnote{Difference with \cite[Thm 6.3]{yangScalingLimitsWide2019arXiv.org}: We have gotten rid of the ``rank convergence'' assumption (which we call ``rank stability'' in this paper) by showing that it comes for free.
See \cref{thm:rankstabilitymain} and see \ref{IH:coreSet} and \cref{lemma:rankStability} in \cref{sec:proofMainTheorem}.}
\end{thm}
\cref{thm:NetsorTMasterTheorem} says that each ``coordinate slice'' $(h^1_\alpha, \ldots, h^k_\alpha)$ can be thought of as an iid copy of $(Z^{h^1},\ldots, Z^{h^k})$. Indeed, as consequences of the Master Theorem, \cref{thm:NetsorTConvergenceInDistribution,{thm:NetsorTMultiConvergenceInDistribution}} formalize this intuition into a convergence in distribution.
Versions of \cref{thm:NetsorTMasterTheorem} with convergence in mean (instead of almost sure) are also available (\cref{thm:NetsorTMeanConvergence,thm:NetsorTMeanConvergenceLinearlyBounded}).
\cref{thm:NetsorTMasterTheorem} strictly generalizes the Master Theorem in \citet{yangTP2}.
The \netsor{} Master Theorem in \citet{yangTP1} allows nonlinearities growing faster than polynomially, but otherwise is also a special case of \cref{thm:NetsorTMasterTheorem}.
We outline the proof of \cref{thm:NetsorTMasterTheorem} in \cref{sec:proofsketch} and give a full proof in \cref{sec:proofMainTheorem}.

\begin{rem}[Partial derivative]\label{rem:PartialDer}
    The partial derivative in \refZdot{} should be interpreted as follows.
    By a simple inductive argument, $Z^x$ for every vector $x$ in the program is defined \emph{uniquely} as a deterministic function $\varphi(\Zhat^{x^1}, \ldots, \Zhat^{x^k})$ of some $x^1, \ldots, x^k$ in $\mathcal V$ or introduced by \refMatMul{}.%
    \footnote{In the context of \netsortplus{} introduced later in \cref{sec:netsortplus}, this is still true, but $\varphi$ here will take the form of a parametrized nonlinearity $\varphi(-;\mathring \theta_1, \ldots, \mathring \theta_l)$ with some deterministic parameters $\mathring \theta_1, \ldots, \mathring \theta_l$.}
    For instance, in \cref{exmp:ATAvZdot}, $Z^x = \Zhat^x + \Zhat^v$ if $v \in \mathcal V$, so $\varphi$ is given by $\varphi(a, b) = a + b$.
    Then
    \begin{equation*}
        \partial Z^x / \partial \Zhat^{x^i} \defeq \partial_i \varphi(\Zhat^{x^1}, \ldots, \Zhat^{x^k}), \quad \text{and} \quad \partial Z^x / \partial \Zhat^z \defeq 0 \text{ for any } z \not\in\{x^1, \ldots, x^k\}.
    \end{equation*}
    Note this definition depends on the precise way the program is written, not just on the underlying mathematics.
    For example, if $y,z \in \mathcal V$ and $x = \phi(W(y+z))$, then $Z^x = \phi(\Zhat^{W(y+z)})$ so that $\partial Z^x/\partial \Zhat^{Wy} = \partial Z^x/\partial \Zhat^{Wz} = 0$.
    If instead, we have $x = \phi(Wy+Wz)$, then $Z^x = \phi(\Zhat^{Wy} + \Zhat^{Wz})$ so that $\partial Z^x/\partial \Zhat^{W(x+y)} = 0$.
    However, in both cases, $\Zdot^{W^\trsp x} = (Z^y + Z^z)\EV \phi'(\Zhat^{W(y+z)})$.
\end{rem}

\begin{rem}[Partial derivative expectation]\label{rem:ExpectationPartialDer}
    The quantity $\EV\frac{\partial Z^{x}}{\partial\hat{Z}^{W^{\trsp}y}}$ is well defined if $Z^{x}$ is differentiable in $\hat{Z}^{W^{\trsp}y}$. However, even if this is not the case, e.g. if $x=\theta(W^\trsp y)$ where $\theta$ is the Heavyside step function, we can still define this expectation by leveraging Stein's lemma (\cref{lemma:stein}):
    
    In \refZdot{}, suppose $\{W^{\trsp}y^{i}\}_{i=1}^{k}$ are all elements of $\mathcal V_{W^\trsp}$ introduced before $x$.
    Define the matrix $C\in\R^{k\times k}$ by $C_{ij}\defeq \EV Z^{y^{i}}Z^{y^{j}}$ and define the vector $b\in\R^{k}$ by $b_{i}\defeq \EV\hat{Z}^{W^{\trsp}y^{i}}Z^{x}$. If $a=C^{+}b$ (where $C^{+}$ denotes the pseudoinverse of $C$), then in \refZdot{} we may set
    \begin{equation}
    \sigma_{W}^{2}\EV\frac{\partial Z^{x}}{\partial\hat{Z}^{W^{\trsp}y^{i}}}=a_{i}.\label{eq:ExpectationPartialDer}
    \end{equation}
    This definition agrees with the partial derivative expectation by Stein's lemma (\cref{lemma:stein}) when the latter is well defined.
    \cref{thm:NetsorTMasterTheorem} holds with this broader definition of partial derivative expectation.
\end{rem}

\paragraph{Extensions}
In \cref{sec:netsortplus}, we describe \netsortplus{}, an extension to \netsort{} by allowing programs to compute the average coordinate of a vector, and use such scalars in \refNonlin{}.
In \cref{appendix:VarDim}, we also describe modification to the Master Theorems if, instead of requiring all matrices in $\mathcal W$ to be square, we allow rectangular matrices.

%% file: semicircleLaw.tex
\section{Semicircle Law}
\label{sec:semicircle}

The \emph{Semicircle Law} \citep{wigner_characteristic_1955,wigner_distribution_1958} is a classical result, of central importance in statistics, physics, and engineering \citep{kivelson_global_1992,von_oppen_conductivity_2000,couillet_random_2011,mezzadri_recent_2005}, on the spectrum of a random Hermitian matrix with independent, zero-mean entries.
It says that, as the size of the matrix tends to infinity, the distribution of its eigenvalues tends to a \emph{semicircle distribution.}
\begin{defn}
    The semicircle distribution $\musc$ is the distribution with density $\propto \sqrt{4-x^2}$.
\end{defn}
\begin{thm}[Semicircle Law for GOE]\label{thm:semicircleLawGOE}
    For each $n \ge 1$, define the random symmetric matrix%
    \footnote{This is known as the Gaussian Orthogonal Ensemble (GOE). While we will only talk about the GOE case here, the semicircle law holds for generic random matrices with iid entries.
    We hope to show universality of \netsort{} Master Theorem in the future, which would automatically generalize our proof here to such cases.}
    $ A = A(n) = W+W^{\trsp}$ for iid Gaussian matrix $W\in\R^{n\times n}$, $W_{\alpha\beta}\sim\Gaus(0,1/{2n})$.
    Let $\lambda_1, \ldots, \lambda_n \in \R$ be the eigenvalues of $A$; these are random variables.
    Then for every compactly supported, continuous $\varphi: \R \to \R$, as $n\to\infty$, we have
    \[\f 1 n \sum_{\alpha=1}^n \varphi(\lambda_\alpha) \asto \EV_{\lambda\sim \musc}\varphi(\lambda).\]
\end{thm}
In this section, we give a new proof of this beautiful result using the \netsort{} Master Theorem. Our purpose is two-fold: we 1) give concrete examples of how to compute with the Master Theorem, especially the new $\Zdot^{Wx}$ rule, and 2) demonstrate our framework is at least powerful enough to prove this cornerstone result.
In \cref{sec:MP}, we also prove the \emph{Marchenko-Pastur Law} with this technique.

By the well-known \emph{Moment Method} (see \cref{{fact:MomentMethod}}), it suffices to prove
\[\inv n \tr A^r \asto \EV_{\lambda \sim \musc} \lambda^r,\quad r=1,2,\ldots\]
It is well known that the semicircle distribution has moments given by the Catalan numbers $C_{k}$
\[
\EV_{\lambda\sim\musc}\lambda^{r}=\begin{cases}
C_{k} & \text{if \ensuremath{r=2k}}\\
0 & \text{otherwise}
\end{cases}
\]
where the Catalan numbers $C_{k}$ are the unique numbers satisfying
\begin{equation}
C_{0}=1,\quad C_{k+1}=\sum_{i=0}^{k}C_{i}C_{k-i}.
\label{eqn:Catalan}
\end{equation}
The first few Catalan numbers are $C_{0}=1,C_{1}=1,C_{2}=2,C_{3}=5,C_{4}=14$. For more background, see \citet{tao_topics_2012}. 

\subsection{The Main Proof Idea}

\newcommand{\semicirc}{\put(2.5,2){\oval(4,4)[t]}\put(0.5, 1.9){\line(1,0){4}}\phantom{\circ}}

We need to show for all integers $k$,
\[
\inv n \tr A^{2k}\asto C_{k},\quad\text{and}\quad\inv n \tr A^{2k+1}\asto0.
\]
To do so, we use a trivial but useful equality: for any $M\in\R^{n\times n}$
\begin{equation}
\tr M=\EV_z z^{\trsp}Mz,\quad\text{for}\quad z\sim\Gaus(0,I).
\label{eqn:zdef}
\end{equation}
Then
\[
\tr A^{2k}=\EV_z z^{\trsp}A^{2k}z,
\]
which we can express as a \netsort{} program: Let $\mathcal V=\{z\}$, and $\mathcal W = \{W\}$ (with sampling data $Z^{z}=\Gaus(0,1), \sigma_W^2 = 1/2$).
With $z^{0}=z$, we define recursively
\begin{equation}
\boxed{x^{t}=Wz^{t-1},\quad y^{t}=W^{\trsp}z^{t-1},\quad z^t=x^{t}+y^{t}.}
\label{eqn:semicircleTP}
\end{equation}

Then, mathematically, we have computed
\[
z^{t}=A^{t}z,\quad\text{and thus}\quad\tr A^{t}=\EV_z z^{\trsp}z^{t}.
\]
Note $A^t$ is the $t$th power of $A$ but the $t$ in $x^t, y^t, z^t$ appear as indices.
By the Master Theorem (and some additional arguments below), it then suffices to show
\begin{align*}
    \EV Z^z Z^{z^{2k}} = C_k,\ \EV Z^z Z^{z^{2k+1}}=0
    \quad \text{so that}\quad
    \inv{n}\EV_z z^{\trsp}z^{2k}\asto C_{k},\ 
    \inv{n}\EV_z z^{\trsp}z^{2k+1}\asto0
\end{align*}

\subsection{Examples of First Few Moments}

We first construct the random variables $Z^{z^{t}},Z^{x^{t}},Z^{y^{t}}$, for $z^t, x^t, y^t$ defined in \cref{eqn:semicircleTP}. While we can do the proof much more succinctly, in the pedagogical spirit, let's do the first few manually to get a feel for it. 

\paragraph*{First Moment}

First we have $Z^{z^{0}} =\Gaus(0,1)$ by definition.
Then $Z^{x^{1}} = \hat Z^{x^{1}} + \Zdot^{x^{1}} = \hat Z^{x^{1}}$ because we have not used $W^\trsp$ yet, so $\Zdot^{x^1} = 0$.
Now, by \refZhat{}, $\EV (\hat Z^{x^1})^2 = \sigma_W^2 \EV (Z^{z^0})^2 = 1/2 \cdot 1 = 1/2.$
Therefore, $Z^{x^1} = \hat Z^{x^{1}}$ is a Gaussian with zero mean and variance $1/2$, independent from $Z^{z^0}$.

A similar reasoning shows $\hat Z^{y^1} = \Gaus(0, 1/2)$ as well, independent from $Z^{z^0}$ and $Z^{x^1}$.
Next we apply \refZdot{} to calculate $\Zdot^{y^1}$.
But $y^1$ does not depend on any vector of the form $W^\trsp \bullet$, so $\Zdot^{y^1} = 0$.
Thus, $Z^{y^1} = \hat Z^{y^1}$,  as a summary,
\begin{align*}
Z^{z^{0}} & =\Gaus(0,1),\quad
Z^{x^{1}} =\Gaus(0,1/2),\quad
Z^{y^{1}} =\Gaus(0,1/2),
\end{align*}
all independent from each other.
In general, $Z^{y^i}$ and $Z^{x^i}$ are symmetric, as illustrated here.
Then
\[
Z^{z^{1}}=Z^{x^{1}}+Z^{y^{1}}
\implies
\EV\left(Z^{z^{1}}\right)^{2}=1\quad\text{and}\quad
\inv n \tr A \asto \EV Z^{z^{1}}Z^{z^{0}}=0.
\]

\paragraph*{Second Moment}

Next,
\[
Z^{x^{2}}=\hat{Z}^{x^{2}}+\Zdot^{x^{2}}
\]
with $\hat{Z}^{x^{2}}$ independent from $Z^{y^1}, Z^z$, but jointly Gaussian with $\hat{Z}^{x^{1}}=Z^{x^{1}}$ with (co-)variance
\[
\EV(\hat{Z}^{x^{2}})^{2}=\frac{1}{2}\EV(Z^{z^{1}})^{2}=\frac{1}{2},\quad
\Cov(\hat{Z}^{x^{2}},\hat{Z}^{x^{1}})=\frac{1}{2}\EV Z^{z^{1}}Z^{z^{0}}=0
\quad\text{(i.e. independent).}
\]
On the other hand, since $y^1 = W^\trsp z^0$ is the only usage of $W^\trsp$ in defining $x^2$,
\[
\Zdot^{x^{2}}=\frac{1}{2}Z^{z^{0}}\EV\frac{\partial Z^{z^{1}}}{\partial \hat Z^{y^{1}}}=\frac{1}{2}Z^{z^{0}}\EV1=\frac{1}{2}Z^{z^{0}}.
\]
Altogether, we have
\[
Z^{x^{2}}=\hat{Z}^{x^{2}}+\Zdot^{x^{2}}=\hat{Z}^{x^{2}}+\frac{1}{2}Z^{z^{0}},\quad
\text{
and symmetrically,}
\quad
Z^{y^{2}}=\hat{Z}^{y^{2}}+\Zdot^{y^{2}}=\hat{Z}^{y^{2}}+\frac{1}{2}Z^{z^{0}}.
\]
Since $\hat Z^{x^2}$ and $\hat Z^{y^2}$ are zero-mean and independent from $Z^{z^0}$, we have
\[
Z^{z^{2}}=Z^{x^{2}}+Z^{y^{2}}=\hat{Z}^{x^{2}}+\hat{Z}^{y^{2}}+Z^{z^{0}}
\implies
\inv n \tr A^{2} \asto \EV Z^{z^{2}}Z^{z^{0}} = \EV (Z^{z^{0}})^2 = 1 =C_{1}.
\]

\subsection{Proof for General Moments}
\label{sec:semicircleProofGeneralK}

\paragraph{Overview}
Notice how $Z^{z^t}, t=1,2,$ above are of the form $Z^{Z^t} = \tau_t Z^{z^0} + S$ for some $\tau_t\in\R$ and some zero-mean $S$ independent from $Z^{z^0}$.
We will show this is the case for all $Z^{z^t}$, so that $\EV Z^{z^t} Z^{z^0} = \EV (\tau_t Z^{z^0} + S) Z^{z^0} = \tau_t$.
By \cref{{thm:MasterTheoremmeanConvergenceMainText}} below (which is just the Master Theorem plus a standard truncation argument to turn the almost sure convergence into almost sure convergence of conditional expectations), this means $\inv n \tr A^t = \inv n \EV_z z^\trsp z^t \asto \tau_t$.
Finally, we will show $\tau_{2k+1}=0$ and $\tau_{2k}$ satisfies the same recurrence as $C_{k}$, which then yields the desired result.

The following theorem is a direct consequence of the more general \cref{thm:NetsorTMeanConvergenceLinearlyBounded}.
\begin{thm}\label{thm:MasterTheoremmeanConvergenceMainText}
    With the same premise as in \cref{thm:NetsorTMasterTheorem}, suppose further $\psi$ is quadratically bounded and all nonlinearities used in \refNonlin{} are linearly bounded.
    Let $\mathcal S \sbe \mathcal V$ be a subcollection of initial vectors.
    Then
    \begin{equation}
    \f 1n \EV_{\mathcal S} \sum_{\alpha=1}^{n}\psi(h_{\alpha}^{1},\ldots,h_{\alpha}^{k})
    \asto
    \EV\psi(Z^{h^{1}},\ldots,Z^{h^{k}}).
    \label{eqn:condmeanconvergenceMainText}
    \end{equation}
    where $\EV_{\mathcal S}$ denotes conditional expectation given the values of vectors in $\mathcal V \setminus \mathcal S$ and of matrices in $\mathcal W$.
\end{thm}

\paragraph{Calculations}
In general, $\{\hat{Z}^{x^{s}}\}_{s}$, $\{\hat{Z}^{y^{s}}\}_{s}$, $Z^{z^0}$ are zero-mean and independent from one another.
We have $\{\hat{Z}^{x^{s}}\}_{s}$ is jointly Gaussian with covariance
\[
\Cov(\hat{Z}^{x^{s}},\hat{Z}^{x^{r}})=\frac{1}{2}\EV Z^{z^{s-1}}Z^{z^{r-1}}
\]
and $\{\hat{Z}^{y^{s}}\}_{s}$ satisfies symmetric covariance identities.
In addition, by \refZdot{},
\begin{align*}
\Zdot^{x^{t+1}} & =\frac{1}{2}\sum_{s=0}^{t-1}Z^{z^{s}}\EV\frac{\partial Z^{z^{t}}}{\partial\hat{Z}^{y^{s+1}}},\quad
\Zdot^{y^{t+1}} =\frac{1}{2}\sum_{s=0}^{t-1}Z^{z^{s}}\EV\frac{\partial Z^{z^{t}}}{\partial\hat{Z}^{x^{s+1}}}
\end{align*}
It's easy to see that there are deterministic coefficients $b_{s}^{t} \in \R$ (independent of $n$) so that 
\begin{equation}
Z^{z^{t}}=\sum_{s=1}^{t}b_{s}^{t}(\hat{Z}^{x^{s}}+\hat{Z}^{y^{s}})+b_{0}^{t}Z^{z^{0}}.
\label{eqn:Zzt}
\end{equation}
Then
\[\inv n \tr A^t \asto \EV Z^{z^t} Z^{z^0} = b^t_0.\]
Note
\[
\frac{\partial Z^{z^{t}}}{\partial\hat{Z}^{x^{s}}}=\frac{\partial Z^{z^{t}}}{\partial\hat{Z}^{y^{s}}}=b_{s}^{t}.
\]
Consequently,
\[
\Zdot^{x^{t+1}}=\Zdot^{y^{t+1}}=\frac{1}{2}\sum_{s=0}^{t-1}Z^{z^{s}}b_{s+1}^{t}
\]
and
\begin{align*}
Z^{z^{t+1}} & =\hat{Z}^{x^{t+1}}+\hat{Z}^{y^{t+1}}+\Zdot^{x^{t+1}}+\Zdot^{y^{t+1}}
  =\hat{Z}^{x^{t+1}}+\hat{Z}^{y^{t+1}}+\sum_{s=0}^{t-1}Z^{z^{s}}b_{s+1}^{t}\\
 & =\hat{Z}^{x^{t+1}}+\hat{Z}^{y^{t+1}}+\sum_{s=0}^{t-1}b_{s+1}^{t}\left(\sum_{r=1}^{s}b_{r}^{s}(\hat{Z}^{x^{r}}+\hat{Z}^{y^{r}})+b_{0}^{s}Z^{z^{0}}\right)\\
 & =\hat{Z}^{x^{t+1}}+\hat{Z}^{y^{t+1}}+\sum_{r=1}^{t-1}(\hat{Z}^{x^{r}}+\hat{Z}^{y^{r}})\sum_{s=r}^{t-1}b_{r}^{s}b_{s+1}^{t}+\sum_{s=0}^{t-1}b_{0}^{s}b_{s+1}^{t}Z^{z^{0}}
\end{align*}
Matching coefficients with \cref{eqn:Zzt}, this implies
\begin{align*}
b_{t+1}^{t+1} & =1,\quad
b_{r}^{t+1} =\sum_{s=r}^{t-1}b_{r}^{s}b_{s+1}^{t},\forall r\le t-1.
\end{align*}
Using the Catalan identity \cref{eqn:Catalan}, we can check that the solution is
\[
b_{r}^{t}=\begin{cases}
C_{(t-r)/2} & \text{if \ensuremath{t-r} is even}\\
0 & \text{otherwise.}
\end{cases}
\]
Then
\begin{align*}
\inv n \tr A^{2k} & \asto b_{0}^{2k}=C_{k},\quad
\inv n \tr A^{2k+1} \asto b_{0}^{2k+1}=0
\end{align*}
as desired.

\subsection{Comparison with the Classical Proof}

The classical way of calculating the expected moments $\EV\tr A^{k}$ is to expand this trace as a sum of products $A_{\alpha_{1}\alpha_{2}}\cdots A_{\alpha_{k}\alpha_{1}}$ of entries of $A$, and then notice that all summands contribute vanishingly other than those that have each unique term $A_{\alpha_{i}\alpha_{i+1}}$ appear with power 2. Then the computation of $\EV\tr A^{k}$ to top order can be seen to boil down to a counting problem of \emph{non-crossing partitions}. Since the solution of such counting problem is exactly the Catalan numbers, we have the desired result.

In contrast to this classical proof, our proof by Tensor Programs is purely symbolic (i.e. does not require manually identifying leading and subleading terms and doing the combinatorics).
The role of the mathematician here has been mostly to express the moment computation as a \netsort{} program, and the rest follows from the Master Theorem (and can be done by a computer).
While perhaps after unwinding, this technique may be implicitly doing a sort of combinatorics similar to the classical proof, we believe this particular way of \emph{repackaging} is useful.
Indeed, when the matrix ensemble in question involves nonlinear dependencies, such as in a neural network Jacobian, our techique applies readily (see \cref{sec:jac}), while the classical proof is hard to transfer as we can no longer expand the matrix power in terms of the matrix entries because of the nonlinear dependencies\footnote{One can still naively expand through the nonlinearities by Taylor expansion, but 1) this requires the nonlinearities to be smooth, and 2) this asks for significant effort for the mathematician to count and bound the lower order terms. In contrast, the proof by Tensor Programs will be, for the most part, mechanical, following the Master Theorem, and allows nonsmooth nonlinearities.}.

\paragraph{Universality}
On the other hand, a current drawback of this Tensor Programs proof is the limitation of the Master Theorem to Gaussian matrices.
However, we expect universality will hold in our case, and the Master Theorem can be proven for general, iid matrices, for example through some version of the Lindeberg Replacement Trick.
We leave this to future work.

%% file: FreeIndependence.tex
\section{The Free Independence Principle of \netsort{} Programs}
\label{sec:FIPMain}

A powerful analogue of \emph{independence} (of scalar random variables) in random matrix theory is called \emph{Asymptotic Free Independence}, or \emph{Asymptotic Freeness}, defined below in \cref{defn:asfree}.
\begin{defn}\label{defn:asfree}
    Fix $k$.
Consider collections of random matrices $\mathcal{W}_{n}^{1},\ldots,\mathcal{W}_{n}^{k}\subseteq\R^{n\times n}$ for each $n\ge1$, of constant cardinalities (with $n$).
(For example, each $\mathcal{W}_{n}^{i}$ can be $\{W, W^{\trsp}\}$ for some weight matrix $W$ in a neural network).
We say $\mathcal{W}_{n}^{1},\ldots,\mathcal{W}_{n}^{k}$ are \emph{almost surely asymptotically free}%
\footnote{
    We can also speak of asymptotically free in expectation, in which case, we want the expectation of the trace to converge to 0.
    In most scenarios (where we have tail bounds on the matrix operator norms), this is a weaker notion than almost sure asymptotic freeness.
}, if
\[
\inv n \tr\left(\prod_{i=1}^{k}\left(P_{i}(\mathcal{W}_{n}^{j_{i}})-\tau_{i}\right)\right)\asto 0
\]
where $\tau_{i}=\inv n \tr(P_{i}(\mathcal{W}_{n}^{j_{i}}))$, $P_{i}$ is a (noncommutative) polynomial in $|\mathcal W^{j_i}_n|$ variables, and $j_{1},\ldots,j_{k}\in[k]$ are indices with no two adjacent $j_{i}$ equal,
with $\{P_i\}_i$, $\{j_i\}_i$ independent of $n$.%
\footnote{
    Note here $\prod$ is a non-commutative product, where in its expansion, $i$ goes from right to left.
}
\footnote{One can also formulate the asymptotic freeness of subalgebras of a non-commutative algebra, in which case our definition here is equivalent to the freeness of the subalgebras generated by the respective collections of random matrices.}
\end{defn}

Whereas the independence of \emph{scalar} random variables allows one to compute the expectation of their sum and product easily, the asymptotic freeness of random \emph{matrices} allows one to compute the asymptotic spectral distributions of their sum and (matrix) product easily.
Independence implies that two random variables are in ``general position'' and thus cannot conspire to fluctuate in the same direction. Likewise, asymptotic freeness of two random matrices, intuitively, implies that their respective eigenvectors and eigenvalues lie in general position to each other, so that their spectral distributions would combine in predictable ways were they to be summed or multiplied. For more background, see \citep{tao_topics_2012}.

\emph{A priori}, because the activations and preactivations of a neural network depend in a highly nonlinear and complex manner in the weight matrices (and biases), one can hardly suspect that activations can be ``independent'' from the weight matrices in some way. 
However, we will show, in a randomly initialized neural network of any architecture, the weight matrices are asymptotically free from (the diagonal matrices formed from) the preactivations of the network.
This will follow from the much more general \emph{Free Independence Principle} of \netsort{} programs, \cref{thm:FreeIndependencePrinciple}.
We give a proof in \cref{{sec:FIP}} that follows the same overall strategy as the proof of the Semicircle Law in \cref{sec:semicircle}.
\begin{thm}[Free Independence Principle, for Tensor Programs]
\label{thm:FreeIndependencePrinciple} Consider any \netsort{} program $\pi$ with polynomially bounded nonlinearities
and \cref{setup:netsort}. Then the random matrix collections $\{W,W^{\trsp}\}$ for every $W\in\mathcal{W}$, along with the collection of diagonal matrices $\mathcal D(\pi)$ (defined immediately below in \cref{defn:DMatrices}) are asymptotically free as $n\to\infty$.
\end{thm}
\begin{defn}\label{defn:DMatrices}
    Suppose $\mathcal X$ is a subset of ($\R^n$) vectors in a program.
    Let $\mathcal D(\mathcal X)$ denote the (infinite) collection of diagonal matrices formed from bounded, coordinatewise images of $\mathcal X$:
    \begin{equation}\mathcal D(\mathcal X) \defeq \{\Diag(\psi(x^1,\ldots, x^k)): k \ge 0;\ \ x^1, \ldots, x^k \in \mathcal X;\ \ \psi: \R^k \to \R \text{ bounded}\} \sbe \R^{n\times n}.
        \label{eqn:DMat}
    \end{equation}
    If $\pi$ is a program, we write $\mathcal D(\pi)$ to denote $\mathcal D(\{\text{all vectors in $\pi$}\})$.
\end{defn}

Note $\psi$ in \cref{eqn:DMat} is distinct from the nonlinearities in the program $\pi$.
For example, if $\pi$ expresses the forward pass of a ReLU MLP, then $\pi$ has unbounded nonlinearity (ReLU) but $\psi$ in \cref{eqn:DMat} can be the step function, which is bounded.
We have kept $\psi$ bounded in \cref{eqn:DMat} for technical reasons (similar to the appearance of \emph{bounded} continuous functions in the definition of convergence in distribution).
But we believe \cref{thm:FreeIndependencePrinciple} holds when $\psi$ is more generally polynomially bounded.
This generalization would follow if \cref{thm:NetsorTMasterTheorem} holds for almost sure convergence of conditional means; see \cref{conj:NetsorTMeanConvergence}.

\paragraph*{Intuition and Discussion}
One can perhaps accept \netsort{} programs as a formalization of ``reasonable ways'' to compute vectors (and their diagonal matrices) from a set of random matrices.
Then FIP says that a random Gaussian matrix is asymptotically free from any diagonal matrix that ``depends on it in a reasonable way.''
This formalizes the intuition that singular vectors of Gaussian matrices are in general position to singular vectors of diagonal matrices, so one may expect these matrices to be asymptotically free.

We shall see next that \cref{thm:FreeIndependencePrinciple} allows us to easily compute the asymptotic Jacobian singular value distribution of a randomly initialized neural network. 

\paragraph{Extension to \netsortplus{} programs with variable dimensions}
\cref{thm:FreeIndependencePrinciple} holds as stated for \netsort{} programs with variable dimensions.
It also holds for \netsortplus{} programs with variable dimensions if nonlinearities are parameter-controlled and rank stability (\cref{assm:asRankStab}) is satisfied.

%% file: NNjacobian.tex
\newcommand{\mump}{\mu_{\mathrm{mp}}}

\newcommand{\Spec}[1]{\mu_{#1}}

\section{Jacobian Singular Values of a Randomly Initialized Neural Network}
\label{sec:jac}

\paragraph{Notation}
We denote the empirical spectral distribution of a random matrix $W$ by $\Spec{W}$.%
\footnote{i.e.\ $\Spec{W} = \f 1 n \sum_{\alpha=1}^n \delta_{\lambda_\alpha}$, where $\delta_{\lambda_\alpha}$ is the Dirac Delta distribution centered on the $\alpha$th eigenvalue $\lambda_\alpha$.}
We write $\mu_W \boxtimes \mu_V$ to denote the free multiplicative convolution of $\mu_W$ and $\mu_V$.%
\footnote{
  If $W$ and $V$ are asymptotically free random matrices, then $\mu_W \boxtimes \mu_V$ converges to the asymptotic spectral distribution of $WV$.
  See \citet{speicher_free_2009} for more details on free probability.
}

\paragraph*{Review of semirigorous computation of Jacobian singular value distribution in prior works.}

Analyses in previous works \citep{pennington_resurrecting_2017} are mostly semirigorous and proceed, for example, as follows:

If $f(\xi)$ is an MLP as in \cref{eqn:MLP} with width $n$, then the Jacobian $J=\partial h^L/\partial {h^1} \in \R^{n\times n}$, on a fixed input $\xi$, can be written as%
\footnote{
    Note we study the Jacobian of the \emph{body} of the network, whose dimensions $n$ tend to infinity.
    The perhaps more natural input-output Jacobian has finite dimensions, so there's no asymptotic distribution to speak of.
}
\[
J=W^{L}D^{L-1}W^{L-1}D^{L-2}\cdots W^{2}D^{1},
\]
where $W^{l}$ are its weight matrices and $D^{l}=\Diag(\phi'(h^{l}))$ are the diagonal matrices with activation derivatives on the diagonals. Then the singular values of $J$ are the square roots of the spectrum of 
\[
J^{\trsp}J=D^{1}W^{2\trsp}\cdots D^{L-1}W^{L\trsp}W^{L}D^{L-1}\cdots W^{2}D^{1}.
\]
Now here's the non-rigorous part: 
With the random initialization $W_{\alpha\beta}^1\sim\Gaus(0,1/d)$ and $W_{\alpha\beta}^2, \ldots, W_{\alpha\beta}^L \sim\Gaus(0,1/n),$
prior works assume that the random matrix collections 
\begin{equation}
\text{$\{\ensuremath{W^{2}},\ensuremath{W^{2\trsp}}\},\ensuremath{\ldots},\{\ensuremath{W^{L}},\ensuremath{W^{L\trsp}}\},\{\ensuremath{D^{1}}\},\ensuremath{\ldots},\{\ensuremath{D^{L-1}}\}$ are asymptotically free.}\label{stmt:AsymptoticFreenessAssumption}
\end{equation}
Then, with this assumption, noting that the spectrum of $AB$ and $BA$ agree for any two matrices $A,B$ of appropriate sizes%
\footnote{This can be seen by applying Sylvester's Determinant Theorem $\det(zI-AB)=\det(zI-BA)$ on the characteristic polynomials of $AB$ and $BA$.}, we have
\begin{align*}
\Spec{J^{\trsp}J} & =\Spec{D^{1}W^{2\trsp}\cdots D^{L-1}W^{L\trsp}W^{L}D^{L-1}\cdots W^{2}D^{1}}
  =\Spec{(D^{1})^{2}W^{2\trsp}\cdots D^{L-1}W^{L\trsp}W^{L}D^{L-1}\cdots W^{2}}
\end{align*}
so that, by the freeness asumption of $D^{1}$ from the other matrices, we have
\[
\lim_{n\to\infty}\Spec{J^{\trsp}J}=\lim_{n\to\infty}\Spec{(D^{1})^{2}}\boxtimes\lim_{n\to\infty}\Spec{W^{2}\cdots D^{L-1}W^{L\trsp}W^{L}D^{L-1}\cdots W^{2}}
\]
where $\boxtimes$ denotes multiplicative free convolution. Repeating this logic yields
\[
\lim_{n\to\infty}\Spec{J^{\trsp}J}=\lim_{n\to\infty}\Spec{(D^{1})^{2}}\boxtimes\lim_{n\to\infty}\Spec{W^{2}W^{2\trsp}}\boxtimes\cdots\boxtimes\lim_{n\to\infty}\Spec{W^{L}W^{L\trsp}}.
\]
Since $\lim_{n\to\infty}\Spec{W^{i}W^{i\trsp}}$ is just the Marchenko-Pastur distribution (\cref{eqn:MPdefn}) and $\lim_{n\to\infty}\Spec{(D^{l})^{2}}$ is (at least, at the time, heuristically) distributed like $\phi'(Z^{h^{l}})$, the standard S-transform technique (see \citet{speicher_free_2009}) allows one to explicitly compute $\lim_{n\to\infty}\Spec{J^{\trsp}J}$.

\paragraph*{Our contribution}

Of course, by \cref{thm:FreeIndependencePrinciple} and the \netsort{} program \cref{eqn:MLP}, \cref{stmt:AsymptoticFreenessAssumption} is now completely rigorous.
In fact, \cref{thm:FreeIndependencePrinciple} implies a much more general result.
\begin{cor}[Free Independence Principle]
\label{cor:NNFreeIndependence}
In any randomly initialized neural network expressible in \netsort{} program $\pi$ with polynomially bounded \refNonlin{}, the weight matrices are asymptotically free from (the diagonal matrices formed from) bounded images of all preactivations:
the random weight matrix collections $\{W,W^{\trsp}\}$ for matrices $W \in \mathcal W$, along with $\mathcal D(\pi)$ (defined in \cref{defn:DMatrices}), are asymptotically free. Furthermore, for any mutually independent partition $\mathcal{X}_{1},\ldots,\mathcal{X}_{k}$ of $\{Z^{x}:x\in\pi\}$, the random diagonal matrix collections
$\mathcal D(\{x: Z^{x}\in\mathcal{X}_{1}\}),\ldots,\mathcal D(\{x:Z^{x}\in\mathcal{X}_{k}\})$
are asymptotically free.
\end{cor}
Since almost all neural networks can be written in \netsort{}, as shown in \citet{yangTP1,yangTP2}, \cref{cor:NNFreeIndependence} is a very universal result.
By this corollary, the entire computation above in fact (after easily checking $Z^{h^{1}},\ldots,Z^{h^{L-1}}$ are mutually independent) yields a proof of the following almost sure convergence:
\begin{thm}\label{thm:JacSVMLP}
    Consider an MLP $f(\xi)$ as in \cref{eqn:MLP} with $L$ hidden layers, width $n$, and nonlinearity $\phi$ with bounded weak derivative%
    \footnote{i.e.\ $\phi$ is almost everywhere differentiable and $\phi'$ is a function that agrees with this derivative almost everywhere. ReLU is a typical example of $\phi$, whose weak derivative is the step function and is bounded.}
     $\phi'$.
Then its Jacobian $J=\partial h^L/\partial {h^1} \in \R^{n\times n}$ on a fixed input $\xi \in \R^d$ has the $n\to\infty$ limit
\[
\Spec{J^{\trsp}J}\asto\mu_{\phi'(Z^{h^{1}})}\boxtimes\cdots\boxtimes\mu_{\phi'(Z^{h^{L-1}})}\boxtimes\mump^{\boxtimes(L-1)}
\]
where $\mump$ is the Marchenko-Pastur distribution with shape ratio 1 (see \cref{eqn:MPdefn}), $\mu_{\phi'(Z^{h^{l}})}$ is the distribution of the random variable $\phi'(Z^{h^{l}})$, and $\asto$ denotes almost sure convergence of random measures (\cref{defn:ASConvergenceESD}).

\end{thm}
Most nonlinearities $\phi$ in deep learning practice have bounded weak derivatives (such as tanh or ReLU), and thus is covered by \cref{thm:JacSVMLP}.
When $\phi$ is identity, \cref{thm:JacSVMLP} just recovers the Marchenko-Pastur Law.
Note here the dependence of $\mu_{J^\trsp J}$ on input $\xi$ comes purely through $\mu_{\phi'(Z^{h^1})}, \ldots, \mu_{\phi'(Z^{h^L-1})}$.

\paragraph{Generalizations}
\cref{thm:JacSVMLP} generalizes straightforwardly to the case when the MLP has non-unit width ratios, in which case the $\mump^{\boxtimes(L-1)}$ should be replaced by free multiplicative convolution of Marchenko-Pastur distributions of different shape ratios (see \cref{eqn:MPdefn}).
Also, like the remark below \cref{thm:FreeIndependencePrinciple}, we expect \cref{thm:JacSVMLP} holds if $\phi'$ is just polynomially bounded and \cref{cor:NNFreeIndependence} holds if $\mathcal D$ is defined using polynomially bounded (instead of bounded) $\psi$. We leave this to future work.

\paragraph{Computing the Jacobian Singular Value Distribution of Any Neural Architecture}
More generally, \cref{cor:NNFreeIndependence} allows us to compute the Jacobian singular value of neural networks of any architecture (such as residual networks, recurrent networks, convolutional networks, etc).

Indeed, the Jacobian can always be expressed as a polynomial in the matrices and (the diagonal matrices formed from) vectors of an appropriate \netsort{} program.
By \cref{cor:NNFreeIndependence}, these random matrix collections are asymptotically free.
The asymptotic spectral distribution of such a polynomial in asymptotically free matrices can be computed via operator-valued free probability \citep[Chapter 10]{mingo_free_2017}.
Thus our result here yields a general algorithm for computing the asymptotic singular value distribution of the Jacobian of a randomly initialized neural network of any architecture.
Alternatively, one can always resort to explicit moment computations via Tensor Programs, as in \cref{sec:semicircle}.

\paragraph{Simple GIA Check Implies GIA}
It turns out that the Neural Tangent Kernel depends on the computation of backpropagation only through quantities that can be expressed as 2nd moment of Jacobian singular values, if Simple GIA Check (\cref{assm:simpleGIACheck}) holds.
By the results of this section, we thus can replace the matrices $W^\trsp$ in backpropagation with copies that are independent from $W$ the forward pass.
See \cref{sec:GIAFIP} for more details.

%% file: proofsketch.tex
\section{Proof Sketch of the Master Theorem}
\label{sec:proofsketch}

Here we explain the main ideas of the proof of \cref{thm:NetsorTMasterTheorem}.
The Master Theorems of \citet{yangTP1,yangTP2} all use some subset of these ideas for their proofs, so this section also summarizes the core insights there.
We start by proving an easier version of \cref{thm:NetsorTMasterTheorem} that assumes all nonlinearities in \refNonlin{} are sufficiently smooth (\cref{sec:First-Attempt}).
Then we show how to remove the smoothness assumption (\cref{sec:First-Order-Correction,{sec:rankStability}}).
Finally, we give an outline in \cref{sec:Zdot} for proving the form of $\Zdot$ in \cref{defn:Z}.

\subsection{Master Theorem Proof Sketch with Sufficient Regularity Assumptions\label{sec:First-Attempt}}

A natural intuition for proving the Master Theorem is to perform induction on the number of vectors. Suppose further for simplicity that all matrices $W\in\mathcal{W},W\in\R^{n\times n}$ are sampled like $W_{\alpha\beta}\sim\Gaus(0,1/n)$ . Here we sketch the proof of the simpler statement that \cref{{eqn:NetsorTMasterTheorem}} converges to \emph{some} limit, when all nonlinearities $\phi$ used in \refNonlin{} are smooth enough. We will discuss the form of the limit itself in another section (\cref{sec:Zdot}). Under these assumptions, the core idea is similar to the proof of \citet{bayati_dynamics_2011} for Approximate Mesage Passing (but we will not require knowledge of this proof below).
\begin{defn}   
We say a vector is a \emph{G-var} if it is in $\mathcal{V}$ or it is introduced by \refMatMul{}.%
\footnote{
    The letter \emph{G} elicits the intuition that the vector is roughly \emph{Gaussian} plus a correction term.
    We will not use the terms \emph{A-var} and \emph{H-var} from \citet{yangScalingLimitsWide2019arXiv.org,yangTP1,yangTP2}, in favor of more intuitive terms \emph{matrix} and \emph{vector} in the program.
}
\end{defn}
Suppose the program has G-vars $g^{1},\ldots,g^{m}$ introduced in that order. Since all other vectors can be expressed as a \refNonlin{} image of them, it suffices to prove \cref{eqn:NetsorTMasterTheorem} for $g^{1},\ldots,g^{m}$, i.e. for any \emph{sufficiently smooth $\psi:\R^{m}\to\R$, }we have
\begin{equation}
\f 1n\sum_{\alpha=1}^{n}\psi(g_{\alpha}^{1},\ldots,g_{\alpha}^{m})\asto\EV\psi(Z^{g^{1}},\ldots,Z^{g^{m}}).\label{eq:MomentGvars}
\end{equation}

The base case is when $g^{1},\ldots,g^{m}\in\mathcal{V}$ are the initial vectors. Then \cref{eq:MomentGvars} converges by law of large numbers.

For the inductive case, suppose $g^{m+1}=Ah$, where $h=\phi(g^{1},\ldots,g^{m})$, for some \emph{sufficiently smooth} $\phi:\R^{m}\to\R$, and WLOG for $A \in \mathcal W$.
We want to show \cref{eq:MomentGvars} for $m\gets m+1$. The \emph{key idea is to condition} $A$ on $g^{1},\ldots,g^{m}$, and then try to reduce \cref{eq:MomentGvars} for $m+1$ to \cref{eq:MomentGvars} for $m$ through a law of large numbers on the remaining randomness in $g^{m+1}$.
This conditioning puts a linear constraint on $A$ in the form of
\[
\Xme=A\Yme,\quad\Ume=A^{\trsp}\Vme
\]
for matrices $\Xme,\Yme,\Ume,\Vme$ with previous vectors as columns.
For example, if the program is $\{g^2 = A g^1, g^3 = A^\trsp g^2\}$, then conditioning on $g^1, g^2, g^3$, we have $\Xme = g^2, \Yme = g^1, \Ume=g^3, \Vme=g^2$.
By standard formulas for Gaussian conditioning (see \cref{lemma:condTrick}), we can derive the following conditional distribution%
\footnote{see \cref{sec:probfacts} for definition of $\disteq$}

\[
A\disteq_{g^{1},\ldots,g^{m}}E+\Pi_{1}\tilde{A}\Pi_{2},\quad g^{m+1}\disteq_{g^{1},\ldots,g^{m}}\left(E+\Pi_{1}\tilde{A}\Pi_{2}\right)h,
\]
where $\tilde{A}$ is an iid copy of $A$, $E\in\R^{n\times n}$ is the ``conditional mean'', and $\Pi_{1},\Pi_{2}\in\R^{n\times n}$ are two orthogonal projection matrices into subspaces of dimension $n-O(1)$. Then we can see each coordinate $g_{\alpha}^{m+1}$ is conditionally distributed like $(Eh)_{\alpha}+\sigma\Pi_{1}\zeta$ where $\sigma^{2}=\|\Pi_{2}h\|^{2}/n$ and $\zeta\sim\Gaus(0,I)$. Now we make several approximations that can be made rigorous using the smoothness of $\psi$:
\begin{enumerate}
    \item Since $\Pi_{1}$ has small corank, we approximate $\Pi_{1}\approx I$.\footnote{this is roughly because $\Pi_1$ is multiplied to $\tilde A$, which generically sends a vanishing amount of its image to the subspace represented by $\Pi_1$.} \label{item:projI}
    \item It turns out $\sigma^{2}$ can be rewritten as a continuous function of quantities in the form of \cref{eq:MomentGvars}, so by induction hypothesis, $\sigma\asto\mathring{\sigma}$ for some deterministic limit $\mathring{\sigma}\ge0$ (see \cref{lemma:sigmaConverges}). We make the approximation $\sigma\approx\mathring{\sigma}$. \label{item:sigmaApprox}
    \item Similarly, it turns out $Eh=\sum_{i=1}^{r}a_{i}h^{i}$ where $h^{i}$ are some previous vectors (which are necessarily \refNonlin{} images of $g^{1},\ldots,g^{m}$), and each $a_{i}$ is a (continuous function of) average of \cref{eq:MomentGvars}'s form (see \cref{lemma:omegaExpansion}). They converge $a_{i}\asto\mathring{a}_{i}$ by induction hypothesis, so we approximate $a_{i}\approx\mathring{a}_{i}$.
    Noet that the specific forms of $a_i$ and $h^i$ here will dictate the form of $\Zdot^{g^{m+1}}$; see \cref{sec:Zdot}.
    \label{item:coefApprox}
\end{enumerate}
In summary, we have now approximated, for each $\alpha\in[n]$,
\begin{equation}
g_{\alpha}^{m+1}\overset{d}{\approx}_{g^{1},\ldots,g^{m}}\sum_{i=1}^{r}\mathring{a}_{i}h_{\alpha}^{i}+\mathring{\sigma}\zeta_{\alpha},\quad\zeta_{\alpha}\sim\Gaus(0,1).
\label{eqn:gexpansionSketch}
\end{equation}
Thus, as $g_{\alpha}^{m+1}$ is iid in $\alpha\in[n]$ with this approximation, by law of large numbers, we should expect \cref{eq:MomentGvars} for $m\gets m+1$ to concentrate around its conditional expectation for large $n$:
\begin{align*}
\left|\f 1n\sum_{\alpha=1}^{n}\psi(g_{\alpha}^{1},\ldots,g_{\alpha}^{m},g_{\alpha}^{m+1})-S\right|&\asto0,\\
\text{where}\quad &S\defeq\frac{1}{n}\sum_{\alpha=1}^{n}\EV_{z\sim\Gaus(0,1)}\psi\left(g_{\alpha}^{1},\ldots,g_{\alpha}^{m},\mathring{\sigma}z+\sum_{i=1}^{r}\mathring{a}_{i}h_{\alpha}^{i}\right).
\end{align*}

Now $S$ is in the form of \cref{eq:MomentGvars} so we can apply induction hypothesis. This finishes the proof sketch.

In this proof sketch, we have substituted many quantities for their limits (that are inductively proven to exist). This is only possibly because we assume all nonlinearities are sufficiently smooth. What if we don't have this assumption? (This is important, for example, for expressing the backpropagation of a ReLU neural network, since ReLU's (weak) derivative is not continuous).

\subsection{Getting Rid of Smoothness Assumption}
\label{sec:First-Order-Correction}

The key insight into the removal of smoothness assumption on \refNonlin{} is that, \emph{we get smoothness for free from averaging}.
Here's the main strategy: If we can show the conditional concentration
\begin{align*}
\left|\f 1n\sum_{\alpha=1}^{n}\psi(g_{\alpha}^{1},\ldots,g_{\alpha}^{m},g_{\alpha}^{m+1})-\bar{S}\right|&\asto0,\\
\text{where}\quad&\bar{S}\defeq\EV\left[\left.\f 1n\sum_{\alpha=1}^{n}\psi(g_{\alpha}^{1},\ldots,g_{\alpha}^{m},g_{\alpha}^{m+1})\right| g^{1},\ldots,g^{m}\right],\numberthis\label{eq:conditionalConcentration}
\end{align*}
then $\bar{S}$ can be expressed as an average of \emph{smooth }functions of $g_{\alpha}^{1},\ldots,g_{\alpha}^{m}$ and some scalars $a_{1},\ldots a_{r},\sigma$ (as in \cref{item:sigmaApprox,item:coefApprox} above) that converge to deterministic limits. 
The smoothness of these functions come from the Gaussian averaging inside the conditional expectation, even if $\psi$ itself is not smooth.
This works as long as $\sigma > 0$; we shall discuss this assumption more thoroughly in \cref{sec:rankStability}.
With this smoothness, we can again replace $a_{1},\ldots,a_{r},\sigma$ with their limits $\mathring a_{1},\ldots,\mathring a_{r},\mathring{\sigma}$, and then the proof is finished by the induction hypothesis, as in \cref{sec:First-Attempt}.

Thus, if we can prove \cref{eq:conditionalConcentration} without using smoothness of $\psi$, then we would be done. Revisiting the approximations we made in our first-attempt proof, we see that we need to remake our law of large number argument (\cref{item:projI}) without substituting $\Pi_{1}\approx I$. This substitution has been quite convenient, as it made $g_{\alpha}^{m+1}$ conditionally iid in $\alpha$, so that the usual law of large numbers can apply. Now, we have to manually wrestle with the correlations between $g_{\alpha}^{m+1}$ and $g_{\beta}^{m+1}$ for pairs $\alpha,\beta\in[n]$.

\subsubsection{Law of Large Numbers for Images of Weakly Correlated Gaussians}
We thus prove a new law of large numbers for this case. It says that the average of images of weakly correlated Gaussians will converge deterministically.
\begin{thm}[LLN for Images of Weakly Correlated Gaussians (Simplified)]
 \label{thm:LLNGaussianImage} Consider a triangular array $\{\zeta_{1}^{n},\ldots,\zeta_{n}^{n}\}_{n\ge1}$ of Gaussian variables, where each row is given by $\zeta^{n}\sim\Gaus(0,\Sigma^{n})$ and the covariance matrix $\Sigma^{n}$ satisfies $\sum_{\alpha\ne\beta}(\Sigma_{\alpha\beta}^{n})^{2}/(\Sigma_{\alpha\alpha}^n \Sigma_{\beta\beta}^n)=O(1)$. Consider any polynomially-bounded $\phi:\R\to\R$. Then the triangular array $\{\phi(\zeta_{1}^{n}),\ldots,\phi(\zeta_{n}^{n})\}_{n\ge1}$ satisfies a strong law of large numbers.\footnote{The full theorem (\cref{thm:StrongLLNGaussianImage}) allows each $\zeta_{\alpha}^{n}$ to have its own $\phi_{\alpha}:\R\to\R$ and this is in fact what's needed to finish the proof of \cref{thm:NetsorTMasterTheorem}. However, for conveying the main insights, we will be content with the simplified statement here.}
\end{thm}

This theorem will be applicable to $\Sigma^{n}=\Pi_{1}$ as $\Pi_{1}$ is a projection matrix with low corank and can therefore be seen to have small off-diagonal entries (see \cref{lemma:projectionCorrelation}).
This would then finish the proof of \cref{eq:conditionalConcentration} without assuming smoothness of \refNonlin{}.

Let us then sketch a proof of \cref{thm:LLNGaussianImage}. It is instructive to first show a \emph{weak} law of large numbers (\cref{thm:WeakLLNGaussianImage}) by bounding the variance of the fluctuation around the mean (\cref{thm:projectVariance}). Suppose, for simplicity, $\phi$ is even, so that $\EV\phi(\zeta_{\alpha}^{n})=0$. Then we need to bound $\EV\left(\frac{1}{n}\sum_{\alpha=1}^{n}\phi(\zeta_{\alpha}^{n})\right)^{2}$. Expanding the square, we have $n$ diagonal terms $\frac{1}{n^{2}}\EV\phi(\zeta_{\alpha}^{n})^{2},\alpha\in[n],$ and $n(n-1)$ cross terms $\frac{1}{n^{2}}\EV\phi(\zeta_{\alpha}^{n})\phi(\zeta_{\beta}^{n}),\alpha\ne\beta$. The former contributes $O(1/n)$. We shall show the latter is, too.

Let $\phi(x)=b_{1}H_{1}(x)+b_{2}H_{2}(x)+\cdots$ be the Hermite expansion of $\phi$, with $H_{i}$ denoting the $i$th Hermite polynomial. Note that there's no $b_{0}$ term because $\phi$ has mean 0. Then a neat identity \cref{fact:hermiteInnerproduct} says, for any jointly Gaussian $(z_{1},z_{2})$ with zero-mean, unit variance, and covariance $c$, we have
\begin{equation}
\EV\phi(z_{1})\phi(z_{2})=b_{1}^{2}c+b_{2}^{2}c^{2}+\cdots.\label{eq:hermiteGaussianInnerProduct}
\end{equation}
If we assume $\Sigma^{n}$ has unit diagonal, then by this identity, we have
\begin{align*}
\frac{1}{n^{2}}\sum_{\alpha\ne\beta}\EV\phi(\zeta_{\alpha}^{n})\phi(\zeta_{\beta}^{n})
&=\frac{1}{n^{2}}\sum_{i\ge1}b_{i}^{2}\sum_{\alpha\ne\beta}\left(\Sigma_{\alpha\beta}^{n}\right)^{i}
\overset{1}{\le}\frac{1}{n^{2}}\sum_{i\ge1}b_{i}^{2}n\sqrt{\sum_{\alpha\ne\beta}\left(\Sigma_{\alpha\beta}^{n}\right)^{2i}}\\
&\overset{2}{\le}\frac{1}{n}\left(\sum_{i\ge1}b_{i}^{2}\right)\sqrt{\sum_{\alpha\ne\beta}\left(\Sigma_{\alpha\beta}^{n}\right)^{2}}
\overset{3,4}{=} O(1/n).
\end{align*}
Here we used 1) power mean inequality, 2) $\Sigma_{\alpha\beta}^{n}\le1\implies(\Sigma_{\alpha\beta}^{n})^{2i}\le(\Sigma_{\alpha\beta}^{n})^{2}$ for $i\ge1$, 3) $\sum_{i\ge1}b_{i}^{2}=\EV_{z\sim\Gaus(0,1)}\phi(z)^2=O(1)$, and 4) $\sum_{\alpha\ne\beta}\left(\Sigma_{\alpha\beta}^{n}\right)^{2}=O(1)$ by assumption. This finishes the proof sketch of the weak version of \cref{thm:LLNGaussianImage} under the simplifying assumptions of even $\phi$ and $\Sigma^{n}$ having unit diagonal (which can be removed easily by complicating the proof a bit).

To prove the strong version of \cref{thm:LLNGaussianImage}, we need to bound the higher moments $\EV\left(\frac{1}{n}\sum_{\alpha=1}^{n}\phi(\zeta_{\alpha}^{n})\right)^{p},p\ge4$. This requires a similar analysis as the above, but much more technically involved. For example, \cref{eq:hermiteGaussianInnerProduct} generalizes to higher-order cross terms $\EV\phi(z_{1})\cdots\phi(z_{k})$ for jointly Gaussian $(z_{1},\ldots,z_{k})$, but the resulting expression \cref{thm:multivariateHermiteExpectation} is difficult to use directy. Instead, we need to divide into cases and bound it: either all pairs $(z_{i},z_{j})$ have uniformly weak correlations (\cref{lemma:smallRhoMomentBound}), or some pair has really large correlation comparatively (\cref{lemma:largeRhoMomentBound}). For full details, check \cref{thm:controlHighMoments}.

\subsubsection{Rank Stability}
\label{sec:rankStability}

We have glossed over two important points in the above sketches:
A) The scalars $\sigma$ and $a_i$ in \cref{eqn:gexpansionSketch} depend on the pseudo-inverse of some Gram matrix of vectors in the program.
Even though this Gram matrix will converge almost surely, its rank could drop suddenly in the limit, causing its pseudo-inverse to diverge, so that $\sigma$ and $a_i$ also diverge (see \cref{prop:pseudoinverseLambda}).
B) if $\sigma=0$ (the ``conditional standard deviation of $g^{m+1}$ given $g^1, \ldots, g^m$''), then the conditional expectation involves no ``averaging'' so the argument of ``getting smoothness for free'' does not work (see \cref{sec:inductiveCoreSet}). More precisely, there are several scenarios for $\sigma$ and its limit $\mathring \sigma$:
\begin{enumerate}
    \item $\mathring{\sigma}>0$ so, since $\sigma\asto\mathring{\sigma}$, $\sigma>0$ almost surely as well. 
    \item $\mathring{\sigma}=0$ and $\sigma=0$ a.s. for large $n$.
    \item $\mathring{\sigma}=0$ and $\sigma>0$ a.s., only converging to 0 at $n=\infty$.
\end{enumerate}
In the case of 1), the arguments of \cref{sec:First-Order-Correction} go through, but this most likely won't work in the cases of 2) and 3). Intuitively, case 2) can perhaps allow us to reduce $g^{m+1}$ to a deterministic function of $g^1, \ldots, g^m$, and
show $\f 1n\sum_{\alpha=1}^{n}\psi(g_{\alpha}^{1},\ldots,g_{\alpha}^{m},g_{\alpha}^{m+1})$ is a.s. equal to $\f 1n\sum_{\alpha=1}^{n}\bar{\psi}(g_{\alpha}^{1},\ldots,g_{\alpha}^{m})$ for an appropriate $\bar{\psi}$, so we can apply the induction hypothesis. But this argument cannot apply to case 3), because of the randomness in $g^{m+1}$ even after conditioning.

It turns out the intuition for case 2) is correct and case 3) doesn't happen! This will follow from the property of rank stability, which will simultaneously solve both problem A) and B):
\begin{thm}[Rank Stability]
\label{thm:rankstabilitymain} Let $y,y^{1},\ldots,y^{k}$ be any collection of vectors in a \netsort{} program. If $Z^{y}=b_{1}Z^{y^{1}}+\ldots+b_{k}Z^{y^{k}}$, then almost surely, for large enough $n$, $y=b_{1}y^{1}+\cdots+b_{k}y^{k}$.
\end{thm}

Indeed, if $\mathring{\sigma}=0$, then it turns out we can show $Z^{g^{m+1}}$ is a linear combination of $Z^{g^{1}},\ldots,Z^{g^{m}}$, so rank stability tells us $g^{m+1}$ is the \emph{same} linear combination of $g^{1},\ldots,g^{m}$ (almost surely, for large $n$), i.e. we are in case 2). Thereafter, we can straightforwardly reduce to induction hypothesis. On the other hand, this also shows case 3) can never occur. Now it just suffices to show \cref{thm:rankstabilitymain}, which can be done essentially by a simultaneous induction with the main induction hypothesis. See \cref{sec:proofMainTheorem} for more details.

\subsection{The Calculation of \texorpdfstring{$\Zdot$}{Zdot}}
\label{sec:Zdot}

As remarked in \cref{{rem:ZIntuition}}, the adjunction relation $\langle y,Wx\rangle=\langle W^{\trsp}y,x\rangle$ is importantly related to the existence and form of $\Zdot^{Wx}$. If we assume the Master Theorem, then this adjunction implies the identity $\EV Z^{y}Z^{Wx}=\EV Z^{W^{\trsp}y}Z^{x}$ for any $x,y$ in the program. In fact, a very similar identity exists as well for $\Zdot$:
\[
\EV Z^{y}\Zdot^{Wx}=\EV\hat{Z}^{W^{\trsp}y}Z^{x}.
\]
This is proved in \cref{lemma:DtrxAdjunction}.
With this identity, we can verify (\cref{lemma:hpartReExpression}) that, in the notation of \cref{eqn:gexpansionSketch} (using the precise form of $\mathring a_i$ omitted here),
\[
\sum_{i=1}^{r}\mathring{a}_{i}\Zdot^{h^{i}}=\Zdot^{g^{m+1}}.
\]
Additionally, straightforward computation (\cref{lemma:YgConditionalDistribution}) shows
\[
\sum_{i=1}^{r}\mathring{a}_{i}\hat{Z}^{h^{i}}+\mathring{\sigma}z\disteq_{g^1, \ldots, g^m}\hat{Z}^{g^{m+1}},z\sim\Gaus(0,1).
\]
Then we can rewrite \cref{eqn:gexpansionSketch} heuristically as
\begin{align*}
    g_{\alpha}^{m+1}
    &\overset{d}{\approx}_{g^{1},\ldots,g^{m}}
        \sum_{i=1}^{r}\mathring{a}_{i}h_{\alpha}^{i}+\mathring{\sigma}\zeta_{\alpha},\quad\zeta_{\alpha}\sim\Gaus(0,1).
        \\
    &\overset{d}{\approx}_{g^{1},\ldots,g^{m}}
        \sum_{i=1}^{r}\mathring{a}_{i}Z^{h^i}+\mathring{\sigma}\zeta_{\alpha}
    =
        \lp \sum_{i=1}^{r}\mathring{a}_{i}\Zdot^{h^i}\rp + \lp \sum_{i=1}^{r}\mathring{a}_{i}\Zhat^{h^i}+\mathring{\sigma}\zeta_{\alpha}\rp
    =
        \Zdot^{g^{m+1}} + \Zhat^{g^{m+1}},
\end{align*}
as desired.

%% file: extraTheorems.tex
\section{\netsort{} Master Theorem for Convergence in Mean and in Distribution}
\label{sec:OtherModesConvergence}

\subsection{Convergence in Mean}

\begin{thm}
\label{thm:NetsorTMeanConvergence}
For the same premise as in \cref{thm:NetsorTMasterTheorem}, if $\psi$ is bounded, then we have the convergence in mean
\begin{equation}
\f 1n\sum_{\alpha=1}^{n}\psi(h_{\alpha}^{1},\ldots,h_{\alpha}^{k})\meanto\EV\psi(Z^{h^{1}},\ldots,Z^{h^{k}}).
\label{eqn:meanconvergence}
\end{equation}
This in particular means the convergence of expectations
\begin{equation}
\f 1n\EV\sum_{\alpha=1}^{n}\psi(h_{\alpha}^{1},\ldots,h_{\alpha}^{k})=\EV\psi(h_{\beta}^{1},\ldots,h_{\beta}^{k})\to\EV\psi(Z^{h^{1}},\ldots,Z^{h^{k}}),\quad\text{for any \ensuremath{\beta\ge1}}.
\label{eqn:expectationsConverge}
\end{equation}
In fact, we have almost sure convergence and convergence in mean of \emph{conditional} expectations as well:
For each $n$, let $\Bb$ be a sub-$\sigma$-algebra of the $\sigma$-algebra induced by random sampling in \cref{setup:netsort}.
Then
\begin{equation}
\left|\f 1n \EV \left[\sum_{\alpha=1}^{n}\psi(h_{\alpha}^{1},\ldots,h_{\alpha}^{k}) \mid \Bb \right] - \EV\psi(Z^{h^{1}},\ldots,Z^{h^{k}}) \right|
\asto
0, \meanto 0.
\label{eqn:condmeanconvergence}
\end{equation}
\end{thm}

\begin{proof}
    \cref{{eqn:condmeanconvergence},eqn:meanconvergence} follow from \cref{thm:NetsorTMasterTheorem}, the boundedness of $\psi$, and (conditional) dominated convergence.
    The first equality of \cref{eqn:expectationsConverge} follows from the symmetry in $\alpha$, and the convergence of expectations follows from convergence in mean.
\end{proof}

In \cref{thm:NetsorTMeanConvergenceLinearlyBounded} below, we also show convergence in mean when $\psi$ is quadratically bounded and all nonlinearities are linearly bounded.
We say a function $\phi: \R^k \to \R$ is \emph{linearly bounded} (resp.\ \emph{quadratically bounded}) if for all $x_1, \ldots, x_k \in \R$, $\phi(x_1, \ldots, x_k) \le C(1 + |x_1| + \cdots + |x_k|)$ (resp.\ $C(1 + |x_1|^2 + \cdots + |x_k|^2)$) for some constant $C > 0$.
Note that such a linearly bounded $\phi$, applied coordinatewise to vectors $h^1, \ldots, h^k \in \R^n$, preserves $\ell_2$ norm:
\begin{align*}
    \f 1 n \|\phi(h^1, \ldots, h^k)\|^2 
    &\le \f {C^2}n \sum_{\alpha=1}^n (1 + |h^1_\alpha| + \cdots + |h^k_\alpha|)^2 
    \le \f {(k+1) C^2}n \sum_{\alpha=1}^n 1 + |h^1_\alpha|^2 + \cdots + |h^k_\alpha|^2\\
    &\le (k+1) C^2 \lp 1 + \f{\|h^1\|^2}{n} + \cdots + \f{\|h^k\|^2}{n}\rp.
    \numberthis\label{eqn:linearlyboundedPreservesL2}
\end{align*}

\begin{thm}
    \label{thm:NetsorTMeanConvergenceLinearlyBounded}
    \cref{thm:NetsorTMeanConvergence} also holds if $\psi$ is quadratically bounded and all nonlinearities $\phi$ used in \refNonlin{} are linearly bounded.
\end{thm}
\begin{proof}
    As in the proof of \cref{thm:NetsorTMeanConvergence}, it suffices to prove the convergence in mean.
    Below, we say ``random variable $R$ is bounded with high probability'' if there exist absolute constants $C, c>0$ such that, for all $r \ge C$, we have $R \le r$ with probability at least $1 - C\exp(-cn)$.

    Because each Gaussian matrix $W \in \mathcal W$ has bounded operator norm with high probability (\cref{fact:opnormTailBound}), each application of \refMatMul{} preserves $\ell_2$ norm with high probability.
    Likewise, each application of \refNonlin{} preserves $\ell_2$ norm by \cref{eqn:linearlyboundedPreservesL2}.
    Finally, by classic concentration of measure, each $v \in \mathcal V$ has bounded $\ell_2$ norm with high probability.
    Thus, by induction, all vectors in the program have bounded $\ell_2$ norm with high probability.
    Because $\psi$ is quadratically bounded, this implies
    \[
        Q \defeq \f 1n\sum_{\alpha=1}^{n}\psi(h_{\alpha}^{1},\ldots,h_{\alpha}^{k})\]
    is bounded with high probability.
    In particular, this quantity has a subexponential tail
    \[
        \Pr(Q > r) \le C e^{-crn}, \forall r > C \implies \EV [Q \ind(Q > C)] \le \f C {cn} e^{-cCn} \to 0.\]
    Thus we may apply standard truncation technique to $Q$, decomposing it as $Q = Q\ind(Q \le C) + Q\ind(Q >C)$.
    The latter converges in mean to 0, as shown above, while the former converges in mean by dominated convergence.
\end{proof}

\begin{fact}[Upper tail estimate for iid random matrix, Corollary 2.3.5 of \citet{tao_topics_2012}]\label{fact:opnormTailBound}
    For $W \in \R^{n \times n}$ with $W_{\alpha\beta}\sim\Gaus(0, 1/n)$, there exist absolute constants $C, c > 0$ such that
    \[\Pr(\|W\|_{op} > r) \le C e^{-crn}\]
    for all $r \ge C$.
\end{fact}

We believe
\begin{conj}\label{conj:NetsorTMeanConvergence}
\cref{thm:NetsorTMeanConvergence} holds for any polynomially bounded $\psi$.
\end{conj}
As in the proof of \cref{thm:NetsorTMeanConvergenceLinearlyBounded}, this amounts to proving a tail bound on the LHS of \cref{eqn:meanconvergence}.
Intuitively, because the source of the randomness is from Gaussian sampling and the nonlinearities are all polynomially bounded, this quantity should have a sub-Weibull tail beyond some constant upper bound.
But making this rigorous is subtle, and we leave this for future work.

\subsection{Convergence in Distribution of Coordinates}

\begin{thm}[Convergence in Distribution of Coordinates]
\label{thm:NetsorTConvergenceInDistribution}
Assume the same premise as in \cref{thm:NetsorTMasterTheorem}. 
For any $\alpha\ge1$, we have the convergence in distribution of
\[
(h_{\alpha}^{1},\ldots,h_{\alpha}^{k})\distto(Z^{h^{1}},\ldots,Z^{h^{k}}).
\]
\end{thm}

\begin{proof}
For any bounded continuous function $\psi$, $\EV\psi(h_{\alpha}^{1},\ldots,h_{\alpha}^{k})\to\EV\psi(Z^{h^{1}},\ldots,Z^{h^{k}})$ by \cref{thm:NetsorTMeanConvergence}.
\end{proof}
A slightly more involved version of this symmetry argument also shows that different coordinate slices are independent from one another in the large $n$ limit.

\begin{thm}\label{thm:NetsorTMultiConvergenceInDistribution}
Assume the same premise as in \cref{thm:NetsorTMasterTheorem}. 
For any $\alpha_{1},\ldots,\alpha_{r}\ge1$, all different for each other, we have the convergence in distribution of
\[
\begin{pmatrix}h_{\alpha_{1}}^{1} & \cdots & h_{\alpha_{1}}^{k}\\
\vdots & \ddots & \vdots\\
h_{\alpha_{r}}^{1} & \cdots & h_{\alpha_{r}}^{k}
\end{pmatrix}\distto\begin{pmatrix}Z_{1}^{h^{1}} & \cdots & Z_{1}^{h^{k}}\\
\vdots & \ddots & \vdots\\
Z_{r}^{h^{1}} & \cdots & Z_{r}^{h^{k}}
\end{pmatrix},
\]
where $(Z_{1}^{h^{1}},\ldots,Z_{1}^{h^{k}}),\ldots,(Z_{r}^{h^{1}},\ldots,Z_{r}^{h^{k}})$ are iid copies of $(Z^{h^{1}},\ldots,Z^{h^{k}})$.
\end{thm}

\begin{proof}
We proceed by induction on $r$. The base case of $r=1$ has already been proven in \cref{thm:NetsorTConvergenceInDistribution}. Now assume the inductive hypothesis is proven for $r-1$ and we seek to show it is also true for $r$. It suffices to prove, for all bounded continuous functions $f_{1},\ldots,f_{r}:\R^{k}\to\R$, \footnote{or we can just consider the families $f_{i}(x)=e^{it_{i}x}$ (or their real and imaginary parts), where $\{t_{i}\}_{i=1}^{r}$ varies over $\R^{r}$.}
\[
\EV\prod_{i=1}^{r}f_{i}(h_{\alpha_{i}}^{1},\ldots,h_{\alpha_{i}}^{k})\to\prod_{i=1}^{r}\EV f_{i}(Z^{h^{1}},\ldots,Z^{h^{k}}).
\]
By \cref{thm:NetsorTMasterTheorem}, we have, for all $i\in[r]$,
\[
\frac{1}{n}\sum_{\beta=1}^{n}f_{i}(h_{\beta}^{1},\ldots,h_{\beta}^{k})\asto\EV f_{i}(Z^{h^{1}},\ldots,Z^{h^{k}}).
\]
Thus, taking the product over all $i\in[r]$ and then taking expectation, we have
\[
\frac{1}{n^{r}}\sum_{\beta_{1},\ldots,\beta_{r}}\EV\prod_{i=1}^{r}f_{i}(h_{\beta_{i}}^{1},\ldots,h_{\beta_{i}}^{k})\to\prod_{i=1}^{r}\EV f_{i}(Z^{h^{1}},\ldots,Z^{h^{k}}),
\]
by dominated convergence and the boundedness of $f_{i}$. Now in the LHS, only $o(n^{r})$ of the summands have some pair from $\beta_{1},\ldots,\beta_{r}$ equal to each other. Each such $o(n^{r})$ summands, by induction hypothesis, converges to some finite quantity. Thus their total contribution to the LHS is vanishing with $n$. Hence, we have
\[
\EV_{\substack{\beta_{1},\ldots,\beta_{r}\\
\text{all distinct}
}
}\EV\prod_{i=1}^{r}f_{i}(h_{\beta_{i}}^{1},\ldots,h_{\beta_{i}}^{k})\asto\prod_{i=1}^{r}\EV f_{i}(Z^{h^{1}},\ldots,Z^{h^{k}}).
\]
Finally, we note that the inner expectation in the LHS here is symmetric in all such distinct $\beta_{1},\ldots,\beta_{r}$, so 
\[
\EV\prod_{i=1}^{r}f_{i}(h_{\beta_{i}}^{1},\ldots,h_{\beta_{i}}^{k})\asto\prod_{i=1}^{r}\EV f_{i}(Z^{h^{1}},\ldots,Z^{h^{k}})
\]
for any distinct $\beta_{1},\ldots,\beta_{r}$, and in particular, for $\beta_{1},\ldots,\beta_{r}=\alpha_{1},\ldots,\alpha_{r}$, as desired.
\end{proof}

\subsection{Extensions to \netsortplus{} and Programs with Variable Dimensions}

All of the theorems above hold as stated for programs with variable dimensions (\cref{appendix:VarDim}), if these programs are setup as in \cref{setup:netsortVarDim}.
Likewise, \cref{thm:NetsorTMeanConvergence,{thm:NetsorTConvergenceInDistribution},{thm:NetsorTMultiConvergenceInDistribution}} hold for \netsortplus{} programs (\cref{sec:netsortplus}) as well, if rank stability (\cref{assm:asRankStab}) is satisfied and all nonlinearities $\varphi^u(-;-)$ are parameter-controlled at $\mathring\bigtheta^u$ (see \cref{thm:PCNetsorT+MasterTheorem,thm:PCNetsorT+MasterTheoremVarDim}).
Similarly, 
\begin{thm}\label{thm:NetsorT+MeanConvergenceLinearlyBounded}
    \cref{thm:NetsorTMeanConvergenceLinearlyBounded} holds for \netsortplus{} programs if rank stability is satisfied and the following conditions on nonlinearities hold:
    \begin{itemize}
        \item for every vector $h$ with defining nonlinearity $\varphi^h(-;-)$ and limit parameters $\mathring\bigtheta^h$, $\varphi^h(\vec x; \mathring \bigtheta^h)$ is linearly bounded in $\vec x$, and for any $\bigtheta$,
        \begin{align}
            |\varphi^h(\vec x; \bigtheta) - \varphi^h(\vec x; \mathring \bigtheta^h)| \le f(\bigtheta - \mathring \bigtheta^h) \phi(\vec x)
            \label{eqn:linearParameterControl}
        \end{align}
        for some linearly bounded $\phi$ and some continuous function $f$, taking values in $\R$, with $f(0) = 0$.
        \item for every scalar $c$ with defining nonlinearity $\varphi^c(-;-)$ and limit parameters $\mathring\bigtheta^c$, $\varphi^c(\vec x; \mathring \bigtheta^c)$ is quadratically bounded in $\vec x$ and \cref{eqn:linearParameterControl} is satisfied for some quadratically bounded $\phi$ and some continuous function $f$, taking values in $\R$, with $f(0) = 0$.
    \end{itemize}
\end{thm}

%% file: Generalized_GP4A_NTK4A.tex
\section{Generalized Architectural Universality of Neural Network-Gaussian Process}
\label{sec:NNGP}

Wide, randomly initialized neural networks are distributed like Gaussian processes (GP) \citep{neal_bayesian_1995,lee_deep_2018,matthews_gaussian_2018_arxiv,novak_bayesian_2018,garriga-alonso_deep_2018,yangTP1}.
\citet{yangTP1} systematically generalized this result from toy neural networks to all modern neural networks.
This result was proved by 1) showing that the kernel of input embeddings converge almost surely to a deterministic kernel $K$, and 2) because the readout layer(s) are independently initialized from the rest of the network, the distribution of random neural network function converges to a GP with this kernel $K$. Part 1 assumed that no weight matrix is the transpose of another weight matrix, and with this assumption, it follows from the \netsor{} Master Theorem \citep{yangTP1}. However, with the \netsort{} Master Theorem, we can straightforwardly get rid of this assumption, and the logic above still holds. We thus conclude
\begin{thm}
Consider any neural network whose forward propagation is expressible in a \netsort{} program, and whose output layer(s) are independently sampled from other parameters and not used in the interior of the network. Then we have the convergence in distribution of the neural network function to a Gaussian Process, as the network widths tend to infinity.%
\footnote{Of course, to compute the GP kernel requires going through the \netsort{} Master Theorem (in particular, computing $\Zdot$ and $\hat{Z}$), and it would be incorrect to just assume all instances of $W^{\trsp}$ to be independent from $W$.
}
\end{thm}

We can also consider what happens when the output layer(s) are not independent from other parameters. For example, we can consider the case where the output can be expressed as the first coordinate $y_{1}$ of a vector $y=Wu\in\R^{n}$ for some embedding $u$ of the input. Here $W$ could have been used in the interior of the network. Then in general, $\Zdot^{y}$ is nonzero, and $y_{1}\distto Z^{y}$ converges to a non-Gaussian distribution, which can still nevertheless be calculated using \cref{defn:Z} and \cref{thm:NetsorTMasterTheorem}.

\section{Generalized Architectural Universality of Neural Tangent Kernel}
\label{sec:generalizedNTK4A}

\newcommand{\NTK}{\Theta}

\citet{jacot_neural_2018} showed that, in the limit of large width, a neural network undergoing training by gradient descent evolves like a linear model with a kernel, called the \emph{Neural Tangent Kernel (NTK)}.
\citet{yangTP2} showed how to calculate this the infinite-width limit of this NTK for any architecture, when a commonly satisfied condition, called \emph{Simple GIA Check}, is satisfied.
This condition allows us to assume $W^{\trsp}$ to be independent from $W$ in the computation of NTK (this heuristic is called the \emph{Gradient Independence Assumption}, or \emph{GIA}).
\begin{cond}[Simple GIA Check]\label{assm:simpleGIACheck}
    \label{text:simpleGIACheck}
    No weight matrix is the transpose of another weight matrix;
    the output layer is sampled independently and with zero mean from all other parameters and is not used anywhere else in the interior of the network%
    \footnote{i.e.\ if the output weight is $v$ and the output is $v^\trsp x$, then $x$ does not depend on $v$.}.
\end{cond}
Using \cref{thm:NetsorTMasterTheorem}, we may now generalize the result of \citet{yangTP2} to when Simple GIA Check is not satisfied (\cref{thm:NTK4A}). 
Before that, let's gather some concrete intuition for this generalization.

\paragraph{Example}
\citet{yangTP2} demonstrated a simple neural network not satisfying Simple GIA Check, and, if we assume $W^{\trsp}$ to be independent from $W$, then the resulting NTK computation would be wrong. Now, with the help of \netsort{} Master Theorem, we may finally perform the correct computation.

The neural network in question computes
\begin{equation}
x^{1}=W^{1}\xi+1,\quad h^{2}=W^{2}x^{1},\quad x^{2}=\phi(h^{2}),\quad y=1^{\trsp}x^{2}/n
\label{eqn:forwardGIABreak}
\end{equation}
with $\phi(z)=z^{2}$ being the square function, $\xi=0\in\R^{d},y\in\R,x^{1},h^{2},x^{2}\in\R^{n},W^{1}\in\R^{n\times d},W^{2}\in\R^{n\times n},W_{\alpha\beta}^{1}\sim\Gaus(0,1/d),W_{\alpha\beta}^{2}\sim\Gaus(0,1/n)$. If we set $dx^{2}=n\frac{\partial y}{\partial x^{2}}$, then backprop yields 
\begin{equation}
dx^{2}=1,\quad dh^{2}=2h^{2}\odot1=2h^{2},\quad dx^{1}=W^{2\trsp}dh^{2}=2W^{2\trsp}h^{2}=2W^{2\trsp}W^{2}x^{1}.
\label{eqn:backwardGIABreak}
\end{equation}
\citet{yangTP2} showed that $\EV dx_{\alpha}^{1}=2\EV x_{\alpha}^{1}=1$ but if we were to assume $W^{2\trsp}$ be independent from $W^{2}$, then we would get the erroneous answer $\EV dx_{\alpha}^{1}=0$. Here, let us compute $Z^\bullet$ for each vector above using \cref{defn:Z}, and see we get a consistent result with $\EV Z^{dx_{\alpha}^{1}}=2$.

In the forward pass, we have $Z^{h^{2}}=\Gaus(0,1)$ and likewise in the backward pass, $Z^{dh^{2}}=2Z^{h^{2}}=\Gaus(0,4)$. Next, we need to compute $\hat{Z}^{dx^{1}}$ and $\Zdot^{dx^{1}}$. Just like in the Master Theorem of \citet{yangTP2}, $\hat{Z}^{dx^{1}}$ is the random variable that we would get if we assume $W^{\trsp}$ be independent from $W$: $\hat{Z}^{dx^{1}}=\Gaus(0,\EV(Z^{dh^{2}})^{2})=\Gaus(0,4)$ and is independent from $Z^{dh^{2}}$ and $Z^{h^{2}}$. On the other hand, $\Zdot^{dx^{1}}$ is a scalar multiple $\Zdot^{dx^{1}}=x^{1}\EV\frac{\partial Z^{dh^{2}}}{\partial h^{2}}$ of $x^{1}.$ The multiple is $\EV\frac{\partial Z^{dh^{2}}}{\partial h^{2}}=\EV2=2$. Putting them all together, we get
\[
Z^{dx^{1}}=\hat{Z}^{dx^{1}}+\Zdot^{dx^{1}}=\Gaus(0,4)+2x^{1}
\]
which has mean $\EV Z^{dx^{1}}=2\EV Z^{x^{1}}=2$.

More generally, by \cref{thm:NetsorTMasterTheorem}, we trivially have the following result:
\begin{thm}\label{thm:NTK4A}
Consider any neural network whose forward and backpropagation are expressible in a \netsort{} program%
\footnote{as shown in \citet{yangTP2}, this includes practically all architectures used in modern deep learning}, and which does \emph{not} need to satisfy Simple GIA Check.
Suppose the network is parametrized in the \emph{NTK parametrization}.
If all of the network's nonlinearities have polynomially bounded weak derivatives, then its NTK, on standard Gaussian initialization of the network parameters, converges to a deterministic kernel as its widths tend to infinity, over any finite set of inputs.

\end{thm}
As with the NNGP in \cref{sec:NNGP}, this infinite-width NTK can be computed in a straightforward way using \cref{thm:NetsorTMasterTheorem}, following the examples of \citet{yangTP2}.

\section{Simple GIA Check Implies Gradient Independence Assumption}
\label{sec:GIAFIP}

We give a new proof of the following in this section through FIP.
\begin{thm}\label{thm:GIA}
    If a neural network expressible%
    \footnote{i.e. its forward and backward propagations at initialization can be written in a \netsort{} program. As shown in \citet{yangTP1,yangTP2}, this covers almost all classic and modern neural networks.}
    in a \netsort{} program satisfies Simple GIA Check (\cref{assm:simpleGIACheck}), then its NTK can be computed assuming that $W^\trsp$ used in backpropagation is independent from $W$ used in forward propagation.
\end{thm}

\citet{yangTP2} showed that, for any pair of inputs $\xi, \bar \xi \in \R^d$, the Neural Tangent Kernel of a NN $f$ depends on backpropagation only through the quantities $\langle {\nabla_{ y(\xi)} f(\xi)}, {\nabla_{ y(\bar \xi)} f(\bar \xi)} \rangle$.
If the network satisfied Simple GIA Check (\cref{assm:simpleGIACheck}), i.e.\ the network output is computed like $f(\xi) = n^{-1/2} v^\trsp e(\xi)$ for some embedding $e(\xi)$ of $\xi$ and $v_\alpha \sim\Gaus(0, 1)$ independent of $e$, then we can rewrite ${\nabla_{ y(\xi)} f(\xi)} = n^{-1/2} v^\trsp J(\xi)$ where $J(\xi) = \partial e(\xi)/\partial y(\xi)$.
Therefore,
\begin{align*}
  \langle {\nabla_{ y(\xi)} f(\xi)}, {\nabla_{ y(\bar \xi)} f(\bar \xi)} \rangle
  = \inv n v^\trsp J(\xi) J(\bar\xi)^\trsp v.
\end{align*}
By reversing the trace trick (\cref{{eqn:zdef}}), this quantity has the same limit as $\inv n \tr J(\xi) J(\bar\xi)^\trsp$.
If $\pi$ is the program expressing the forward and backward propagations of $f(\xi)$ and $f(\bar \xi)$, then this is a moment in $\mathcal D(\pi)$ and $\{W, W^\trsp\}, W \in \mathcal W$.
By FIP (\cref{thm:FreeIndependencePrinciple}), this moment stays the same if all $W \in \mathcal W$ are assumed independent from $\mathcal D(\pi)$, i.e.\ if transposed weight matrices $W^\trsp$ in the backward pass are independent from $W$ used in the forward pass (which produced the diagonals of $\mathcal D(\pi)$).

%% file: advancedNetsorT.tex
\section{\netsortplus{}: Adding Scalars to \netsort{}}
\label{sec:netsortplus}

In this section, we discuss the extension of \netsort{} with scalars that can be computed from (essentially) averaging some vector in the program.
This is not used substantially until \cref{sec:subprogramIndependence}, so the reader should feel free to skip ahead and come back only as needed.
\begin{defn}\label{defn:netsortplus}
A \netsortplus{} program\footnote{What we refer to as \netsortplus{} program is the same as ``simplified \netsortplus{}'' in \citet{yangTP2}} is just a sequence of $\R^{n}$ vectors and scalars in $\R$ inductively generated via one of the following ways from an initial set $\mathcal{C}$ of random scalars, an initial set $\mathcal{V}$ of random $\R^{n}$ vectors, and a set $\mathcal{W}$ of random $\R^{n\times n}$ matrices

\begin{description}
\item [\texttt{Nonlin$^+$\label{instr:nonlin+}}] Given $\phi:\R^{k}\times\R^{l}\to\R$, previous scalars $\theta_{1},\ldots,\theta_{l}\in\R$ and vectors $x^{1},\ldots,x^{k}\in\R^{n}$, we can generate a new vector 
\[
\phi(x^{1},\ldots,x^{k};\theta_{1},\ldots,\theta_{l})\in\R^{n}
\]
where $\phi(-;\theta_{1},\ldots,\theta_{l})$ applies coordinatewise to each ``$\alpha$-slice'' $(x_{\alpha}^{1},\ldots,x_{\alpha}^{k})$. 
\item [\texttt{Moment\label{instr:moment}}] Given same setup as above, we can also generate a new scalar 
\[
\f 1n\sum_{\alpha=1}^{n}\phi(x_{\alpha}^{1},\ldots,x_{\alpha}^{k};\theta_{1},\ldots,\theta_{l})\in\R.
\]
\item [\texttt{MatMul\label{instr:matmul+}}] Given $W\in\R^{n\times n}$ and $x\in\R^{n}$, we can generate $Wx\in\R^{n}$ or $W^{\trsp}x\in\R^{n}$ 
\end{description}
We say a vector is a \emph{G-var} if it is in $\mathcal{V}$ or it is introduced by \refMatMulPlus{}.
\end{defn}
We will typically discuss \netsortplus{} programs when $\mathcal{C}$, $\mathcal{V}$, and $\mathcal{W}$ are sampled as follows.
Note $\mathcal{V}$ and $\mathcal{W}$ are sampled the same way here as in \cref{setup:netsort}.

\begin{setup}[\netsortplus{}]\label{setup:netsortplus}
    1) each random scalar $c$ in $\mathcal{C}$ converges to a deterministic limit $\mathring{c}\in\R$ as $n\to\infty$; 2) for each initial $W\in\mathcal{W}$, $W_{\alpha\beta}\sim\Gaus(0,\sigma_{W}^{2}/n)$ for an associated variance $\sigma_{W}^{2}$; 3) there is a multivariate Gaussian $Z^{\mathcal{V}}=\left\{ Z^{g}:g\in\mathcal{V}\right\} \in\R^{|\mathcal{V}|}$ such that the initial set of vectors $\mathcal{V}$ are sampled like $\left\{ g_{\alpha}:g\in\mathcal{V}\right\} \sim Z^{\mathcal{V}}$ iid for each $\alpha\in[n]$.
\end{setup}

\begin{defn}[Key Intuition for \netsortplus{}]
    \label{defn:netsortplusKeyIntuit}
 Just like in the \netsort{} case, each vector $h$ in the program has roughly iid coordinates when $n\gg1$, each of which is distributed like a random variable $Z^{h}$. In addition, each scalar $\theta$ in the program will converge to a deterministic limit $\mathring{\theta}$. We'll recursively define $Z^{h}$ and $\mathring{\theta}$ as follows.
\begin{description}
\item [\texttt{ZInit}]
If $h\in\mathcal{V}$, then $Z^{h}$ is defined as the distribution of each coordinate of $h$ given in \cref{setup:netsortplus}. We also set $\hat{Z}^{h}\defeq Z^{h}$ and $\Zdot^{h}\defeq 0$.
  Likewise, if $\theta\in\mathcal{C}$, then $\mathring{\theta}$ is defined as the limit of $\theta$ specified in \cref{setup:netsortplus}.
\item [\texttt{ZMatMul}] Same as in \refZMatMul{} in \netsort{}
\item [\texttt{ZNonlin$^+$}] Given $\phi:\R^{k}\times\R^{l}\to\R$, previous scalars $\theta_{1},\ldots,\theta_{l}\in\R$ and vectors $x^{1},\ldots,x^{k}\in\R^{n}$, we have
\[
Z^{\phi(x^{1},\ldots,x^{k};\theta_{1},\ldots,\theta_{l})}\defeq\phi(Z^{x^{1}},\ldots,Z^{x^{k}};\mathring{\theta}_{1},\ldots,\mathring{\theta}_{l}).
\]
\item [\texttt{ZMoment}] Given same setup as above and scalar $\theta = \f 1n\sum_{\alpha=1}^{n}\phi(x_{\alpha}^{1},\ldots,x_{\alpha}^{k};\theta_{1},\ldots,\theta_{l})$, then
\[
\mathring{\theta}\defeq\EV\phi(Z^{x^{1}},\ldots,Z^{x^{k}};\mathring{\theta}_{1},\ldots,\mathring{\theta}_{l}).
\]
Here $\mathring{\theta}_{1},\ldots,\mathring{\theta}_{l}$ are deterministic, so the expectation is taken over $Z^{x^{1}},\ldots,Z^{x^{k}}$.
\end{description}
\end{defn}

\subsection{\netsortplus{} Master Theorem for Pseudo-Lipschitz Nonlinearities}
\paragraph{Pseudo-Lipschitz functions}

are, roughly speaking, functions whose weak derivatives are polynomially bounded.
\begin{defn}\label{defn:pseudoLipschitz}
    A function $f: \R^k \to \R$ is called \emph{pseudo-Lipschitz} of degree $d$ if $|f(x) - f(y)| \le C\|x-y\|(1 + \sum_{i=1}^k |x_i|^d + |y_i|^d)$ for some $C$.
\end{defn}
Here are some basic properties of pseudo-Lipschitz functions:
\begin{itemize}
    \item The norm $\|\cdot\|$ in \cref{defn:pseudoLipschitz} can be any norm equivalent to the $\ell_2$ norm, e.g. $\ell_p, p\ge1,$ norms. Similarly, $\sum_{i=1}^k |x_i|^d + |y_i|^d$ can be replaced by $\|x\|^d_p + \|y\|^d_p$, for any $p \ge 1$.
    \item A pseudo-Lipschitz function is polynomially bounded.
    \item A composition of pseudo-Lipschitz functions of degrees $d_1$ and $d_2$ is pseudo-Lipschitz of degree $d_1 + d_2$.
    \item A pseudo-Lipschitz function is Lipschitz on any compact set.
\end{itemize}

\paragraph{Master Theorem}
We state the Master Theorem below assuming a generic regularity condition called \emph{rank stability} (\cref{assm:asRankStab}), which we shall describe shortly.
The proof will follow from a more general, but more wordy Master Theorem in the next section (\cref{thm:PCNetsorT+MasterTheorem}).
\begin{thm}[Pseudo-Lipschitz \netsortplus{} Master Theorem]
\label{thm:PLNetsorT+MasterTheorem} Fix a \netsortplus{} program. Suppose the initial matrices $\mathcal{W}$, vectors $\mathcal{V}$, and scalars $\mathcal{C}$ are sampled in the fashion of \cref{setup:netsortplus}.
Suppose the program satisfies the \emph{rank stability assumption} below (\cref{assm:asRankStab}).
Assume all $\phi$ used in \refNonlinPlus{} are pseudo-Lipschitz (or, we can assume the slightly weaker \cref{assm:MasterTheoremSmoothness}).
Then 
\begin{itemize}
\item For any fixed $k$ and any polynomially-bounded $\psi:\R^{k}\to\R$, as $n\to\infty$,
\[
\f 1n\sum_{\alpha=1}^{n}\psi(h_{\alpha}^{1},\ldots,h_{\alpha}^{k})\asto\EV\psi(Z^{h^{1}},\ldots,Z^{h^{k}}),
\]
for any vectors $h^{1},\ldots,h^{k}$ in the program, where $Z^{h^{i}}$ are as defined in \cref{{defn:netsortplusKeyIntuit}}.
\item Any scalar $\theta$ in the program tends to $\mathring{\theta}$ almost surely, where $\mathring{\theta}$ is as defined in \cref{{defn:netsortplusKeyIntuit}}.
\end{itemize}
\end{thm}

\begin{assm}
    \label{assm:MasterTheoremSmoothness}
    Suppose
    \begin{enumerate}
        \item If a function $\phi(;-): \R^{0 + l} \to \R$ with only parameter arguments is used in \refMoment{}, then $\phi$ is continuous in those arguments.
        \item Any other function $\phi(-;-): \R^{k + l} \to \R$ with parameters (where $k>0$) used in \refNonlinPlus{} or \refMoment{} is pseudo-Lipschitz in all of its arguments (both inputs and parameters).
    \end{enumerate}
\end{assm}
Statement 1 in \cref{assm:MasterTheoremSmoothness} essentially says that if we have scalars $\theta_1, \ldots, \theta_l$ in the program, then we can produce a new scalar by applying a continuous function (a weaker restriction than a pseudo-Lipschitz function) to them.
Indeed, if $\theta_1, \ldots, \theta_l$ converge almost surely, then this new scalar does too.

\paragraph*{Rank Stability}

The following assumption says that the vectors in a program should not change any linear dependence relations abruptly in the infinite $n$ limit.

\begin{assm}[Rank Stability]\label{assm:asRankStab}
    Fix a \netsortplus{} program that is setup by \cref{setup:netsortplus}.
    We say this program satisfies \emph{rank stability} if for any matrix $W\in\R^{n\times n}$ in the program%
    \footnote{i.e.\ either $W\in\mathcal{W}$ or $W^{\trsp}\in\mathcal{W}$}
    and any collection $\mathcal{H}$ of vectors $h$ such that $Wh$ appears in the program, we have $\rank \mathcal{H} = \rank \{Z^{h}:h\in\mathcal{H}\}$, almost surely, for sufficiently large $n$.%
    \footnote{In the case of variable dimension \netsortplus{} programs, this is the same except $W\in \R^{n\times m}$ can have unequal dimensions and the limit is taken in the manner of \cref{setup:netsortVarDim}.}
\end{assm}

Most commonly, $\{Z^{h}:h\in\mathcal{H}\}$ will be linearly independent for all such $\mathcal H$, and therefore, by the lower semicontinuity of rank, \cref{assm:asRankStab} is automatically satisfied.
We shall discuss the necessity of \cref{assm:asRankStab} below (\cref{{remk:necessityRankStab}}).

\subsection{\netsortplus{} Master Theorem for Parameter-Controlled Nonlinearities}

\paragraph{Parameter-Control}
\cref{thm:PLNetsorT+MasterTheorem} will follow from the more general Master Theorem we state in this section, which allows for more general nonlinearities which only needs to be \emph{mildly} smooth in the scalar parameters $\theta_1,\ldots, \theta_l$, but not necessarily in $x^1_\alpha, \ldots, x^k_\alpha$:

\begin{defn}[Parameter-Control]\label{defn:parameterControlled} 
    We say a parametrized function $\phi(-;-):\R^{k}\times\R^{l}\to\R$ is \emph{polynomially parameter-controlled} or just \emph{parameter-controlled} for short\footnote{This overloads the meaning of \emph{parameter-controlled} from \citet{yangTP1}, where the definition replaces the ``polynomially bounded'' in the definition here with ``bounded by $e^{C\|\cdot\|^{2-\epsilon}+c}$ for some $C,c,\epsilon>0$.'' In this paper, we shall never be concerned with the latter (more generous) notion of boundedness, so there should be no risk of confusion.} , at $\mathring{\bigtheta}\in\R^{l}$ if 
\begin{enumerate}
\item $\phi(-;\mathring{\bigtheta})$ is polynomially bounded, and\label{item:parameterControlled1} 
\item there are some polynomially bounded $\bar{\phi}:\R^{k}\to\R$ and some function $f:\R^{l}\to\R^{\ge0}\cup\{\infty\}$ that has $f(\mathring{\bigtheta})=0$ and that is continuous at $\mathring{\bigtheta}$, such that, for all $x^{1},\ldots,x^{k}\in\R$ and $\bigtheta\in\R^{l}$, 
\[
|\phi(x^{1},\ldots,x^{k};\bigtheta)-\phi(x^{1},\ldots,x^{k};\mathring{\bigtheta})|\le f(\bigtheta)\bar{\phi}(x^{1},\ldots,x^{k}).
\]
\label{item:parameterControlled2} 
\end{enumerate}
 \end{defn} 

Note that $f$ and $\bar{\phi}$ here can depend on $\mathring{\bigtheta}$. The following examples come from \citet{yangTP1}. \begin{exmp}\label{exmp:parameterControl} 
    Any function that is (pseudo-)Lipschitz in $x^{1},\ldots,x^{k}$ and $\bigtheta$ is polynomially parameter-controlled.
    An example of a discontinuous function that is polynomially parameter-controlled is $\phi(x;\theta)=\mathrm{step}(\theta x)$: For $\mathring{\theta}\ne0$, we have
\[
|\phi(x;\theta)-\phi(x;\mathring{\theta})|\le\f{|\mathring{\theta}-\theta|}{|\mathring{\theta}|},
\]
so we can set $f(\theta)=\f{|\mathring{\theta}-\theta|}{|\mathring{\theta}|}$ and $\bar{\phi}=1$ in \cref{defn:parameterControlled}. \end{exmp}

Next, note that we can always express a vector in a \netsortplus{} program as a nonlinear image of previous G-vars.
For example, if $z = \phi(x^1, x^2; \theta_1), x^1 = W v, x^2 = \psi(y; \theta_2), y = W u$, then $z$ can be directly expressed in terms of G-vars: $z = \hat \phi(W v, W u; \theta_1, \theta_2) \defeq \phi(Wv, \psi(W u; \theta_2); \theta_1)$.
Therefore, we can make the following definition
\begin{defn}[$\varphi^{\bullet},\bigtheta^{\bullet}$ Notation] For any $\R^n$ vector $x$ in a \netsortplus{} program (\cref{defn:netsortplus}), let $\varphi^{x}$ and $\bigtheta^{x}$ be the parametrized nonlinearity and the scalars such that $x=\varphi^{x}(z^{1},\ldots,z^{k};\bigtheta^{x})$ for some G-vars $z^{1},\ldots,z^{k}$. Likewise, for any scalar $c$ in a \netsortplus{} program, let $\varphi^{c}$ and $\bigtheta^{c}$ be the parametrized nonlinearity and the scalars such that $c=\f 1n\sum_{\alpha=1}^{n}\varphi^{x}(z_{\alpha}^{1},\ldots,z_{\alpha}^{k};\bigtheta^{c})$ for some G-vars $z^{1},\ldots,z^{k}$. \end{defn}

\begin{thm}[Parameter-Controlled \netsortplus{} Master Theorem]
    \label{thm:PCNetsorT+MasterTheorem} Fix a \netsort{} program. Suppose the initial matrices $\mathcal{W}$, vectors $\mathcal{V}$, and scalars $\mathcal{C}$ are sampled in the fashion of \cref{setup:netsortplus}.
    Suppose the program satisfies the \emph{rank stability assumption} (\cref{assm:asRankStab}).
    Assume $\varphi^u(-;-)$ is parameter-controlled at $\mathring \bigtheta^u$ for all vectors and scalars $u$.
    Then 
    \begin{itemize}
        \item 
        For any random vector $\bigtheta \in \R^l$ that converges almost surely to a deterministic vector $\mathring{\bigtheta}$ as $n \to \infty$, and for any $\psi(-; -): \R^k \times \R^l \to \R$ parameter-controlled at $\mathring{\bigtheta}$,
        \begin{align*}
            \f 1 n \sum_{\alpha=1}^n \psi(g^1_\alpha, \ldots, g^k_\alpha; \bigtheta) \asto \EV\psi(Z^{g^1}, \ldots, Z^{g^k}; \mathring{\bigtheta}).
        \end{align*}
        for any G-vars $g^1, \ldots, g^k$, 
        where $Z^{g^{i}}$ are as defined in \cref{{defn:netsortplusKeyIntuit}}.
        \item Any scalar $\theta$ in the program tends to $\mathring{\theta}$ almost surely, where $\mathring{\theta}$ is as defined in \cref{{defn:netsortplusKeyIntuit}}.
    \end{itemize}
\end{thm}

Since pseudo-Lipschitz parametrized functions and parameterless polynomially bounded functions are both parameter-controlled, \cref{thm:PCNetsorT+MasterTheorem} implies \cref{thm:PLNetsorT+MasterTheorem} trivially.
Proof of \cref{thm:PCNetsorT+MasterTheorem} can be found in \cref{sec:netsortplusMasterTheoremProof}

\paragraph{Necessity of the Master Theorems' Assumptions}

The following remarks from \citep{yangTP1} show the necessity of parameter control and of rank stability in \cref{thm:PCNetsorT+MasterTheorem}.
\begin{rem}
[Necessity of parameter-control] Suppose $\psi(x;\theta)=\ind(\theta x\ne0)$. For $\theta\ne0$, $\psi$ is 1 everywhere except $\psi(0;\theta)=0$. For $\theta=0$, $\psi$ is identically 0. Thus it's easily seen that $\psi$ is not parameter-controlled at $\theta=0$.

Now, if $g\in\R^{n}$ is sampled like $g_{\alpha}\sim\Gaus(0,1)$ and if $\theta=1/n$ so that $\theta\to\mathring{\theta}=0$, then 
\[
\f 1n\sum_{\alpha=1}^{n}\psi(g_{\alpha};\theta)\asto1
\]
but 
\[
\EV\psi(Z^{g};\mathring{\theta})=\EV0=0.
\]
So our Master Theorem can't hold in this case. 
\end{rem}

\begin{rem}
[Necessity of Rank Stability \cref{assm:asRankStab}]
\label{remk:necessityRankStab} Suppose we have two Initial G-vars $g^{1},g^{2}\in\R^{n}$ which are sampled independently as $g_{\alpha}^{1},g_{\alpha}^{2}\sim\Gaus(0,1)$. Let $W\in\R^{n\times n}$ be sampled as $W_{\alpha\beta}\sim\Gaus(0,1/n)$. Then we can define $h^{2}=\theta g^{2}$ where $\theta=\exp(-n)$ as a function of $n$, using \refNonlinPlus{}, so that $h_{\alpha}^{2}\asto0$. Additionally, let $\bar{g}^{1}=Wg^{1}$ and $\bar{g}^{2}=Wh^{2}$. Again, $\bar{g}_{\alpha}^{2}\asto0$ but for any finite $n$, $\bar{g}^{2}$ is linearly independent from $\bar{g}^{1}$. Thus rank stability does not hold here, since the rank of $\{\bar{g}^{1},\bar{g}^{2}\}$ is 2 for all finite $n$, yet $\{Z^{\bar{g}^{1}},Z^{\bar{g}^{2}}\}=\{\Gaus(0,1),0\}$ has rank 1.

Now consider the (parameterless) nonlinearity $\psi(x,y)$ that is 1 except on the line $y=0$, where it is 0. Then 
\[
\f 1n\sum_{\alpha=1}^{n}\psi(\bar{g}_{\alpha}^{1},\bar{g}_{\alpha}^{2})\asto1
\]
since $\bar{g}_{\alpha}^{2}$ is almost surely nonzero, but 
\[
\EV\psi(Z^{\bar{g}^{1}},Z^{\bar{g}^{2}})=\EV\psi(Z^{\bar{g}^{1}},0)=\EV0=0.
\]
\end{rem}

Note this example uses the non-smoothness of $\psi$ in an essential way.
Therefore, we conjecture that rank stability assumption can be removed in \cref{thm:PLNetsorT+MasterTheorem}.

\begin{rem}
[Rank stability already holds for \netsort{} programs] \label{remk:rankStabilityNetsor} It turns out that, as long as we only have parameterless nonlinearities, we get rank stability \cref{assm:asRankStab} for free. This is formulated explicitly in \cref{lemma:rankStability}. It is as a result of our proof of \cref{thm:NetsorTMasterTheorem} that interleaves an inductive proof of this rank stability (more generally, the inductive hypothesis \ref{IH:coreSet}) with an inductive proof of the ``empirical moment'' convergence (the inductive hypothesis \ref{IH:MomConv}). 
\end{rem}

\paragraph{Relative Utilities of \cref{thm:PLNetsorT+MasterTheorem} and \cref{thm:PCNetsorT+MasterTheorem}}
    While pseudo-Lipschitz functions are closed under composition (at the expense of increasing degree), this is in general not true for parameter-control.
    Thus, the assumptions of \cref{thm:PCNetsorT+MasterTheorem} are relatively more cumbersome to check: one has to manually unwind the nonlinearities into $\varphi^x$ and then check this nested function is parameter-controlled.
    However, this increased flexibility of the parameter-control condition allows us to prove useful theorems that is not possible with the just \cref{thm:PLNetsorT+MasterTheorem}.
    For example, if $\varphi^x(z; \theta) = z / \theta$, then $\varphi^x$ is not pseudo-Lipschitz, but it is parameter-controlled if $\mathring \theta > 0$.
    This fact is used in \citet{yangTP1,yangTP2} to compute the Gaussian Process kernel and the Neural Tangent Kernel of a randomly initialized neural network with layernorm.

\subsection{Getting Rid of Rank Stability Assumption through More Test Function Smoothness}

Sometimes, the rank stability assumption \cref{assm:asRankStab} can be difficult to check, so it is convenient to have a version of the Master Theorem without it.
As shown in \cref{remk:necessityRankStab}, to do so, it's not enough to only have parameter-control; we need to have some smoothness in the nonlinearities and the test functions.
It turns out we can obtain a version of \cref{thm:PLNetsorT+MasterTheorem} without assuming rank stability.

\begin{thm}[Pseudo-Lipschitz \netsortplus{} Master Theorem without Rank Stability]
    \label{thm:PLNetsorT+MasterTheoremNoRS} Fix a Tensor Program initialized accordingly to \cref{setup:netsortplus}.
    Assume all nonlinearities are pseudo-Lipschitz (or, we can assume the slightly weaker \cref{assm:MasterTheoremSmoothness}).
    Then 
    \begin{enumerate}
    \item For any fixed $k$ and any pseudo-Lipschitz $\psi:\R^{k}\to\R$, as $n\to\infty$,
    \begin{equation*}
    \f 1n\sum_{\alpha=1}^{n}\psi(h_{\alpha}^{1},\ldots,h_{\alpha}^{k})\asto\EV\psi(Z^{h^{1}},\ldots,Z^{h^{k}}),
    \end{equation*}
    for any vectors $h^{1},\ldots,h^{k}$ in the program, where $Z^{h^{i}}$ are as defined in \cref{{defn:netsortplusKeyIntuit}}.
    \item Any scalar $\theta$ in the program tends to $\mathring{\theta}$ almost surely, where $\mathring{\theta}$ is as defined in \cref{{defn:netsortplusKeyIntuit}}.
    \end{enumerate}
\end{thm}

\paragraph{Difference between \cref{thm:PLNetsorT+MasterTheorem} and \cref{thm:PLNetsorT+MasterTheoremNoRS}}
is 1) \cref{thm:PLNetsorT+MasterTheorem} needs rank stability \cref{assm:asRankStab} and 2) the test function $\psi$ in \cref{thm:PLNetsorT+MasterTheoremNoRS}(1) is pseudo-Lipschitz.

\paragraph{Proving \cref{thm:PLNetsorT+MasterTheoremNoRS}}
The main difficulty with proving \cref{thm:PLNetsorT+MasterTheorem} without rank stability is that even if the covariance matrix of the $Z$ random variables converges, its pseudo-inverse may not.
This is a crucial step in the Gaussian conditioning trick, so we need to more carefully modify the proof skeleton of \cref{thm:NetsorTMasterTheorem} (in contrast to the proof of \cref{thm:PCNetsorT+MasterTheorem}, which just needed to modify some minor bounds in the proof skeleton).
The main idea is to carefully extend the core-set argument of \cref{sec:proofMainTheorem}.
See \cref{sec:proofNetsorTPlusNoRS} for the proof.

\section{Programs with Variable Dimension}
\label{appendix:VarDim}

\newcommand{\highlight}[1]{{\color{red}#1}}

We can also allow the matrices and vectors to have variable dimensions (not all equal to $n$). 
This is useful in reasoning about neural networks of varying widths in each layer, and in proving the Marchenko-Pastur law for arbitrary rectangular shape ratio (see \cref{{sec:MP}}).
The reader should feel free to skip ahead and read this section only when approaching these topics.

Here we need to spend a few words on the right analogue of ``large-$n$'' limit we should take. The basic idea is that, every time we apply one of the \refMatMul{}, \refNonlinPlus{}, or \refMoment{} rules, we implicitly set equal some dimensions of the matrices and/or vectors involved. These equalities partitions the vectors into classes sharing common dimensions. The limit we shall take is one in which the dimension of each class tends to infinity, such that the dimension ratio of each pair of distinct classes tends to a fixed number strictly in $(0,\infty)$.

\paragraph*{Notation}

In this section, we let $\dim(x)$ denote the dimension of a vector $x$.
We present a relatively self-contained exposition, but inevitably this will be highly similar to the nonvariable-dimension case, so we highlight important differences in \highlight{red}.
\begin{defn}\label{defn:netsortVarDim}
A \netsort{} program (with variable dimensions) is just a sequence of vectors inductively generated via one of the following ways from an initial set $\mathcal{V}$ of random vectors and a set $\mathcal{W}$ of random matrices 
\end{defn}

\begin{description}
\item [\texttt{Nonlin}\label{instr:nonlinVarDim}] (Same as in \cref{defn:netsort}) Given $\phi:\R^{k}\to\R$ and $x^{1},\ldots,x^{k}\in\R^{n}$ in the same CDC, we can generate $\phi(x^{1},\ldots,x^{k})\in\R^{n}$ 
\item [\highlight{\texttt{MatMul}}\label{instr:matmulVarDim}] Given $W\in\R^{n\times m}$ (resp. $W\in\R^{m\times n}$) and $x\in\R^{m}$, we can generate $Wx\in\R^{n}$ (resp. $W^{\trsp}x\in\R^{n}$)
\end{description}
\begin{defn}
A \netsortplus{} program (with variable dimension) is just a sequence of vectors inductively generated via one of the following ways from an initial set $\mathcal{C}$ of random scalars, an initial set $\mathcal{V}$ of random vectors, and a set $\mathcal{W}$ of random matrices 
\begin{description}
\item [\texttt{Nonlin$^+$}\label{instr:nonlin+VarDim}] (Same as in \cref{defn:netsortplus}) Given $\phi:\R^{k}\times\R^{l}\to\R$, previous scalars $\theta_{1},\ldots,\theta_{l}\in\R$ and vectors $x^{1},\ldots,x^{k}\in\R^{n}$ in the same CDC, we can generate a new vector 
\[
\phi(x^{1},\ldots,x^{k};\theta_{1},\ldots,\theta_{l})\in\R^{n}
\]
where $\phi(-;\theta_{1},\ldots,\theta_{l})$ applies coordinatewise to each $\alpha$-slice $(x_{\alpha}^{1},\ldots,x_{\alpha}^{k})$. 
\item [\texttt{Moment}\label{instr:momentVarDim}] (Same as in \cref{defn:netsortplus}) Given same setup as above, we can also generate a new scalar 
\[
\f 1n\sum_{\alpha=1}^{n}\phi(x_{\alpha}^{1},\ldots,x_{\alpha}^{k};\theta_{1},\ldots,\theta_{l})\in\R
\]
\item [\highlight{\texttt{MatMul}}\label{instr:matmul+VarDim}] Same as in \cref{defn:netsortVarDim} above.
\end{description}
\end{defn}
Note that in both definitions above, the essential change is that the matrix in \refMatMulVarDim{} is now allowed to have different dimensions on two sides.

The initial vectors in $\mathcal{V}$ can have varying dimensions. When we take the ``large dimension'' limit, we may seek to hold certain pairs of such dimensions to be equal. We formalize this as an equivalence relation between vectors in $\mathcal{V}$: $x\simeq y$ if we hold $\dim(x)=\dim(y)$ as this dimension tends to infinity.
In addition, we have the following natural constraints on the dimensions of vectors involved in each rule.
\begin{equation}
\begin{cases}
\text{If \ensuremath{y=\phi(x^{1},\ldots,x^{k})}, then \ensuremath{\dim(y)=\dim(x^{i}),\forall i}.}\\
\text{If \ensuremath{y=Wx} and \ensuremath{\bar{y}=W\bar{x}}, then \ensuremath{\dim(x)=\dim(\bar{x})} and \ensuremath{\dim(y)=\dim(\bar{y})}.}
\end{cases}\label{eqn:dimconstraint}
\end{equation}
This allows us to extend the equivalence relation $\simeq$ on $\mathcal{V}$ as discussed above to all vectors in the program.
\begin{defn}
\label{def:CDC}
Given an equivalence relation $\simeq$ on the vectors of a program, we extend this to an equivalence relation on all vectors of the program as the smallest equivalence relation containing the relation
\begin{equation}
h\equiv h'\iff h\simeq h'\text{ OR \ensuremath{h} and \ensuremath{h'} are constrained to have the same dimension by (\ref{eqn:dimconstraint})}.
\end{equation}
We call any such equivalence class a \emph{Common Dimension Class}, or CDC. We denote this common dimension of vectors in a CDC $\cdc$ by $\dim(\cdc)$.
\end{defn}

Intuitively, the dimensions of vectors in each CDC are all the same but can be different in different CDCs. The CDCs form a partition of all the vectors in a program. Each matrix in $\mathcal{W}$ ``straddles'' between two (possible the same) CDCs.

\begin{exmp}
    Consider the MLP in \cref{eqn:MLP} but with variable widths:
     $f(\xi; \theta) = W^{L+1} x^L(\xi)$ with input $\xi \in \R^{n^0}$ and output dimension $n^{L+1} = 1$, where we recursively define, for $l = 2, \ldots , L$,
\begin{align*}
    h^{l}(\xi)=W^{l}x^{l-1}(\xi)\in\R^{n^{l}},\quad x^{l}(\xi)=\phi(h^{l}(\xi)),\quad h^{1}(\xi)=W^{1}\xi\in\R^{n^{1}}
\end{align*}
where each $W^l \in \R^{n^l \times n^{l-1}}$.
Suppose we sample $W^l_{\alpha\beta} \sim \Gaus(0,1/n^{l-1})$.
The vectors in the program are $h^1(\xi), x^1(\xi), \ldots, h^l(\xi), x^l(\xi)$.
In \cref{def:CDC}, if $\simeq$ is empty, then the CDCs will just be $\{h^1(\xi), x^1(\xi)\}, \ldots, \{h^l(\xi), x^l(\xi)\}$ (each with common dimension $n^l$).
If instead we set $h^1(\xi) \simeq \cdots \simeq h^l(\xi)$, then there is only one CDC with common dimension $n^1 = \cdots = n^L$.
\end{exmp}

\begin{setup}[Variable Dimension \netsort{} or \netsortplus{}]\label{setup:netsortVarDim}
    We will consider \netsort{} or \netsortplus{} programs initialized as follows.
    \begin{description}
        \item[Matrix sampling] For each initial $W\in\R^{n\times m}$ in $\mathcal{W}$, we sample $W_{\alpha\beta}\sim\Gaus(0,\sigma_{W}^{2}/m)$ for an associated variance $\sigma_{W}^{2}$. In this setting, we also set $\sigma_{W^{\trsp}}^{2}\defeq\frac{n}{m}\sigma_{W}^{2}$ so that $(W^{\trsp})_{\alpha\beta}\disteq\Gaus(0,\sigma_{W^{\trsp}}^{2}/n)$.
        \item[Vector sampling] For every CDC $\cdc$, there is a multivariate Gaussian $Z^{\cdc\cap\mathcal{V}}=\left\{ Z^{g}:g\in\cdc\cap\mathcal{V}\right\} \in\R^{\cdc\cap\mathcal{V}}$ such that the vectors in $\cdc\cap\mathcal{V}$ are sampled like $\left\{ g_{\alpha}:g\in\cdc\cap\mathcal{V}\right\} \sim Z^{\cdc\cap\mathcal{V}}$ iid for each $\alpha\in[\dim(\cdc)]$. 
        \item[Scalar sampling] For a \netsortplus{} program, the scalars in $\mathcal{C}$ are sampled the same way as in \cref{setup:netsortplus}.
        \item[How the limit is taken] We shall consider a limit where for every matrix $W\in\R^{n\times m}$ in $\mathcal W$, the dimensions $n,m$ (which are dimensions of corresponding CDCs) tend to $\infty$ such that their ratio $n/m\to\rho\in(0,\infty)$ for some finite but nonzero $\rho$. Note this implies $\sigma_{W^{\trsp}}^{2}=\frac{n}{m}\sigma_{W}^{2}\to\mathring{\sigma}_{W^{\trsp}}^{2}\defeq\rho\sigma_{W}^{2}$. We also define $\mathring{\sigma}_{W}^{2}\defeq\sigma_{W}^{2}$ for convenience.
    \end{description}
\end{setup}

\begin{defn}\label{defn:ZVarDim}
    Given this setup, each vector $h$ again has roughly iid coordinates distributed like $Z^{h}$, which is defined as follows.
    Likewise, for \netsortplus{} programs, each scalar $\theta$ will tend to a deterministic limit $\mathring \theta$, as defined below.
This is exactly the same as in \cref{defn:Z} or \cref{defn:netsortplusKeyIntuit} except that $\sigma_W$ should be replaced by $\mathring \sigma_W$.
We highlight the places where this occurs in \highlight{red} below.
\begin{description}
\item [\texttt{ZInit}] Same as in \cref{defn:Z} or \cref{defn:netsortplusKeyIntuit}.
\item [\texttt{ZMatMul}] For every $W_{\alpha\beta}\sim\Gaus(0,\sigma_{W}^{2}/n)$ and vector $x$ in the program, we set $Z^{Wx}\defeq \hat{Z}^{Wx}+\Zdot^{Wx}$ where
    \begin{description}
    \item [\texttt{ZHat}] {$\hat{Z}^{Wx}$} is a Gaussian variable with zero mean. Let $\mathcal V_W$ denote the set of all vectors in the program of the form $W y$ for some $y$.
    Then $\{\hat Z^{W y}: W y \in \mathcal V_W\}$ is defined to be jointly Gaussian with zero mean and covariance
    \[
    \Cov\left(\hat{Z}^{Wx},\hat{Z}^{W y}\right)\defeq \highlight{\mathring\sigma_{W}^{2}}\EV Z^{x}Z^{y},\quad\text{for any \ensuremath{Wx,W y\in\mathcal{V}_W}.}
    \]
    Furthermore, $\{\hat Z^{W y}: W y \in \mathcal V_W\}$ is mutually independent from $\{\hat Z^{v }: v \in \mathcal V \cup \bigcup_{\bar W \ne W} \mathcal V_{\bar W}\}$, where $\bar{W}$ ranges over $\mathcal{W}\cup\{A^{\trsp}:A\in\mathcal{W}\}$.
    \item [\texttt{ZDot}]
    With partial derivative interpreted naively (but see \cref{rem:PartialDer} and \ref{rem:ExpectationPartialDerVarDim}),
    \[\Zdot^{Wx}\defeq \highlight{\mathring\sigma_{W}^{2}}\sum Z^{y}\EV\frac{\partial Z^{x}}{\partial\hat{Z}^{W^{\trsp}y}},\]
    summing over $W^\trsp y \in \mathcal V_{W^\trsp}$ introduced before $x$.
    \end{description}
\item [\texttt{ZNonlin}] Same as in \cref{defn:Z} or \cref{defn:netsortplusKeyIntuit}.
\item [\texttt{ZMoment}] Same as in \cref{defn:netsortplusKeyIntuit}.
\end{description}
\end{defn}

\begin{rem}[Partial derivative expectation]\label{rem:ExpectationPartialDerVarDim}
    The partial derivative expectation $\EV\frac{\partial Z^{x}}{\partial\hat{Z}^{W^{\trsp}x^{i}}}$ can be defined without derivatives as in \cref{rem:ExpectationPartialDer}, with the only change being the dependence on the dimension ratio, which we highlight in \highlight{red} below:
    In \texttt{ZDot} above, suppose $\{W^{\trsp}y^{i}\}_{i=1}^{k}$ are all elements of $\mathcal V_{W^\trsp}$ introduced before $x$.
    Define the matrix $C\in\R^{k\times k}$ by $C_{ij}\defeq \EV Z^{y^{i}}Z^{y^{j}}$ and define the vector $b\in\R^{k}$ by $b_{i}\defeq \EV\hat{Z}^{W^{\trsp}y^{i}}Z^{x}$. If $a=\highlight{\rho^{-1}}C^{+}b$ (where $C^{+}$ denotes the pseudoinverse of $C$ and $\rho=\lim m/n$ is the limit ratio of dimensions of $W\in\R^{m\times n}$), then in \refZdot{} we may set
    \begin{equation}
    \sigma_{W}^{2}\EV\frac{\partial Z^{x}}{\partial\hat{Z}^{W^{\trsp}y^{i}}}=a_{i}.\label{eq:ExpectationPartialDerNetsorT+}
    \end{equation}
\end{rem}

Then by straightforward modifications of the proofs of the nonvariable-dimension cases, we can prove the following Master Theorems.
\begin{thm}[Variable Dimension \netsort{} Master Theorem]
\label{thm:NetsorTMasterTheoremVarDim} Fix a \netsort{} program (with variable dimensions). Suppose the initial matrices $\mathcal{W}$ and vectors $\mathcal{V}$ are sampled in the fashion of \cref{setup:netsortVarDim}.
Assume all $\phi$ used in \refNonlinVarDim{} are polynomially bounded. Then for any fixed $k$ and any \emph{polynomially bounded} $\psi:\R^{k}\to\R$, as the dimensions of the vectors tend to $\infty$ as specified in \cref{setup:netsortVarDim}, we have

\[
\f 1n\sum_{\alpha=1}^{n}\psi(h_{\alpha}^{1},\ldots,h_{\alpha}^{k})\asto\EV\psi(Z^{h^{1}},\ldots,Z^{h^{k}}),
\]
for any collection of vectors $h^{1},\ldots,h^{k} \in \R^n$ in the program, where $Z^{h^{i}}$ are defined in \cref{defn:ZVarDim}.
\end{thm}

\begin{thm}[Variable Dimension Pseudo-Lipschitz \netsortplus{} Master Theorem]
    \label{thm:PLNetsorT+MasterTheoremVarDim} Fix a \netsort{} program (with variable dimensions). Suppose the initial matrices $\mathcal{W}$, vectors $\mathcal{V}$, and scalars $\mathcal{C}$ are sampled in the fashion of \cref{setup:netsortVarDim}.
    Suppose the program satisfies the \emph{rank stability assumption} (\cref{assm:asRankStab}).
    Assume each nonlinearity of \refNonlinPlusVarDim{} is pseudo-lipschitz.
    Then 
    \begin{itemize}
        \item For any fixed $k$ and any polynomially-bounded $\psi:\R^{k}\to\R$, in the limit described in \cref{setup:netsortVarDim},
        \[
        \f 1n\sum_{\alpha=1}^{n}\psi(h_{\alpha}^{1},\ldots,h_{\alpha}^{k})\asto\EV\psi(Z^{h^{1}},\ldots,Z^{h^{k}}),
        \]
        for any vectors $h^{1},\ldots,h^{k} \in \R^n$ in the same CDC, where $Z^{h^{i}}$ are as defined in \cref{defn:ZVarDim}.
        \item Any scalar $\theta$ in the program tends to $\mathring{\theta}$ almost surely, where $\mathring{\theta}$ is as defined in \cref{defn:ZVarDim}.
    \end{itemize}
\end{thm}

Again, \cref{thm:PLNetsorT+MasterTheoremVarDim} strictly generalized by the following Master Theorem for Parameter-Controlled nonlinearities.

\begin{thm}[Variable Dimension Parameter-Controlled \netsortplus{} Master Theorem]
    \label{thm:PCNetsorT+MasterTheoremVarDim} Fix a \netsort{} program (with variable dimensions). Suppose the initial matrices $\mathcal{W}$, vectors $\mathcal{V}$, and scalars $\mathcal{C}$ are sampled in the fashion of \cref{setup:netsortVarDim}.
    Suppose the program satisfies the \emph{rank stability assumption} (\cref{assm:asRankStab}).
    Assume $\varphi^u(-;-)$ is parameter-controlled at $\mathring \bigtheta^u$ for all vectors and scalars $u$.
    Then 
    \begin{itemize}
        \item 
        For any random vector $\bigtheta \in \R^l$ that converges almost surely to a deterministic vector $\mathring{\bigtheta}$, and for any $\psi(-; -): \R^k \times \R^l \to \R$ parameter-controlled at $\mathring{\bigtheta}$, we have
        \begin{align*}
            \f 1 n \sum_{\alpha=1}^n \psi(g^1_\alpha, \ldots, g^k_\alpha; \bigtheta) \asto \EV\psi(Z^{g^1}, \ldots, Z^{g^k}; \mathring{\bigtheta}).
        \end{align*}
        in the limit described in \cref{setup:netsortVarDim},
        for any G-vars $g^1, \ldots, g^k \in \R^n$ in the same CDC,
        where $Z^{g^{i}}$ are as defined in \cref{defn:ZVarDim}.
        \item Any scalar $\theta$ in the program tends to $\mathring{\theta}$ almost surely, where $\mathring{\theta}$ is as defined in \cref{defn:ZVarDim}.
    \end{itemize}
\end{thm}

%% file: RMTBackground.tex
\section{Random Matrix Theory Background}

In random matrix theory, we often are concerned with \emph{what the spectrum of a random matrix looks like} when the matrix is large.
The answer to this question is typically phrased in a convergence theorem which says \emph{the spectrum as a histogram converges to some distribution}.
We single out the notion of convergence we are concerned with in this paper.
\begin{defn}\label{defn:ASConvergenceESD}
    Let $\mu_n$ be a \emph{random} measure (such as the spectral distribution of a random matrix) on $\R$ for each $n=1,2,\ldots$.
    Let $\mu$ be a \emph{deterministic} measure on $\R$.
    We say $\mu_n$ converges almost surely to $\mu$, written $\mu_n \asto \mu$, if for every compactly supported continuous function $\varphi$, we have $\int_\R \varphi(x)\dd \mu_n(x) \asto \int_\R \varphi(x) \dd \mu(x)$ (where the almost sure convergence is over the randomness of $\mu_n$).
\end{defn}

The \emph{Moment Method} is a standard technique in probability theory to establish convergence in distribution.
In the context of random matrix theory, this method goes like the following:
\begin{itemize}
\item Under general conditions (see Carleman's condition \cref{fact:carleman}), a distribution is ``pinned down'' by its moments: if all moments of a distribution $\mu$ converges to the moment of a target distribution $\mu^{*}$, then we can in general say that $\mu\to\mu^{*}$ for some notion of convergence $\to$.
\item For a Hermitian matrix $A$ with eigenvalues $\lambda_1, \ldots, \lambda_n$, the empirical distribution $\mu$ of its eigenvalues has moments given by $\EV_{\lambda\sim\mu}\lambda^{r} = \f 1 n \sum_{\alpha=1}^n (\lambda_\alpha)^r =\f 1 n \tr A^{r}$.
\item Thus, to prove that the spectral distribution $\mu$ of a random matrix converges to some distribution $\mu^*$, we need to show, for all $r\ge0$, we have $\tr A^{r}\to\EV_{\lambda\sim\mu^*}\lambda^{r}$, for some notion of convergence $\to$.
\end{itemize}
This method implies that, to check the convergence in \cref{defn:ASConvergenceESD}, it suffices to check the convergence of moments:
\begin{fact}[Moment Method]\label{fact:MomentMethod}
    In the context of \cref{defn:ASConvergenceESD} where $\mu_n$ and $\mu$ are probability measures, $\mu_n \asto \mu$ if 1) $\mu$ satisfies Carleman's Condition (\cref{fact:carleman}) and so does $\mu_n$ almost surely for large enough $n$, and 2) for every $k = 1, 2, \ldots$,
    \[
        \EV_{x\sim \mu_n} x^k \asto \EV_{x \sim \mu} x^k
        \]
    where the convergence $\asto$ is over the randomness of the measure $\mu_n$.
\end{fact}
For more background on the Moment Method, see \cite{tao_topics_2012}.

\begin{fact}[Carleman's Condition]\label{fact:carleman}
    Let $\mu$ be a measure on $\R$ such that all moments
    \[M_k = \int_{-\infty}^\infty x^n \dd \mu(x),\quad k=0,1,2,\ldots\]
    are finite.
    If
    \[\sum_{k=1}^\infty M_{2k}^{-\f{1}{2k}} = \infty,\]
    then $\mu$ is the only measure on $\R$ with $\{M_k\}_k$ as its moments.
\end{fact}

%% file: MarchenkoPastur.tex
\global\long\def\mump{\mu_{\mathrm{mp}}}%

\global\long\def\musc{\mu_{\mathrm{sc}}}%
\global\long\def\relu{\mathrm{relu}}%

\section{Proving the Marchenko-Pastur Law by Tensor Programs}
\label{sec:MP}

Marchenko-Pastur law \citep{marcenko_distribution_1967} is another cornerstone random matrix result on the level of semicircle law. It says that the spectrum $\mu_{AA^\trsp}$ of (what's known as Wishart) matrices $AA^{\trsp}$, where $A\in\R^{m\times n},A_{\alpha\beta}\sim\Gaus(0,1/n)$, converges in distribution if the shape ratio $m/n\to\rho$ for a finite but nonzero $\rho\in(0,\infty)$:
(see \cref{defn:ASConvergenceESD} for meaning of $\asto$)
\begin{equation}
    \mu_{AA^\trsp} \asto
\mump(x)\defeq\relu(1-\rho^{-1})\delta(x)+\frac{1}{\rho2\pi x}\sqrt{(b-x)(x-a)}\ind_{[a,b]}(x)\dd x
\label{eqn:MPdefn}
\end{equation}
where $\delta(x)$ is the Dirac Delta, $\relu(x) = x \ind(x > 0)$, $a=(1-\sqrt{\rho})^{2}$, and $b=(1+\sqrt{\rho})^{2}$. This distribution also yields the (square of) of the singular value distributions of $A$.

In this section, we use \netsort{} with variable dimensions (\cref{appendix:VarDim}) to prove this law. Like for the semicircle law, our purpose is to 1) demonstrate concrete examples of computing with \netsort{} with variable dimensions, and 2) ``benchmark'' and show our framework powerful enough to prove this nontrivial result.

Like in \cref{sec:semicircle}, we adopt the Moment Method (\cref{fact:MomentMethod}). This means that we need to show that the moments of $AA^{\trsp}$ converge to the corresponding moments of $\mu$:
\begin{equation}
\frac{1}{n}\tr\left(AA^{\trsp}\right)^{k}\to\EV_{\lambda\sim\mump}\lambda^{k},\quad \text{for all $k=1,2, \ldots$}\label{eq:MPMomentConverge}
\end{equation}
To calculate this moment $\tr\left(AA^{\trsp}\right)^{k}$, we use the trace trick to rewrite it as
\[
\frac{1}{n}\tr\left(AA^{\trsp}\right)^{k}=\frac{1}{n}\EV_{v\sim\Gaus(0,I)}v^{\trsp}\left(AA^{\trsp}\right)^{k}v.
\]
We can then express the RHS in a \netsort{} program with variable dimensions: Let $\mathcal{V}=\{v\},\mathcal{W}=\{A\}$, and introduce new vectors via \refMatMul{} like so
\[
u^{i}=A^{\trsp}v^{i-1},\quad v^{i}=Au^{i},\quad i=1,\ldots,k,
\]
where we have set $v^{0}=v$ for convenience. Then mathematically, $v^{i}=(AA^{\trsp})^{i}v$. By \cref{thm:MasterTheoremmeanConvergenceMainText},
\[
    \frac{1}{n}\tr\left(AA^{\trsp}\right)^{k}
    = \EV_v \frac{v^{\trsp}v^{k}}{n}\asto\EV Z^{v}Z^{v^{k}}.
\]
Thus, we are done if can demonstrate
\[
\EV Z^{v}Z^{v^{k}}=\EV_{\lambda\sim\mump}\lambda^{k}
\]
Now, all that is left is to \emph{compute} $Z^{v^{i}}$.

\subsection{Explicit Calculations of the First Few Moments}

To give a concrete feel for the proof, we first calculate the first few moments as examples.
Before we begin, note that $\mathring{\sigma}_{A}^{2}=1$ and $\mathring{\sigma}_{A^{\trsp}}^{2}=\rho$ (where $\mathring{\sigma}_{A^{\trsp}}^{2}$ is as defined in \cref{setup:netsortVarDim}).

\paragraph*{First Moment}

First we have $Z^{v}=\Gaus(0,1)$, $Z^{u^{1}}=\Gaus(0,\mathring{\sigma}_{A^{\trsp}}^{2})=\Gaus(0,\rho),$ independent of each other. Then
\[
Z^{v^{1}}=\hat{Z}^{v^{1}}+\Zdot^{v^{1}},
\]
where $\Zdot^{v^{1}}=\mathring{\sigma}_{A}^{2}Z^{v}\EV\frac{\partial Z^{u^{1}}}{\partial\hat{Z}^{u^{1}}}=Z^{v}$ and $\hat{Z}^{v^{1}}=\Gaus(0,\rho)$ independent of everything else. Thus, the first moment of $AA^{\trsp}$ has limit
\[
\frac{1}{n}\tr AA^{\trsp}\asto\EV Z^{v}Z^{v^{1}}=\EV Z^{v}\Zdot^{v^{1}}=\EV(Z^{v})^{2}=1=\EV_{\lambda\sim\mump}\lambda
\]

\paragraph*{Second Moment}

Next, we have $Z^{u^{2}}=Z^{A^{\trsp}v^{1}}=\hat{Z}^{u^{2}}+\Zdot^{u^{2}}$ where 
\[
\Zdot^{u^{2}}=\mathring{\sigma}_{A^{\trsp}}^{2}\left(Z^{u^{1}}\EV\frac{\partial Z^{v^{1}}}{\partial\hat{Z}^{v^{1}}}\right)=\rho Z^{u^{1}},
\]
and $\hat{Z}^{u^{2}}$ is zero-mean and jointly Gaussian with $\hat{Z}^{u^{1}}$, with
\begin{align*}
\Var\left(\hat{Z}^{u^{2}}\right) & =\mathring{\sigma}_{A^{\trsp}}^{2}\EV\left(Z^{v^{1}}\right)^{2}=\rho\left(\EV\left(\hat{Z}^{v^{1}}\right)^{2}+\EV\left(\Zdot^{v^{1}}\right)^{2}\right)=\rho(\rho+1)=\rho^{2}+\rho\\
\Cov\left(\hat{Z}^{u^{2}},\hat{Z}^{u^{1}}\right) & =\mathring{\sigma}_{A^{\trsp}}^{2}\EV Z^{v^{1}}Z^{v}=\rho\EV\Zdot^{v^{1}}Z^{v}=\rho.
\end{align*}
Then $Z^{v^{2}}=Z^{Au^{2}}=\hat{Z}^{v^{2}}+\Zdot^{v^{2}}$, where
\[
\Zdot^{v^{2}}=\mathring{\sigma}_{A}^{2}\left(Z^{v^{1}}\EV\frac{\partial Z^{u^{2}}}{\partial\hat{Z}^{u^{2}}}+Z^{v}\EV\frac{\partial Z^{u^{2}}}{\partial\hat{Z}^{u^{1}}}\right)=Z^{v^{1}}+\rho Z^{v}.
\]
Thus, the second moment of $AA^{\trsp}$ has limit
\[
\frac{1}{n}\tr\left(AA^{\trsp}\right)^{2}\asto\EV Z^{v}Z^{v^{2}}=\EV Z^{v}\left(Z^{v^{1}}+\rho Z^{v}\right)=1+\rho=\EV_{\lambda\sim\mump}\lambda^{2}.
\]

\subsection{Proof for General Moments}

In general, we have
\begin{itemize}
\item $\{\hat{Z}^{v^{i}}\}_{i}$ is jointly Gaussian with covariance
\[
\Cov\left(\hat{Z}^{v^{i}},\hat{Z}^{v^{j}}\right)=\EV Z^{u^{i}}Z^{u^{j}},
\]
\item $\{\hat{Z}^{u^{i}}\}_{i}$ is jointly Gaussian with covariance
\[
\Cov\left(\hat{Z}^{u^{i}},\hat{Z}^{u^{j}}\right)=\rho\EV Z^{v^{i-1}}Z^{v^{j-1}},
\]
\item and
\begin{equation}
\Zdot^{v^{i}}=\sum_{j=0}^{i-1}Z^{v^{j}}\EV\frac{\partial Z^{u^{i}}}{\partial\hat{Z}^{u^{j+1}}},\quad\Zdot^{u^{i}}=\rho\sum_{j=1}^{i-1}Z^{u^{j}}\EV\frac{\partial Z^{v^{i-1}}}{\partial\hat{Z}^{v^{j}}}.\label{eq:MPcheckZRecursion}
\end{equation}
\end{itemize}
Expanding all $Z^{\bullet}$ into $\hat{Z}^{\bullet}+\Zdot^{\bullet}$ recursively, it is easy to see that there are coefficients $\{a_{j}^{i}\}_{i,j},\{b_{j}^{i}\}_{i,j}$ such that
\begin{equation}
Z^{v^{i}}=\sum_{j=0}^{i}a_{j}^{i}\hat{Z}^{v^{j}},\quad Z^{u^{i}}=\sum_{j=1}^{i}b_{j}^{i}\hat{Z}^{u^{j}}.\label{eq:MPZvExpansion}
\end{equation}
Since $\hat{Z}^{v}=\hat{Z}^{v^{0}}$ is independent from all other $\hat{Z}^{\bullet}$, we thus seek to prove
\[
a_{0}^{k}=\EV Z^{v^{k}}Z^{v}=\EV_{\lambda\sim\mump}\lambda^{k}.
\]
Given \cref{eq:MPZvExpansion}, we see
\[
a_{j}^{i}=\EV\frac{\partial Z^{v^{i}}}{\partial\hat{Z}^{v^{j}}},\quad b_{j}^{i}=\EV\frac{\partial Z^{u^{i}}}{\partial\hat{Z}^{u^{j}}}.
\]
Plugging \cref{eq:MPZvExpansion} into \cref{eq:MPcheckZRecursion}, we see
\begin{align*}
\Zdot^{v^{i}} & =\sum_{j=0}^{i-1}Z^{v^{j}}\EV\frac{\partial Z^{u^{i}}}{\partial\hat{Z}^{u^{j+1}}}=\sum_{j=0}^{i-1}Z^{v^{j}}b_{j+1}^{i}\\
 & =\sum_{j=0}^{i-1}\left(\sum_{l=0}^{j}a_{l}^{j}\hat{Z}^{v^{l}}\right)b_{j+1}^{i}=\sum_{l=0}^{i-1}\hat{Z}^{v^{l}}\sum_{j=l}^{i-1}a_{l}^{j}b_{j+1}^{i}.
\end{align*}
Then 
\[
\sum_{l=0}^{i}a_{l}^{i}\hat{Z}^{v^{l}}=Z^{v^{i}}=\hat{Z}^{v^{i}}+\Zdot^{v^{i}}=\hat{Z}^{v^{i}}+\sum_{l=0}^{i-1}\hat{Z}^{v^{l}}\sum_{j=l}^{i-1}a_{l}^{j}b_{j+1}^{i}.
\]
Matching coefficients, we obtain the recurrence relation
\begin{equation}
a_{i}^{i}=1,\quad\text{and for all \ensuremath{l=0,\ldots,i-1,}}\quad a_{l}^{i}=\sum_{j=l}^{i-1}a_{l}^{j}b_{j+1}^{i}.\label{eq:MParec}
\end{equation}
Similarly
\begin{align*}
\Zdot^{u^{i}} & =\rho\sum_{j=1}^{i-1}Z^{u^{j}}\EV\frac{\partial Z^{v^{i-1}}}{\partial\hat{Z}^{v^{j}}}=\rho\sum_{j=1}^{i-1}Z^{u^{j}}a_{j}^{i-1}\\
 & =\rho\sum_{j=1}^{i-1}\left(\sum_{l=1}^{j}b_{l}^{j}\hat{Z}^{u^{l}}\right)a_{j}^{i-1}=\rho\sum_{l=1}^{i-1}\hat{Z}^{u^{l}}\sum_{j=l}^{i-1}b_{l}^{j}a_{j}^{i-1}.
\end{align*}
Then
\[
\sum_{l=1}^{i}b_{l}^{i}\hat{Z}^{u^{l}}=Z^{u^{i}}=\hat{Z}^{u^{i}}+\Zdot^{u^{i}}=\hat{Z}^{u^{i}}+\rho\sum_{l=1}^{i-1}\hat{Z}^{u^{l}}\sum_{j=l}^{i-1}b_{l}^{j}a_{j}^{i-1}.
\]
Matching coefficients, we get
\begin{equation}
b_{i}^{i}=1,\quad\text{and for all \ensuremath{l=1,\ldots,i-1},}\quad b_{l}^{i}=\rho\sum_{j=l}^{i-1}b_{l}^{j}a_{j}^{i-1}.\label{eq:MPbrec}
\end{equation}
Let $M_{r}\defeq\EV_{\lambda\sim\mump}\lambda^{r}$. Then we claim that the solution to the recurrence \cref{eq:MParec} and \cref{eq:MPbrec} is given by
\[
a_{j}^{i}=M_{i-j},\quad b_{i}^{i}=1,\quad\text{and}\quad b_{j}^{i}=\rho M_{i-j}\quad\text{for all \ensuremath{i-j\ge1}}.
\]
Plugging into \cref{eq:MParec} and \cref{eq:MPbrec}, we see it remains to show the following Catalan-like identity
\begin{equation}
M_{s}=\rho\sum_{r=1}^{s-2}M_{r}M_{s-1-r}+(1+\rho)M_{s-1}.\label{eq:MPCatalanLikeIdentity}
\end{equation}
This can be done by a change of variables to express $\EV_{\lambda\sim\mump}\lambda^{r}$ as an expectation over the semicircle law $\musc$:
\[
M_{r}=\EV_{\lambda\sim\mump}\lambda^{r}=\EV_{\lambda\sim\musc}(1+\rho+\lambda\sqrt{\rho})^{r-1}.
\]
Expanding the power and using the Catalan identity \cref{eqn:Catalan} for the moments of the semicircle law, we have
\begin{equation}
M_{r}=\sum_{k=0}^{\lfloor(r-1)/2\rfloor}\alpha^{k}(1+\alpha)^{r-1-2k}\binom{r-1}{2k}C_{k}\label{eq:MPMomentCatalan}
\end{equation}
where $C_{k}$ is the $k$th Catalan number. Then one can verify \cref{eq:MPCatalanLikeIdentity} by expanding $M_{r}$ into Catalan numbers using \cref{eq:MPMomentCatalan}, and repeatedly applying the Catalan identity $C_{k}=\sum_{i=0}^{k-1}C_{i}C_{k-1-i}$. This finishes the proof of the Marchenko-Pastur law.

%% file: FIPProof.tex
\section{Subprogram and Subprogram Independence}

\label{sec:subprogramIndependence}

To prove \cref{thm:FreeIndependencePrinciple}, we need to introduce a notion of \emph{subprograms} and then the \emph{independence} of subprograms.
Our key lemma in this section is \cref{lem:uncorrelationIndependence}, which shows that, if a subprogram is sufficiently uncorrelated from the previous subprogram, then it is also independent from it.
\begin{defn}
A program can be written down formally as a sequence of lines, each assigning a new vector generated from \refNonlin{} or \refMatMul{} to a new variable, like the examples above. A \emph{subprogram} of a program is a span of such lines. Given a program $\pi$ and subprograms $\pi_{1},\ldots,\pi_{k}$, we write $\pi=\pi_{1}|\cdots|\pi_{k}$ if $\pi$ consists of $\pi_{1}$ followed by $\pi_{2}$ followed by $\pi_{3}$, and so on, ending with $\pi_{k}$. Each vector in $\pi$ is introduced in a unique subprogram $\pi_{i}$, for which we write $v\in\pi_{i}$. By convention, all initial vectors in $\mathcal V$ are included in the first subprogram $\pi_1$.

For example, the simple \netsort{} program represented by \cref{eqn:MLP} can be split into two subprograms, one which introduces $x^l, h^l$ for $l=1,\ldots, R$, and another with $l=R+1,\ldots, L$.
As another example, \cref{eqn:forwardGIABreak} and \cref{eqn:backwardGIABreak} are two subprograms that comprise the program expressing the forward and backward propagations of the network.
In general, forward propagation and backward propagation form natural subprograms, where the latter depends on (the preactivations computed in) the former.

Next, we introduce the notion of subprogram independence, which generalizes the main property of Simple GIA Check (\cref{assm:simpleGIACheck}).
Before that, we first recall the notation for distributional equality.
\paragraph{Notations}
Given two random variables $X, \Zz$, and a $\sigma$-algebra $\Aa$, the notation $X \disteq_\Aa \Zz$ means that for any integrable function $\phi$ and for any random varible $Z$ measurable on $\Aa$, $\EV \phi(X) Z = \EV \phi(\Zz)Z$.
We say that $X$ is distributed as (or is equal in distribution to) $\Zz$ conditional on $\Aa$.
In case $\Aa$ is the trivial $\sigma$-algebra, we just write $X \disteq \Zz$.
If $X$ and $Z$ agrees almost surely, then we write $X \aseq Z$.
\end{defn}

\begin{defn}
\label{def:subprogramIndep}Consider a program divided into two subprograms: $\pi=\pi_{1}|\pi_{2}$. For each $W\in\mathcal{W}$, we create an iid copy $\bar{W}$. Let $\bar{\pi}_{2}$ be the same as $\pi_{2}$ except that each matrix $W$ used in $\pi_{2}$ is replaced with $\bar{W}$ in $\bar{\pi}_{2}$. Let $x^{1},\ldots,x^{k}$ denote the vectors of $\pi_{1}$, $y^{1},\ldots,y^{p}$ denote the vectors of $\pi_{2}$, and $\bar{y}^{1},\ldots,\bar{y}^{p}$ denote the vectors of $\bar{\pi}_{2}$. We say \emph{$\pi_{2}$ is independent from $\pi_{1}$} if we have the distributional equality
\[
(Z^{x^{1}},\ldots,Z^{x^{k}},Z^{y^{1}},\ldots,Z^{y^{p}})\disteq(Z^{x^{1}},\ldots,Z^{x^{k}},Z^{\bar{y}^{1}},\ldots,Z^{\bar{y}^{p}})
\]
or equivalently, the conditional distributional equality
\[
(Z^{y^{1}},\ldots,Z^{y^{p}})\disteq_{Z^{x^{1}},\ldots,Z^{x^{k}}}(Z^{\bar{y}^{1}},\ldots,Z^{\bar{y}^{p}}).
\]
\end{defn}

For example, if $\pi_{1}$ expresses the forward computation of a network satisfying Simple GIA Check (\cref{assm:simpleGIACheck}), and $\pi_{2}$ expresses the backward computation, then we can assume $W^\trsp$ used in $\pi_2$ (backpropagation) is independent from $W$ used in $\pi_1$ (forward propagation).
This then implies $\pi_{2}$ is independent from $\pi_{1}$.
On the other hand, the subprogram given by \cref{eqn:backwardGIABreak} is \emph{not} independent from the subprogram given by \cref{eqn:forwardGIABreak}, as we calculated.

Note \cref{def:subprogramIndep} is purely about the structure of the subprograms (which is the only thing the random variables $Z^\bullet$ depend on), not the values of the vectors or matrices in the program.

We will talk about polynomials of matrices below, e.g.\ $WW^\trsp W^2 (W^\trsp)^2$ if $W$ is a square matrix.
\begin{defn}[Noncommutative Polynomial]
    Given a matrix $W \in \R^{m\times n}$, a \emph{(noncommutative) monomial} in $\{W, W^\trsp\}$ is a product $a X_1 X_2 \cdots X_k$ for some $a\in \R$, $k \ge 0$, where each $X_i \in \{W, W^\trsp\}$ (with appropriate matching shapes).
    Its degree is $k$.
    A \emph{(noncommutative) polynomial} in  $\{W, W^\trsp\}$ is a linear combination of such monomials.
    Its degree is the maximum degree of all of its nonzero monomials.
\end{defn}
\begin{rem}\label{rem:calculatingPolynomial}
    If $W$ and $v$ are a matrix and a vector of a program, and $A$ is a polynomial in $\{W,W^\trsp\}$, then one can naturally express the product $Av$ in \netsort{}.
    For example, if $A = W (W^\trsp)^2 W$, then $Av = u$ where
    \[
        v^1 = Wv,\quad
        v^2 = W^\trsp v^1,\quad
        v^3 = W^\trsp v^2,\quad
        u = W v^3.\]
    Below, when we say a program \emph{calculates $Av$}, we mean a program of this form.
\end{rem}

The following is the key workhorse of this section. It says that, in a program $\pi_{1}|\pi_{2}$, if $\pi_{2}$ 1) consists of only \refMatMul{} and 2) only depends on $\pi_{1}$ through a single vector $v$ \emph{uncorrelated} with most vectors of $\pi_{1}$, then $\pi_{2}$ is independent from $\pi_{1}$. As we will comment shortly, this gives another proof of the Gradient Independence Assumption when Simple GIA Check (\cref{assm:simpleGIACheck}) is satisfied.

\begin{lemma}[Independence from Uncorrelation]
\label{lem:uncorrelationIndependence}Consider a \netsort{} program $\pi=\pi_{1}|\pi_{2}$, and $\pi_{2}$ calculates $Av$ (in the sense of \cref{rem:calculatingPolynomial}) where $v$ is a vector of $\pi_{1}$ and $A$ is a polynomial in $\{W,W^{\trsp}\}$ for some $W\in\mathcal{W}$. If $\EV Z^{v}Z^{x}=\EV Z^{v}\hat{Z}^{y}=0$ for every $x,y\in\pi_{1}$ with $y=Wx$ or $y=W^{\trsp}x$, then $\pi_{2}$ is independent from $\pi_1$.
\end{lemma}

Note if $Wv$ or $W^{\trsp}v$ appears in $\pi_{1}$, then the premise implies $Z^{v}=0$ and the theorem becomes vacuously true. So to get nontrivial implications, $Wv$ and $W^{\trsp}v$ should not appear in $\pi_{1}$.
Given \cref{lem:uncorrelationIndependence}, we can obtain a more explicit form of $Z^{Av}$ by some straightforward calculation.
\begin{lemma}
\label{lem:uncorrelatedDecomp}In the setting of \cref{lem:uncorrelationIndependence}, we have
\[
Z^{Av}=\tau Z^{v}+S
\]
where $S$ is zero-mean and independent from $\{Z^{u}:u\in\pi_{1}\}$, and $\tau=\lim_{n\to\infty}\frac{1}{n}\tr A$.
\end{lemma}

\begin{proof}
By \cref{lem:uncorrelationIndependence}, we can assume that $W$ (where $A$ is a polynomial in $\{W,W^{\trsp}\}$) is independent from everything in $\pi_{1}$. Then a simple induction on the degree of $A$ in $W$ shows that 
\begin{equation}
Z^{Av}=\tau Z^{v}+S\label{eq:dontknowtau}
\end{equation}
 where $S$ is zero-mean and independent from $\{Z^{u}:u\in\pi_{1}\}$.

However, at this point, we don't know what $\tau$ would be. To understand $\tau$, we proceed as follows. Let $\bar{v}$ be sampled iid according to $\bar{v}_{\alpha}\sim Z^{v}$ (which can be constructed with a \netsort{} program\footnote{The vector $v$ can always be expressed as $\phi(g^{1},\ldots,g^{k})$ for some G-vars $g^{1},\ldots,g^{k}$ and some function $\phi:\R^{k}\to\R$. Then $\bar{v}$ can be constructed as $\phi(\bar{g}^{1},\ldots,\bar{g}^{k})$ where $\{\bar{g}_{\alpha}^{1},\ldots,\bar{g}_{\alpha}^{k}\}\sim\Gaus(\mu,\Sigma)$ with $\mu_{i}=\EV Z^{g^{i}}$ and $\Sigma_{ij}=\EV Z^{g^{i}}Z^{g^{k}}$.}), so that $Z^{\bar{v}}\disteq Z^{v}$. Then we can easily see from the definition of $Z$ that $Z^{A\bar{v}}=\tau Z^{\bar{v}}+\bar{S}$ where $\tau$ here is the same as $\tau$ in \cref{eq:dontknowtau} and $(Z^{\bar{v}},\bar{S})\disteq(Z^{v},S)$. Then by the Master Theorem,
\[
\tau\EV(Z^{\bar{v}})^{2}=\EV Z^{A\bar{v}}Z^{\bar{v}}=\lim_{n\to\infty}\frac{1}{n}\bar{v}^{\trsp}A\bar{v}=\lim_{n\to\infty}\frac{1}{n}\tr A\EV(Z^{\bar{v}})^{2}
\]
we get $\tau=\lim_{n\to\infty}\frac{1}{n}\tr A$ as desired.
\end{proof}

\paragraph{Extension to \netsortplus{} and Variable Dimensions}
Both \cref{lem:uncorrelatedDecomp,lem:uncorrelationIndependence} hold as stated for \netsortplus{} programs (see \cref{sec:netsortplus}) and programs with variable dimensions (see \cref{appendix:VarDim}).
Note that, in programs with variable dimensions, the $\tau$ in \cref{lem:uncorrelatedDecomp} vanishes if $v$ and $Av$ are in different CDCs.

\paragraph{Another Look at How Simple GIA Check Implies GIA}

The lemmas above give us another proof that the gradient independence assumption leads to correct calculations given Simple GIA Check (\cref{thm:GIA}).
To demonstrate \cref{lem:uncorrelationIndependence,lem:uncorrelatedDecomp} in action, before proving FIP, we give a proof of \cref{thm:GIA} for MLP.
\begin{proof}[Proof sketch for an MLP]
Consider the MLP $f(\xi)=W^{L+1}x^{L}(\xi)$ with input $\xi\in\R^{d}$ and output dimension $1$, where we recursively define, for $l=1,\ldots,L$, 
\begin{align*}
g^{l}(\xi) & =W^{l}x^{l-1}(\xi) & dx^{l-1}(\xi) & =W^{l\trsp}dg^{l}(\xi) \\
x^{l}(\xi) & =\phi(g^{l}(\xi)) & dg^{l}(\xi) & =\phi'(g^{l}(\xi))\odot dx^{l}(\xi).
\end{align*}
Here, the dimensionalities are $W^1 \in \R^{n \times d}; W^2, \ldots, W^L \in \R^{n \times n}; W^{L+1} \in \R^{1 \times n}$, and for all $l\in[L]$, $g^l, x^l, dx^l, dg^l \in \R^n$.

Let $\pi_{F}$ be the program that computes $g^l, x^l, l=1,\ldots, L$, on an input $\xi$.
Below, we abbreviate $g^l = g^l(\xi)$ and so on.
In this program, $\mathcal V = \{g^1, dx^L = \sqrt n W^{L+1}\}$,
\footnote{
    Here $dx^l$ should be interpreted as $\sqrt n \partial f/\partial x^l$, and likewise for $dg^l$.
}%
and $\mathcal W = \{W^2, \ldots, W^L\}$.%
\footnote{
    For concreteness (which will not matter much below): we sample $W^l_{\alpha\beta} \sim \Gaus(0, 1/n)$ for $l = 2, \ldots, L+1$, and suppose $g^1_\alpha \disteq \Gaus(0, 1)$, induced by appropriate sampling of $W^1$.
}

Now let's see how to apply \cref{lem:uncorrelatedDecomp,lem:uncorrelationIndependence} to show \cref{thm:GIA}.
The first step of the backward pass is $dg^{L}=\phi'(g^{L})\odot dx^{L}$. Let $\pi_{1}$ denote $\pi_{F}$ plus this line. Let $\pi_{2}$ be the next line $dx^{L-1}=W^{L\trsp}dg^{L}$. Then $\pi_{1},\pi_{2}$ along with $v=dg^{L}$ satisfies the condition of \cref{lem:uncorrelationIndependence}.
Indeed, $Z^{dg^L}$ is uncorrelated from $Z^{g^L}$ and $Z^{x^{L-1}} = \hat Z^{x^{L-1}}$ because $Z^{dg^L}$ is linear in $Z^{dx^L}$, which is by definition sampled independently from all $W^1, \ldots, W^L$.
Thus, by \cref{lem:uncorrelationIndependence}, we may \emph{rigorously} treat $W^{L\trsp}$ as independent from $W^{L}$ for the purpose of computing any quantity of the form of the \cref{thm:NetsorTMasterTheorem} for the program $\pi_{1}|\pi_{2}$.
In particular, by \cref{lem:uncorrelatedDecomp}, $Z^{dx^{L-1}} = \tau Z^{dg^L} + S$, where $S$ is zero-mean and independent from $\{Z^{u}:u\in\pi_{1}\}$, and $\tau=\lim_{n\to\infty}\frac{1}{n}\tr W^{L\trsp} = 0$, i.e.\ $Z^{dx^{L-1}} = S$.

Then we can repeat this reasoning inductively: $Z^{dg^{L-1}} = \phi'(Z^{g^{L-1}}) \cdot Z^{dx^{L-1}} = \phi'(Z^{g^{L-1}}) \cdot S$ is uncorrelated from $Z^{g^{L-1}}$ and $Z^{x^{L-2}} = \hat Z^{x^{L-2}}$ because $Z^{dg^{L-1}}$ is linear in $S$.
This lets us apply \cref{lem:uncorrelationIndependence,lem:uncorrelatedDecomp} to $\pi_2$ being the line $dx^{L-2}=W^{L-1\trsp}dg^{L-1}$, $\pi_1$ being all previous lines, and $v$ being $dg^{L-1}$.
So on and so forth.
\end{proof}

\subsection{Proving \texorpdfstring{\cref{lem:uncorrelationIndependence}}{Key Lemma}}

Here we prove \cref{lem:uncorrelationIndependence}.
It will use the following trivial but useful facts.
\begin{prop}
    \label{prop:hatZRandomSource}For any $x\in\pi$, $Z^{x}$ is measurable against the $\sigma$-algebra generated by $\{\hat{Z}^{y}:\text{G-var }y\in\pi\}$.
    .
\end{prop}
    
\begin{lemma}
    \label{lemma:uncorImpliesPartialMeanZero}Fix a matrix $W$. If $\EV Z^{v}\hat{Z}^{Wh}=0$ for all G-var $Wh$ introduced before $v$, then $\EV\partial Z^{v}/\partial\hat{Z}^{Wh}=0$ for all such $Wh$ as well.
    
\end{lemma}
\cref{prop:hatZRandomSource} follows from a simple inductive argument, and \cref{lemma:uncorImpliesPartialMeanZero} follows from \cref{eq:ExpectationPartialDer}.

\begin{proof}[Proof of \cref{lem:uncorrelationIndependence}]
    It suffices to prove the case when we assume $A=\prod_{i=1}^{p}W^{t_{i}},t_{i}\in\{1,\trsp\}$, can be expressed as a monomial of degree $p$ in $\{W,W^{\trsp}\}$. Let $\bar{W}$ be an iid copy of $W$ as in \cref{def:subprogramIndep} and likewise let $\bar{\pi}_{2}$, $y^{1},\ldots,y^{p}$, $\bar{y}^{1},\ldots,\bar{y}^{p}$ be as in \cref{def:subprogramIndep}. By our assumption, $\pi_{2}$ computes $y^{1}=W^{t_{1}}v$ and $y^{i}=W^{t_{i}}y^{i-1}$ for $i=2,\ldots,p$, and likewise $\bar{\pi}_{2}$ computes $\bar{y}^{1}=\bar{W}^{t_{1}}v$ and $\bar{y}^{i}=\bar{W}^{t_{i}}\bar{y}^{i-1}$ for $i=2,\ldots,p$.
    
    Define $\mathcal{X}$ to be the $\sigma$-algebra generated by $\{Z^{x}:x\in\pi_{1}\}$. Note that by \cref{prop:hatZRandomSource}, $\mathcal{X}$ is also the $\sigma$-algebra generated by $\{\hat{Z}^{x}:\text{G-var }x\in\pi_{1}\}$. Then we need to show
    \[
    (Z^{y^{1}},\ldots,Z^{y^{p}})\disteq_{\mathcal{X}}(Z^{\bar{y}^{1}},\ldots,Z^{\bar{y}^{p}}).
    \]
    We proceed by induction on $p$. The base case of $p=0$ is vacuously true.
    
    Suppose the induction hypothesis holds for $p=q-1$ 
    \[
    (Z^{y^{1}},\ldots,Z^{y^{q-1}})\disteq_{\mathcal{X}}(Z^{\bar{y}^{1}},\ldots,Z^{\bar{y}^{q-1}}).
    \]
    and we shall prove it for $p=q$. A simple inductive argument (as in our proof of the semicircle law; see \cref{sec:semicircleProofGeneralK}) shows that for each $i\in[q-1]$, we have 
    \begin{equation}
    Z^{\bar{y}^{i}}=\tau_{i}Z^{v}+\bar{S}_{i}\label{eq:barYDecomp}
    \end{equation}
    for some $\tau_{i}\in\R$ and some zero-mean $\bar{S_{i}}$ independent from $\mathcal{X}$ (and thus also independent from $Z^{v}$). %
    
    Since we have assumed $A$ to be a monomial in $\{W,W^{\trsp}\}$, we can suppose WLOG that $t_{q}=1$ so that $y^{q}=Wy^{q-1}$. By definition, $Z^{y^{q}}=\hat{Z}^{y^{q}}+\Zdot^{y^{q}}$ and $Z^{\bar{y}^{q}}=\hat{Z}^{\bar{y}^{q}}+\Zdot^{\bar{y}^{q}}$. We shall show
    \[
    (Z^{y^{1}},\ldots,Z^{y^{q-1}},\hat{Z}^{y^{q}},\Zdot^{y^{q}})\disteq_{\mathcal{X}}(Z^{\bar{y}^{1}},\ldots,Z^{\bar{y}^{q-1}},\hat{Z}^{\bar{y}^{q}},\Zdot^{\bar{y}^{q}})
    \]
    which would imply the IH for $p=q$.
    
    \textbf{Case $\hat{Z}^{y^{q}}$: }By definition, $\hat{Z}^{y^{q}}$ (resp. $\hat{Z}^{\bar{y}^{q}}$) is zero-mean, jointly Gaussian with $\{\hat{Z}^{Wx}:x,Wx\in\pi\}$ (resp. $\{\hat{Z}^{\bar{W}x}:x,\bar{W}x\in\bar{\pi}_{2}\}$), and independent from $\{\hat{Z}^{Qx}:x,Qx\in\pi\}$ for any $Q\ne W$ (resp. for any $Q\ne\bar{W}$). By \cref{prop:hatZRandomSource}, these facts fully determine the distributions of $Z^{y^{1}},\ldots,Z^{y^{q-1}},\hat{Z}^{y^{q}}$ and of $Z^{\bar{y}^{1}},\ldots,Z^{\bar{y}^{q-1}},\hat{Z}^{\bar{y}^{q}}$ conditioned on $\mathcal{X}$. \emph{We thus need to show that A) $\Cov(\hat{Z}^{y^{q}},\hat{Z}^{y^{i}})=\Cov(\hat{Z}^{\bar{y}^{q}},\hat{Z}^{\bar{y}^{i}})$ for all $i=1,\ldots,q-1$, and B) $\Cov(\hat{Z}^{y^{q}},\hat{Z}^{g})=\Cov(\hat{Z}^{\bar{y}^{q}},\hat{Z}^{g})$ for all G-var $g\in\pi_{1}$.}
    
    \textbf{A) }For any $i=2,\ldots,q-1$, we either have $y^{i}=Wy^{i-1}$ or $y^{i}=W^{\trsp}y^{i-1}$. 1) In the latter case, $\hat{Z}^{y^{i}}$ is independent from $\hat{Z}^{y^{q}}$, and likewise $\hat{Z}^{\bar{y}^{i}}$ is independent from $\hat{Z}^{\bar{y}^{q}}$, so trivially \emph{$\Cov(\hat{Z}^{y^{q}},\hat{Z}^{y^{i}})=\Cov(\hat{Z}^{\bar{y}^{q}},\hat{Z}^{\bar{y}^{i}})=0$. }2) In the former case,
    \begin{align*}
    \Cov(\hat{Z}^{y^{q}},\hat{Z}^{y^{i}}) & =\Cov(\hat{Z}^{Wy^{q-1}},\hat{Z}^{Wy^{i-1}})\\
     & =\sigma_{W}^{2}\EV Z^{y^{q-1}}Z^{y^{i-1}}=\sigma_{W}^{2}\EV Z^{\bar{y}^{q-1}}Z^{\bar{y}^{i-1}}\quad\text{by IH}\\
     & =\Cov(\hat{Z}^{\bar{y}^{q}},\hat{Z}^{\bar{y}^{i}})
    \end{align*}
    as desired. For $i=1$, we either have $y^{1}=Wv$ or $y^{1}=W^{\trsp}v$, and a similar logic shows $\Cov(\hat{Z}^{y^{q}},\hat{Z}^{y^{1}})=\Cov(\hat{Z}^{\bar{y}^{q}},\hat{Z}^{\bar{y}^{1}})$ as well.
    
    \textbf{B) }If $g\in\pi_{1}$ is a G-var with $g=Qx$ for some $Q\ne W$ and $x\in\pi_{1}$, then by definition, $\hat{Z}^{g}$ is independent from both $\hat{Z}^{y^{q}}$ and $\hat{Z}^{\bar{y}^{q}}$. Now suppose instead $g=Wx$ for some $x\in\pi_{1}$. Then
    \begin{align*}
    \Cov(\hat{Z}^{y^{q}},\hat{Z}^{g}) & =\sigma_{W}^{2}\EV Z^{y^{q-1}}Z^{x}=\sigma_{W}^{2}\EV Z^{\bar{y}^{q-1}}Z^{x}\quad\text{by IH}\\
     & =\sigma_{W}^{2}(\tau_{q-1}\EV Z^{v}Z^{x}+\EV\bar{S}_{q-1}Z^{x})=0\quad\text{by \cref{eq:barYDecomp}}\\
     & =\Cov(\hat{Z}^{\bar{W}\bar{y}^{q-1}},\hat{Z}^{Wx})=\Cov(\hat{Z}^{\bar{y}^{q}},\hat{Z}^{g}).
    \end{align*}
    Here $\EV\bar{S}_{q-1}Z^{x}=0$ because $\bar{S}_{q-1}$ is zero-mean and independent from $Z^{x}$ as $x\in\pi_{1}$, and $\EV Z^{v}Z^{x}=0$ by the premise of this theorem. 
    
    \textbf{Case $\Zdot^{y^{q}}$: }By definition, $\Zdot^{y^{q}}$ is a linear combination $\Zdot^{y^{q}}=\sigma_{W}^{2}\sum_{(y,x)\in\mathcal{P}}Z^{x}\EV\frac{\partial Z^{y^{q-1}}}{\partial\hat{Z}^{y}}$, where $\mathcal{P}$ is the set of all $y=W^{\trsp}x$ introduced before $y^{q}$ in $\pi$. Likewise, $\Zdot^{\bar{y}^{q}}$ is a linear combination $\Zdot^{\bar{y}^{q}}=\sigma_{W}^{2}\sum_{i\in\mathcal{I}}Z^{\bar{y}^{i-1}}\EV\frac{\partial Z^{\bar{y}^{q-1}}}{\partial\hat{Z}^{\bar{y}^{i}}}$ where $\mathcal{I}$ is the set of all $i\in[q-1]$ such that $\bar{y}^{i}=\bar{W}^{\trsp}\bar{y}^{i-1}$, and for convenience here we have set $y^{0}=\bar{y}^{0}=v$. Clearly, by IH, 
    \[
    (Z^{\bar{y}^{1}},\ldots,Z^{\bar{y}^{q-1}},\hat{Z}^{\bar{y}^{q}},\Zdot^{\bar{y}^{q}})\disteq_{\mathcal{X}}\left(Z^{y^{1}},\ldots,Z^{y^{q-1}},\hat{Z}^{y^{q}},\sigma_{W}^{2}\sum_{i\in\mathcal{I}}Z^{y^{i-1}}\EV\frac{\partial Z^{y^{q-1}}}{\partial\hat{Z}^{y^{i}}}\right).
    \]
    Note that $\{(y^{i},y^{i-1}):i\in\mathcal{I}\}\subseteq\mathcal{P}$. So it suffices to show that $\EV\frac{\partial Z^{y^{q-1}}}{\partial\hat{Z}^{y}}=0$ for all $(y,x)\in\mathcal{P}\setminus\{(y^{i},y^{i-1}):i\in\mathcal{I}\}$. Fix one such $y$, which must have been introduced in $\pi_{1}$.
    
    Now by \cref{eq:barYDecomp}, $\EV\frac{\partial Z^{y^{q-1}}}{\partial\hat{Z}^{y}}=\EV\frac{\partial Z^{\bar{y}^{q-1}}}{\partial\hat{Z}^{y}}=\EV\tau_{q-1}\frac{\partial Z^{v}}{\partial\hat{Z}^{y}}+\frac{\partial\bar{S}_{q-1}}{\partial\hat{Z}^{y}}$. Since $\bar{S}_{q-1}$ is zero-mean and independent from $\mathcal{X}\ni\hat{Z}^{y}$, we have $\EV\frac{\partial\bar{S}_{q-1}}{\partial\hat{Z}^{y}}=0$. At the same time, by \cref{lemma:uncorImpliesPartialMeanZero}, $\EV\frac{\partial Z^{v}}{\partial\hat{Z}^{y}}=0$ because $\EV Z^{v}\hat{Z}^{W^{\trsp}h}=0$ for all G-var $W^{\trsp}h\in\pi_{1}$ by the premise of this theorem. Combining these results together, we get $\EV\frac{\partial Z^{y^{q-1}}}{\partial\hat{Z}^{y}}=0$ as desired. %
\end{proof}

\section{Proving \texorpdfstring{Free Independence Principle}{Free Independence Principle}}
\label{sec:FIP}

\begin{proof}[Proof of \cref{thm:FreeIndependencePrinciple}]

For each $i=1,\ldots,t$, let $A^{i}$ be a polynomial in $\{W,W^{\trsp}\}$ for some $W\in\mathcal{W}$ or in $\{\Diag(x):x\in\pi\}$, such that consecutive $A^{i}$ and $A^{i+1}$ are always polynomials in different collections. (Here superscripts always denote indices).
Then we need to show
\[
\frac{1}{n}\tr(A^{t}-I\frac{1}{n}\tr A^{t})\cdots(A^{1}-I\frac{1}{n}\tr A^{1})\asto0.
\]
We apply the trace trick to re-express this as
\begin{equation}
\frac{1}{n}\EV v^{\trsp}(A^{t}-I\frac{1}{n}\EV u^{t\trsp}A^{t}u^{t})\cdots(A^{1}-I\frac{1}{n}\EV u^{1\trsp}A^{1}u^{1})v \asto 0
\label{eq:MixedMomentConverges}
\end{equation}
where expectation is taken over $v,u^{1},\ldots,u^{t}\sim\Gaus(0,I)$.
We will express this as a \netsortplus{} program and apply \cref{thm:NetsorT+MeanConvergenceLinearlyBounded} to show this convergence.
This \netsortplus{} program takes the form of $\pi|\pi'$ where $\pi$ is the program in the theorem statement defining $W\in\mathcal{W}$ and $x\in\pi$, and $\pi'$ computes $\frac{1}{n}v^{\trsp}(A^{t}-\frac{1}{n}u^{t\trsp}A^{t}u^{t})\cdots(A^{1}-\frac{1}{n}u^{1\trsp}A^{1}u^{1})v$, as we describe in more detail below%
\footnote{For concreteness (which won't matter in the proof below): The initial set of vectors $\mathcal{V}_{\pi|\pi'}$ is then the initial vectors of $\pi$ along with $\{v,u^{1},\ldots,u^{t}\}$, $\mathcal{V}_{\pi|\pi'}=\mathcal{V}_{\pi}\cup\{v,u^{1},\ldots,u^{t}\}$.
Likewise, the initial matrices $\mathcal W_{\pi |\pi'}$ are the same as those of $\pi$, $\mathcal W_{\pi}$.}.
.

Inside $\pi'$ , we let $v^{0}=v$ and compute $v^{i}$ as follows in subprograms of $\pi'$:
\begin{align*}
v^{i} & =(A^{i}-\frac{1}{n}u^{i\trsp}A^{i}u^{i})v^{i-1},\quad\forall i\in[t].
\end{align*}
Then the quantity in \cref{eq:MixedMomentConverges} is just $\frac{1}{n}v^{\trsp}v^{t}$. %

\textbf{Example} If $A^{i}=(W^{\trsp})^{2}W$ for some $W\in\mathcal{W}$ then we can unpack $v^{i}=(A^{i}-\frac{1}{n}u^{i\trsp}A^{i}u^{i})v^{i-1}$ into the subprogram

\begin{align*}
g^{1} & =Wv^{i-1}&
g^{2} & =W^{\trsp}g^{2}&
g^{3} & =W^{\trsp}g^{3}=A^{i}v^{i-1}&
h^{1} & =Wu^{i}\\
h^{2} & =W^{\trsp}h^{1}&
h^{3} & =W^{\trsp}h^{2}&
c & =\frac{1}{n}u^{i\trsp}h^{3}&
v^{i} & =g^{3}-c\cdot v^{i-1}.
\end{align*}
Here $c$ is computed using \refMoment{} and $v^i$ is computed using \refNonlinPlus{}, and everything else uses \refMatMul{}.
Similarly, if $A^{i}=\Diag(x)$ for some $x\in\pi$ with bounded coordinates, then we can unpack $v^{i}=(A^{i}-\frac{1}{n}u^{i\trsp}A^{i}u^{i})v^{i-1}$ into
\begin{align*}
g & =x\odot v^{i-1}&
h & =x\odot u^{i}&
c & =\frac{1}{n}u^{i\trsp}h&
v^{i} & =g-c\cdot v.
\end{align*}
Here $c$ is computed using \refMoment{} and everything else is computed using \refNonlinPlus{}.
Note that, in both examples, the nonlinearities involved satisfy the nonlinearity conditions of \cref{thm:NetsorT+MeanConvergenceLinearlyBounded}.
One can easily see this is true in general for the entire subprogram $\pi'$.
By induction on $t$, we prove the following claim, which would imply \cref{eq:MixedMomentConverges}.
\begin{claim}
\emph{For each $v^{i},i\ge1$, the associated random variable $Z^{v^{i}}$ is zero-mean and 1) independent from $\mathcal{X}^{i-1}\defeq\{Z^{x}:\text{\ensuremath{x} introduced before or is \ensuremath{v^{i-1}}\}}$ if $A^{i}$ is not diagonal or 2) uncorrelated from $\mathcal{\mathcal{X}}^{i-1}$ and $\hat{\mathcal{X}}^{i-1}\defeq\{\hat{Z}^{x}:\text{G-var \ensuremath{x} introduced before or is \ensuremath{v^{i-1}}\}}$ if $A^{i}$ is diagonal.}
\end{claim}

Here, ``uncorrelated'' means $\EV Z^{v^{i}}R=0$ for all $R\in\mathcal{X}^{i-1}\cup\hat{\mathcal{X}}^{i-1}$. Applying this claim to $v^{t}$, we get
\[
\EV_{v, u^1, \ldots, u^t} v^{\trsp}(A^{t}-u^{t\trsp}A^{t}u^{t})\cdots(A^{1}-u^{1\trsp}A^{1}u^{1})v=v^{\trsp}v^{t}\asto\EV Z^{v}Z^{v^{t}}=0
\]
by \cref{thm:NetsorT+MeanConvergenceLinearlyBounded}%
\footnote{
    While \cref{thm:NetsorT+MeanConvergenceLinearlyBounded} is stated for a program (and not a subprogram), its proof can be readily adapted to the subprogram case.
}
, as desired, assuming rank stability (\cref{assm:asRankStab}). We shall check rank stability after proving this claim.
\begin{proof}
We proceed by induction on $t$, starting with $t=1$. Our key tool is \cref{lem:uncorrelatedDecomp}.
\begin{description}
\item [{BaseCase:}] By the cyclic property of trace, we can WLOG suppose $A^{1}$ is a polynomial in $\{W,W^{\trsp}\}$ for some $W\in\mathcal{W}$. Furthermore, we can assume WLOG that $A^{1}$ is a monomial in $W$ and $W^{\trsp}$. Then $\frac{1}{n}u^{1\trsp}A^{1}u^{1}\asto\tau^{1}\defeq\lim_{n\to\infty}\frac{1}{n}\tr A^{1}$. by something? By \cref{lem:uncorrelatedDecomp}, $Z^{v^{1}}=Z^{A^{1}v}=\tau^{1}Z^{v}+S$ where $S$ is zero-mean and independent from $\mathcal{X}^{0}$. Then $Z^{v^{1}}=Z^{A^{1}v-\left(\frac{1}{n}u^{1\trsp}A^{1}u^{1}\right)v}=S$ satisfies the required zero-mean and independence property.
\item [{Induction:}]~
\begin{description}
\item [{NondiagonalCase}] By IH, $Z^{v^{i-1}}$ is uncorrelated from $\mathcal{X}^{i-2}\cup\hat{\mathcal{X}}^{i-2}$. So we can apply \cref{lem:uncorrelatedDecomp} in the same fashion as in the base case to obtain the desired result.
\item [{DiagonalCase:}] Suppose $A^{i}=\Diag(x)$ for some $x\in\pi$. By the non-consecutive assumption, $A^{i-1}$ is not diagonal. So IH tells us $Z^{v^{i-1}}$ is independent from $\mathcal{X}^{i-2}$. One can see easily that $Z^{v^{i}}=Z^{v^{i-1}}(Z^{x}-\EV Z^{x}).$ We need to prove $\EV Z^{v^{i}}R=0$ for all $R\in\mathcal{X}^{i-1}\cup\mathcal{\hat{X}}^{i-1}$. We divide into cases:
\begin{enumerate}
\item Suppose $R\in\mathcal{X}^{i-2}\cup\hat{\mathcal{X}}^{i-2}$. Then $\EV Z^{v^{i}}R=\EV Z^{v^{i-1}}(Z^{x}-\EV Z^{x})R=\EV Z^{v^{i-1}}\EV(Z^{x}-\EV Z^{x})R=0$ because $Z^{v^{i-1}}$ is independent from $Z^{x}$ and $R$ and is zero-mean.
\item Suppose $R\in\mathcal{\hat{X}}^{i-1}\setminus\mathcal{\hat{X}}^{i-2}$. Then $R$ is possibly correlated with $Z^{v^{i-1}}$ but $\{R,Z^{v^{i-1}}\}$ is independent from $\mathcal{\mathcal{X}}^{i-2}\ni x$ by \cref{lem:uncorrelationIndependence}. Therefore, $\EV Z^{v^{i}}R=\EV Z^{v^{i-1}}(Z^{x}-\EV Z^{x})R=\EV Z^{v^{i-1}}R\EV(Z^{x}-\EV Z^{x})=0$.
\item Suppose $R\in\mathcal{X}^{i-1}\setminus\mathcal{X}^{i-2}$. Then $R$ decomposes into $R=\tau Z^{v^{i-2}}+S$ by \cref{lem:uncorrelatedDecomp} for some $\tau\in\R$ and $S$ zero-mean and independent from $\mathcal{X}^{i-2}$. Then 
\begin{align*}
\EV Z^{v^{i}}R & =\EV Z^{v^{i-1}}=\EV Z^{v^{i-1}}(Z^{x}-\EV Z^{x})R\\
 & =\EV Z^{v^{i-1}}(Z^{x}-\EV Z^{x})(\tau Z^{v^{i-2}}+S)\\
 & =\tau\EV Z^{v^{i-1}}(Z^{x}-\EV Z^{x})Z^{v^{i-2}}+\EV Z^{v^{i-1}}(Z^{x}-\EV Z^{x})S\\
 & =0.
\end{align*}
Here $\EV Z^{v^{i-1}}(Z^{x}-\EV Z^{x})S=\EV Z^{v^{i-1}}S\EV(Z^{x}-\EV Z^{x})=0$ because $\{Z^{v^{i-1}},S\}$ is independent from $Z^{x}$. Similarly, $\EV Z^{v^{i-1}}(Z^{x}-\EV Z^{x})Z^{v^{i-2}}=\EV Z^{v^{i-1}}\EV(Z^{x}-\EV Z^{x})Z^{v^{i-2}}=0$ because by IH $Z^{v^{i-1}}$ is zero-mean and independent from $\mathcal{X}^{i-2}\ni Z^{x},Z^{v^{i-2}}$.
\end{enumerate}
\end{description}
\end{description}
\end{proof}

Finally, to see rank stability (\cref{assm:asRankStab}), we simply note two points: 1) The subprogram $\pi$ is a \netsort{} program and thus it has rank stability automatically (\cref{remk:rankStabilityNetsor}). 2) $\hat Z^x$ has nonzero variance for every G-var $x$ computed in $\pi'$.
Since each vector used in \refMatMulPlus{} in $\pi'$ depends linearly on some unique $\hat Z^x$, this shows that the rank in \cref{assm:asRankStab} is always full, and thus rank stability holds.

\end{proof}

%% file: proofs.tex
\section{Mathematical Tools}

\label{sec:proofs}

We will use the following trivial but useful fact repeatedly.

\begin{lemma}\label{lem:powerbound}
For an integer $m$, and complex numbers $a_i \in \C$, $i \in [k]$,
\[
\left| \sum_{i=1}^k a_i \right|^m
\le 
k^{m-1} \sum_{i=1}^k \left|a_i\right|^m
.
\]
\end{lemma}
\begin{proof}
Expand the power in the LHS using the multinomial theorem, apply AM-GM to each summand, and finally aggregate using triangle inequality.
\end{proof}

\subsection{Probability Facts}
\label{sec:probfacts}

\paragraph{Notations}
Given two random variables $X, \Zz$, and a $\sigma$-algebra $\Aa$, the notation $X \disteq_\Aa \Zz$ means that for any integrable function $\phi$ and for any random varible $Z$ measurable on $\Aa$, $\EV \phi(X) Z = \EV \phi(\Zz)Z$.
We say that $X$ is distributed as (or is equal in distribution to) $\Zz$ conditional on $\Aa$.
In case $\Aa$ is the trivial $\sigma$-algebra, we just write $X \disteq \Zz$.
If $X$ and $Z$ agrees almost surely, then we write $X \aseq Z$.
The expression $X \distto \Zz$ (resp. $X \asto \Zz$) means $X$ converges to $\Zz$ in distribution (resp. almost surely).

\begin{lemma}\label{lemma:momentBoundASConvergence}
Let $\{X_n\}_{n \ge 1}$ be a sequence of random variables with zero mean.
If for some $p \in \N$ and for all $n$, $\EV X_n^{2p} \le c n^{-1-\lambda}$, for some $\lambda > 0$, then $X_n \to 0$ almost surely.
\end{lemma}
\begin{proof}
By Markov's inequality, for any $\epsilon > 0$,
\begin{align*}
    \Pr(|X_n| > \epsilon)
        &=
            \Pr(X_n^{2p} > \epsilon^{2p})
        \le
            \EV X_n^{2p}/\epsilon^{2p}
        \le c n^{-1-\lambda}/\epsilon^{2p}
        \\
    \sum_n \Pr(|X_n| > \epsilon)
        &\le
            \sum_n c n^{-1-\lambda}/\epsilon^{2p}
        <    
            \infty.
\end{align*}
By Borel-Cantelli Lemma, almost surely, $|X_n| \le \epsilon$ for all large $n$.
Then, if we pick a sequence $\{\epsilon_k > 0\}_k$ converging to 0, we have that, almost surely, for each $k$, $|X_n| \le \epsilon_k$ for large enough $n$ --- i.e. almost surely, $X_n \to 0$.
\end{proof}

The following is a standard fact about multivariate Gaussian conditioning
\begin{prop}\label{prop:GaussianCondition}
Suppose $\R^{n_1 + n_2} \ni x \sim \Gaus(\mu, K)$, where we partition $x = (x_1, x_2) \in \R^{n_1} \times \R^{n_2}, \mu = (\mu_1, \mu_2) \in \R^{n_1} \times \R^{n_2}$, and $K = \begin{pmatrix} K_{11} & K_{12}\\ K_{21} & K_{22}\end{pmatrix}$.
Then
$x_1 \disteq_{x_2} \Gaus(\mu|_{x_2}, K|_{x_2})$
where
\begin{align*}
    \mu|_{x_2}
        &=
            \mu_1 + K_{12} K_{22}^+ (x_2 - \mu_2)\\
    K|_{x_2}
        &=
            K_{11} - K_{12} K_{22}^+ K_{21}.
\end{align*}

\end{prop}

\begin{lemma}[Stein's Lemma]\label{lemma:stein}
For jointly Gaussian random variables $Z_1, Z_2$ with means $\overline Z_1, \overline Z_2$, and any differentiable function $\phi: \R \to \R$ where both $\EV \phi'(Z_1)$ and $\EV Z_1 \phi(Z_2)$ exist, we have
\[\EV (Z_1 - \overline Z_1) \phi(Z_2) = \Cov(Z_1, Z_2) \EV \phi'(Z_2).\]

More generally, for jointly Gaussian random variables $Z_1,\ldots, Z_k$ with means $\overline Z_1, \ldots, \overline Z_k$, and any differentiable function $\phi: \R^k \to \R$, we have
\[\EV (Z_1 - \overline Z_1) \phi(Z_1, \ldots, Z_k) =
\sum_{j=1}^k \Cov(Z_1, Z_j) \EV \pd_j \phi(Z_1, \ldots, Z_k)\]
whenever both sides are finite.
\end{lemma}

\begin{lemma}\label{lemma:EXf}
Let $X = (X_1, \ldots, X_k) \in \R^k$ be a multivariate Gaussian with 0 mean and nondegenerate covariance $\Omega$.
Let $X_{\bar i}$ denote the vector $(X_1, \ldots, X_{i-1}, X_{i+1}, \ldots, X_k) \in \R^{k-1}$.
Likewise, let $\Omega_{i \bar i} \in \R^{k-1}$ (resp.\ $\Omega_{\bar i i} \in \R^{k-1}$) be the $i$th row (resp.\ column) with entry $i$ removed,
and $\Omega_{\bar i \bar i} \in \R^{(k-1) \times (k-1)}$ be the submatrix of $\Omega$ obtained by removing the $i$th row and $i$th column.
Then for any $i \in [k]$ and any $f: \R^k \to \R$,
\begin{align*}
\EV X_i f(X)
  &=
    \sum_{j=1}^k a_j \Omega_{ji}
\end{align*}
whenever both sides are defined,
where
\begin{align*}
a_j
  &\defeq
    \f{\EV (X_j - \EV[X_j \mid X_{\bar j}]) f(X)}
    {\Var(X_j \mid X_{\bar j})}
  =
    \f{\EV (X_j - \Omega_{j\bar j} \Omega_{\bar j\bar j}^+ X_{\bar j}) f(X)}
    {\Omega_{jj} - \Omega_{j\bar j} \Omega_{\bar j\bar j}^+ \Omega_{\bar j j}}
    .
\end{align*}
Note that, for any $j$, $a_j$ does not depend on $i$.
\end{lemma}
\begin{proof}
By a standard density argument, we may assume $f$ is differentiable.
Then by multivariate Stein's Lemma (\cref{lemma:stein}),
\begin{align*}
\EV X_i f(X)
  &=
    \sum_{j=1}^k \Omega_{ij} \EV \pd_j f(X)
    .
\end{align*}
By applying bivariate Stein's Lemma (\cref{lemma:stein}) to conditional distributions, we get
\begin{align*}
\EV \pd_j f(X)
  &=
    \EV_{X_{\bar j}} \EV_{X_j | X_{\bar j}} \pd_j f(X)
  =
    \EV_{X_{\bar j}}
    \EV_{X_j | X_{\bar j}}
      \f{(X_j - \EV[X_j \mid X_{\bar j}]) f(X)}
      {\Var(X_j \mid X_{\bar j})}
  =
    \EV_{X}
      \f{(X_j - \EV[X_j \mid X_{\bar j}]) f(X)}
      {\Var(X_j \mid X_{\bar j})}
      .
\end{align*}
The other equality follows from straightforward calculations.
\end{proof}

\begin{lemma}\label{lemma:gaussianDer}
Let $\Phi: \R^n \to \R$ be measurable.
Then for $z \sim \Gaus(\zeta, \Sigma)$, the following Hessian and gradient matrices are equal:
\begin{align*}
    \Jac{^2}{\zeta^2} \EV \Phi(z)
        &=
            2\Jac{}{\Sigma} \EV \Phi(z)
\end{align*}
whenever both sides exist.
\end{lemma}
\begin{proof}
First assume $\Sigma$ is invertible.
We check
\begin{align*}
    \Jac{}{\zeta} e^{-\f 1 2 (\zeta-z)\inv \Sigma (\zeta-z)}
        &=
            - \inv \Sigma (\zeta - z)
            e^{-\f 1 2 (\zeta-z)\inv \Sigma (\zeta-z)}
            \\
    \Jac{^2}{\zeta^2} e^{-\f 1 2 (\zeta-z)\inv \Sigma (\zeta-z)}
        &=
            \left[
                - \inv \Sigma
                + \inv \Sigma (\zeta-z) (\zeta-z)^\trsp \inv \Sigma
            \right]
            e^{-\f 1 2 (\zeta-z)\inv \Sigma (\zeta-z)}
            \\
    \Jac{}{\Sigma} \f{e^{-\f 1 2 (\zeta-z)\inv \Sigma (\zeta-z)}}{
                    \det(2\pi \Sigma)^{1/2}}
        &=
            \f 1 2 \left[
                - \inv \Sigma
                + \inv \Sigma (\zeta-z) (\zeta-z)^\trsp \inv \Sigma
            \right]
            \f{e^{-\f 1 2 (\zeta-z)\inv \Sigma (\zeta-z)}}{
                    \det(2\pi \Sigma)^{1/2}}
            \\
        &=
            \f 1 2
            \Jac{^2}{\zeta^2} e^{-\f 1 2 (\zeta-z)\inv \Sigma (\zeta-z)}
            .
\end{align*}
Integrating against $\Phi$ gives the result.
For general $\Sigma$, apply a continuity argument, since the set of invertible $\Sigma$s is dense inside the set of all PSD $\Sigma$.
\end{proof}

\subsection{Gaussian Conditioning Trick}

\paragraph{Review of Moore-Penrose Pseudoinverse}

We first recall Moore-Penrose pseudoinverse and some properties of it.
\begin{defn}\label{defn:pseuodoinverse}
For $A \in \R^{n \times m}$, a pseudoinverse of $A$ is defined as a matrix $A^+ \in \R^{m \times n}$ that satisfies all of the following criteria
\begin{align*}
  A A^+ A &= A,&
  A^+ A A^+ &= A^+,&
  (AA^+)^\trsp &= AA^+,&
  (A^+ A)^\trsp &= A^+ A
  .
\end{align*}
\end{defn}

The following facts are standard.
\begin{itemize}
    \item If $A$ has real entries, then so does $A^+$.
    \item The pseudoinverse always exists and is unique.
    \item When $A$ is invertible, $A^+ = \inv A$.
    \item $(A^\trsp)^+ = (A^+)^\trsp$, which we denote as $A^{+\trsp}$.
    \item $A^+ = (A^\trsp A)^+ A^\trsp = A^\trsp (A A^\trsp)^+$.
    \item $AA^+$ is the orthogonal projector to the column space of $A$;
        $I - A^+ A$ is the orthogonal project to the null space of $A$.
    \item If $A$ has singular value decomposition $A = U\Lambda V$ where $U$ and $V$ are orthogonal and $\Lambda$ has the singular values on its diagonal, then $A^+ = V^\trsp \Lambda^+ U^\trsp$ where $\Lambda^+$ inverts all nonzero entries of $\Lambda$.
    \item For any collection of vectors $\{v_i\}_{i=1}^n$ in a Hilbert space, $w \mapsto \sum_{i,j=1}^n v_i (\Sigma^+)_{ij} \la v_j, w \ra $, where $\Sigma_{ij} = \la v_i, v_j \ra$, is the projection operator to the linear span of $\{v_i\}_{i=1}^n$.
\end{itemize}

We present a slightly more general versions of lemmas from \citet{bayati_dynamics_2011} that deal with singular matrices.
\begin{lemma}\label{lemma:condTrickVec}
Let $z \in \R^n$ be a random vector with i.i.d. $\Gaus(0, v^2)$ entries and let $D \in \R^{m\times n}$ be a linear operator.
Then for any constant vector $b \in \R^n$ the distribution of $z$ conditioned on $Dz = b$ satisfies:
\begin{align*}
    z
        &\disteq_{Dz = b}
            D^+ b + \Pi \tilde z
\end{align*}
where $D^+$ is the (Moore-Penrose) pseudoinverse, $\Pi$ is the orthogonal projection onto subspace $\{z: Dz = 0\}$, and $\tilde z$ is a random vector of i.i.d. $\Gaus(0, v^2)$.
\end{lemma}
\begin{proof}
When $D = [I_{m \times m} | 0_{m \times {n-m}}]$, this claim is immediate.
By rotational symmetry, this shows that, for any vector space $\mathcal V$ and $v$ orthogonal to it, conditioning $z$ on $\mathcal V + v$ yields a Gaussian centered on $v$ with covariance determined by $\Pi_{\mathcal V} z$.
Then the lemma in the general case is implied by noting that $\{z: Dz = b\}$ can be decomposed as $\{z: Dz = 0 \} + D^+ b$.
\end{proof}

\begin{lemma}\label{lemma:condTrick}
Let $A \in \R^{n \times m}$ be a matrix with random Gaussian entries, $A_{ij} \sim \Gaus(0, \sigma^2)$.
Consider fixed matrices $Q \in \R^{m \times q}, \Zz \in \R^{n \times q}, P \in \R^{n \times p}, X \in \R^{m \times p}$.
Suppose there exists a solution in $A$ to the equations $\Zz = AQ$ and $X = A^\trsp P$.
Then the distribution of $A$ conditioned on $\Zz = AQ$ and $X = A^\trsp P$ is
\begin{align*}
    A &\disteq_{\Zz=AQ, X=A^\trsp P} E + \Pi_P^\perp \tilde A \Pi_Q^\perp
\end{align*}
where
\begin{align*}
    E
        &=
            \Zz Q^+
            + P^{+\trsp} X^\trsp
            - P^{+\trsp} P^\trsp
                YQ^+,
\end{align*}
$\tilde A$ is an iid copy of $A$,
and $\Pi_P^\perp = I - \Pi_P$ and $\Pi_Q^\perp = I - \Pi_Q$ in which $\Pi_P = PP^+$ and $\Pi_Q = QQ^+$ are the orthogonal projection to the space spanned by the column spaces of $P$ and $Q$ respectively.
\end{lemma}
\begin{proof}
We apply \cref{lemma:condTrickVec} to $D: A \mapsto (AQ, P^\trsp A)$.
The pseudoinverse of $D$ applied to $(\Zz, X^\trsp)$ can be formulated as the unique solution of
\begin{align*}
    \argmin_A \left\{ \|A\|^2_F : AQ = \Zz, P^\trsp A = X^\trsp \right\}
\end{align*}
where $\|-\|_F$ denotes Frobenius norm.
We check that $E$ is a 1) a solution to $AQ = \Zz, P^\trsp A = X^\trsp$ and 2) the minimal norm solution.

We have $EQ = 
        \Zz Q^+Q
            + P^{+\trsp} X^\trsp Q
            - P^{+\trsp} P^\trsp
                YQ^+Q$.
Note that $YQ^+Q = \Zz$ because $\Zz=AQ \implies YQ^+ Q = AQQ^+Q = AQ = \Zz$.
So $EQ = \Zz + P^{+T} (X^\trsp Q - P^\trsp \Zz)$.
But $X^\trsp Q = P^\trsp A Q = P^\trsp \Zz$, so $EQ = \Zz$ as desired.
A similar, but easier reasoning, gives $P^\trsp E = X^\trsp$.
This verifies that $E$ is a solution.

To check that $E$ is minimal norm, we show that it satisfies the stationarity of the Lagrangian
\begin{align*}
    L(A, \Theta, \Gamma)
        &=
            \|A\|^2_F + \la \Theta, \Zz - AQ \ra + \la \Gamma, X - A^\trsp P\ra.
\end{align*}
So $\pdf{L}{A} = 0 \implies 2A = \Theta Q^\trsp + P \Gamma^\trsp$ for some choices of $\Theta \in \R^{n \times q}$ and $\Gamma \in \R^{m \times p}$.
For $\Theta = 2 \Zz (Q^\trsp Q)^+$ and $\Gamma^\trsp = 2(P^\trsp P)^+ [ X^\trsp - P^\trsp \Zz Q^\trsp]$, we can check that
\begin{align*}
    \Theta Q^\trsp + P \Gamma^\trsp
        &=
            2 \Zz (Q^\trsp Q)^+ Q^\trsp + 2P(P^\trsp P)^+ [ X^\trsp - P^\trsp \Zz Q^+] \\
        &=
            2 \Zz Q^+ + 2 P^{+\trsp} X^\trsp - 2P^{+\trsp} P^\trsp \Zz Q^+\\
        &=
            2E
\end{align*}
as desired.
\end{proof}

\subsection{Hermite Polynomials}
We follow a presentation roughly given by \citet{odonnell_analysis_2014}.

\newcommand{\pHerm}{\mathrm{He}}
\newcommand{\Herm}{H}

\begin{defn}
Let $\pHerm_n(x)$ be the {\it probabilist's Hermite polynomial}, given by the generating function $e^{xt - \f 1 2 t^2} = \sum_{n=0}^\infty \pHerm_n(x) \f{t^n}{n!}.$
Let $L^2(\R; \Gaus(0, 1))$ be the space of square-integrable functions against the standard Gaussian measure, equipped with inner product $\la \phi, \psi \ra_G = \EV_{x \sim \Gaus(0, 1)} \phi(x) \psi(x)$ and norm $\|\phi\|_G^2 = \la \phi, \phi \ra_G$.
Let $\Herm_n(x) = \pHerm_n(x) / \|\pHerm_n\|_G$ be the normalized versions.
\end{defn}

\begin{fact}
$\{\pHerm_n(x)\}_{n \ge 0}$ form an orthogonal basis for $L^2(\R; \Gaus(0, 1))$ and $\{\Herm_n(x)\}_{n \ge 0}$ form an orthonormal basis for $L^2(\R; \Gaus(0, 1))$.
\end{fact}
\begin{fact}
$\|\pHerm_n\|_G^2 = n!$ so that $\Herm_n(x) = \pHerm_n(x)/\sqrt{n!}$.
\end{fact}

\begin{fact}\label{fact:hermiteInnerproduct}
  Let $\phi, \psi: \R \to \R$ be square integrable against $\Gaus(0, 1)$.
  Suppose we have the expansions in the orthonormal Hermite basis
  \begin{align*}
    \phi(x) = a_0 H_0(x) + a_{1} H_1(x) + \cdots,\quad
    \psi(x) = b_0 H_0(x) + b_{1} H_1(x) + \cdots.
  \end{align*}
  Let $(z_1, z_2) \sim \Gaus(0, C)$ where $C_{11} = C_{22} = 1$ and $C_{12} = \rho \in [-1, 1]$.
  Then we have the absolutely convergence series
  \begin{align*}
    \EV \phi(z_1)\psi(z_2) = a_0 b_0 + a_1 b_1 \rho + a_2 b_2 \rho^2 + \cdots.
  \end{align*}
\end{fact}

Suppose $u^1, \ldots, u^k$ are unit vectors in $\R^k$, and let $\lambda_{ij} \defeq \la u^i, u^j\ra$.
Construct a zero mean Gaussian vector $z = (z_1, \ldots, z_k)$ such that $\EV z_i z_j = \lambda_{ij}$.
Note that $z \disteq U g$ where $g = (g_1, \ldots, g_k)$ is a standard Gaussian vector and $U = (u^i_j)_{i,j=1}^k$ is the matrix with $u^i$ as rows.
Then for any $s = (s_1, \ldots, s_k)$ we can compute
\begin{align*}
    \EV \exp(\la s, z \ra)
        &=
            \EV \exp(s^\trsp U g)
        = 
            \EV \prod_i \exp(g_i (U^\trsp s)_i)
            \\
        &=
            \prod_i \EV \exp(g_i (U^\trsp s)_i)
        &
            \pushright{\text{by independence of $\{g_i\}_i$}}
            \\
        &=
            \prod_i \exp\left(\f 1 2 (U^\trsp s)_i^2\right)
        =
            \exp\left(\f 1 2 \sum_i (U^\trsp s)_i^2\right)
            \\
        &=
            \exp\left(\f 1 2 \|U^\trsp s\|^2\right)
        =
            \exp \lp \f 1 2 \sum_{i,j} \la u^i, u^j \ra s_i s_j \rp
            \\
        &=
            \exp \lp \f 1 2 \sum_{i,j} \lambda_{ij} s_i s_j \rp
            .
\end{align*}
Dividing by $\exp\lp \f 1 2 \sum_i s_i^2 \rp$, we obtain
\begin{align*}
    \EV \exp\left(\sum_i s_i z_i - s_i^2\right)
        &=
            \exp \lp \sum_{i< j} \lambda_{ij} s_i s_j \rp
            \\
    \EV \prod_i \sum_m \pHerm_m(z_i) (m!)^{-1} s_i^m
        &=
            \prod_{i< j} \sum_n (n!)^{-1}\lp \lambda_{ij} s_i s_j\rp^n
            \\
    \sum_{(m_i)_{i=1}^k}
        \prod_i \f{s_i^{m_i}}{m_i!} \EV \prod_i \pHerm_{m_i}(z_i)
        &=
            \sum_{(n_{(ij)})_{i<j}}
                \prod_i s_i^{\sum_{j \ne i} n_{(ij)}}
                \prod_{i<j} \f{\lambda_{ij}^{n_{(ij)}}}
                                {n_{(ij)}!}
\end{align*}
where $m_i \ge 0$ for all $i$, and $n_{(ij)} = n_{(ji)} \ge 0$ are indexed by unordered sets $\{i,j\}$.
Matching coefficients of $s$, we get
\begin{thm}
For any sequence $(m_i \ge 0)_{i=1}^k$,
\begin{align*}
    \EV \prod_i \pHerm_{m_i}(z_i)
        &=
            \lp \prod_r m_r! \rp 
            \lp \prod_{i<j} \f{\lambda_{ij}^{n_{(ij)}}}
                                {n_{(ij)}!}
                \rp
            \\
    \EV \prod_i \Herm_{m_i}(z_i)
        &=
            \lp \prod_r \sqrt{m_r!} \rp 
            \lp \prod_{i<j} \f{\lambda_{ij}^{n_{(ij)}}}
                                {n_{(ij)}!}
                \rp
\end{align*}
whenever there are $(n_{(ij)} \ge 0)_{i<j}$ such that, for all $i$, $m_i = \sum_{j\ne i} n_{(ij)}$.
$\EV \prod_i \pHerm_{m_i}(z_i) = 0$ otherwise.
\end{thm}

In particular, 
\begin{thm}\label{thm:multivariateHermiteExpectation}
If $\phi_i: \R \to \R$ has Hermite expansion $\phi_i(z) = \sum_{u=0}^\infty a_{iu} \Herm_u(z) = \sum_{u=0}^\infty b_{iu} \pHerm_u(z)$ where $b_{iu} = a_{iu}/\sqrt{u!}$, then
\begin{align*}
    \EV \prod_i \phi_i(z_i)
        &=
            \sum_{(n_{(ij)})_{i<j}}
                \lp \prod_r b_{r m_r} m_r! \rp 
                \lp \prod_{i<j} \f{\lambda_{ij}^{n_{(ij)}}}
                                    {n_{(ij)}!}
                    \rp
            \\
        &=
            \sum_{(n_{(ij)})_{i<j}}
                \lp \prod_r a_{r m_r} \sqrt{m_r!} \rp 
                \lp \prod_{i<j} \f{\lambda_{ij}^{n_{(ij)}}}
                                    {n_{(ij)}!}
                    \rp
            \\
        &=
            \sum_{(n_{(ij)})_{i<j}}
                \lp \prod_r a_{r m_r} \sqrt{\binom{m_i}{\{n_{(ij)}\}_{j \ne i}}} \rp 
                \lp \prod_{i<j} \lambda_{ij}^{n_{(ij)}}
                    \rp
\end{align*}
where $m_i = \sum_{j\ne i} n_{(ij)}$, whenever the RHS is absolutely convergent.
\end{thm}

\begin{lemma}\label{lemma:smallRhoMomentBound}
Suppose $\phi_i, i\in[k]$ are as in \cref{thm:multivariateHermiteExpectation}, with additionally the constraint that we have an index set $I \sbe [k]$ such that $b_{i0} = a_{i0} = 0$ (i.e. $\EV \phi_i(z_i) = 0$) for all $i \in I$.
Assume that, for some $\lambda < 1/2 $, $|\lambda_{ij}| \le \lambda/(k-1)$ for all $i \ne j$.
Then
\begin{align*}
    \left|\EV \prod_{i=1}^k \phi_i(z_i)\right| \le
    C_{k,|I|} \lp \prod_{r=1}^k \|\phi_r\|_G\rp 
        \lambda^{\lceil |I|/2\rceil}
\end{align*}
for some constant $C_{k, |I|}$ depending on $k$ and $|I|$ but independent of $\{\phi_i\}_i$ and $\lambda$.
\end{lemma}
\begin{proof}
In the notation of \cref{thm:multivariateHermiteExpectation}, $\binom{m_i}{\{n_{(ij)}\}_{j \ne i}} \le (k-1)^{m_i}$ by the multinomial theorem.
Thus
\begin{align*}
    \left|\EV \prod_{i=1}^k \phi_i(z_i)\right|
    &\le
        \sum_{\substack{(n_{(ij)})_{i<j}:\\\forall r\in I, m_r \ge 1}}
        \left|
        \lp \prod_{r=1}^k a_{r m_r} \sqrt{\binom{m_r}{\{n_{(rj)}\}_{j \ne r}}} \rp 
        \lp \prod_{i<j} \lambda_{ij}^{n_{(ij)}}
            \rp
        \right|
        \\
    &\le
        \sum_{\substack{(n_{(ij)})_{i<j}:\\\forall r \in I, m_r \ge 1}}
        \lp \prod_{r=1}^k \|\phi_r\|_G \sqrt{(k-1)^{m_r}} \rp 
        \lp \prod_{i<j} \lp \f \lambda{k-1} \rp^{n_{(ij)}}
            \rp
        \\
    &\le
        \sum_{\substack{(n_{(ij)})_{i<j}:\\\forall r \in I, m_r \ge 1}}
        \lp \prod_{r=1}^k \|\phi_r\|_G  \rp 
        \lambda^{\sum_{i<j} n_{(ij)}}
        \\
    &=
        \lp \prod_{r=1}^k \|\phi_r\|_G\rp 
        \lp B_{|I|} \lambda^{\lceil |I|/2\rceil} (1 + o(1)) \rp
        .
\end{align*}
where $B_V$ is the number of ways to cover $V$ vertices with $\lceil V/2 \rceil$ edges, and
$o(1)$ is a term that goes to 0 as $\lambda \to 0$ and is bounded above by a function of $k$ whenever $\lambda < 1/2$.
Then an appropriate $C_{k, |I|}$ can be chosen to obtain the desired result.
\end{proof}

\begin{lemma}\label{lemma:largeRhoMomentBound}
Suppose $\phi_i, i\in[k]$ are as in \cref{thm:multivariateHermiteExpectation}, with additionally  the constraint that, we have some index set $I \sbe [3, k]$ such that for all $i \in I$, $b_{i0} = a_{i0} = 0$ (i.e. $\EV \phi_i(z_i) = 0$).
Assume that $|\lambda_{12}| \le 1/2$, for some $\lambda < 1/\sqrt 8 $, $|\lambda_{ij}| \le \lambda/(k-1)$ for all $i \ne j$ and $\{i,j\} \ne \{1, 2\}$.
Then
\begin{align*}
    \left|\EV \prod_{i=1}^k \phi_i(z_i)\right| \le
    C'_{k, |I|} \lp \prod_{r=1}^k \|\phi_r\|_G\rp 
        \lambda^{\lceil |I|/2\rceil}
\end{align*}
for some constant $C'_k$ depending on $k$ and $I$ but independent of $\{\phi_i\}_i$ and $\lambda$.
\end{lemma}
\begin{proof}
\newcommand{\suchthat}{\mathrm{s.t.}}
Define $\mathcal P = \{(i,j): 1\ne i<j \ne 2\}$ and $\mathcal Q = \{(i,j): i<j \text{ and } (\text{$i=1$ XOR $j=2$})\}$.
Also write $R = \prod_{r=1}^k \|\phi_r\|_G.$
As in the above proof,
\begin{align*}
    &\phantomeq
      |\EV \prod_{i=1}^k \phi_i(z_i)|\\
    &\le
        \sum_{\substack{(n_{(ij)})_{i<j}:\\\forall r \in I, m_r \ge 1}}
        \left|
        \lp \prod_{r=1}^k a_{r m_r} \sqrt{\binom{m_r}{\{n_{(rj)}\}_{j \ne r}}} \rp 
        \lp \prod_{(i,j) \in \mathcal P} \lambda_{ij}^{n_{(ij)}}
            \rp
        2^{-n_{(12)}}
        \right|
        \\
    &\le
        \sum_{\substack{(n_{(ij)})_{i<j}:\\\forall r \in I, m_r \ge 1}}
        R
        \sqrt{\prod_{r=1}^2 \binom{m_r}{n_{(12)}}
                    \binom{m_r - n_{(12)}}{\{n_{(rj)}\}_{j \not \in \{1,2\}}}
            }
        \prod_{r=3}^k \sqrt{(k-1)^{m_r}}
        \lp \prod_{(i,j) \in \mathcal P} \lp \f \lambda{k-1} \rp^{n_{(ij)}}
            \rp
        2^{-n_{(12)}}
        \\
    &\le
        R
        \sum_{\substack{(n_{(ij)})_{i<j}:\\\forall r \in I, m_r \ge 1}}
        \sqrt{\prod_{r=1}^2 \binom{m_r}{n_{(12)}}
                    (k-1)^{m_r - n_{(12)}}
            }
        (k-1)^{\f 1 2 \sum_{r=3}^k m_r}
         \lp \f \lambda{k-1} \rp^{\sum_{i<j} n_{(ij)} - n_{(12)}}
        2^{-n_{(12)}}
        \\
    &=
        R
        \sum_{\substack{(n_{(ij)})_{i<j}:\\\forall r \in I, m_r \ge 1}}
        \sqrt{\prod_{r=1}^2 \binom{m_r}{n_{(12)}}
            }
         \lambda^{\sum_{i<j} n_{(ij)} - n_{(12)}}
         2^{-n_{(12)}}
        \\
    &\le
        R
        \sum_{\substack{(n_{(ij)})_{i<j}:\\\forall r \in I, m_r \ge 1}}
        \binom{\f{m_1+m_2}2}{n_{(12)}}
         2^{-n_{(12)}}
         \lambda^{\sum_{i<j} n_{(ij)} - n_{(12)}}
        \\
    &\le
        R
        \sum_{\substack{(n_{(ij)})_{i<j}:\\\forall r \in I, m_r \ge 1}}
        \binom{n_{(12)} + \f 1 2 m_{(12)}}{n_{(12)}}
         2^{-n_{(12)}}
         \lambda^{\sum_{i<j} n_{(ij)} - n_{(12)}}
        \\
    &\qquad {\text{where $m_{(12)} = \sum_{(i,j) \in \mathcal Q} n_{(ij)}$}}
        \\
    &\le
        2R
        \sum_{\substack{(n_{(ij)})_{(i,j) \in \mathcal P}:\\\forall r \in I, m_r \ge 1}}
        \lp \f 1 {1 - 1/2} \rp^{1 + \f 1 2 m_{(12)}}
         \lambda^{\sum_{(i,j) \in \mathcal P} n_{(ij)}}
        \\
    &\le
        2R
        \sum_{\substack{(n_{(ij)})_{(i,j) \in \mathcal P}:\\\forall r \in I, m_r \ge 1}}
         (\sqrt 2 \lambda)^{\sum_{(i,j) \in \mathcal Q} n_{(ij)}}
         \lambda^{\sum_{(i,j)\in \mathcal P \setminus \mathcal Q} n_{(ij)}}
        \\
    &\le
        2R
        \lp 2^{|I|/4} B_{|I|} \lambda^{\lceil |I|/2\rceil} (1 + o(1)) \rp
\end{align*}
where $B_{|I|}$ is the number of ways of covering $|I|$ vertices with $\lceil \f{|I|}2 \rceil$ edges, and
$o(1)$ is a term that goes to 0 as $\lambda \to 0$ and is upper bounded by a function of $k$ for all $\lambda < 1/\sqrt 8$.
Choosing the appropriate constant $C'_{k, |I|}$ then gives the result.
\end{proof}

\subsection{Bounding the Off-Diagonal Correlations of a Projection Matrix}

\begin{lemma}\label{lemma:projectionDiagonal}
Let $\Pi \in \R^{n \times n}$ be an orthogonal projection matrix.
Then each diagonal entry $\Pi_{ii} \in [0, 1].$
\end{lemma}
\begin{proof}
Because $\Pi = \Pi^2$, we have for each $i$, $\Pi_{ii} = \sum_{j} \Pi_{ij}^2 \implies \Pi_{ii} (1 - \Pi_{ii}) = \sum_{j \ne i} \Pi_{ij}^2 \ge 0 \implies \Pi_{ii} \in [0, 1].$
\end{proof}

\begin{lemma}\label{lemma:projectionCorrelation}

Let $\Pi \in \R^{n \times n}$ be an orthogonal projection matrix of rank $k$.
Suppose its diagonal entries are strictly positive.
Consider the correlation matrix $C \defeq D^{-1/2} \Pi D^{-1/2}$ where $D = \Diag(\Pi)$.
Then the off-diagonal entries of $C$ satisfy $\sum_{i < j} C_{ij}^2 \le 0.5 (r^2 + r),$ where $r = n-k$.
\end{lemma}
\noindent\textit{Proof.}
Because $\Pi = \Pi^2$, we have for each $i$, $\Pi_{ii} = \sum_{j} \Pi_{ij}^2 = \sum_j \Pi_{ii}\Pi_{jj} C_{ij}^2
\implies
1 - \Pi_{ii} = \sum_{j \ne i} \Pi_{jj} C_{ij}^2.$
At the same time, each $C_{ij}^2 \in [0, 1]$.
Thus we seek an upper bound on the following linear program in the $n(n-1)/2$ variables $C_{(ij)}^2$ (which identiy $C_{ij} = C_{ji} = C_{(ij)}$).
\begin{align*}
    \text{Maximize }&
        \sum_{i \ne j} C_{(ij)}^2
        \\
    \text{s.t. }&
        \forall i, 1 - \Pi_{ii} = \sum_{j \ne i} \Pi_{jj} C_{(ij)}^2
        \\
        &\forall i < j, C_{(ij)}^2 \in [0, 1].
\end{align*}

This LP has the dual
\begin{align*}
    \text{Minimize }&
        \sum_{i < j} \tau_{ij} + \sum_{i=1}^n (1 - \Pi_{ii}) \zeta_i\\
    \text{s.t. }&
        \forall i < j,
            \tau_{ij} + \zeta_i \Pi_{jj} + \zeta_j \Pi_{ii} \ge 1\\
        &\forall i < j,
            \tau_{ij} \ge 0
            \\
        &\forall i, \zeta_i \in \R.
\end{align*}

Any feasible value of the dual LP is an upper bound on the original LP.
We now set the dual variables.

We will set $\tau_{ij} = 1 - \zeta_i \Pi_{jj} - \zeta_j \Pi_{ii}$ for all $i < j$, so that the objective is now
\begin{align*}
  \binom n 2 - (k-1)\sum_{i=1}^n \zeta_i.
\end{align*}
It remains to 1) set the variables $\zeta_i$ and 2) verify that $1 - \zeta_i \Pi_{jj} + \zeta_j \Pi_{ii} \ge 0$ for all $i < j$.

\begin{figure}[h]
  \centering
  \includegraphics[width=0.7\textwidth]{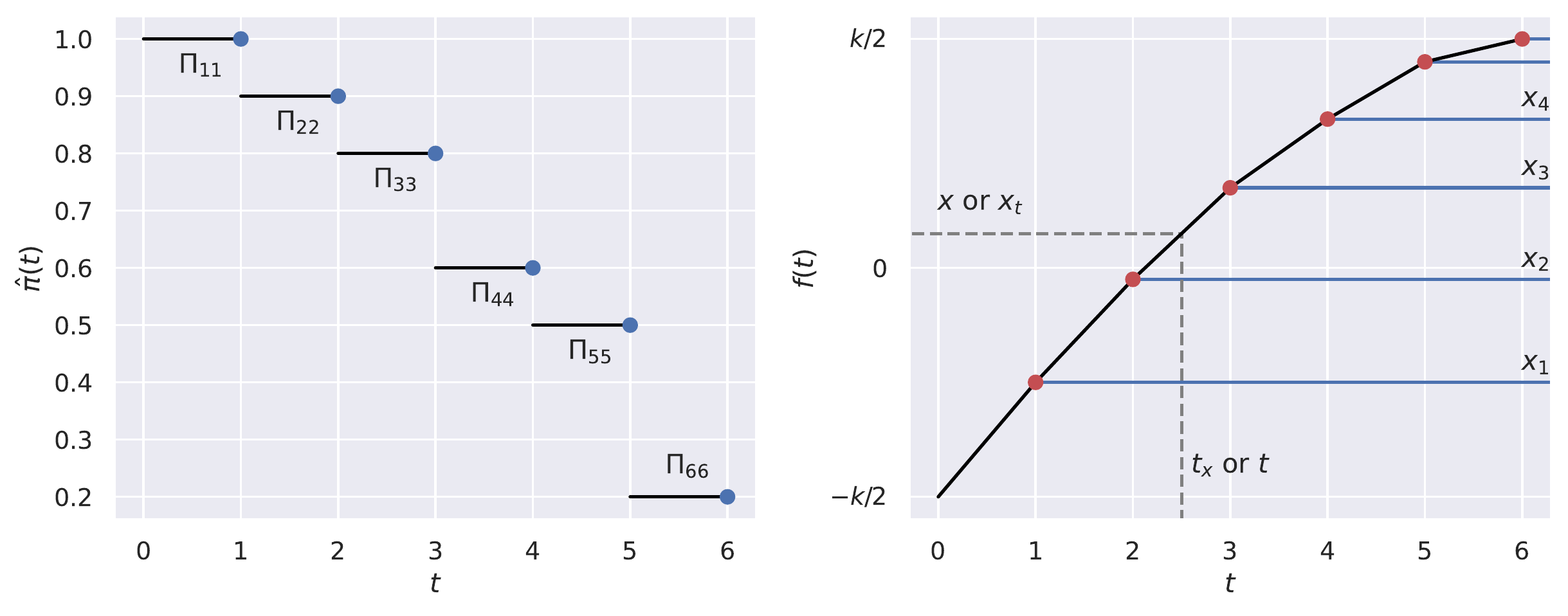}
  \caption{Illustration of $\hat \pi$, $x_t$, and $t_x$.}
  \label{fig:pihat}
\end{figure}

\global\long\def\zzeta{\hat{\zeta}}%

\global\long\def\ppi{\hat{\pi}}%
WLOG, assume $\Pi_{11}\ge\cdots\ge\Pi_{nn}>0$. Define the function $\pi:(0,n]\to\R$ as the piecewise constant extension of the diagonal of $\Pi$ to a real function on $(0,n]$:
\[
\pi(t)\defeq\Pi_{ii},\quad\text{where}\quad i=\lceil t\rceil.
\]
Note $\pi$ is nonincreasing and takes values in $(0,1]$. Define $f$ as the integral of $\pi$:
\[
f(s)\defeq\int_{0}^{s}\pi(t)\dd t.
\]
Note $f$ is increasing and concave because $\pi$ is positive and nonincreasing, so $f$ has an inverse $f^{-1}$. Additionally, $f(0)=0,f(n)=k$.

Let
\[
t_{x}\defeq f^{-1}(k/2+x),\quad x_{t}\defeq f(t)-k/2.
\]

Now define $\ppi:[-k/2,k/2]\to(0,1]$ by
\[
\ppi(x)=\pi(t_{x})=\pi(\lceil t_{x}\rceil),\quad\text{so that for all \ensuremath{t\in(0,n],}}\quad\ppi(x_{t})=\pi(t).
\]
\cref{fig:pihat} illustrates the above definitions.
Suppose 
\begin{equation}
\text{\ensuremath{\{x_{1},\ldots,x_{n}\}} and \ensuremath{\{-x_{1},\ldots,-x_{n}\}} are disjoint and do not contain 0.}\label{assm:grid}
\end{equation}

We can assume this WLOG as the set of $\Pi$ satisfying this property is dense and the function $\Pi\mapsto\sum C_{ij}^{2}$ is continuous so our bound applies to all $\Pi$ be continuity. This assumption implies that any line segment $[-\delta,+\delta]$ has at most one endpoint in $\{x_{1},\ldots,x_{n}\}$.

\begin{wrapfigure}{r}{0.5\textwidth}
  \centering
  \includegraphics[width=0.5\textwidth]{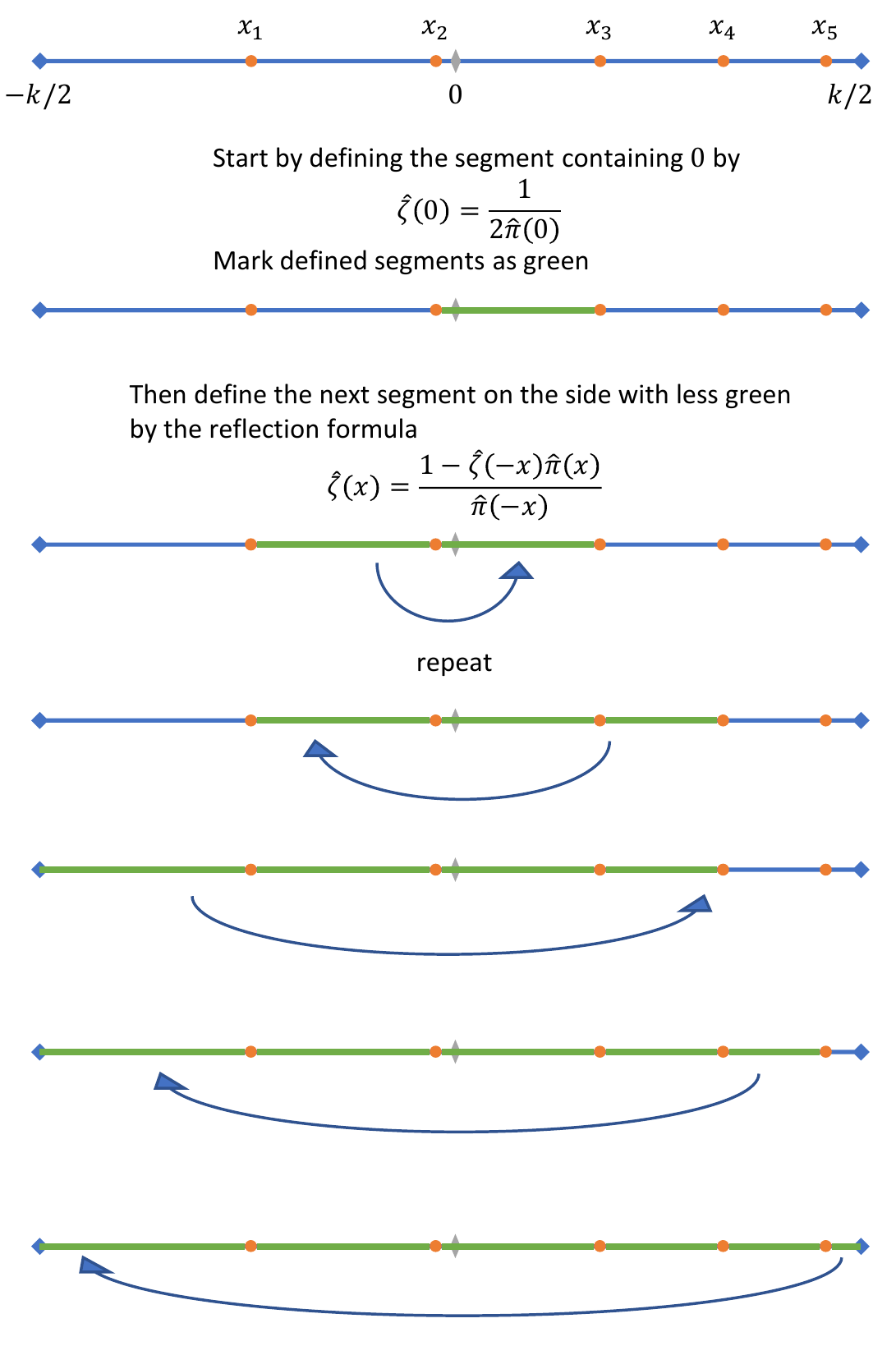}
  \caption{Illustration of how to define $\hat \zeta$.}
  \label{fig:zetahat}
\end{wrapfigure}
Now we define $\zzeta:[-k/2,k/2]\to\R$ as the unique function satisfying
\begin{equation}
\zzeta(x)\ppi(-x)+\zzeta(-x)\ppi(x)=1.\label{eq:reflection}
\end{equation}
and 
\begin{equation}
\zzeta(x)=\zzeta(y)\quad\text{if $\lceil t_{x}\rceil=\lceil t_{y}\rceil$}\label{eq:zzetapiecewise}
\end{equation}
Note \cref{eq:reflection} implies $\zzeta(0)=\frac{1}{2\ppi(0)}$ and
\begin{equation}
\frac{\zzeta(x)}{\ppi(x)}+\frac{\zzeta(-x)}{\ppi(-x)}=\frac{1}{\ppi(x)\ppi(-x)}.\label{eq:reflection2}
\end{equation}
By \cref{eq:zzetapiecewise}, $\zzeta$ is constant on each subinterval of $[-k/2,k/2]$ in the partition induced by $\{x_{1},\ldots,x_{n}\}$. A second of thought shows there is a unique $\zzeta$ satisfying \cref{eq:reflection} and \cref{eq:zzetapiecewise}; this is illustrated by by \cref{fig:zetahat}.

Thus we may set the dual variables
\[
\zeta_{i}\defeq\zzeta(x),\quad\text{where}\quad\lceil t_{x}\rceil=i.
\]
We check this is a valid assignment of dual variables (in that $\tau_{ij}\ge0$ are satisfied) in \cref{claim:tauconstraint}. 

Let $r=n-k$. It turns out, given \cref{claim:sumzeta} and \cref{claim:intInverseProd}, we have
\[
\sum_{i=1}^{n}\zeta_{i}\ge\frac{n+r}{2}
\]
so that the objective value given the above dual variables is
\[
\binom{n}{2}-\frac{n+r}{2}(n-r-1)=\frac{1}{2}(n^{2}-n)-\frac{1}{2}(n^{2}-r^{2}-n-r)=\frac{1}{2}(r^{2}+r)
\]
as desired.
\begin{claim}
\label{claim:sumzeta}
\[
\sum_{i=1}^{n}\zeta_{i}=\int_{-k/2}^{k/2}\frac{\zzeta(x)}{\ppi(x)}\dd x=\int_{0}^{k/2}\frac{\dd x}{\ppi(x)\ppi(-x)}.
\]
\end{claim}

\begin{proof}
The first equality follows from 
\[
\int_{x:\lceil t_{x}\rceil=i}\frac{\zzeta(x)}{\ppi(x)}\dd x=\int_{f(i-1)-k/2}^{f(i)-k/2}\frac{\zzeta(x)}{\ppi(x)}\dd x=\int_{f(i-1)}^{f(i)}\frac{\zeta_{i}}{\pi(i)}\dd x=(f(i)-f(i-1))\frac{\zeta_{i}}{\pi(i)}=\zeta_{i}.
\]

The second equality follows from folding the integral around $x=0$ and applying \cref{eq:reflection}.
\end{proof}
\begin{claim}
\label{claim:intInverseProd}With $r=n-k$,
\[
\int_{0}^{k/2}\frac{\dd x}{\ppi(x)\ppi(-x)}\ge\frac{n+r}{2}
\]
\end{claim}

\begin{proof}
Note $\int_{-k/2}^{k/2}\frac{\dd x}{\ppi(x)}=\sum_{i=1}^{n}1=n$ by same reasoning as in \cref{claim:sumzeta}. We also have $\ppi(x)\le1\implies\frac{1}{\ppi(x)}\ge1$ for all $x$. Thus $\int_{0}^{k/2}\frac{\dd x}{\ppi(x)\ppi(-x)}$ is at least
\[
\frac{1}{2}\inf\left\{ \int_{-k/2}^{k/2}g(x)g(-x)\dd x\mid g:[-k/2,k/2]\to[1,\infty),\int_{-k/2}^{k/2}g(x)\dd x=n,g\text{ nondecreasing}\right\} .
\]
It's not hard to see this infimum is achieved by $g^{*}(x)=1$ for all $x\in[-k/2,k/2)$ with a delta jump of mass $n-k=r$ at the point $k/2$. The value for this $g^{*}(x)$ is $\frac{1}{2}(k+2r)=\frac{n+r}{2}$.
\end{proof}
\begin{claim}
\label{claim:tauconstraint}For any $x,y\in[-k/2,k/2]$,
\[
\zzeta(x)\ppi(y)+\zzeta(y)\ppi(x)\le1.
\]
\end{claim}

\begin{proof}
We show the equivalent claim that
\[
\frac{\zzeta(x)}{\ppi(x)}+\frac{\zzeta(y)}{\ppi(y)}\le\frac{1}{\ppi(x)\ppi(y)}.
\]

Suppose $-x<y$, then by telescoping and \cref{claim:deltazeta},
\begin{align*}
\frac{\zzeta(x)}{\ppi(x)}+\frac{\zzeta(y)}{\ppi(y)} & =\frac{\zzeta(x)}{\ppi(x)}+\frac{\zzeta(-x)}{\ppi(-x)}+\sum_{i=\lceil t_{-x}\rceil}^{\lceil t_{y}\rceil}\left(\frac{\zeta_{i+1}}{\pi(i+1)}-\frac{\zeta_{i}}{\pi(i)}\right)\\
 & =\frac{1}{\ppi(x)\ppi(-x)}+\sum_{i=\lceil t_{-x}\rceil}^{\lceil t_{y}\rceil}\left(\frac{1}{\pi(i+1)}-\frac{1}{\pi(i)}\right)\frac{1}{\ppi(-x_{i})}.
\end{align*}
Note that because $x_{i}\ge-x$ for all $i$ in the sum, we have $-x_{i}\le x$ and $\frac{1}{\ppi(-x_{i})}\le\frac{1}{\ppi(x)}$. Therefore, by telescoping again,
\begin{align*}
\frac{\zzeta(x)}{\ppi(x)}+\frac{\zzeta(y)}{\ppi(y)} & \le\frac{1}{\ppi(x)\ppi(-x)}+\sum_{i=\lceil t_{-x}\rceil}^{\lceil t_{y}\rceil}\left(\frac{1}{\pi(i+1)}-\frac{1}{\pi(i)}\right)\frac{1}{\ppi(x)}\\
 & =\frac{1}{\pi(\lceil t_{y}\rceil)\ppi(x)}=\frac{1}{\ppi(y)\ppi(x)}
\end{align*}
as desired. 

Now suppose $y<-x$. Then by telescoping and \cref{claim:deltazeta},
\begin{align*}
\frac{\zzeta(x)}{\ppi(x)}+\frac{\zzeta(y)}{\ppi(y)} & =\frac{\zzeta(x)}{\ppi(x)}+\frac{\zzeta(-x)}{\ppi(-x)}-\sum_{i=\lceil t_{y}\rceil}^{\lceil t_{-x}\rceil}\left(\frac{\zeta_{i+1}}{\pi(i+1)}-\frac{\zeta_{i}}{\pi(i)}\right)\\
 & =\frac{1}{\ppi(x)\ppi(-x)}-\sum_{i=\lceil t_{y}\rceil}^{\lceil t_{-x}\rceil}\left(\frac{1}{\pi(i+1)}-\frac{1}{\pi(i)}\right)\frac{1}{\ppi(-x_{i})}.
\end{align*}
Note that because $x_{i}\le-x$ for all $i$ in the sum, we have $-x_{i}\ge x$ and $\frac{1}{\ppi(-x_{i})}\ge\frac{1}{\ppi(x)}$. Therefore, by telescoping again,
\begin{align*}
\frac{\zzeta(x)}{\ppi(x)}+\frac{\zzeta(y)}{\ppi(y)} & \le\frac{1}{\ppi(x)\ppi(-x)}-\sum_{i=\lceil t_{y}\rceil}^{\lceil t_{-x}\rceil}\left(\frac{1}{\pi(i+1)}-\frac{1}{\pi(i)}\right)\frac{1}{\ppi(x)}\\
 & =\frac{1}{\pi(\lceil t_{y}\rceil)\ppi(x)}=\frac{1}{\ppi(y)\ppi(x)}
\end{align*}
as desired.
\end{proof}
\begin{claim}
\label{claim:deltazeta}For any $i=1,\ldots n-1$, we have
\[
\frac{\zeta_{i+1}}{\pi(i+1)}-\frac{\zeta_{i}}{\pi(i)}=\frac{1}{\ppi(-x_{i})}\left(\frac{1}{\pi(i+1)}-\frac{1}{\pi(i)}\right).
\]
\end{claim}

\begin{proof}
By \cref{eq:reflection2}, for any $x\in\{x_{1},\ldots,x_{n-1}\}$, we have
\[
\frac{\zzeta(x+\epsilon)}{\ppi(x+\epsilon)}-\frac{\zzeta(x-\epsilon)}{\ppi(x-\epsilon)}=\frac{1}{\ppi(x+\epsilon)\ppi(-x)}-\frac{1}{\ppi(x-\epsilon)\ppi(-x)}\ge0
\]
for sufficiently small $\epsilon>0$. This is because, by \cref{assm:grid}, $\ppi(-x+\epsilon)=\ppi(-x-\epsilon)$ and $\zzeta(-x+\epsilon)=\zzeta(-x-\epsilon)$, and we obtain the above by subtracting \cref{eq:reflection2} for $x+\epsilon$ and $x-\epsilon$. In particular, letting $i=\lceil t_{x-\epsilon}\rceil$ so that $i+1=\lceil t_{x+\epsilon}\rceil$, we have the desired claim.
\end{proof}

\qed

\begin{rem}\label{lemma:projectionCorrelation}
Bobby He provided a much simpler argument to show that $\sum_{i < j} C_{ij}^2 \le 2 r^2 + 6r$ (which is slightly weaker than \cref{lemma:projectionCorrelation} but suffices for our downstream needs):

Let $J \defeq \{i: \Pi_{ii} \ge 1/2\}$ and $\bar J = [n] \setminus J$.
Then because $\tr \Pi_{ii} = n - r$, $|\bar J| \le 2 r$.
We split the squared sum of off-diagonal entries into
\begin{align*}
  \sum_{i\ne j} C_{ij}^2
  &=
    \sum_{\substack{i\ne j\\ i,j \in J}} C_{ij}^2
    + 2\sum_{\substack{i \in J \\ j \in \bar J}} C_{ij}^2
    + \sum_{\substack{i\ne j\\ i, j \in \bar J}} C_{ij}^2
    .
\end{align*}
We bound each of the terms separately.
\begin{align*}
  \sum_{\substack{i\ne j\\ i,j \in J}} C_{ij}^2
  \le 4 \sum_{\substack{i\ne j\\ i,j \in J}} \Pi_{ij}^2
  \le 4 \sum_{\substack{i\ne j}} \Pi_{ij}^2
  = 4 (\tr \Pi ^ 2 - \sum_{i} \Pi_{ii}^2)
  \le 4 (k - n (k/n)^2)
  = 4(n - k)k/n \le 4r.
\end{align*}
\begin{align*}
  2\sum_{\substack{i \in J \\ j \in \bar J}} C_{ij}^2
  = 2\sum_{\substack{i \in J \\ j \in \bar J}} \Pi_{ij}^2/\Pi_{ii} \Pi_{jj}
  \le 4\sum_{\substack{i \in J \\ j \in \bar J}} \Pi_{ij}^2/  \Pi_{jj}
  = 4\sum_{\substack{j \in \bar J}} \Pi_{jj}/  \Pi_{jj}
  = 4 |\bar J|
  \le 8 r.
\end{align*}
Above, we used the fact that $\Pi^2 = \Pi \implies \Pi_{jj} + \sum_{i} \Pi_{ij}^2$.
Finally, because $|C_{ij}| \le 1$, we have
\begin{align*}
  \sum_{\substack{i\ne j\\ i, j \in \bar J}} C_{ij}^2
  \le |\bar J|^2 \le 4r^2.
\end{align*}
Summing the above and dividing by two yields the desired result.
\end{rem}

\subsection{Law of Large Numbers for Images of Weakly Correlated Gaussians}

\begin{thm}\label{thm:projectVariance}
  Let $z \sim \Gaus(0, \Pi)$ where $\Pi \in \R^{n \times n}$ is a matrix with nonzero diagonal entries and
  satisfying $\sum_{i<j} \Pi_{ij}^2 / (\Pi_{ii} \Pi_{jj}) \le R$
  for some constant $R$.
  Consider functions $\phi_i: \R \to \R, i\in[n],$ with finite variance $\Var\lp \phi_i(x) : x \sim \Gaus(0, \Pi_{ii})\rp < \infty$.
  Then
  \begin{align*}
      \Var\lp\sum_{i=1}^n \phi_i(z_i)\rp
          &\le
              \lp R\sqrt 2 + 1 \rp \sum_i \Var\lp \phi_i(x) : x \sim \Gaus(0, \Pi_{ii})\rp.
  \end{align*}
\end{thm}

Note, by \cref{lemma:projectionCorrelation}, if $\Pi$ is an orthogonal projection matrix of rank $k$, then we may take $R = 0.5 ((n - k)^2 + (n-k))$.
By \cref{thm:projectVariance} and Chebyshev's inequality, we have the following.
\begin{cor}[Weak Law of Large Numbers for Images of Weakly Correlated Gaussians]
  \label{thm:WeakLLNGaussianImage} Consider a triangular array $\{\zeta_{1}^{n},\ldots,\zeta_{n}^{n}\}_{n\ge1}$ of Gaussian variables, where each row is given by $\zeta^{n}\sim\Gaus(0,\Sigma^{n})$ and the covariance matrix $\Sigma^{n}$ has nonzero diagonal entries and satisfies
  \[\sum_{\alpha<\beta}(\Sigma_{\alpha\beta}^{n})^{2} / (\Sigma_{\alpha\alpha}^n \Sigma_{\beta\beta}^n)\le R(n)\]
  for some $R(n) > 0$.
  Let $\phi^n_\alpha:\R\to\R, \alpha\in[n], n \ge1,$ be functions with mean $\mu_\alpha^n \defeq \EV_{\zeta_\alpha^n} \phi^n_\alpha(\zeta^n_\alpha)$.
  Then, as $n\to\infty$, we have the following convergence in probability
  \begin{align*}
    \f 1 n \sum_{\alpha=1}^n (\phi^n_\alpha(\zeta^n_\alpha) - \mu^n_\alpha) \probto 0
  \end{align*}
  if the following holds
  \begin{align*}
    \f{R(n) \sqrt 2 + 1}{n^2}\sum_{\alpha=1}^n \Var_{\zeta^n_\alpha}(\phi^n_\alpha(\zeta^n_\alpha)) \to 0.
  \end{align*}
\end{cor}
Note that, if the variance $\Var_{\zeta^n_\alpha}(\phi^n_\alpha(\zeta^n_\alpha))$ is uniformly bounded by some $T>0$, then $R(n)$ can grow like $n^{1-\varepsilon}$ and we still have convergence in probability.

\begin{proof}[Proof of \cref{thm:projectVariance}]
Let $C = D^{-1/2} \Pi D^{-1/2}, D = \Diag(\Pi),$ be the correlation matrix of $\Pi$.
The premise of the theorem implies $\sum_{i<j} C_{ij}^2 \le R$.
Define functions $\psi_i$ by $\psi_i(y) = \phi_i(\sqrt{\Pi_{ii}}y) - \EV_{x \sim \Gaus(0, 1)}[\phi_i(\sqrt{\Pi_{ii}}x)]$.
Then
\begin{align*}
\Var_{z \sim \Gaus(0, \Pi)}\lp \sum_{i=1}^n \phi_i(z_i) \rp
    &=
        \EV_{z \sim \Gaus(0, C)}
            \lp \sum_{i=1}^n \psi_i(z_i) \rp^2
    =
        \EV_{z \sim \Gaus(0, C)}
            \sum_{i,j} \psi_i(z_i) \psi_j(z_j).
\end{align*}
Expand $\psi_i$ in the Hermite orthonormal basis,
\begin{align*}
    \psi_i(x) = a_{i1} H_1(x) + a_{i2} H_2(x) + \cdots
\end{align*}
where $H_j(x)$ is the $j$th Hermite polynomial, normalized so that $\EV_{z \sim \Gaus(0, 1)} H_j(z)^2 = 1$ (note that $H_0(x) = 1$ and does not appear here because $\EV_{x \sim \Gaus(0, 1)}\psi_i(x) = 0$ by construction).
For any locally integrable $\phi: \R \to \R$, let $\|\phi\|_G^2 \defeq \EV_{z \sim \Gaus(0, 1)} \phi(z)^2$, so that $\|\psi_i\|_G^2 = \sum_k a_{ik}^2 = \Var\lp \phi_i(x) : x \sim \Gaus(0, \Pi_{ii})\rp.$
Then,
\begin{align*}
    \sum_{i < j} \EV_{z \sim \Gaus(0, C)} \psi_i(z_i) \psi_j(z_j)
        &=
            \sum_{i<j} \sum_{k=1}^\infty a_{ik} a_{jk} C_{ij}^k
            \\
        &\le
            \sum_{k=1}^\infty \sqrt{\lp \sum_{i<j} a_{ik}^2 a_{jk}^2\rp
                                    \lp \sum_{i<j} C_{ij}^{2k}\rp}
            \\
        &\le
            \sum_{k=1}^\infty \sqrt{\f 1 2 \lp \sum_{i} a_{ik}^2\rp^2
                                    \lp \sum_{i<j} C_{ij}^2\rp}
        &
            \pushright{\text{since $|C_{ij}| \le 1$}}
            \\
        &\le
            2^{-1/2}\sum_{k=1}^\infty \lp \sum_{i} a_{ik}^2\rp R
        &
            \pushright{\text{by premise}}
            \\
        &=
            \f R {\sqrt 2} \sum_i \|\psi_i\|_G^2
\end{align*}
On the other hand, $\sum_i \EV_{x \sim \Gaus(0, 1)} \psi_i(x)^2 = \sum_i \|\psi\|_G^2$, so that
\begin{align*}
    \sum_{i,j}\EV_{z \sim \Gaus(0, C)} \psi_i(z_i) \psi_j(z_j)
        &\le
            \lp R \sqrt 2 + 1 \rp \sum_i \|\psi_i\|_G^2
            \\
        &=
            \lp R \sqrt 2 + 1 \rp \sum_i \Var\lp \phi_i(x) : x \sim \Gaus(0, \Pi_{ii})\rp.
\end{align*}

\end{proof}

\begin{thm}\label{thm:controlHighMoments}
\newcommand{\LL}{2p}
\newcommand{\CC}{\mathcal{C}}
Let $z \sim \Gaus(0, \Pi)$ where $\Pi \in \R^{n \times n}$ is a matrix with nonzero diagonal entries and
satisfying $\sum_{i<j} \Pi_{ij}^2 / (\Pi_{ii} \Pi_{jj}) \le R$
for some constant $R$ independent of $n$.
Consider functions $\phi_i: \R \to \R$ for each $i \in [n]$ with mean $\mu_i \defeq \EV_{z_i} \phi_i(z_i)$.
Suppose each $\phi_i$ has finite ($\LL$)th centered moment $\EV_{z_i} (\phi_i(z_i) - \mu_i)^{\LL}$, where $p \ge 6$.
Then for $Q \defeq \f 1 n \sum_{i=1}^n \phi_i(z_i),$ %
\begin{align*}
    \EV[(Q - \EV Q)^{2p}]
        &\le
           \CC
            n^{-1.5} \max_{i \in [n]}
                \EV
                    \left(\phi_{i}(z_i) - \mu_i \right)^{2p}
\end{align*}
for some constant $\CC$ depending on $p$ and $R$, but not on $n$ or the functions $\phi_i$.
If in addition, each $\phi_i$ has finite centered moments of order $2p L$ for some $L > 1$, then
\begin{align*}
    \EV[(Q - \EV Q)^{2p}]
        &\le
           \CC
            n^{-1.5+1/L}
                \sqrt[L]{\f 1 n \sum_{i=1}^n 
                    \EV
                    \left(\phi_{i}(z_i) - \mu_i \right)^{2p L}
                    }
            .
\end{align*}
\end{thm}

Note, by \cref{lemma:projectionCorrelation}, an orthogonal projection matrix of rank $n - O(1)$ satisfies the off-diagonal condition of \cref{thm:controlHighMoments}.
Combining \cref{thm:controlHighMoments} and \cref{lemma:momentBoundASConvergence}, we obtain the following.
\begin{cor}[Strong Law of Large Numbers for Images of Weakly Correlated Gaussians]
  \label{thm:StrongLLNGaussianImage} Consider a triangular array $\{\zeta_{1}^{n},\ldots,\zeta_{n}^{n}\}_{n\ge1}$ of Gaussian variables, where each row is given by $\zeta^{n}\sim\Gaus(0,\Sigma^{n})$ and the covariance matrix $\Sigma^{n}$ has nonzero diagonal entries and satisfies
  \[\sum_{\alpha<\beta}(\Sigma_{\alpha\beta}^{n})^{2} / (\Sigma_{\alpha\alpha}^n \Sigma_{\beta\beta}^n)\le R\]
  for some $R > 0$ independent of $n$.
  Let $\phi^n_\alpha:\R\to\R, \alpha\in[n], n \ge1,$ be functions with mean $\mu_\alpha^n \defeq \EV_{\zeta_\alpha^n} \phi^n_\alpha(\zeta^n_\alpha)$.
  Then, as $n\to\infty$,
  \begin{align*}
    \f 1 n \sum_{\alpha=1}^n (\phi^n_\alpha(\zeta^n_\alpha) - \mu^n_\alpha) \asto 0
  \end{align*}
  if one of the following condition holds:
  \begin{itemize}
    \item For some $p \ge 6$ and some $S$ independent of $n$,
    \[\EV_{\zeta^n_\alpha} (\phi^n_\alpha(\zeta^n_\alpha) - \mu^n_\alpha)^{2p} \le S.\]
    \item For some $q > 12$ and some $S$ independent of $n$,
    \[\f 1 n \sum_{\alpha=1}^n \EV_{\zeta^n_\alpha} (\phi^n_\alpha(\zeta^n_\alpha) - \mu^n_\alpha)^{2q} < S.\]
  \end{itemize}

 \end{cor}

\begin{proof}[Proof of \cref{thm:controlHighMoments}]
\newcommand{\Cijt}[3]{C_{\p{(#1 #2)}{#3}}}
Define the correlation matrix $C = D^{-1/2} \Pi D^{-1/2}, D = \Diag(\Pi)$.
The premise of the theorem implies $\sum_{i<j} C_{ij}^2 \le R$.
Let $\psi_i(y) = \phi_i(\sqrt{\Pi_{ii}}y) - \EV_{x \sim \Gaus(0, 1)}[\phi_i(\sqrt{\Pi_{ii}}x)]$ be the scaled, centered version of $\phi_i$.

Order the off-diagonal entries of the correlation matrix in the order of decreasing squared value:
$$C_{(ij)^{(1)}}^2 \ge C_{(ij)^{(2)}}^2 \ge \ldots \ge C_{(ij)^{(N)}}^2,$$
where $N = \binom n 2$, and $(ij)^{(t)} = (\p i t \p j t)$ are unordered pairs of distinct indices $\p i t \ne \p j t$.
Since $\sum_t C_{\p{(ij)}{t}}^2 \le R$, by \cref{lemma:projectionCorrelation}, we deduce that
\begin{equation}
    |\Cijt i j t| \le n^{-1/4} \qquad \text{for all $t > R \sqrt n$}.
    \label{eqn:CijtBound}
\end{equation}

Consider the $(2p)$th centered moment 
\begin{equation*}
\EV \lp \f 1 n \sum_{i=1}^n \psi_i(y_i)\rp^{2p}
= n^{-2p} \EV \sum_{\sigma: [2p] \to [n]} \prod_{a=1}^{2p} \psi_{\sigma(a)}(y_{\sigma(a)}),
\end{equation*}
where $y \sim \Gaus(0, C)$.
We shall bound the sum to show that this moment is not too large.

First note the naive bound via AM-GM,
\begin{align*}
    \EV \left|\prod_{a=1}^{2p} \psi_{\sigma(a)}(y_{\sigma(a)})\right|
    &\le
        \EV \f 1 {2p} \sum_{a=1}^{2p} \psi_{\sigma(a)}(y_{\sigma(a)})^{2p}
    \le
        \max_{i \in [n]} \EV_{y \sim \Gaus(0, 1)}[\psi_i(y)^{2p}]
        \\
    &=
        \max_{i \in [n]} \EV[(\phi_{i}(z_i) - \EV \phi_i(z_i))^{2p}: 
                        z_i \sim \Gaus(0, \Pi_{ii})]
    \defeq
        B_{2p}.
        \numberthis
        \label{eqn:naiveAMGM}
\end{align*}
Now, for any collection of numbers $\{x_i \in \R \}_{i=1}^m$ and any $L > 0$, we have the trivial bound $\max_i |x_i| \le \lp \sum_{j=1}^m |x_j|^L \rp^{1/L}$, and this bound is tighter the larger $L$ is.
Thus
\begin{equation*}
B_{2p} \le n^{1/L} \sqrt[L]{\f 1 n \sum_{i=1}^n (\EV[\psi_i(y)^{2p}])^L} \le n^{1/L} B_{2p,L},
\end{equation*}
where $B_{2p,L} \defeq \sqrt[L]{\f 1 n \sum_{i=1}^n \EV[\psi_i(y)^{2pL}]}$, for any $L$.

\newcommand{\CC}{\mathcal{C}}

We can categorize the $n^{2p}$ terms of 
\begin{equation}
\sum_{\sigma: [2p] \to [n]} \EV\prod_{a=1}^{2p} \psi_{\sigma(a)}(y_{\sigma(a)})
\label{eqn:momentSum}
\end{equation}
as follows.
\begin{itemize}
    \item Suppose $\sigma$ is injective.
    \begin{itemize}
        \item
            Suppose for each $a\ne b$, $(\sigma(a)\sigma(b)) = \p{(ij)}{t}$ for some $t > R\sqrt n$, so that $|C_{\sigma(a)\sigma(b)}| \le n ^{-1/4}$ by \cref{eqn:CijtBound}.
            By \cref{lemma:smallRhoMomentBound}, 
            \[\EV \prod_{a=1}^{2p} \psi_{\sigma(a)}(y_{\sigma(a)}) \le \CC_1 \lp \prod_{r=1}^{2p} \|\psi_{\sigma(r)}\|_G\rp \lp n^{-1/4} \rp ^{p}\]
            for some constant $\CC_1$ dependent on $p$ but \emph{not} on $R$, $\{\psi_r\}_r$, or $n$.
            Thus the contribution of all such $\sigma$ to the sum \cref{eqn:momentSum} is
            \begin{align*}
                \sum_{\text{all such }\sigma} \CC_1 \lp \prod_{r=1}^{2p} \|\psi_{\sigma(r)}\|_G\rp n^{-p/4}
                    &\le
                        \sum_{\text{all }\sigma} \CC_1 \lp \prod_{r=1}^{2p} \|\psi_{\sigma(r)}\|_G\rp n^{-p/4} 
                        \\
                    &\le 
                        \CC_1 n^{-p/4} \lp \sum_{i=1}^{n} \|\psi_{\sigma(r)}\|_G \rp^{2p}
                        \\
                    &=
                        \CC_1 n^{1.75 p} B'_{2p}
            \end{align*}
            where we have set
            \[B'_{2p} \defeq \lp \f 1 n \sum_{i=1}^{n} \|\psi_{i}\|_G \rp^{2p}.
            \]
        \item
            Suppose for some $a, b \in [2p]$, $(\sigma(a)\sigma(b)) = \p{(ij)}t$ for $t \le R\sqrt n$.
            There are at most $2\binom{2p}2 R \sqrt n \cdot n^{2p-2} \le \CC_2 n^{2p-1.5}$ such $\sigma$, for some $\CC_2$ depending on $p$ and $R$ but not $n$ (or $\{\psi_r\}_r$).
            Indeed, there are $R \sqrt n$ of choosing such a $t$, $2\binom {2p} 2$ ways of choosing their preimages under $\sigma$ out of $2p$, and $\le n^{2p-2}$ ways of choosing the rest of the values of $\sigma$.
            By \cref{eqn:naiveAMGM}, the contribution of all such $\sigma$ to the sum is at most $\CC_2 n^{2p-1.5} B_{2p}.$
            
    \end{itemize}
    \item
        Suppose for some $a^* \ne b^*$ in $[2p]$, $\sigma(a^*) = \sigma(b^*)$, but $\sigma|_{[n] \setminus \{a^*, b^*\}}$ is injective and takes range outside $\{\sigma(a^*)\}$.
        There are $\binom {2p} 2 n \binom{n-1}{2p-2} \le \CC_3 n^{2p-1}$ such $\sigma$, where $\CC_3$ depends only on $p$ (but not on $R$, $\{\psi_r\}_r$, and $n$).
        \begin{itemize}
            \item 
                Suppose for each $a\ne b$, $(\sigma(a)\sigma(b)) = \p{(ij)}{t}$ for some $t > R\sqrt n$, so that $|C_{\sigma(a)\sigma(b)}| \le n^{-1/4}.$
                We apply \cref{lemma:smallRhoMomentBound} to the $2p-1$ functions $\{\psi_{\sigma(a^*)}^2\} \cup \{\psi_{\sigma(a)}\}_{a \not \in \{a^*, b^*\}}$, with $\psi_{\sigma(a^*)}^2$ being the sole function whose expectation is not 0, so that the $I$ of \cref{lemma:smallRhoMomentBound} has size $2p-2$, and the $\lambda$ of \cref{lemma:smallRhoMomentBound} is $(2p-2)n^{-1/4}.$
                Then \cref{lemma:smallRhoMomentBound} gives
                \begin{align*}
                    \EV \prod_{a=1}^{2p} \psi_{\sigma(a)}(z_{\sigma(a)})
                        &\le
                            \CC_4
                            \|\psi^2_{\sigma(a^*)}\|_G
                            \lp \prod_{a\not\in\{a^*, b^*\}} \|\psi_{\sigma(a)}\|_G \rp
                            (n^{-1/4})^{(2p-2)/2}
                            \\
                        &=
                            \CC_4
                            \|\psi^2_{\sigma(a^*)}\|_G
                            \lp \prod_{a\not\in\{a^*, b^*\}} \|\psi_{\sigma(a)}\|_G \rp
                            n^{-(p-1)/4}
                \end{align*}
                for some constant $\CC_4$ depending on $p$ but \emph{not} on $n$, $R$, or $\{\psi_r\}_r$.
                Thus the collective contribution of such $\sigma$ to the sum is at most
                \begin{align*}
                    &\phantomeq
                        \sum_{\text{all such }\sigma} \CC_4
                            \|\psi^2_{\sigma(a^*)}\|_G
                            \lp \prod_{a\not\in\{a^*, b^*\}} \|\psi_{\sigma(a)}\|_G \rp
                            n^{-(p-1)/4}
                        \\
                    &\le
                        \CC_4 n^{-(p-1)/4} \binom p 2
                        \sum_{i=1}^n \|\psi^2_i\|_G
                            \sum_{\pi: [2p-2] \to [n] \setminus \{i\}}
                                \prod_{a=1}^{2p-2}
                                \|\psi_{\pi(a)}\|_G
                        \\
                    &\le
                        \CC_4 n^{-(p-1)/4} \binom p 2
                        \sum_{i=1}^n \|\psi^2_i\|_G
                            \sum_{\pi: [2p-2] \to [n]}
                                \prod_{a=1}^{2p-2}
                                \|\psi_{\pi(a)}\|_G
                        \\
                    &=
                        \CC_5
                        n^{-(p-1)/4}
                        \lp \sum_{i=1}^n \|\psi_i^2\|_G\rp
                        \lp \sum_{i=1}^n \|\psi_i\|_G\rp^{2p-2}
                        \\
                    &=
                        \CC_5 n^{1.75(p-1)} 
                        B''_{2p}
                        ,
                \end{align*}
                where $\CC_5 = \CC_4 \binom p 2$ and depends only on $p$, but \emph{not} on $n$, $R$, or $\{\psi_r\}_r$, and where we have set
                \[
                    B''_{2p}
                        =
                            \lp \f 1 n \sum_{i=1}^n \|\psi_i^2\|_G\rp
                            \lp \f 1 n \sum_{i=1}^n \|\psi_i\|_G\rp^{2p-2}.
                \]
            \item
                 Suppose for some $a, b \in [2p]$, $(\sigma(a)\sigma(b)) = \p{(ij)}t$ for $t \le R\sqrt n$.
                 There are at most $\binom{2p}{2} n \cdot R\sqrt n \cdot \binom{n-2}{2p-3} = \CC_6 n^{2p-1.5}$ such $\sigma$, where $\CC_6$ depends only on $p$ and $R$, but \emph{not} on $n$ or $\{\psi_r\}_r$.
                 Using \cref{eqn:naiveAMGM} again, we can upper bound the contribution of such $\sigma$ by $\CC_6 n^{2p-1.5} B_{2p}.$
        \end{itemize}
        \item Otherwise, there are more than one pair of inputs that collide under $\sigma$. 
        There are at most $\CC_7 n^{2p-2}$ such $\sigma$, where $\CC_7$ depends only on $p$, but \emph{not} on $n$, $R$, or $\{\psi_r\}_r$.
        Using \cref{eqn:naiveAMGM}, we upper bound their contributions by $\CC_7 n^{2p-2} B_{2p}$.
\end{itemize}
To summarize, using $O(-)$ to hide the constants $\CC_*$ which don't depend on $n$ or the functions $\psi_i$:
\begin{align*}
    \EV \sum_{\sigma: [2p] \to [n]} \prod_{a=1}^{2p} \psi_{\sigma(a)}(y_{\sigma(a)})
    &\le
        O\lp
        n^{1.75p} B'_{2p}
        + n^{1.75(p-1)} B''_{2p}
        + n^{2p-1.5} B_{2p}
        \rp
        \\
    &\le
        O\lp
        n^{1.75p} B'_{2p}
        + n^{1.75(p-1)} B''_{2p}
        + n^{2p-1.5+1/L} B_{2p, L}
        \rp
        \\
    \EV \lp \f 1 n \sum_{i=1}^n \phi_i(z_i) - \mu_i)\rp^{2p}
        &\le
           O\lp
            n^{-0.25 p} B'_{2p}
            + n^{-0.25 p - 1.75} B''_{2p}
            + n^{-1.5} B_{2p}
            \rp
            \\
        &\le
           O\lp
            n^{-0.25 p} B'_{2p}
            + n^{-0.25 p - 1.75} B''_{2p}
            + n^{-1.5+1/L} B_{2p, L}
            \rp
\end{align*}

By the power mean inequality, we get that $B'_{2p}, B''_{2p} \le B_{2p} \le n^{1/L} B_{2p,L}$.
Substitution then gives the desired result.

\end{proof}

This theorem can be significantly strengthened with more careful case work and applying more involved versions of \cref{lemma:smallRhoMomentBound,lemma:largeRhoMomentBound}, but we will not be concerned with this here.

\section{Proof of \netsort{} Master Theorem}
\label{sec:proofMainTheorem}
\renewcommand{\nu}{\eta}
\newcommand{\rankext}[1]{{\color{green}#1}}

\newcommand{\coreset}{{\mathcal{M}}}
\newcommand{\basespace}{\mathsf{U}}
\newcommand{\MM}{m}

In this section, we will give the proof for our main theorem.

\paragraph{A Bit of Notation and Terminology}
Note that, for each $n$, the randomness of our program specified by \cref{thm:NetsorTMasterTheorem} comes from the sampling of the initial vectors $\mathcal V$ and matrices $\mathcal W$.
Let $\basespace$ be the product space obtained from multiplying together the corresponding probability space for each $n$.
Each sample from this product probability space thus correspond to a sequence $\{S(n)\}_n$ of instantiatiations of $\mathcal V$ and $\mathcal W$.
Below, when we say ``almost surely'' (often abbreviated ``a.s.''), we mean ``almost surely over the probability of $\basespace{}$.''
We will also often make statements of the form 
\begin{equation*}
\text{\emph{almost surely (or, a.s.), for all large $n$, \quad $\mathcal A(n)$ is true}}
\end{equation*}
where $\mathcal A(n)$ is a claim parametrized by $n$.
This means that for all but a $\basespace{}$-probability-zero set of sequences $\{S(n)\}_n$, $\mathcal A(n)$ is true for large enough $n$.
Note that the order of the qualifiers is very important here.

Let $g^1, \ldots, g^M$ be all of the G-vars (including those of $\mathcal V$) in the program.
It suffices to prove \cref{thm:NetsorTMasterTheorem} where $k=M$ and the vector $\{h^i\}_i$ are the G-vars $g^1, \ldots, g^M$, since all other vectors are \refNonlin{} images of them.
For convenience, we also assume that for all $g\in \mathcal V$, we have $\EV Z^g = 0$; this is without loss of generality (WLOG) since any nonzero mean can be absorbed into applications of \refNonlin{}.

\paragraph{We induct, but on what?}
A natural way of going about proving \cref{thm:NetsorTMasterTheorem} is by inducting on the number of variables in a program.
While this would be the main backbone of the proof, it turns out such a proof would require us to assume a certain rank condition (\cref{lemma:rankStability}) on $g^1, \ldots, g^M$ which is not so easy to check.
Thus, it would be more fruitful to perform a simultaneous induction on our claim (\ref{IH:MomConv}) along with another statement, parametrized by $\MM$, that would prove such a condition for us.
\begin{description}
\item[Moments\label{IH:MomConv}]\!\!\!$(\MM)$\ \ \ 
    For any polynomially-bounded $\psi: \R^\MM \to \R$, as $n \to \infty$,
\begin{align*}
    \f 1 n \sum_{\alpha=1}^n \psi(g^1_\alpha, \ldots, g^\MM_\alpha) &\asto 
    \EV
      \psi\left(
      \Zz^{g^1}, \ldots, \Zz^{g^\MM} \right)
      .
\end{align*}

\item[CoreSet\label{IH:coreSet}]\!\!\!$(\MM)$\ \ \ 
    There exists a ``core set'' $\coreset \sbe [\MM]$ such that, 
  \begin{description}
    \item[Basis\label{prop:basis}]\!\!\!$(\MM)$\ \ \ 
      almost surely, for large enough $n$, for every $i \in [\MM]$, there exist \emph{unique} constants (not depending on $n$) $\{a_j\}_{j \in \coreset}$ such that $g^i = \sum_{j \in \coreset} a_j g^j$.
      Note the uniqueness implies that $\{g^i\}_{i \in \coreset}$ is linearly independent.
    \item[Density\label{prop:density}]\!\!\!$(\MM)$\ \ \ 
      The distribution of the random vector
      \[\{\Zz^{g^j}\}_{j \in \coreset} \in \R^{\coreset}\]
      is absolutely continuous w.r.t.\ the Lebesgue measure in $\R^{\coreset}$, and vice versa.%
      \footnote{
        {Compared to the proof of the Master Theorem in \citet{yangTP2}, this is a new inductive hypothesis.
        This is trivially true in the NTK case since the distribution in question is Gaussian with full support.}}
      In other words, for every measurable set $U \sbe \R^{\coreset}$,
      \[\Pr[\{\Zz^{g^j}\}_{j \in \coreset} \in U] = 0 \iff \text{$U$ has Lebesgue measure 0.}\]
    \item[NullAvoid\label{prop:nullAvoid}]\!\!\!$(\MM)$\ \ \ 
      for every triangular array of Lesbegue measure zero sets $\{U_{n\alpha} \in \R^{\coreset} \}_{n \in \N, \alpha \in [n]}$, almost surely for all large enough $n$, for all $\alpha \in [n]$, we have
      \[\{g^i_\alpha\}_{i \in \coreset} \not \in U_{n\alpha}.\]
      In other words, the values $\{g^i_\alpha\}_{\alpha \in \coreset}$ of the core set ``avoid'' Lebesgue measure zero sets asymptotically.
      Intuitively, this says that the distribution of these values are not singular.
      (Note the LHS depends on $n$ although we are suppressing it notationally)
  \end{description}
\end{description}

Let us explain in brief why we need to consider \ref{IH:coreSet} satisfying \ref{prop:basis} and \ref{prop:nullAvoid}.
\begin{itemize}
\item
    \ref{prop:basis} reduces the consideration of \ref{IH:MomConv} to only the core set G-vars, since every other G-var is asymptotically a linear combination of them.
\item
    When we apply the Gaussian conditioning technique \cref{prop:GaussianCondition}, we need to reason about the pseudo-inverse $\Lambda^+$ of some covariance matrix $\Lambda$.
    Each entry of $\Lambda$ is of the form $\f 1 n \sum_{\alpha=1}^n \phi_i(g^1_\alpha, \ldots, g^{\MM-1}_\alpha) \phi_j(g^1_\alpha, \ldots, g^{\MM-1}_\alpha)$ for a collection of polynomially bounded scalar functions $\{\phi_i\}_i$.
    This $\Lambda$ will be a random variable which converges a.s.\ to a determinstic limit $\mathring \Lambda$  as $n \to \infty$.
    It should be generically true that $\Lambda^+ \asto \mathring \Lambda^+$ as well, which is essential to make the Gaussian conditioning argument go through.
    But in general, this is guaranteed only if $\Lambda$'s rank doesn't drop suddenly in the $n \to \infty$ limit.
    We thus need to guard against the possibility that $g^1, \ldots, g^\MM$, in the limit, suddenly concentrate on a small set on which $\{\phi_i(g^1, \ldots, g^\MM)\}_i$ are linearly dependent.
    This is where \ref{prop:nullAvoid} comes in.
    It tells us that $g^1, \ldots, g^\MM$ will avoid any such small set asymptotically, so that indeed the rank of $\Lambda$ will not drop in the limit.
\end{itemize}

\paragraph{Proof organization}

We will show that \ref{IH:MomConv} and \ref{IH:coreSet} are true for initial G-vars in $\mathcal V$, as the base case, and
\begin{equation*}
\text{\ref{IH:MomConv}}(\MM-1) \text{ and } \text{\ref{IH:coreSet}}(\MM-1) \implies \text{\ref{IH:MomConv}}(\MM) \text{ and } \text{\ref{IH:coreSet}}(\MM)
\end{equation*}
as the inductive step.
By induction, we obtain \ref{IH:MomConv}$(M)$, which is \cref{thm:NetsorTMasterTheorem}.

The base cases are easy and we will dispatch with them immediately after this in \cref{sec:basecases}, but the inductive step is much more complicated, and we will need to set up notation in \cref{sec:inductiveSetup}.
During this setup, we prove some basic limit theorems using the induction hypothesis.
However, the full generality of these claims requires some consequences of \ref{IH:coreSet}, which we call ``rank stability'' and ``zero stability.''
These notions are introduced and proved in \cref{sec:rankStabilityZeroStability}.

We would then finally be able to handle the inductive steps at this point.
We first prove
\begin{equation*}
\text{\ref{IH:MomConv}$(\MM-1)$ and \ref{IH:coreSet}$(\MM-1)$} \implies \text{\ref{IH:coreSet}}(\MM)
\end{equation*}
in \cref{sec:inductiveCoreSet} because it is easier.
Then we prove
\begin{equation*}
\text{\ref{IH:MomConv}$(\MM-1)$ and \ref{IH:coreSet}$(\MM-1)$} \implies \text{\ref{IH:MomConv}}(\MM)
\end{equation*}
in \cref{sec:inductiveMoments}.

Before we proceed with the induction, let us set up establish some matrix notations that will make the proof significantly easier to express.

\subsection{Preliminaries: Square-Integrable Random Variables and Matrix Notation}

\newcommand{\LsqZ}{\mathcal{L}}

\begin{defn}
\label{defn:LsqZ}
Consider the space $\LsqZ$ of square-integrable random variables.
This space has inner product given by $(X, Y) \mapsto \EV XY$ for any $X, Y \in \LsqZ$.
We will often use matrix notation in \emph{square brackets} to express certain sum of (inner) products between elements of $\LsqZ$ and real scalars in $\R$.
For example, for $X^1, \ldots, X^k, Y^1, \ldots, Y^k, X, Y \in \LsqZ$ and $a_1, \ldots, a_k, a \in \R$, we write
\begin{align*}
[X^1, \ldots, X^k]
\begin{bmatrix}
Y^1\\
\vdots\\
Y^k
\end{bmatrix}
  &=
    \sum_{i=1}^k \EV X^i Y^i
      \in \R
    &
[Y^1, \ldots, Y^k]
\begin{bmatrix}
a_1\\
\vdots\\
a_k
\end{bmatrix}
  &=
    \sum_{i=1}^k a_i Y^i \in \LsqZ
    \\
\begin{bmatrix}
X^1\\
\vdots\\
X^k
\end{bmatrix}
[Y]
  &=
    \begin{bmatrix}
    \EV X^1 Y\\
    \vdots\\
    \EV X^k Y
    \end{bmatrix}
      \in \R^{k}
    &
\begin{bmatrix}
X^1\\
\vdots\\
X^k
\end{bmatrix}
[a]
  &=
    \begin{bmatrix}
    aX^1\\
    \vdots\\
    aX^k
    \end{bmatrix}
    \in \LsqZ^k.
\end{align*}
In addition,
\begin{align*}
  \begin{bmatrix}
  X^1\\
  \vdots\\
  X^k
  \end{bmatrix}
  [Y^1, \ldots, Y^k]
  =
  \begin{bmatrix}
    \EV X^1 Y^1 & \cdots & \EV X^1 Y^k\\
    \vdots & \ddots & \vdots\\
    \EV X^k Y^1 & \cdots & \EV X^k Y^k
  \end{bmatrix} \in \R^{k\times k},
\end{align*}
but note that this ``outer product'' is not a rank-1 matrix, but rather full rank typically.
In general, a matrix in $\LsqZ^{k \times l}$ multiplied by a matrix $\LsqZ^{l \times m}$ on the right produces a matrix $\R^{k \times m}$, where the elementwise product is provided by the inner product of $\LsqZ$;
a matrix in $\LsqZ^{k \times l}$ multiplied by a matrix $\R^{l \times m}$ on the right produces a matrix $\LsqZ^{k \times m}$, where the elementwise product is provided by the scalar multiplication of $\LsqZ$.
In general, this mixed-type matrix multiplication is not associative, i.e. $A(BC) \ne (AB)C$.
We will always read a series of matrix multiplication from the right: $ABCD = A(B(CD))$.

\end{defn}

\newcommand{\phz}{\phantom{0}}
\newcommand{\sqmat}[1]{
    \begin{bmatrix}
    \phz
      &\phz
        &\phz\\
    \phz
      &#1
        &\phz\\
    \phz
      &\phz
        &\phz
    \end{bmatrix}}
\newcommand{\colmat}[1]{
    \begin{bmatrix}
    \phz\\
    #1\\
    \phz
    \end{bmatrix}}

\begin{fact}\label{fact:LsqProjection}
Just like in a finite-dimensional inner product space, given a finite collection $\Ss = \{X^i\}_{i=1}^k \sbe \LsqZ$, the orthogonal projection operator $\Pi_\Ss$ to the span of $\Ss$ (inside $\LsqZ$) is given by
\[\Pi_\Ss Y \defeq \sum_{i=1}^k a_i X^i,
\]
for any $Y \in \LsqZ$, where
\begin{align*}
a &= \Lambda^+ b \in \R^k,\quad
b_j = \EV X^j Y,\ b \in \R^k,\quad
\Lambda_{ij} = \EV X^i X^j,\ \Lambda \in \R^{k \times k}.
\end{align*}
In the matrix notation of \cref{defn:LsqZ}, this can be expressed as
\begin{align*}
\Pi_\Ss Y
  &=
    [X^1, \ldots, X^k]
    \begin{bmatrix}
    \phz
      &\phz
        &\phz\\
    \phz
      &\Lambda^+
        &\phz\\
    \phz
      &\phz
        &\phz
    \end{bmatrix}
    \begin{bmatrix}
    X^1\\
    \vdots\\
    X^k
    \end{bmatrix}
    [Y]
    \\
  &=
    [X^1, \ldots, X^k]
    \left(
      \begin{bmatrix}
      X^1\\
      \vdots\\
      X^k
      \end{bmatrix}
      [X^1, \ldots, X^k]
    \right)^+
    \begin{bmatrix}
      X^1\\
      \vdots\\
      X^k
    \end{bmatrix}
    [Y]
    .
    \numberthis\label{eqn:matrixformProjection}
\end{align*}
We denote the the projection to the orthogonal complement of the linear span of $\Ss$ by
\[\Pi_\Ss^\perp \defeq I - \Pi_\Ss.\]

\end{fact}

\subsection{Preliminaries: Adjunction Relation Between \texorpdfstring{$\Zhat$}{Z hat} and \texorpdfstring{$\Zdot$}{Z check}}
A sort of ``adjunction'' holds between $\Zhat$ and $\Zdot$, as we describe below.
This is a crucial calculation needed to prove that the ``Onsager correction term'' $\Zdot$ is right.
\begin{lemma}\label{lemma:DtrxAdjunction}
For any vectors $x, h \in \R^n$ and matrix $W\in \R^{n\times n}$ in the program, we have%
\footnote{
  In the variable dimension case, if $x \in \R^m, h \in \R^n, W \in \R^{m \times n}$, and $n/m \to \rho$, then we have
  $\rho \EV \Zhat^{W^\trsp x} Z^h = \EV Z^{x} \Zdot^{Wh}.$
}
\begin{align*}
  \EV \Zhat^{W^\trsp x} Z^h = \EV Z^{x} \Zdot^{Wh}.
\end{align*}
\end{lemma}
Compare with the identity
\[\EV Z^{W^\trsp x} Z^h = \EV Z^x Z^{Wh}\]
which, assuming \cref{thm:NetsorTMasterTheorem}, would follow from $(W^\trsp x)^\trsp h = x^\trsp (Wh)$.
\begin{proof}
  \newcommand{\xx}{x}
  \newcommand{\Ee}{\mathcal{E}}
  \newcommand{\uu}{u}
  \newcommand{\vv}{v}
  \newcommand{\Ff}{\mathcal{F}}
  Fix $g=Wh$ and we will prove, for all $u = W^\trsp v$ in the program,
  \begin{align*}
    \EV \Zhat^{u} Z^h = \EV Z^{v} \Zdot^g.
  \end{align*}
  Let $\Gg$ be the set of G-vars introduced before $g$ (so $g\not\in\Gg$).
  Note that by \cref{prop:hatZRandomSource}, $Z^h$ is a deterministic function of $\Zhat^\Gg \defeq \{\Zhat^{x}: x \in \Gg\}$.
  Let $\Ee$ be a subset of $\Gg$ such that $\Zhat^{\Ee} \defeq \{\Zhat^{x}: x \in \Ee\}$ has a nonsingular covariance matrix and generates the same $\sigma$-algebra as $\Zhat^\Gg$.
  Then, because $\Zhat^\Gg$ is jointly Gaussian, $\Zhat^\Gg$ is a linear function of $\Zhat^\Ee$, and $Z^h$ is a deterministic function of $\Zhat^\Ee$.

  Suppose $\uu^1, \ldots, \uu^k$ are all G-vars in $\Gg$ introduced via \refMatMul{} with $W^\trsp$
  \[\uu^i = W^\trsp \vv^i \]
  for H-vars $\vv^1, \ldots, \vv^k$.
  Assume WLOG that for $\ell \le k$, we have $\{\uu^1, \ldots, \uu^{\ell}\} = \{\uu^1, \ldots, \uu^{k}\} \cap \Ee$, so that $\Zhat^{\uu^{1}}, \ldots, \Zhat^{\uu^{{\ell}}}$ also have full rank covariance.
  Note that any $\Zhat^{\uu^{j}}, j \in [k]$, has to be a linear combination of $\{\Zhat^{\uu^i}\}_{i=1}^{\ell}$.

  Any G-var $\bar g$ where $\bar g = W^\trsp \bar h$ for some $\bar h$ must be one of $u^i, i\in[k],$ or $\bar g \not\in\Ee$.
  We will divide up the proof of \cref{lemma:DtrxAdjunction} into two cases.

  \emph{Case 1.} For any $i \in [k]$, we have $\EV \Zhat^{\uu^{i}} Z^{h} = \EV Z^{\vv^{i}} \Zdot^{g}$.

  \emph{Case 2.} For any G-var $\bar g \not \in \Ee$ such that $\bar g = W^\trsp \bar h$ for some $\bar h$, we have $\EV  \Zhat^{\bar g} Z^{h}  = \EV   Z^{\bar h} \Zdot^{g} $.

  \emph{Proof of Case 1.}
  By treating $Z^h$ as a deterministic function of $\Zhat^\Ee$, \cref{lemma:EXf} says there are coefficients $\{a_{\xx} \in \R: \xx \in \Ee\}$ such that
  \begin{align*}
  \EV \Zhat^{\uu^i} Z^h
    &=
      \sum_{\xx \in \Ee} a_{\xx} \Cov(\Zhat^{\uu^i}, \Zhat^\xx)
      .
  \end{align*}
  However, if $\xx \in \Ee\setminus \{\uu^1, \ldots, \uu^k\}$ (i.e.\ $\xx \ne W^\trsp y$ for some $y$), then $\Cov(\Zhat^{\uu^{i}}, \Zhat^\xx) = 0$ by \refZhat{} for all $i=1,\ldots,k$.
  Thus the sum above over $x\in\Ee$ reduces to a sum over $x\in \Ee \cap \{\uu^1, \ldots, \uu^k\} = \{\uu^1, \ldots, \uu^\ell\}$:
  \begin{align*}
  \EV  \Zhat^{\uu^i} Z^{h} 
    &=
      \sum_{j=1}^{\ell} a_{\uu^j} \Cov(\Zhat^{\uu^{i}}, \Zhat^{\uu^{j}})
    =
      \sum_{j=1}^{\ell} a_{\uu^j} \sigma_W^2
        \EV  Z^{\vv^i} Z^{\vv^j} 
    =
      \EV
        Z^{\vv^i}
        \sum_{j=1}^{\ell} a_{\uu^j} \sigma_W^2
          Z^{\vv^j}
      .
  \end{align*}
  Note that this argument works for any $\uu^i$ with $i \le \ell$, and the constants $\{a_{\uu^j}\}_{j=1}^{\ell}$ \emph{remain the same} for all $\uu^i$:
  \begin{equation*}
  \forall i \in [\ell],\quad
  \EV  \Zhat^{\uu^i} Z^{h} 
    =
      \EV
        Z^{\vv^i}
        \sum_{j=1}^{\ell} a_{\uu^j} \sigma_W^2
          Z^{\vv^j}
      .
  \end{equation*}

  By the observation above that any $\Zhat^{\uu^{j}}, j \in [k],$ is a linear combination of $\{\Zhat^{\uu^i}\}_{i=1}^{\ell}$, this equality is also extended to all $i \in [k]$:
  \begin{equation}
  \forall i \in [k],\quad
  \EV  \Zhat^{\uu^i} Z^{h} 
    =
      \EV
        Z^{\vv^i}
        \sum_{j=1}^{\ell} a_{\uu^j} \sigma_W^2
          Z^{\vv^j}
      .
      \label{eqn:partialAdjunction}
  \end{equation}

  Set $Y \defeq \sum_{j=1}^{\ell} a_{\uu^j} \sigma_W^2
  Z^{\vv^j}$.
  To prove our claim for case 1, it suffices to show that $\Zdot^{g} = Y$.

  Recall that in \cref{rem:ExpectationPartialDer}, the matrix $C\in \R^{k \times k}$ and $b \in \R^k$, in the matrix notation of \cref{defn:LsqZ}, are given by
  \begin{align*}
  C
    &=
      \begin{bmatrix}
      Z^{\vv^1}\\
      \vdots\\
      Z^{\vv^k}
      \end{bmatrix}
      [Z^{\vv^1}, \ldots,  Z^{\vv^{k}}]
      ,\quad
  b
    =
      \begin{bmatrix}
      \Zhat^{\uu^1}\\
      \vdots\\
      \Zhat^{\uu^k}
      \end{bmatrix}
      [Z^{h}]
      .
  \end{align*}
  By \cref{eqn:partialAdjunction}, we can also re-express $b$ as the following product of a $k \times 1$ vector and a $1 \times 1$ vector,
  \begin{equation*}
  b
    =
      \begin{bmatrix}
      Z^{\vv^1}\\
      \vdots\\
      Z^{\vv^k}
      \end{bmatrix}
      \left[ Y
      \right]
      .
  \end{equation*}
  Thus, from \cref{rem:ExpectationPartialDer},
  \begin{align*}
  \Zdot^{g}
    &=
      [Z^{\vv^1}, \ldots,  Z^{\vv^k}]
      \sqmat{C^+}
      \colmat{b}
    =
      [Z^{\vv^1}, \ldots,  Z^{\vv^k}]
      \sqmat{C^+}
      \begin{bmatrix}
      Z^{\vv^1}\\
      \vdots\\
      Z^{\vv^k}
      \end{bmatrix}
      [Y]
    =
      \Pi Y
  \end{align*}
  where $\Pi$ is the orthogonal projection in $\LsqZ$ to the subspace spanned by $\{ Z^{\vv^1}, \ldots,  Z^{\vv^k}\}$, as discussed in \cref{fact:LsqProjection}.
  Since $Y$ is already in this subspace, we just get
  \begin{align*}
  \Zdot^{g}
    &=
      Y
  \end{align*}
  as desired.

  \emph{Proof of Case 2.}
  Let $U$ denote the random column vector $(Z^{\uu^1}, \ldots, Z^{\uu^k})^\trsp$.
  Then, conditioned on $U$, the random variable $Z^h$ has randomness coming only from $\{\Zhat^\xx: \xx \in \Ee \setminus \{\uu^1, \ldots, \uu^k\}\}$.
  So conditioned on $U$, $Z^h$ is independent from $\Zhat^{\bar g}$, by \refZhat{}.
  Therefore,
  \begin{align}
    \EV \Zhat^{\bar g} Z^h
    &=
      \EV_{U}\lp
      \EV [ Z^h \mid U]
      \EV [ \Zhat^{\bar g} \mid U]
      \rp
      .  \label{eqn:E1}
  \end{align}
  We now will massage the RHS into the desired form.
  Let $C \in \R^{k\times k}, C_{ij} = \Cov(\Zhat^{u^i}, \Zhat^{u^j})$ be the covariance matrix of $\Zhat^{u^i}$, and let $b \in \R^k, b_i = \Cov(\Zhat^{\bar g}, \Zhat^{u^i})$.
  Then by standard Gaussian conditioning formula (\cref{prop:GaussianCondition}), we have
  \begin{align*}
    \EV [ \Zhat^{\bar g} \mid U] = b^\trsp C^+ U,
  \end{align*}
  as a linear function of $U$.
  Plugging into \cref{eqn:E1}, in the matrix notation of \cref{defn:LsqZ}, we have
  \begin{align*}
    \EV \Zhat^{\bar g} Z^h = b^\trsp C^+ 
    \begin{bmatrix}
    \Zhat^{\uu^1}\\
    \vdots\\
    \Zhat^{\uu^k}
    \end{bmatrix}
    [Z^{h}]
    .
  \end{align*}
  By Case 1, we have $\EV \Zhat^{\uu^{i}} Z^{h} = \EV Z^{\vv^{i}} \Zdot^{g}$, so we can write
  \begin{align*}
    \EV \Zhat^{\bar g} Z^h = b^\trsp C^+ 
    \begin{bmatrix}
    \Zhat^{\vv^1}\\
    \vdots\\
    \Zhat^{\vv^k}
    \end{bmatrix}
    [\Zdot^{g}]
    = \EV \lp b^\trsp C^+ V \rp \Zdot^g
    ,
  \end{align*}
  where $V$ is the random column vector $(Z^{v^1}, \ldots, Z^{v^k})^\trsp$.
  Now, because $C_{ij} = \sigma_W^2 \EV Z^{v^i} Z^{v^j}$ and $b_i = \sigma_W^2 \EV Z^{\bar h} Z^{v^i}$, we see
  \[b^\trsp C^+ V
  = \Pi Z^{\bar h},\]
  where $\Pi$ is the orthogonal projection to the subspace of $\LsqZ$ spanned by $\{Z^{v^1}, \ldots, Z^{v^k}\}$.
  Putting this all together, we have, by the self-adjoint property of $\Pi$,
  \begin{align*}
    \EV \Zhat^{\bar g} Z^h = \EV (\Pi Z^{\bar h}) \Zdot^g = \EV Z^{\bar h} (\Pi \Zdot^g).
  \end{align*}
  However, since $\Zdot^g$ is already a linear combination of $\{Z^{v^1}, \ldots, Z^{v^k}\}$ by definition (\refZdot), we have $\Pi \Zdot^g = \Zdot^g$.
  Thus we obtain
  \begin{align*}
    \EV \Zhat^{\bar g} Z^h = \EV Z^{\bar h} \Zdot^g,
  \end{align*}
  as desired.
\end{proof}

\begin{lemma}\label{lemma:conjugatePairDtrxh}
  Consider a G-var $g = Wh$ in our program.
  Suppose $g^i = W^\trsp h^i, i=1,\ldots,\ell$, includes all G-vars of the form $W^\trsp x$ introduced before $h$.
  If we set $ Z^{H}$ be the column vector $[Z^{h^1},
  \ldots,
  Z^{h^\ell}]^\trsp$ and define $C \in\R^{\ell\times \ell}, b \in \R^\ell$ by
  \[C_{ij} \defeq \EV Z^{h^i} Z^{h^j},\quad b_i \defeq \EV \Zhat^{g^i} Z^h,\]
  then, using the matrix notation of \cref{defn:LsqZ}, we have
  \begin{align}
  \Zdot^{g} &= 
     Z^{H}{}^\trsp C^+ b =
  [Z^{h^1},
  \ldots,
   Z^{h^\ell}]
  \begin{bmatrix}
  \phz
    &\phz
      &\phz\\
  \phz
    &C^+
      &\phz\\
  \phz
    &\phz
      &\phz
  \end{bmatrix}
  \begin{bmatrix}
  \phz\\
  b\\
  \phz
  \end{bmatrix}
  \label{eqn:expandedZdotSum}
  .
  \end{align}
\end{lemma}
Note, by assumption, $\ell$ here is at least $k$ in \cref{rem:ExpectationPartialDer}.
If $\{g^i\}_i$ are exactly the G-vars of the form $W^\trsp x$ introduced before $h$, then \cref{eqn:expandedZdotSum} is equivalent to \cref{eq:ExpectationPartialDer} in \cref{rem:ExpectationPartialDer}.
So this lemma says this expression always evaluates to $\Zdot^g$ as long as $\{g^i\}_i$ is no smaller.
\begin{proof}
  By \cref{lemma:DtrxAdjunction}, we can re-express
  \[b_i = \EV   Z^{h^i} \Zdot^{g}.\]
  Then by \cref{fact:LsqProjection},
  \begin{align*}
  Z^{H\trsp } C^+ b
    &=
      Z^{H}{}^\trsp C^+ \EV Z^{H} \Zdot^{g}
    =
      \Pi_H \Zdot^{g}
  \end{align*}
  where $\Pi_H$ is the orthogonal projection operator to the span of $\{ Z^{h^1},
  \ldots,
  Z^{h^\ell}\}$; see \cref{fact:LsqProjection}.
  But by \refZdot{}, $\Zdot^{g}$ is already a linear combination of $\{Z^{h^i}: g^i\text{ introduced before $h$}\}$.
  Thus
  \[ \Pi_H \Zdot^{g} = \Zdot^{g}\]
  as desired.
\end{proof}

\subsection{Base Cases: \ref{IH:MomConv} and \ref{IH:coreSet} for Initial Vectors}
\label{sec:basecases}

\paragraph{Base case: \ref{IH:MomConv}($\mathcal V$)}
Since the vectors in $\mathcal V$ are sampled iid coordinatewise, \ref{IH:MomConv}($\mathcal V$) just follows from the strong law of large numbers.

\paragraph{Base Case: \ref{IH:coreSet}($\mathcal V$)}
Pick the core set $\coreset$ to be (the indices of) any subset $\mathcal U \sbe \mathcal V$ such that $Z^{\mathcal U}$ forms a linear basis of $Z^{\mathcal V}$.
Then it's straightforward to verify \ref{prop:basis}, \ref{prop:density}, and \ref{prop:nullAvoid}.

\subsection{Inductive Case: Setup}
\label{sec:inductiveSetup}

We now assume \ref{IH:MomConv}$(\MM-1)$ and \ref{IH:coreSet}$(\MM-1)$ and want to reason about $g^\MM$ to show \ref{IH:MomConv}$(\MM)$ and \ref{IH:coreSet}$(\MM)$.
In this section, we will set up the notation and helper lemmas toward this goal. 
We need to (unfortunately) introduce a large number of symbols.
To alleviate possible confusion, we provide an index of all such symbols in \cref{sec:indexSymbols}.

\subsubsection{Definitions}
Suppose
\begin{align}
    g^{\MM} = A h \quad \text{and} \quad h = \phi(g^{1}, \ldots, g^{\MM - 1})
    \label{eqn:gAh}
\end{align}
for some polynomially-bounded $\phi$.
For brevity, we will just write $g = g^{\MM}$.
Consider all previous instances where $A$ or $A^\trsp$ is used: 
\begin{equation}
  \xme^{i} = A \yme^{i}, i = 1, \ldots, r, \quad\text{and}\quad
  \ume^{j} = A^\trsp \vme^{j}, j = 1, \ldots, s.
  \label{eqn:xAy&uATv}
\end{equation}
Define the matrices $\Xme \in \R^{n\times r}, \Ume \in \R^{n \times s}, \Yme \in \R^{n \times r}, \Vme \in \R^{n \times s}$ with $\xme^i, \ume^i, \yme^i, \vme^i$ as columns
\begin{align}
\Xme &\defeq [\xme^1| \ldots| \xme^{r}],&
\Ume &\defeq [\ume^1| \ldots| \ume^{s}],&
\Yme &\defeq [\yme^1| \ldots| \yme^r],&
\Vme &\defeq [\vme^1| \ldots| \vme^s]
.
\label{eqn:XUYZ}
\end{align}
Let $\Bb$ be the $\sigma$-algebra spanned by all previous G-vars $g^1, \ldots, g^{\MM-1}$; since all previous vectors are deterministic images of these G-vars, $\Bb$ is also the $\sigma$-algebra generated by all previous vectors before $g$.
Conditioning on $\Bb$, $A$ is linearly constrained by $\Xme = A\Yme, \Ume = A^\trsp \Vme$.
Thus we have, by the Gaussian conditioning trick (\cref{lemma:condTrick}),
\begin{align}
    g \disteq_{\Bb} (E + \PP_{\Vme}^\perp \tilde A \PP_{\Yme}^\perp) h
    \label{eqn:gCondTrick}
\end{align}
where
$\tilde A$ is an independent copy of $A$, and $\PP_{\Yme} \defeq \Yme \Yme^+ = \Yme(\Yme^\trsp \Yme)^+ \Yme^\trsp$ is the projection to the column space of $\Yme$ (likewise for $\PP_{\Vme}$), and
\begin{equation}
  \begin{aligned}
  E
      &\defeq
          \Xme \Yme^+
          + \Vme^{+\trsp} \Ume^{\trsp}
          - \Vme^{+\trsp} \Ume^\trsp \Yme \Yme^+
          \\
      &=
          \Xme (\Yme^\trsp \Yme)^+ \Yme^{\trsp}
          + \Vme (\Vme^{\trsp} \Vme)^+ \Ume^{\trsp}
          - \Vme (\Vme^{\trsp} \Vme)^+ \Ume^\trsp \Yme (\Yme^\trsp \Yme)^+ \Yme^\trsp.
  \end{aligned}
  \label{eqn:Eexpansion}
\end{equation}

We can make obvious the conditional distribution of $g$ if we rewrite \cref{eqn:gCondTrick} as
\begin{align}
  g &\disteq_\Bb \omega + \sigma \PP_{\Vme}^\perp z,\quad \text{with $z \sim \Gaus(0, I_n)$}
  \label{eqn:gConditionedOnB}
  \\
  \text{with}\quad
  \omega
  &\defeq
      E h \in \R^n,
      \quad
  \sigma \defeq
    \sigma_A
  \sqrt{\|\PP_{\Yme}^\perp h\|^2/n} \in \R
  .
  \label{eqn:meanvardef}
\end{align}
where $\omega, \sigma, \PP_\Vme^\perp$ are all deterministic conditioned on $\Bb$, and the only randomness after conditioning comes from $z$.
We will see below that $\omega$ can be expressed explicitly as a linear combination in $\Xme$ and $\Vme$ (\cref{lemma:omegaExpansion}), and $\sigma$ converges to a deterministic limit $\mathring\sigma$ (\cref{lemma:sigmaConverges}).
But to get there, we need to first define a few useful matrices and vectors of fixed dimensions
\begin{equation}
\begin{aligned}
    \Upsilonme 
        &\defeq
            \Yme^\trsp \Yme/n \in \R^{r \times r}
            &
    \Lambdame 
        &\defeq
            \Vme^\trsp \Vme/n \in \R^{s \times s}
            &
    \Gamma
        &\defeq
            \Ume^\trsp \Yme/n \in \R^{s \times r}
            \\
    \gammame
        &\defeq
            \Yme^\trsp h/n \in \R^{r}
            &
    \deltame
        &\defeq
            \Ume^\trsp h /n \in \R^s
            .
\end{aligned}
    \label{eqn:momentMatrices}
\end{equation}

By induction hypothesis \ref{IH:MomConv}$(\MM-1)$, $\Upsilonme , \Lambdame , \Gamma, \gammame, \deltame$ all converge a.s.\ to corresponding deterministic limits $\mathring{\Upsilonme }, \mathring{\Lambdame }, \mathring \Gamma, \mathring{\gammame}, \mathring{\deltame}$:
\begin{equation}
\begin{aligned}
  \Upsilonme_{ij}
    &\asto
      \mathring{\Upsilonme }_{ij}
    \defeq
      \EV  Z^{\yme^i}  Z^{\yme^j}
    =
      (\sigma_A)^{-2} 
      \Cov(\Zhat^{\xme^i}, \Zhat^{\xme^j})
      \\
    \Lambdame_{ij}
      &\asto
        \mathring{\Lambdame }_{ij}
      \defeq
        \EV  Z^{\vme^i}  Z^{\vme^j}
      =
        (\sigma_A)^{-2} \Cov(\Zhat^{\ume^{i}}, \Zhat^{\ume^{j}})
      \\
  \gammame_i
    &\asto
      \mathring{\gammame}_i
    \defeq
      \EV  Z^{\yme^i} Z^h
    =
      (\sigma_A)^{-2} \Cov(\Zhat^{\xme^i}, \Zhat^{g})
      \\
  \Gamma_{ij}
    &\asto
      \mathring \Gamma_{ij}
    \defeq
      \EV  Z^{\ume^i} Z^{\yme^j}
      \\
  \deltame_i
    &\asto
      \mathring{\deltame}_i
    \defeq
      \EV  Z^{\ume^i} Z^h
\end{aligned}
\label{eqn:limitMatrices}
\end{equation}

It turns out that, as a consequence of \cref{lemma:rankStability} below, we have: a.s.\ for all large enough $n$, $\rank \Upsilonme  = \rank \mathring {\Upsilonme }$ and $\rank \Lambdame  = \rank \mathring{\Lambdame }$.
Therefore, as pseudoinverse is continuous on matrices of fixed rank,
we get the following proposition
\begin{prop}\label{prop:pseudoinverseLambda}
$\Upsilonme ^+ \asto \mathring{\Upsilonme }^+$ and $\Lambdame ^+ \asto \mathring{\Lambdame }^+$.
\end{prop}

Using this proposition, we compute the limit of $\sigma^2$.
\begin{lemma}\label{lemma:sigmaConverges}
The quantity $\sigma$, defined in \cref{eqn:meanvardef}, converges to a deterministic limit $\mathring \sigma$ almost surely:
\begin{equation}\sigma^2 \asto \mathring \sigma^2 \defeq \sigma_A^2 (\EV (Z^h)^2 - \mathring \gammame^\trsp \mathring \Upsilonme^+ \mathring \gammame).
  \label{eqn:mathringsigma}
\end{equation}
\end{lemma}
\begin{proof}
Note that
\begin{align*}
    \sigma^2 = \f {\sigma_A^2} n (h^\trsp h - h^\trsp \PP_{\Yme} h)
        = \f {\sigma_A^2} n (h^\trsp h - h^\trsp {\Yme} (\Yme^\trsp \Yme)^+ \Yme^\trsp h)
        = \f {\sigma_A^2} n (h^\trsp h - \gammame^\trsp \Upsilonme ^+ \gammame).
\end{align*}
Because $\phi$ is polynomially-bounded, so is $\phi(z)^2$.
By induction hypothesis (\ref{IH:MomConv}($\MM-1$)),
\begin{align*}
    \f 1 n h^\trsp h = \f 1 n \sum_{\alpha = 1}^n \phi(g^1_\alpha, \ldots, g^{\MM - 1}_\alpha)^2
    \asto
    &\EV
      \phi(Z^{g^1}, \ldots, Z^{g^{\MM-1}})^2
    =
      \EV (Z^h)^2
      .
\end{align*}
By \cref{eqn:limitMatrices}, $\gammame \asto \mathring{\gammame}$ and $\Upsilonme  \asto \mathring{\Upsilonme }$.
By \cref{prop:pseudoinverseLambda}, $\Upsilonme ^+ \asto \mathring{\Upsilonme }^+$.
Combining all of these limits together yields the desired claim.
\end{proof}

Using \cref{eqn:momentMatrices,eqn:Eexpansion}, we can re-express $\omega$ as
\begin{align*}
\omega
    &=
        \Xme \Upsilonme ^+ \gammame
        + \Vme \Lambdame ^+ \deltame
        - \Vme \Lambdame ^+ \Gamma \Upsilonme ^+ \gammame
        .
\end{align*}
Define $\dme \in \R^r$ and $\eme \in \R^s$ by
\begin{equation}
  \begin{aligned}
    \dme &\defeq \Upsilonme ^+ \gammame,
        &\text{so that}&&
          \dme &\asto \mathring {\dme} \defeq \mathring{\Upsilonme }^+ \mathring{\gammame}, \quad\text{and}\\
    \eme &\defeq {\Lambdame }^+({\deltame} - \Gamma {\Upsilonme }^+ {\gammame}),
        &\text{so that}&&
          \eme &\asto \mathring {\eme} \defeq \mathring{\Lambdame }^+(\mathring{\deltame} - \mathring \Gamma \mathring{\Upsilonme }^+ \mathring{\gammame}).
  \end{aligned}
  \label{eqn:zetaeta}
\end{equation}
Then the next lemma follows trivially.
\begin{lemma}\label{lemma:omegaExpansion}
For $\omega$ defined in \cref{eqn:meanvardef}, we have
\[\omega = Eh = \Xme \dme + \Vme \eme = \Xme(\mathring {\dme} + \varepsilonhat) + \Vme (\mathring {\eme} + \varepsiloncheck),\]
for some random vectors $\varepsilonhat \in \R^r, \varepsiloncheck \in \R^s$ that go to 0 a.s.\ with $n$,
\end{lemma}

\subsubsection{Helper Lemmas}

Using the matrix notation of \cref{defn:LsqZ} and the expression for the projection operator, we can rewrite $\mathring{\dme}$ and $\mathring{\eme}$ as follows.
The punchline of this section is \cref{lemma:hpartReExpression}, which says
\begin{align*}
  \sum_{i=1}^r \mathring{\dme}_i \Zdot^{\xme^i}
  +
  \sum_{j=1}^s \mathring{\eme}_j  Z^{\vme^j}
    &=
    \Zdot^g.
\end{align*}

\begin{prop}\label{prop:vReExpression}
  Let $\Pi: \LsqZ \to \LsqZ$ be the orthogonal projection operator onto the linear span of $\{ Z^{\yme^1}, \ldots,  Z^{\yme^r}\}$.
  Then,
  \begin{align}
  \mathring{\dme}
    &=
      \begin{bmatrix}
      \phz
        &\phz
          &\phz\\
      \phz
        &\mathring{\Upsilonme }^+
          &\phz\\
      \phz
        &\phz
          &\phz
      \end{bmatrix}
      \begin{bmatrix}
         Z^{\yme^1}\\
        \vdots\\
         Z^{\yme^r}
      \end{bmatrix}
         [Z^{h}]
      ,
      \quad
  \mathring{\eme}
      =
      \begin{bmatrix}
        \phz
          &\phz
            &\phz\\
        \phz
          &\mathring{\Lambdame }^+
            &\phz\\
        \phz
          &\phz
            &\phz
        \end{bmatrix}
        \begin{bmatrix}
          \Zhat^{\ume^1}\\
          \vdots\\
          \Zhat^{\ume^s}
        \end{bmatrix}
        [\Pi^\perp  Z^{h}]
      .
      \label{eqn:xi&zetaLimit}
  \end{align}
  and
  \begin{equation}
    \mathring\sigma^2 = \sigma_A^2 \EV Z^h \Pi^\perp Z^h
      = \sigma_A^2 \EV (\Pi^\perp Z^h)^2
      .
    \label{eqn:sigmalimitPiPerp}
  \end{equation}
  \end{prop}
  
  \begin{proof}
    The identity for $\mathring \dme$ is obvious.
    We derive the identity for $\mathring \eme$, and the proof of \cref{eqn:sigmalimitPiPerp} follows similarly.
    First, we expand the definitions of $\mathring \deltame, \mathring\Gamma, \mathring\Upsilonme, \mathring\gammame$ in matrix notation of \cref{defn:LsqZ}.
    \begin{align*}
      \mathring \deltame &= \begin{bmatrix}
              Z^{\ume^1}\\
            \vdots\\
              Z^{\ume^s}
          \end{bmatrix}
          [Z^h]
          ,&
      \mathring \Gamma &=
          \begin{bmatrix}
              Z^{\ume^1}\\
            \vdots\\
              Z^{\ume^s}
          \end{bmatrix}
          [Z^{\yme^1}, \ldots, Z^{\yme^r}]
          ,&
      \mathring \Upsilonme &=
        \begin{bmatrix}
            Z^{\yme^1}\\
          \vdots\\
            Z^{\yme^r}
        \end{bmatrix}
        [Z^{\yme^1}, \ldots, Z^{\yme^r}]
        ,&
      \mathring \gammame &=
        \begin{bmatrix}
            Z^{\yme^1}\\
          \vdots\\
            Z^{\yme^r}
        \end{bmatrix}
        [Z^h]
        .
    \end{align*}
    Then with the matrix form of $\Pi$ in mind (\cref{eqn:matrixformProjection}), we derive
    \begin{align}
    \mathring{\deltame} - \mathring \Gamma \mathring{\Upsilonme }^+ \mathring{\gammame}
      &=
        \begin{bmatrix}
           Z^{\ume^1}\\
          \vdots\\
           Z^{\ume^s}
        \end{bmatrix}
        [(I - \Pi)  Z^{h}]
      =
        \begin{bmatrix}
           Z^{\ume^1}\\
          \vdots\\
           Z^{\ume^s}
        \end{bmatrix}
        [\Pi^\perp  Z^{h}]
      =
        \begin{bmatrix}
          \Zhat^{\ume^1}\\
          \vdots\\
          \Zhat^{\ume^s}
        \end{bmatrix}
        [\Pi^\perp  Z^{h}]
        .
        \label{eqn:etaGammaUpsilongamma}
    \end{align}
  Here the last equality follows from the following
  \begin{align*}
    \begin{bmatrix}
       Z^{\ume^1}\\
      \vdots\\
       Z^{\ume^s}
    \end{bmatrix}
    [\Pi^\perp
     Z^{h}]
    &=
      \begin{bmatrix}
        \Pi^\perp  Z^{\ume^1}\\
        \vdots\\
        \Pi^\perp  Z^{\ume^s}
      \end{bmatrix}
       [Z^{h}]
    =
      \begin{bmatrix}
        \Pi^\perp \Zhat^{\ume^1}\\
        \vdots\\
        \Pi^\perp \Zhat^{\ume^s}
      \end{bmatrix}
       [Z^{h}]
    =
      \begin{bmatrix}
        \Zhat^{\ume^1}\\
        \vdots\\
        \Zhat^{\ume^s}
      \end{bmatrix}
      [\Pi^\perp
       Z^{h}]
  \end{align*}
  where the middle equality follows because each $ Z^{\ume^j} - \Zhat^{\ume^j} = \Zdot^{\ume^j}$ is a linear combination of $ \{Z^{\yme^i}\}_{i=1}^r$ by \refZdot{}.
  Finally, plugging in \cref{eqn:etaGammaUpsilongamma} to \cref{eqn:zetaeta}, we have
  \begin{align*}
    \mathring{\eme}
      &=
        \mathring{\Lambdame }^+(\mathring{\deltame} - \mathring \Gamma \mathring  {\Upsilonme }^+ \mathring{\gammame})
      =
        \begin{bmatrix}
        \phz
          &\phz
            &\phz\\
        \phz
          &\mathring{\Lambdame }^+
            &\phz\\
        \phz
          &\phz
            &\phz
        \end{bmatrix}
        \begin{bmatrix}
          \Zhat^{\ume^1}\\
          \vdots\\
          \Zhat^{\ume^s}
        \end{bmatrix}
        [\Pi^\perp  Z^{h}]
        .
  \end{align*}
  \end{proof}

Using \cref{eqn:xi&zetaLimit}, we can write the following
\begin{lemma}\label{lemma:DtrxhHatGReExpression}
  Let $\Pi: \LsqZ \to \LsqZ$ be the orthogonal projection operator onto the linear span of $\{ Z^{\yme^1}, \ldots,  Z^{\yme^r}\}$.
  Then
  \begin{align}
  \sum_{i=1}^r \mathring{\dme}_i \Zdot^{\xme^i}
    &=
      [
        Z^{\vme^1},
        \ldots,
        Z^{\vme^s}
      ]
      \begin{bmatrix}
      \phz
        &\phz
          &\phz\\
      \phz
        &\mathring{\Lambdame }^+
          &\phz\\
      \phz
        &\phz
          &\phz
      \end{bmatrix}
      \begin{bmatrix}
      \Zhat^{\ume^1}\\
      \vdots\\
      \Zhat^{\ume^s}
      \end{bmatrix}
      [\Pi  Z^{h}]    
      .
      \label{eqn:zetaZdot}
  \end{align}
\end{lemma}

\begin{proof}
By \cref{lemma:conjugatePairDtrxh}, for each $i \in [r]$,
\begin{align*}
\Zdot^{\xme^i}
  &=  
    [Z^{\vme^1},
    \ldots,
     Z^{\vme^s}]
    \begin{bmatrix}
    \phz
      &\phz
        &\phz\\
    \phz
      &\mathring{\Lambdame }^+
        &\phz\\
    \phz
      &\phz
        &\phz
    \end{bmatrix}
    \begin{bmatrix}
    \Zhat^{\ume^1}\\
    \vdots\\
    \Zhat^{\ume^s}
    \end{bmatrix}
     [Z^{\yme^i}]
\end{align*}
because $\{\ume^j\}_{j=1}^s$ contains all G-vars of the form $W^\trsp\mathrm{vector}$ introduced before $\xme^i$.
By \cref{prop:vReExpression},
\begin{align*}
\mathring{\dme}
  &=
    \begin{bmatrix}
    \phz
      &\phz
        &\phz\\
    \phz
      &\mathring{\Upsilonme }^+
        &\phz\\
    \phz
      &\phz
        &\phz
    \end{bmatrix}
    \begin{bmatrix}
       Z^{\yme^1}\\
      \vdots\\
       Z^{\yme^r}
    \end{bmatrix}
       [Z^{h}]
      .
\end{align*}
Then
\begin{align*}
\sum_{i=1}^r \mathring{\dme}_i \Zdot^{\xme^i}
&=
  [\Zdot^{\xme^1}, \ldots, \Zdot^{\xme^r}]
  \begin{bmatrix}
    \phz\\
    \mathring{\dme}\\
    \phz
  \end{bmatrix}
  \\
&=
  [Z^{\vme^1},
  \ldots,
   Z^{\vme^s}]
  \begin{bmatrix}
  \phz
    &\phz
      &\phz\\
  \phz
    &\mathring{\Lambdame }^+
      &\phz\\
  \phz
    &\phz
      &\phz
  \end{bmatrix}
  \begin{bmatrix}
  \Zhat^{\ume^1}\\
  \vdots\\
  \Zhat^{\ume^s}
  \end{bmatrix}
  [Z^{\yme^1},
    \ldots,
     Z^{\yme^r}
  ]
  \\
&\phantomeq\quad
  \times
  \begin{bmatrix}
  \phz
    &\phz
      &\phz\\
  \phz
    &\mathring{\Upsilonme }^+
      &\phz\\
  \phz
    &\phz
      &\phz
  \end{bmatrix}
  \begin{bmatrix}
     Z^{\yme^1}\\
    \vdots\\
     Z^{\yme^r}
  \end{bmatrix}
     [Z^{h}]
  \\
&=
  ( Z^{\vme^1},
  \ldots,
   Z^{\vme^s})
  \begin{bmatrix}
  \phz
    &\phz
      &\phz\\
  \phz
    &\mathring{\Lambdame }^+
      &\phz\\
  \phz
    &\phz
      &\phz
  \end{bmatrix}
  \begin{bmatrix}
  \Zhat^{\ume^1}\\
  \vdots\\
  \Zhat^{\ume^s}
  \end{bmatrix}
  [\Pi  Z^{h}]
\end{align*}
as desired.
\end{proof}

By combining \cref{lemma:DtrxhHatGReExpression} with \cref{prop:vReExpression}, we get
\begin{lemma}\label{lemma:hpartReExpression}
\begin{align*}
  \sum_{i=1}^r \mathring{\dme}_i \Zdot^{\xme^i}
  +
  \sum_{j=1}^s \mathring{\eme}_j  Z^{\vme^j}
    &=
    \Zdot^g.
\end{align*}
\end{lemma}
\begin{proof}
  By realizing the $\Pi^\perp$ and $\Pi$ in \cref{eqn:xi&zetaLimit} and \cref{eqn:zetaZdot} ``cancel'' to get identity, we have
\begin{align*}
  \sum_{i=1}^r \mathring{\dme}_i \Zdot^{\xme^i}
  +
  \sum_{j=1}^s \mathring{\eme}_j  Z^{\vme^j}
    &=
      [
         Z^{\vme^1},
        \ldots,
         Z^{\vme^s}
      ]
      \begin{bmatrix}
      \phz
        &\phz
          &\phz\\
      \phz
        &\mathring{\Lambdame }^+
          &\phz\\
      \phz
        &\phz
          &\phz
      \end{bmatrix}
      \begin{bmatrix}
      \Zhat^{\ume^1}\\
      \vdots\\
      \Zhat^{\ume^s}
      \end{bmatrix}
       [Z^{h}]
      \\
    &=
      [
         Z^{\vme^1},
        \ldots,
         Z^{\vme^s}
      ]
      \begin{bmatrix}
      \phz
        &\phz
          &\phz\\
      \phz
        &C^+
          &\phz\\
      \phz
        &\phz
          &\phz
      \end{bmatrix}
      \begin{bmatrix}
      \phz\\
      b\\
      \phz
      \end{bmatrix}
      .
  \end{align*}
where $C=\mathring \Lambdame$ and $b$ are as in \cref{eq:ExpectationPartialDer}.
By definition (\refZdot{}), we have the desired result.
\end{proof}

\subsubsection{Index of Symbols}
\label{sec:indexSymbols}
\begin{align*}
  g=g^{\MM}, h, \phi &\cdots\cdots \text{\cref{eqn:gAh}} &
  x^i,y^i, u^j, v^j &\cdots\cdots \text{\cref{eqn:xAy&uATv}} &
  \Xme, \Yme, \Ume, \Vme &\cdots\cdots \text{\cref{eqn:XUYZ}} \\
  \PP_\Vme, \PP_\Yme & \cdots\cdots \text{\cref{eqn:gCondTrick}} &
  E & \cdots\cdots \text{\cref{eqn:Eexpansion}} &
  \omega, \sigma &\cdots\cdots \text{\cref{eqn:meanvardef}} \\
  \Upsilonme, \Lambdame, \Gamma, \gammame, \deltame &\cdots\cdots \text{\cref{eqn:momentMatrices}} &
  \mathring\Upsilonme, \mathring\Lambdame, \mathring\Gamma, \mathring\gammame, \mathring\deltame &\cdots\cdots \text{\cref{eqn:limitMatrices}} &
  \mathring \sigma & \cdots\cdots \text{\cref{eqn:mathringsigma}}\\
  \dme, \eme, \mathring\dme, \mathring\eme &\cdots\cdots \text{\cref{eqn:zetaeta}} &
\end{align*}
In addition, $\Bb$ denotes the $\sigma$-algebra spanned by $g^1, \ldots, g^{\MM-1}$.

\subsection{Rank Stability and Zero Stability}
\label{sec:rankStabilityZeroStability}
In this section, we prove the following consequence of \ref{IH:coreSet}$(\MM-1)$ and \ref{IH:MomConv}$(\MM-1)$.

\begin{lemma}[Rank Stability]\label{lemma:rankStability}
For any collection of polynomially-bounded functions $\{\psi_j: \R^{\MM-1} \to \R\}_{j=1}^l$, let $K \in \R^{l \times l}$ be the random matrix (depending on $n$) defined by
\begin{equation*}
K_{ij} = \f 1 n \sum_{\alpha=1}^n \psi_i(g^1_\alpha, \ldots, g^{\MM-1}_\alpha) \psi_j(g^1_\alpha, \ldots, g^{\MM-1}_\alpha).
\end{equation*}
By \ref{IH:MomConv}$(\MM-1)$,
\[K \asto \mathring K\]
for some matrix $\mathring K \in \R^{l \times l}$.
\begin{enumerate}
\item
    Then, almost surely, for large enough $n$,
    \begin{equation*}
    \ker K = \ker \mathring K, \quad \im K = \im \mathring K, \quad\text{and}\quad \rank K = \rank \mathring K.
    \end{equation*}
    Here $\ker$ denotes null space and $\im$ denotes image space.
\item
    Suppose $I \sbe [l]$ is any subset such that $\mathring K|_I$, the restriction of $\mathring K$ to rows and columns corresponding to $I$, satisfies
    \[|I| = \rank \mathring K|_I = \rank \mathring K.\]
    There are unique coefficients $\{F_{ij}\}_{i \in [l], j \in I}$ that expresses each row of $\mathring K$ as linear combinations of rows corresponding to $I$:
    \[
    \forall i \in [l],\quad 
    \mathring K_i = \sum_{j \in I} F_{ij} \mathring K_j.
    \]
    Then, a.s.\ for all large $n$, for all $\alpha \in [n]$,
    \begin{equation*}
    \psi_i(g^1_\alpha, \ldots, g^{\MM-1}_\alpha)
    = \sum_{j \in I} F_{ij} \psi_j(g^1_\alpha, \ldots, g^{\MM-1}_\alpha).
    \end{equation*}
\end{enumerate}

\end{lemma}

\cref{lemma:rankStability} will be primarily a corollary of the following \cref{lemma:zerofunStability}.

\begin{lemma}[Zero Stability]\label{lemma:zerofunStability}
If $\psi: \R^{\MM-1} \to \R^{\ge 0}$ is a nonnegative function such that
\begin{align*}
\f 1 n \sum_{\alpha = 1}^n \psi(g^1_\alpha, \ldots, g^{\MM-1}_\alpha) \asto 0
\end{align*}
then, almost surely, for large enough $n$,
\[\psi(g^1_\alpha, \ldots, g^{\MM-1}_\alpha) = 0\]
for all $\alpha \in [n]$.
\end{lemma}

We give the proof of \cref{lemma:rankStability} now, assuming \cref{lemma:zerofunStability}.
\begin{proof}
  \newcommand{\myvv}{z}
Let $\myvv \in \R^l$ be in the null space of $\mathring K$, i.e. $\myvv^\trsp \mathring K \myvv = 0$.
Then we also have $\myvv^\trsp K \myvv \asto \myvv^\trsp \mathring K \myvv = 0$.
But
\begin{align*}
\myvv^\trsp K \myvv
    &=
        \f 1 n \sum_{\alpha=1}^n \Psi(g^1_\alpha, \ldots, g^{\MM-1}_\alpha)
        ,
        \quad
        \text{where}
        \quad
        \Psi(g^1_\alpha, \ldots, g^{\MM-1}_\alpha) \defeq
        \lp
            \sum_{i=1}^l \myvv_i \psi_i(g^1_\alpha, \ldots, g^{\MM-1}_\alpha)
        \rp^2
\end{align*}
and $\Psi$ is a nonnegative function.
By \cref{lemma:zerofunStability}, we have that: almost surely, for large enough $n$, 
\[\Psi(g^1_\alpha, \ldots, g^{\MM-1}_\alpha) = 0 \quad \text{for all $\alpha \in [n]$}
\quad
\implies \myvv^\trsp K \myvv = 0.\]
\textit{Proof of Claim 1.}\ \ 
If we apply this argument to a basis $\{\myvv^1, \ldots, \myvv^t\}$ of $\ker \mathring K$, then we get, 
\[\text{a.s.\ for all large $n$,}\quad 
\ker \mathring K \sbe \ker K,\]
so that
\[\text{a.s.\ for all large $n$,}\quad 
\rank \mathring K \ge \rank K.\]
Because the rank function is lower semicontinuous (i.e.\ the rank can drop suddenly, but cannot increase suddenly), and $K \asto \mathring K$, we also have
\[\text{a.s.\ for all large $n$,}\quad 
\rank \mathring K \le \rank K.\]
Combined with the above, this gives the desired result on rank.
The equality of null space then follows from the equality of rank, and the equality of image space follows immediately, as the image space is the orthogonal complement of the null space.

\textit{Proof of Claim 2.}\ \ 
If we apply the above argument to each $\myvv^i$ defined by inner product as
\[\forall x \in \R^l,\quad x^\trsp \myvv^i = x_i - \sum_{j \in I} F_{ij} x_j,
\]
(note that only for $i \not \in I$ is $\myvv^i$ nonzero),
then we have, a.s.\ for large $n$, $\myvv^i{}^\trsp K \myvv^i = 0$, or
\begin{equation*}
    \psi_i(g^1_\alpha, \ldots, g^{\MM-1}_\alpha)
    = \sum_{j \in I} F_{ij} \psi_j(g^1_\alpha, \ldots, g^{\MM-1}_\alpha),
    \quad \forall \alpha \in [n].
\end{equation*}
\end{proof}

In the rest of this section, we prove \cref{lemma:zerofunStability}.
It helps to first show that the linear relations given in \ref{prop:basis} carries over to the $n\to\infty$ limit.
\begin{prop}\label{prop:limitbasis}
By the \ref{prop:basis} property, each $g^i, i \in \coreset \sbe [\MM-1]$, has a set of unique constants $\{a_j\}_{j \in \coreset}$ (independent of $n$) such that, almost surely, for large enough $n$,
\[g^i = \sum_{j \in \coreset} a_j g^j.\]
Then for each $i \in [\MM-1]$,
\[\Zz^{g^i} \aseq \sum_{j \in \coreset} a_j \Zz^{g^j}
.\]
\end{prop}
\begin{proof}

Let $\psi(x^1, \ldots, x^{\MM-1}) \defeq (x^i - \sum_{j \in \coreset} a_j x^j)^2$.
Then by \ref{prop:basis}$(\MM-1)$ and \ref{IH:MomConv}$(\MM-1)$,
\begin{align*}
\f 1 n \sum_{\alpha=1}^n \psi(g^1_\alpha, \ldots, g^{\MM-1}_\alpha) \asto \EV \psi\lp \Zz^{g^1}, \ldots, \Zz^{g^{\MM-1}}\rp = 0.
\end{align*}
This implies that
\begin{align*}
\Zz^{g^i} - \sum_{j \in \coreset} a_j \Zz^{g^j} \aseq 0
\iff
\Zz^{g^i} \aseq \sum_{j \in \coreset} a_j \Zz^{g^j}
\end{align*}
as desired.

\end{proof}

Now we show \cref{lemma:zerofunStability}.

\begin{proof}[Proof of \cref{lemma:zerofunStability}]
By \ref{IH:MomConv}$(\MM-1)$ and the premise of \cref{lemma:zerofunStability},
\begin{align*}
\f 1 n \sum_{\alpha = 1}^n \psi(g^1_\alpha, \ldots, g^{\MM-1}_\alpha) \to \EV \psi(\Zz^{g^1}, \ldots, \Zz^{g^{\MM-1}}) = 0.
\end{align*}
Let $\Zz^\coreset \defeq \{\Zz^{g^i}\}_{i \in \coreset}$.
By \ref{prop:density}$(\MM-1)$, the law of $\Zz^\coreset$ has density, i.e.\ a set is measure zero against its law iff it has measure zero against Lebesgue measure.
Furthermore, \ref{prop:basis} yields a linear function $F$ such that
\[\text{a.s.\ for large enough $n$, for all $\alpha \in [n]$,}\quad
F(\{g^j_\alpha\}_{j \in \coreset}) = (g^1_\alpha, \ldots, g^{\MM-1}_\alpha)
.
\]
By \cref{prop:limitbasis}, the same linear function satisfies
\[F(\Zz^\coreset) = (\Zz^{g^1}, \ldots, \Zz^{g^{\MM-1}})
.\]
Therefore,
\begin{equation*}
0 = \EV \psi(\Zz^{g^1}, \ldots, \Zz^{g^{\MM-1}})
= \EV \psi \circ F(\Zz^\coreset)
.
\end{equation*}
Because $\psi$, and thus $\psi \circ F$, is a nonnegative function, the nullity of the expectation implies that, other than a set $U$ of measure 0 under the distribution of $\Zz^\coreset$, $\psi\circ F$ is 0.
This set $U$ also has Lebesgue measure zero as the law of $\Zz^\coreset$ has density, as discussed above.

If in \ref{prop:nullAvoid}$(\MM-1)$, we set $U_{n\alpha} = U$ for all $n$ and all $\alpha \in [n]$, then we get that: almost surely, for all large enough $n$, for all $\alpha \in [n]$,

\begin{equation*}
\{g^i_\alpha\}_{i \in \coreset} \not\in U
\iff
\psi\circ F(\{g^i_\alpha\}_{i \in \coreset}) = 0
\iff
\psi(g^1_\alpha, \ldots, g^{\MM-1}_\alpha) = 0,
\end{equation*}
as desired.
\end{proof}

\subsection{Inductive Step: \ref{IH:coreSet}\texorpdfstring{$(\MM)$}{(m)}}
\label{sec:inductiveCoreSet}
In this section, we show
\begin{equation*}
\text{\ref{IH:MomConv}$(\MM-1)$ and \ref{IH:coreSet}$(\MM-1)$} \implies \text{\ref{IH:coreSet}}(\MM).
\end{equation*}
More explicitly, we need to think about whether to add $\MM$ to the core set $\coreset$ of $[\MM-1]$ in order to maintain the \ref{prop:basis}, \ref{prop:density}, and \ref{prop:nullAvoid} properties.

We proceed by casework on whether $\mathring \sigma = 0$.

\newcommand{\Lsq}{\mathcal{L}}
\subsubsection{If \texorpdfstring{$\mathring \sigma = 0$}{sigma converges to 0 a.s.}.}

We will show that the core set properties are maintained if we don't add $m$ to the core set.

\paragraph{$\coreset$ can be kept the same}
Recall that $g = Ah$ and $\xme^i = A \yme^i$ where $h$ was introduced by $h = \phi(g^{1}, \ldots, g^{\MM - 1})$, for some polynomially bounded $\phi$.
Suppose we likewise have $\yme^i = \hat \phi^i(g^1, \ldots, g^{\MM-1})$, for each $i \in [r]$, for polynomially bounded $\hat \phi^i: \R^{\MM-1} \to \R$ (where $\hat\phi^i$ only depends on the G-vars that came before $\yme^i$, but we implicitly pad coordinates of $\hat \phi^i$ to allow it to take all of $g^1, \ldots, g^{\MM-1}$ as inputs).
By \ref{prop:basis}, we know that, a.s.\ for large enough $n$, each of $g^1, \ldots, g^{\MM-1}$ is a (unique, constant-in-$n$) linear combination of $\{g^j\}_{j \in \coreset}$.
Therefore, we can express
\begin{equation}h = \psi(\{g^j\}_{j \in \coreset}),\quad\text{and}\quad
\forall i \in [r],\quad \yme^i = \hat {\psi}^i(\{g^j\}_{j \in \coreset})
\label{eqn:psireduce}
\end{equation}
for some polynomially bounded $\psi, \hat {\psi}^i: \R^\coreset \to \R$.
Let $\Pi: \LsqZ \to \LsqZ$ be the orthogonal projection onto the subspace spanned by $Z^{\yme^1}, \ldots, Z^{\yme^r}$.
By \cref{eqn:sigmalimitPiPerp}, we have
\(
\mathring\sigma^2
    =
        \sigma_A^2 \EV (\Pi^\perp Z^h)^2
        .
\)
Therefore, $\mathring \sigma = 0$ implies that 
\[Z^h \aseq \Pi Z^h.\]
Since $Z^h, Z^{\yme^1}, \ldots, Z^{\yme^r}$ are resp.\ deterministic images of $Z^\coreset$ under the functions $\psi, \hat \psi^1, \ldots, \hat \psi^r$, this shows that $\psi$ is a linear combination of $\hat \psi^1, \ldots, \hat \psi^r$ almost surely under the law of $Z^\coreset$.
But by \ref{prop:density}($\MM-1$), the law of $Z^\coreset$ is absolutely continuous w.r.t.\ the Lesbegue measure (and vice versa), so this statement also holds under Lesbegue measure:
For a set $U \sbe \R^\coreset$ of Lesbegue measure zero and a set of coefficients $c_1, \ldots, c_r \in \R$, we have
\begin{equation*}
\forall \vec x \not \in U,\quad \psi(\vec x) = \sum_{i=1}^r c_i {\hat \psi^i}(\vec x).
\end{equation*}
By \ref{prop:nullAvoid}$(\MM-1)$ applied to $U_{n\alpha} = U$ for all $n$ and $\alpha \in [n]$, we also have that: a.s.\ for large enough $n$,
\begin{align*}
\psi(\{g^j\}_{j \in \coreset})
&= \sum_{i=1}^r c_i {\hat \psi^i}(\{g^j\}_{j \in \coreset}),
\quad\text{so that by \cref{eqn:psireduce}}
\\
g = Ah = A \psi(\{g^j\}_{j \in \coreset})
&= \sum_{i=1}^r c_i A{\hat \psi^i}(\{g^j\}_{j \in \coreset}),
= \sum_{i=1}^r c_i A\yme^i
= \sum_{i=1}^r c_i \xme^i.
\end{align*}
This shows that, if we keep the core set as $\coreset$, then \ref{prop:basis} is still satisfied.
Since the core set is not changing, \ref{prop:density} and \ref{prop:nullAvoid} just follows from the induction hypothesis.
For usage later in the proof of \ref{IH:MomConv}$(\MM)$, we record our observation here as a lemma.
\begin{lemma}\label{lemma:limitSigmaIsZero}
If $\mathring \sigma = 0$, then there are coefficients $c_1, \ldots, c_r \in \R$ independent of $n$ such that a.s.\ for large enough $n$,
\[
g = \sum_{i = 1}^r c_i \xme^i.
\]
\end{lemma}

\subsubsection{If \texorpdfstring{$\mathring \sigma > 0$}{sigma converges to nonzero value a.s.}.}

\newcommand{\prvZ}{\mathcal{Z}}

It's clear that $g$ cannot be in the linear span of $\{\xme^i\}_{i\in[r]}$ asymptotically, so we will add $g$ to the core set, and the \ref{prop:basis} property follows immediately.
In the below, we shall write $\coreset$ for the old core set, and $\coreset' \defeq \coreset \cup \{g\}$ for the new one.

It remains to show \ref{prop:density} and \ref{prop:nullAvoid} for $\coreset'$.

\paragraph{\ref{prop:density}$(\MM)$ holds}
By definition (\refZMatMul{}), we have
\begin{align*}
\Zz^g = \Zhat^g + \Zdot^g = \Zhat^g + \sum_{j=1}^s a_j  Z^{\vme^j}
\end{align*}
where $a_j$ are the partial derivative expectations in \refZdot{} (whose specific values we will not care about) and \cref{eq:ExpectationPartialDer}.
Note that for all $j \in [s]$, $ Z^{\vme^j}$ only depends on $\Zhat^{g^1}, \ldots, \Zhat^{g^{\MM-1}}$.
Let $\prvZ$ be the $\sigma$-algebra generated by $Z^{g^1}, \ldots, Z^{g^{\MM-1}}$, which is the same as the $\sigma$-algebra generated by $\Zhat^{g^1}, \ldots, \Zhat^{g^{\MM-1}}$ by \cref{prop:hatZRandomSource}.
Then conditioned on $\prvZ$, we can follow a quick calculation to see that
\begin{align*}
\Zhat^{g}
  &\disteq_\prvZ
    \mathring \sigma z + \sum_{i=1}^r \mathring{\dme}_i \Zhat^{\xme^i}
\end{align*}
where $z \sim \Gaus(0, 1)$ is independent from $\prvZ$, and $\mathring{\dme}$ is as in \cref{lemma:omegaExpansion}.
This allows us to write
\begin{align*}
\Zz^g
  &\disteq_\prvZ
    \mathring \sigma z + F(\Zhat^{g^1}, \ldots, \Zhat^{g^{\MM-1}})
\end{align*}
for some deterministic function $F$.
Since $\mathring \sigma > 0$, this shows that the distribution of $\Zz^g$ conditioned on $\prvZ$ is absolutely continuous w.r.t.\ the 1-dimensional Lebesgue measure, and vice versa.
If we apply \cref{lemma:conditionalDensity} below with $X_1 = \Zz^g, X_2 = \{\Zz^{g^i}\}_{i \in \coreset},$ and $Y = (\Zhat^{g^1}, \ldots, \Zhat^{g^{\MM-1}})$, then the lemma premise \cref{eqn:X1density} follows from the above reasoning, and the lemma premise \cref{eqn:X2density} follows from induction hypothesis \ref{prop:density}$(\MM-1)$ (for $\coreset$).
This then yields \ref{prop:density}$(\MM)$ (for $\coreset'$), as desired.

\begin{lemma}\label{lemma:conditionalDensity}
Consider a random vector $X = (X_1, X_2) \in \R^a \times \R^b$ and another random vector $Y \in \R^c$.
Suppose $X_2$ is deterministic conditioned on $Y$.
Let $\lambda$ denote the Lebesgue measure in any Euclidean space.
If
\begin{itemize}
\item for every measurable set $U \sbe \R^b$,
  \begin{equation}
    \Pr(X_2 \in U) = 0 \iff \lambda(U) = 0, \quad \text{and}
    \label{eqn:X2density}
  \end{equation}
\item for every $y \in \R^c$, and every measurable set $U \sbe \R^a$,
  \begin{equation}
    \Pr(X_1 \in U \mid Y = y) = 0 \iff \lambda(U) = 0,
    \label{eqn:X1density}
  \end{equation}
\end{itemize}
then for every measurable set $V \sbe \R^a \times \R^b$,
\[\Pr(X \in V) = 0 \iff  \lambda(V) = 0
.\]
\end{lemma}
\begin{proof}
  Fix $V \sbe \R^a \times \R^b$.
We have
\begin{align*}
\Pr(X \in V)
  &=
    \int \Pr(X \in V \mid Y = y)\dd\Pr(Y = y)
    \\
  &=
    \int \Pr(X_1 \in V_{x_2(y)} \mid Y = y)\dd\Pr(Y = y)
    \numberthis\label{eqn:expandInY}
,
\end{align*}
where $x_2(y)$ is the deterministic value of $X_2$ conditioned on $Y = y$, and $V_{x_2} = \{x_1: (x_1, x_2) \in V\}$.
Likewise,
\begin{align}
\lambda(V)
  &=
    \int \lambda(V_{x_2}) \dd \lambda(x_2)
.
  \label{eqn:expandLebesgue}
\end{align}
Then
\begin{align*}
\Pr(X \in V) = 0
&\iff
  \Pr_{Y}(\Pr_{X_1}(X_1 \in V_{X_2} \mid Y) > 0) = 0
  &\text{by \cref{eqn:expandInY}}
  \\
&\iff
  \Pr_{Y}(\lambda(V_{X_2}) > 0) = 0
  &\text{by \cref{eqn:X1density}}
  \\
&\iff
  \Pr_{X_2}(\lambda(V_{X_2}) > 0) = 0
  &\text{because $X_2$ is deterministic given $Y$}
  \\
&\iff
  \lambda(x_2: \lambda(V_{x_2}) > 0) = 0
  &\text{by \cref{eqn:X2density}}
  \\
&\iff
  \lambda(V) = 0
  &\text{by \cref{eqn:expandLebesgue}}
\end{align*}
as desired.

\end{proof}

Now we tackle \ref{prop:nullAvoid}.

\paragraph{\ref{prop:nullAvoid}$(\MM)$ holds}

First, let's assume that, a.s.\ for large enough $n$, $\PP_{\Vme}^\perp$ has no zero diagonal entry;
we shall show this fact below in \cref{lemma:PiCheckHDiagonal}.
Because the conditional variance of $g^\MM_\alpha$ given $g^1, \ldots, g^{\MM-1}$ is $\sigma^2 (\PP_{\Vme}^\perp)_{\alpha\alpha}$, and because $\mathring \sigma > 0$ by assumption in this section, this implies that, a.s.\ for all large enough $n$,
\begin{equation}
\text{$g^\MM_\alpha$, conditioned on $g^1, \ldots, g^{\MM-1}$, has density for all $\alpha \in [n]$.}
\label{eqn:conditionalDistributionHasDensity}
\end{equation}
By ``has density'' here, we in particular mean that any Lesbegue measure zero set in $\R$ has zero probability under the conditional distribution of $g^\MM_\alpha$ given $g^1, \ldots, g^{\MM-1}$.

Now, assuming \cref{lemma:PiCheckHDiagonal}, we prove \ref{prop:nullAvoid} holds for $\coreset'$.

Let $\{U_{n\alpha} \sbe \R^{\coreset'}\}_{n\in\N, \alpha \in [n]}$ be a triangular array of Lesbegue measure zero sets.
For each $U_{n\alpha}$, define $B_{n\alpha} \defeq \{\vec x \in \R^{\coreset}: \lambda(U_{n\alpha}|_{\vec x}) \ne 0\}$, where $U_{n\alpha}|_{\vec x} = \{y \in \R: (\vec x, y) \in U_{n\alpha} \sbe \R^{\coreset} \times \R\}$ is the ``slice'' of $U_{n\alpha}$ at $\vec x$, and $\lambda$ is the 1-dimensional Lebesgue measure.
Because each $U_{n\alpha}$ has measure zero in $\R^{\coreset'}$, necessarily each $B_{n\alpha}$ also has measure zero in $\R^{\coreset}$.
Applying \ref{prop:nullAvoid} to the triangular array $\{B_{n\alpha} \sbe \R^{\coreset}\}_{n \in \N, \alpha \in [n]}$, we get that: a.s.\ for large enough $n$,
\[
\forall \alpha \in [n],\quad \{g^i_\alpha\}_{i \in \coreset} \not \in B_{n\alpha}.\]
Therefore, by \cref{eqn:conditionalDistributionHasDensity}, a.s.\ for large enough $n$,
\[
\forall \alpha \in [n],\quad \{g^i_\alpha\}_{i \in \coreset'} \not \in U_{n\alpha}.
\]
This finishes the proof of \ref{prop:nullAvoid} for $\coreset'$, and also \ref{IH:coreSet}$(\MM)$, save for \cref{lemma:PiCheckHDiagonal} below.

\begin{lemma}\label{lemma:PiCheckHDiagonal}
Almost surely, for large enough $n$, $\PP_{\Vme}^\perp$ has no zero diagonal entry.
\end{lemma}
\begin{proof}

WLOG, assume $\mathring{\Lambdame }$ is full rank.
Otherwise, by \cref{lemma:rankStability}(2), we can replace $\vme^1,\ldots, \vme^s$ by a linearly independent spanning set $\vme^{i_1}, \ldots, \vme^{i_k}$ such that 1) each $\vme^j$ is almost surely, for all large $n$, a linear combination of them and such that 2) their 2nd moment matrix is full rank in the limit.
Then the projection matrix associated to $\vme^{i_1}, \ldots, \vme^{i_k}$ is, almost surely, for all large $n$, the same as $\PP_{\Vme}$.

By the Sherman-Morrison formula (\cref{fact:ShermanMorrison}),
\begin{equation*}
(\PP_{\Vme})_{\alpha\alpha}
    =
        f\lp \f 1 n \check h_\alpha{}^\trsp \Lambdame_{-\alpha}^{-1} \check h_\alpha\rp
\end{equation*}
where $f(x) = x/(1+x)$, $\check h_\alpha$ is the column vector $(\vme^1_\alpha, \ldots, \vme^s_\alpha)^\trsp$,
and $\Lambdame_{-\alpha} = \f 1 n \sum_{\beta \ne \alpha} \check h_\beta \check h_\beta{}^\trsp$.
Thus, unless $\Lambdame_{-\alpha}$ is singular for some $\alpha$, all diagonal entries of $\PP_{\Vme}^\perp = I - \PP_{\Vme}$ are nonzero.
So it suffices to show that, 
\begin{equation*}
\text{a.s.\ for large enough $n$, \quad $\Lambdame_{-\alpha}$ is nonsingular for all $\alpha$.}
\end{equation*}

To do this, it pays to note that $\Lambdame_{-\alpha} = \Lambdame  - \f 1 n \check h_\alpha \check h_\alpha^\trsp$, so that
\[|\lambda_{\mathrm{min}}(\Lambdame_{-\alpha}) - \lambda_{\mathrm{min}}(\Lambdame )| \le \|\f 1 n \check h_\alpha \check h_\alpha^\trsp\|_{\mathrm{op}} = \f 1 n \check h_\alpha^\trsp \check h_\alpha.\]

By \cref{lemma:maxbound} below (which bounds the max by a high moment),
\[\max_{\alpha \in n} \f 1 n \check h_\alpha^\trsp \check h_\alpha \asto 0,\]
and consequently
\[\max_{\alpha \in [n]} |\lambda_{\mathrm{min}}(\Lambdame_{-\alpha}) - \lambda_{\mathrm{min}}(\Lambdame )| \asto 0.\]
Because $\Lambdame  \asto \mathring{\Lambdame }$, we know that, a.s.\ for large enough $n$, $\lambda_{\mathrm{min}}(\Lambdame )$ is bounded away from 0 by a constant (independent of $n$).
Altogether, this implies that all $\Lambdame_{-\alpha}$ are nonsingular, as desired.
\end{proof}

\begin{fact}[Sherman-Morrison formula]\label{fact:ShermanMorrison}
For any nonsingular matrix $A \in \R^{l \times l}$ and vector $a \in \R^l$,
we have
\[a^\trsp (A + aa^\trsp)^{-1} a = \f{a^\trsp A^{-1} a}{1 + a^\trsp \inv A a}.
\]
Consequently, for any full rank matrix $H$, the $\alpha$th diagonal entry of its associated projection matrix $\Pi_H = H (H^\trsp H)^{-1} H^\trsp$ can be written as
\[(\Pi_H)_{\alpha\alpha} = \f{H_\alpha (H_{-\alpha}^\trsp H_{-\alpha})^{-1} H_\alpha^\trsp}{1 + H_\alpha (H_{-\alpha}^\trsp H_{-\alpha})^{-1} H_\alpha^\trsp}
\]
where $H_\alpha$ is the $\alpha$th row of $H$, and $H_{-\alpha}$ is $H$ with the $\alpha$th row removed.
\end{fact}

\begin{lemma}\label{lemma:maxbound}
Assume \ref{IH:MomConv}$(\MM-1)$.
Suppose $\psi: \R^{\MM-1} \to \R$ is polynomially bounded.
Then as $n \to \infty,$
\[\f 1 {n^p} \max_{\alpha \in [n]} |\psi(g^1_\alpha, \ldots, g^{\MM-1}_\alpha)| \asto 0\]
for any $p > 0$.
\end{lemma}
\begin{proof}
For any $q > 0$, we have the elementary bound
\begin{equation*}
\max_{\alpha \in [n]} |\psi(g^1_\alpha, \ldots, g^{\MM-1}_\alpha)|
\le
\sqrt[q]{\sum_{\alpha \in [n]}
|\psi(g^1_\alpha, \ldots, g^{\MM-1}_\alpha)|^q}.
\end{equation*}
Thus, for any $q > 0$,
\begin{align*}
\f 1 {n^p} \max_{\alpha \in [n]} |\psi(g^1_\alpha, \ldots, g^{\MM-1}_\alpha)|
&\le
    \f 1 {n^{p-1/q}} 
    \sqrt[q]{\f 1 n
        \sum_{\alpha \in [n]}
        |\psi(g^1_\alpha, \ldots, g^{\MM-1}_\alpha)|^q}.
\end{align*}
Because, by \ref{IH:MomConv}$(\MM-1)$, $\f 1 n
        \sum_{\alpha \in [n]}
        |\psi(g^1_\alpha, \ldots, g^{\MM-1}_\alpha)|^q \asto C$ for some finite constant $C$ as $n \to \infty$,
the RHS above converges a.s.\ to 0 as soon as we take $q > 1/p$, and therefore so does the LHS.

\end{proof}

\subsection{Inductive Step: \ref{IH:MomConv}\texorpdfstring{$(\MM)$}{(m)}}
\label{sec:inductiveMoments}
In this section, we show
\begin{equation*}
\text{\ref{IH:MomConv}$(\MM-1)$ and \ref{IH:coreSet}$(\MM-1)$} \implies \text{\ref{IH:MomConv}}(\MM).
\end{equation*}

More specifically,
we will show that for any polynomially-bounded $\psi: \R^{\MM} \to \R$,
\begin{align*}
    \f 1 n \sum_{\alpha=1}^n \psi(g^1_\alpha, \ldots, g^{\MM}_\alpha) 
    \asto
    \EV \psi(\Zz^{g^1}, \ldots, \Zz^{g^\MM})
    .
\end{align*}
By \cref{lemma:limitSigmaIsZero}, if $\mathring \sigma = 0$, then almost surely, for large enough $n$, $g = g^\MM$ is just a (fixed) linear combination of $g^1, \ldots, g^{\MM-1}$, so \ref{IH:MomConv} is trivially true.
Therefore, in the below, we assume 
\begin{equation}
\mathring \sigma > 0.
\label{assm:mathringSigmaPositive}
\tag{$\star$}
\end{equation}
This assumption will be crucial for our arguments involving smoothness induced by Gaussian averaging.

\newcommand{\probA}{\mathsf{A}}
\newcommand{\probB}{\mathsf{B}}
\newcommand{\probC}{\mathsf{C}}
\newcommand{\probD}{\mathsf{D}}

\newcommand{\EVbr}[2][]{\EV_{#1}\left[#2\right]}
\newcommand{\EVcond}[2]{\EV\left[\left.#1\right| #2 \right]}
\newcommand{\extcom}[1]{{\color{blue}{#1}}}
\newcommand{\exttcom}[1]{{\color{olive}{#1}}}

To clarify notation in the following, we will write $\EVbr[X]{expression}$ to denote the expectation over only the randomness in $X$, $\EVcond{expression}{\Bb}$ to denote the expectation taken over all randomness except those in $\Bb$, and $\EV[expression]$ to denote expectation taken over \emph{all randomness} in $expression$.

\paragraph{Proof Plan}
By triangle inequality, we decompose
\begin{align}
        \left|\f 1 n \sum_{\alpha = 1}^n \psi(g^1_\alpha, \ldots, g^{\MM}_\alpha)
        - \EV \psi(\Zz^{g^1}, \ldots, \Zz^{g^\MM})
        \right|
    \le
        \probA + \probB + \probC
        \label{eqn:decompABC}
\end{align}
where, with $\omega, \sigma$ (\cref{eqn:meanvardef}), $\mathring \sigma$ (\cref{eqn:mathringsigma}), $\PP_{\Vme}$ (\cref{eqn:gCondTrick}), $\mathring \dme, \mathring\eme$ (\cref{eqn:zetaeta}) as in \cref{sec:inductiveSetup}, and with $z  \sim \Gaus(0, 1)$, we have defined
\begin{align*}
    \probA
        &\defeq
            \left|\f 1 n \sum_{\alpha = 1}^n \psi(g^1_\alpha, \ldots, g^{\MM}_\alpha)
            - \EV_z\psi\left(g^1_\alpha, \ldots, g^{\MM-1}_\alpha, \omega_\alpha + \sigma z \sqrt{(\PP^\perp_{\Vme})_{\alpha\alpha}}\right)\right|
            \\
    \probB
        &\defeq
            \left|
            \f 1 n \sum_{\alpha = 1}^n \EV_z\psi\left(g^1_\alpha, \ldots, g^{\MM-1}_\alpha, \omega_\alpha + \sigma z \sqrt{(\PP^\perp_{\Vme})_{\alpha\alpha}}\right)
            \right.
            \\
          &\phantomeq
            \phantom{\f 1 n \sum_{\alpha = 1}^n}\quad
            \left.
            -
            \EV_z
                {
                    \psi\lp
                        g^1_\alpha, \ldots, g^{\MM-1}_\alpha,
                        \sum_{i=1}^r \mathring {\dme}_i \xme^i_\alpha +
                        \sum_{j=1}^s \mathring {\eme}_j \vme^j_\alpha + \mathring \sigma z
                        \rp
                }
            \right|
            \\
    \probC
        &\defeq
            \left|
            \f 1 n \sum_{\alpha = 1}^n
            \EV_z
                {
                    \psi\lp
                        g^1_\alpha, \ldots, g^{\MM-1}_\alpha,
                        \sum_{i=1}^r \mathring {\dme}_i \xme^i_\alpha +
                        \sum_{j=1}^s \mathring {\eme}_j \vme^j_\alpha + \mathring \sigma z
                        \rp
                }
            -
            \EV \psi(\Zz^{g^1}, \ldots, \Zz^{g^\MM})
            \right|
            .
\end{align*}
Note that $\probB$ and $\probC$ are random variables in $\Bb$, but $\probA$ has additional randomness even after conditioning on $\Bb$.
We will show that each of $\probA, \probB, \probC$ goes to 0 almost surely, which would finish the proof of \cref{thm:NetsorTMasterTheorem}.

\paragraph{High Level Logic}
Roughly speaking,
\begin{description}
  \item[$\probA \asto 0$] because of a law of large numbers,
  \item[$\probB \asto 0$] because of the smoothness in $\EV_z \psi$ induced by Gaussian averaging, and
  \item[$\probC \asto 0$] by induction hypothesis.
\end{description}
We start by proving $\probC \asto 0$, since it's the easiest.

\subsubsection{\texorpdfstring{$\probC$}{C} Converges Almost Surely to 0}

In this section we show that $\probC \asto 0$ by a straightforward reduction to the induction hypothesis.

A simple inductive argument with \cref{defn:Z} shows that $Z^{\xme^1}, \ldots, Z^{\xme^r}, Z^{\vme^1}, \ldots, Z^{\vme^s}$ are all deterministic, polynomially-bounded functions of $Z^{g^1}, \ldots, Z^{g^{\MM-1}}$.
Thus we may define the function $\Psi: \R^{\MM-1} \to \R$ by
\[
\Psi(\Zz^{g^1}, \ldots, \Zz^{g^{\MM-1}})
  \defeq
    \EV_{z \sim \Gaus(0, 1)}
      \psi\lp \Zz^{g^1}, \ldots, \Zz^{g^{\MM-1}},
        \sum_{i=1}^r \mathring{\dme}_i \Zz^{\xme^i}
        + \sum_{j=1}^s \mathring{\eme}_j \Zz^{\vme^j}
        + \mathring \sigma z
      \rp
      .
\]
The function $\Psi$ is polynomially bounded since $\psi$ is, and
$Z^{\xme^1}, \ldots, Z^{\xme^r}, Z^{\vme^1}, \ldots, Z^{\vme^s}$ are  polynomially-bounded functions of $Z^{g^1}, \ldots, Z^{g^{\MM-1}}$.
Observe that the function that expresses $Z^{\xme^i}$ in terms of $Z^{g^1}, \ldots, Z^{g^{\MM-1}}$ is the same function that expresses $\xme^i_\alpha$ in terms of $g^1_\alpha, \ldots, g^{\MM-1}_\alpha$ for all $\alpha \in[n]$; likewise for $Z^{\vme^j}$ and $\vme^j_\alpha$.
Applying the induction hypothesis \ref{IH:MomConv}($\MM-1$) to $\Psi$, we obtain
\begin{align*}
    &\phantomeq
        \f 1 n \sum_{\alpha = 1}^n
            \EV_z
                {
                    \psi\lp
                        g^1_\alpha, \ldots, g^{\MM-1}_\alpha,
                        \sum_{i=1}^r \mathring {\dme}_i \xme^i_\alpha +
                        \sum_{j=1}^s \mathring {\eme}_j \vme^j_\alpha + \mathring \sigma z
                        \rp
                }
        \\
    &=
        \f 1 n \sum_{\alpha = 1}^n
                {
                    \Psi\lp
                        g^1_\alpha, \ldots, g^{\MM-1}_\alpha
                        \rp
                }
        \\
    &\asto
        \EV \Psi(\Zz^{g^1}, \ldots, \Zz^{g^{\MM-1}})
        \\
    &\pushright{
      \text{by \ref{IH:MomConv}$(\MM-1)$}
      }
        \\
    &=
        \EV
        \psi\lp
          \Zz^{g^1}, \ldots, \Zz^{g^{\MM-1}},
          \sum_{i=1}^r \mathring {\dme}_i \Zz^{\xme^i} +
            \sum_{j=1}^s \mathring {\eme}_j \Zz^{\vme^j} + \mathring \sigma z
          \rp
          .
\end{align*}
By \cref{lemma:YgConditionalDistribution} below, this is precisely
\begin{align*}
\EV \psi\lp
            \Zz^{g^1}, \ldots, \Zz^{g^{\MM-1}}, \Zz^{g^{\MM}}\rp
\end{align*}
as desired.

\begin{lemma}\label{lemma:YgConditionalDistribution}
  Let $\prvZ$ be the $\sigma$-algebra generated by $\Zz^{g^1}, \ldots, \Zz^{g^{\MM-1}}$.
  Then for $z \sim \Gaus(0, 1)$ sampled independently of $\prvZ$,
  \begin{align*}
  \Zz^{g}
    &\disteq_{\prvZ}
      \mathring \sigma z +
      \sum_{i=1}^r \mathring {\dme}_i \Zz^{\xme^i} +
      \sum_{j=1}^s \mathring {\eme}_j \Zz^{\vme^j}
  \end{align*}
  where $\mathring{\dme}$ and $\mathring{\eme}$ are as in \cref{eqn:zetaeta} and $\mathring \sigma$ is as in \cref{eqn:mathringsigma}.

\end{lemma}
\begin{proof}
  We can split each $ Z^{\xme^i}$ into $\Zhat^{\xme^i} + \Zdot^{\xme^i}$, to obtain
  \begin{align*}
  &\phantomeq
    \mathring \sigma z +
      \sum_{i=1}^r \mathring {\dme}_i  Z^{\xme^i} +
      \sum_{j=1}^s \mathring {\eme}_j  Z^{\vme^j}
    \\
  &=
    \left(\mathring \sigma z +
      \sum_{i=1}^r \mathring {\dme}_i \Zhat^{\xme^i}
      \right)
    +
    \left(
      \sum_{i=1}^r \mathring {\dme}_i \Zdot^{\xme^i}
      +
      \sum_{j=1}^s \mathring {\eme}_j  Z^{\vme^j}
    \right)
    \\
  &=
    \left(\mathring \sigma z +
      \sum_{i=1}^r \mathring {\dme}_i \Zhat^{\xme^i}
    \right)
    +
    \Zdot^{g}
  &{\text{by \cref{lemma:hpartReExpression}}}
    \\
  &\disteq_\prvZ
    \Zhat^{g}
    +
    \Zdot^{g}
  &{\text{see below for justification}}
    \\
  &=
     Z^{g},
  \end{align*}
  as desired.
  It remains to justify the second-to-last equality, which is easily done via the usual formula for Gaussian conditioning (\cref{prop:GaussianCondition}):
  Let $\Zhat$ be the column vector $(\Zhat^{\xme^1}, \ldots, \Zhat^{\xme^r})^\trsp$.
  Then we know that $\Zhat$ and $\Zhat^{g}$ are jointly Gaussian with zero mean%
  \footnote{recall we have assumed WLOG that $\EV Z^g = 0$ for all $g \in \mathcal V$; see discussion at the beginning of \cref{sec:proofMainTheorem}.}%
  .
  The covariance between $\Zhat$ and $\Zhat^{g}$ is $\sigma_A^2 \mathring\gammame$ and $\Zhat$ has covariance matrix $\sigma_A^2 \mathring\Upsilonme$ by \cref{eqn:limitMatrices}.
  Then by \cref{prop:GaussianCondition}, we know that, conditioned on $\prvZ$, $\Zhat^{g}$ is distributed as a Gaussian with mean
  \[\sigma_A^2 \mathring\gammame^\trsp (\sigma_A^2 \mathring\Upsilonme)^+ \Zhat
  =\mathring\dme^\trsp \Zhat
  =\sum_{i=1}^r \mathring {\dme}_i \Zhat^{\xme^i},\]
  by \cref{eqn:zetaeta}
  and variance
  \[
    \EV (Z^{g})^2 - \sigma_A^2 \mathring\gammame^\trsp (\sigma_A^2 \mathring\Upsilonme)^+ \sigma_A^2 \mathring\gammame
    = \mathring\sigma\]
  by \cref{lemma:sigmaConverges}.
  This is exactly what we needed.
\end{proof}

\subsubsection{\texorpdfstring{$\probA$}{A} Converges Almost Surely to 0}
\label{sec:probA}

\newcommand{\rvec}{z}

In this section we show $\probA \asto 0.$

For each $\alpha \in [n]$, define the function $\psi_\alpha: \R \to \R$ by
 \[\psi_\alpha (x) \defeq \psi(g^1_\alpha, \ldots, g^{\MM-1}_\alpha, \omega_\alpha + \sigma x),\]
with $\omega$ and $\sigma$ defined in \cref{eqn:meanvardef}.
This is a random function depending on the random vectors $g^1_\alpha, \ldots, g^{\MM-1}_\alpha$, and it changes with $n$ as well.
We also consider the ``centered version'' $\tilde \psi_\alpha: \R \to \R$ of $\psi_\alpha,$
\[\tilde \psi_\alpha(x) \defeq \psi_\alpha(x) - \EV_{y}\psi_\alpha(y)\]
with expectation taken over $y \sim \Gaus(0, (\PP^\perp_{\Vme})_{\alpha\alpha})$ (but not $g^1_\alpha, \ldots, g^{\MM-1}_\alpha$).
Note by \cref{eqn:gConditionedOnB},
\[ \probA \disteq_\Bb \left|\f 1 n \sum_{\alpha =1}^n \tilde \psi_\alpha(\rvec_\alpha)\right|\]
where $\rvec \sim \Gaus(0, \PP_{\Vme}^\perp)$.

\paragraph{Proof idea}
To prove our claim, we will show that, for almost all (i.e. probability 1 in the probability space $\basespace$ defined in the beginning of \cref{sec:proofMainTheorem}) sequences of $(g^1, \ldots, g^{\MM-1}) = (g^1(n), \ldots, g^{\MM-1}(n))$ in $n$ --- which we shall call \emph{amenable sequences of $g^1, \ldots, g^{\MM-1}$} --- 
we have a moment bound
\begin{equation}
\EV[\probA^{2\lambda} | \Bb] = 
\EV_{\rvec \sim \Gaus(0, \PP^\perp_{\Vme})} 
    \lp \f 1 n \sum_{\alpha =1}^n \tilde \psi_\alpha(\rvec_\alpha)\rp^{2\lambda}
    < C n^{-1.25}
    \label{eqn:highMomentBound}
\end{equation}
for some large $\lambda$ and some constant $C > 0$ depending only on $\lambda$ and the particular sequence of $\{(g^1(n), \ldots, g^{\MM-1}(n))\}_n$.
Then we apply \cref{lemma:momentBoundASConvergence} to show that, conditioned on any amenable sequence, $\probA$ converges to 0 almost surely over all randomness remaining after conditioning.
Since almost all sequences are amenable, this shows that the convergence is also almost sure without the conditioning.

\paragraph{The moment bound}
For $\lambda \ge 6$ and any $q > 1$, we first apply \cref{thm:controlHighMoments} to get the bound
\begin{equation*}
\EV_{\rvec} 
    \lp \f 1 n \sum_{\alpha =1}^n \tilde \psi_\alpha(\rvec_\alpha)\rp^{2\lambda}
    \le c n^{-1.5 + 1/q}
        \sqrt[q]{\f 1 n \sum_{\alpha=1}^n \EV \tilde \psi_\alpha(\rvec_\alpha)^{2\lambda q}}
\end{equation*}
where on both sides $\rvec \sim \Gaus(0, \PP^\perp_{\Vme})$, and $c$ is a constant depending only on $\lambda$ and $\MM$, but not on $n$, the functions $\psi_\alpha$, or $g^1, \ldots, g^{\MM-1}$.
To obtain \cref{eqn:highMomentBound}, we will show that
\begin{equation}
\f 1 n \sum_{\alpha=1}^n \EV \tilde \psi_\alpha(\rvec_\alpha)^{2\lambda q}
\label{eqn:moment}
\end{equation}
is uniformly bounded (in $n$), almost surely over the randomness of the sequences $\{g^1(n), \ldots, g^{\MM -1}(n)\}_n$.
We take all such sequences to be the \emph{amenable sequences}.
For $q > 4$, we then get the desired moment bound \cref{eqn:highMomentBound}.

It remains to show the almost sure uniform boundedness.

\paragraph{Almost sure uniform boundedness}
Intuitively, \cref{eqn:moment} should converge almost surely to a deterministic value by applying some version of the induction hypothesis, so it should be almost surely uniformly bounded in $n$.
The obstacle is that the variance $(\PP_{\Vme}^\perp)_{\alpha\alpha}$ of $\rvec_\alpha$ depends not only on $g^1_\alpha, \ldots, g^{\MM-1}_\alpha$, but also on $g^1_\beta, \ldots, g^{\MM-1}_\beta$ for other indices $\beta \not= \alpha$ as well.
So \textit{a priori} it is not clear how to apply the \ref{IH:MomConv}($\MM-1$) in a straightforward way.
We thus first process \cref{eqn:moment} a bit.
Let
\[\mu_\alpha \defeq \EV_{x \sim \Gaus(0, (\PP^\perp_{\Vme})_{\alpha\alpha})} \psi_\alpha(x).\]
Then, abbreviating $\EV_z$ for expectation taken over $\rvec \sim \Gaus(0, \PP^\perp_{\Vme})$, we have the following inequalities of random variables in $\Bb$:
\begin{align*}
  \f 1 n \sum_{\alpha=1}^n \EV_z \tilde \psi_\alpha(\rvec_\alpha)^{2\lambda q}
  &=
      \f 1 n \sum_{\alpha=1}^n \EV_z (\psi_\alpha(\rvec_\alpha) - \mu_\alpha)^{2\lambda q}
      \\
  &\le
      \f 1 n 2^{2\lambda q - 1}
          \sum_{\alpha=1}^n
              \EV_z 
                  \left[
                      \psi_\alpha(\rvec_\alpha)^{2\lambda q} + \mu_\alpha^{2\lambda q}
                  \right]
  &{\text{by \cref{lem:powerbound}}}
      \\
  &\le
      \f 1 n 2^{2\lambda q}
          \sum_{\alpha=1}^n
              \EV_z 
                  \psi_\alpha(\rvec_\alpha)^{2\lambda q}
  &\text{see below}
      \\
  &=
      \f 1 n 2^{2\lambda q}
      \sum_{\alpha=1}^n
          \EV_z 
              \psi(g^1_\alpha, \ldots, g^{\MM-1}_\alpha, \omega_\alpha + \sigma \rvec_\alpha )^{2\lambda q}
      ,
\end{align*}
where in the second inequality we applied power mean inequality $\mu_\alpha \le \sqrt[2\lambda q]{\EV_z\psi_\alpha(\rvec_\alpha)^{2\lambda q}}$.
Suppose, WLOG, that $\psi$ is polynomially bounded by an inequality $|\psi(x)| \le C \|x\|^p_p + c$ for some $p, C, c > 0$.
In the below, we will silently introduce constants $C_1, C_2, \ldots$ via \cref{lem:powerbound} and merge with old constants, such that they will only depend on $\lambda, p, q$.
Continuing the chain of inequalities above
\begin{align*}
  \f 1 n \sum_{\alpha=1}^n \EV_z \tilde \psi_\alpha(\rvec_\alpha)^{2\lambda q}
  &\le
      c + 
      \f 1 n C 2^{2\lambda q}
      \sum_{\alpha=1}^n
          \EV_z 
              \lp 
                  |g^1_\alpha|^p + \ldots + |g^{\MM-1}_\alpha|^p + |\omega_\alpha + \sigma \rvec_\alpha|^p
              \rp^{2\lambda q}
      \\
  &\le
      c + 
      \f 1 n C_1
      \sum_{\alpha=1}^n
          \EV_z 
              \lp 
                  |g^1_\alpha|^p + \ldots + |g^{\MM-1}_\alpha|^p + |\omega_\alpha|^p + |\sigma \rvec_\alpha|^p
              \rp^{2\lambda q}
      \\
  &\le
      c +
      \f 1 n C_2
      \sum_{\alpha=1}^n
        \EV_z 
          |g^1_\alpha|^{2 \lambda q p} + \ldots + |g^{\MM-1}_\alpha|^{2 \lambda q p} + |\omega_\alpha|^{2 \lambda q p}
          + |\sigma \rvec_\alpha|^{2 \lambda q p}
      .
      \numberthis \label{eqn:momentUpperBoundDecomposition}
  \end{align*}
  
  We now proceed to show that the summands of \cref{eqn:momentUpperBoundDecomposition} are almost surely uniformly bounded, which finishes our proof of $\probA \asto 0$.
  
  \begin{itemize}
      \item
          By \ref{IH:MomConv}($\MM-1$),
          \[
          \f 1 n
              \sum_{\alpha=1}^n
                  \EV_z 
                      |g^1_\alpha|^{2 \lambda q p} + \ldots + |g^{\MM-1}_\alpha|^{2 \lambda q p}
          \]
          almost surely converges to a deterministic value, so it is almost surely uniformly bounded in $n$.
  
      \item
          In addition, $\sigma \asto \mathring \sigma$, so that, almost surely, for large enough $n$, we have $\sigma \le \mathring \sigma + 1$.
          (The order of the qualifiers is important here; in general this statement cannot be made uniformly in $n$).
          Therefore, \emph{almost surely, for large enough $n$},
          \begin{align*}
          \f 1 n
              \sum_{\alpha=1}^n
              \EV_z |\sigma \rvec_\alpha|^{2 \lambda q p}
          &\le
              \f 1 n
              \sum_{\alpha=1}^n
                  |\mathring \sigma + 1|^{2\lambda q p}
                  \EV_z |\rvec_\alpha|^{2 \lambda q p}
                  .
          \end{align*}
          This is almost surely uniformly bounded in $n$ because $\Var(\rvec_\alpha) = (\PP_{\Vme}^\perp)_{\alpha\alpha} \in [0, 1]$ for all $\alpha$ by \cref{lemma:projectionDiagonal}, so that $\EV_z |\rvec_\alpha|^{2 \lambda q p}$ (which is purely a monotonic function of $\Var(z_\alpha)$) is bounded as well.
      \item
          It remains to bound $\f 1 n \sum_{\alpha=1}^n |\omega_\alpha|^{2\lambda q p}$.
          We extract our reasoning here into the \cref{lemma:omegaAlphaUnifomBounded} below, as we will need to reuse this for later.
          This finishes the proof of $\probA\asto 0.$
  
  \end{itemize}

\begin{lemma}\label{lemma:omegaAlphaUnifomBounded}
  For any polynomially bounded function $\varphi: \R \to \R,$
  \[
  \f 1 n \sum_{\alpha=1}^n |\varphi(\omega_\alpha)|
  \]
  is almost surely uniformly bounded in $n$.
\end{lemma}
\begin{proof}
  It suffices to prove this for $\varphi(x) = |x|^d$ for any $d > 0$.
  
  Expanding $\omega$ according to \cref{lemma:omegaExpansion}, we get
  \begin{align*}
      \f 1 n \sum_{\alpha=1}^n |\omega_\alpha|^{d}
      &=
          \f 1 n \sum_{\alpha=1}^n \left|
              \sum_{i=1}^r
                  \xme_\alpha^i (\mathring {\dme}_i + \varepsilonhat_i)
              + \sum_{j=1}^s
                  \vme_\alpha^j (\mathring{\eme}_j + \varepsiloncheck_j)
          \right|^d
  \end{align*}
  for (fixed dimensional) $\varepsilonhat \in \R^r, \varepsiloncheck \in \R^s$ that go to 0 almost surely with $n$.
  Applying \cref{lem:powerbound}, we get
  \begin{align*}
      \f 1 n \sum_{\alpha=1}^n |\omega_\alpha|^d
      &\le
          \f 1 n C_3\sum_{\alpha=1}^n
              \left|
                  \sum_{i=1}^r
                  \xme_\alpha^i \mathring {\dme}_i
              \right|^d
              +
              \left|
                  \sum_{j=1}^s
                  \vme_\alpha^j \mathring {\eme}_i
              \right|^d
              +
              \left|
                  \sum_{i=1}^r
                  \xme_\alpha^i \varepsilonhat_i
              \right|^d
              +
              \left|
                  \sum_{j=1}^s
                  \vme_\alpha^j \varepsiloncheck_j
              \right|^d
              .
  \end{align*}
  We bound each summand separately.
  \begin{itemize}
      \item
          By induction hypothesis, 
          \begin{align*}
              \f 1 n \sum_{\alpha=1}^n
              \left|
                  \sum_{i=1}^r
                  \xme_\alpha^i \mathring {\dme}_i
              \right|^d
          \end{align*}
          converges a.s.\ to a deterministic value, so it is a.s uniformly bounded in $n.$
      \item
          By the a.s.\ decaying property of $\varepsilonhat$, we have
          almost surely, for large enough $n$, $|\sum_{i=1}^r \xme_\alpha^i \varepsilonhat_i| \le \sum_{i=1}^r |\xme_\alpha^i|$ (again, the order of qualifier is very important here).
          By induction hypothesis,
          \begin{align*}
          \f 1 n \sum_{\alpha=1}^n
              \lp \sum_{i=1}^r |\xme_\alpha^i|\rp^d
          \end{align*}
          converges a.s.\ to a deterministic value, yielding the a.s.\ uniform-boundedness of it and of
          \begin{align*}
          \f 1 n \sum_{\alpha=1}^n
              \left| \sum_{i=1}^r \xme_\alpha^i \varepsilonhat_i \right|^d.
          \end{align*}
      \item
          Likewise, because for each $j$, $\vme^j$ is a polynomially-bounded function of $g^1, \ldots, g^{\MM-1}$\ \footnote{This is the most crucial place where we need the assumption that all nonlinearities in the program are polynomially bounded. Otherwise, for faster growing functions, the compositions of such nonlinearities might not be integrable against the Gaussian measure}, the summands of
          \begin{align*}
          \f 1 n \sum_{\alpha=1}^n
              \lp \sum_{j=1}^s |\vme_\alpha^j|\rp^d
          \quad\text{and}\quad
          \f 1 n \sum_{\alpha=1}^n
                      \left|
                          \sum_{j=1}^s
                          \vme_\alpha^j \mathring{\eme}_j
                      \right|^d
          \end{align*}
          are polynomially-bounded functions of $g^1, \ldots, g^{\MM-1}$ too.
          So by induction hypothesis, these sums converge a.s., implying the a.s. uniform-boundedness of them and of
          \begin{align*}
          \f 1 n \sum_{\alpha=1}^n
                      \left|
                          \sum_{j=1}^s
                          \vme_\alpha^j \varepsiloncheck_j
                      \right|^d
              .
         \end{align*}
  \end{itemize}
\end{proof}

\subsubsection{\texorpdfstring{$\probB$}{B} Converges Almost Surely to 0}
\label{sec:probB}
\newcommand{\JJc}{J}
\newcommand{\JJ}{\bar{\JJc}}

In this section we show  $\probB \asto 0.$

\paragraph{Some Notations}
For brevity, we will set $d_\alpha \defeq (\PP^\perp_{\Vme})_{\alpha\alpha}$.
In addition, for each $\alpha \in [n]$, $w \in \R$, $\tau \ge 0$, define the function $\Psi_\alpha(-; -): \R \times \R^{\ge 0} \to \R$ by
\[\Psi_\alpha(w; \tau^2) \defeq
            \EV_{z\sim\Gaus(0, 1)}
            \psi\lp
                g^1_\alpha, \ldots, g^{\MM-1}_\alpha, w + \tau z
                \rp.
\]
(Here and in all that follows, $\tau^2$ is the square of $\tau$, and the $2$ is not an index).
This is a random function, with randomness induced by $g^1, \ldots, g^{\MM-1}$.

\paragraph{Our proof idea} is to write
\begin{align*}
\probB
    &=
        \left| \f 1 n \sum_{\alpha=1}^n
            \Psi_\alpha\lp \omega_\alpha; \sigma^2 d_\alpha \rp
            -
            \Psi_\alpha\lp
              \sum_{i=1}^r \mathring {\dme}_i \xme^i_\alpha +
                \sum_{j=1}^s \mathring {\eme}_j \vme^j_\alpha
              ; \mathring \sigma^2 \rp
        \right|
        \\
    &\le
        \f 1 n
        \sum_{\alpha\in \JJ}
        \left|\Psi_\alpha\lp \omega_\alpha; \sigma^2 d_\alpha\rp\right|
        +
        \left|\Psi_\alpha\lp
              \sum_{i=1}^r \mathring {\dme}_i \xme^i_\alpha +
                \sum_{j=1}^s \mathring {\eme}_j \vme^j_\alpha
                ;
                \mathring \sigma^2
              \rp
        \right|
        \numberthis \label{eqn:smalldiagonal}
        \\
    &\qquad
        +
        \f 1 n \sum_{\alpha\in \JJc}
        \left|
        \Psi_\alpha\lp \omega_\alpha; \sigma^2 d_\alpha\rp
        -
        \Psi_\alpha\lp
              \sum_{i=1}^r \mathring {\dme}_i \xme^i_\alpha +
                \sum_{j=1}^s \mathring {\eme}_j \vme^j_\alpha
                ; \mathring \sigma^2
              \rp
        \numberthis \label{eqn:largediagonal}
        \right|
\end{align*}
where $\JJc \sqcup \JJ = [n]$ is a partition of $[n]$ with $\JJc \defeq \{\alpha: d_\alpha \ge 1/2\}$ and $\JJ$ is its complement.
Note that $|\JJ| \le 2 \rank \Vme \le 2 s$ is uniformly bounded in $n$.
We then show each summand of \cref{eqn:smalldiagonal} goes to 0 a.s.\ individually.
Finally we use the smoothness of $\Psi_\alpha$ (\cref{eqn:PsiSmoothness}) induced by the Gaussian averaging in $z$ to show each summand of \cref{eqn:largediagonal} is almost surely $o(1)$, finishing the proof.

\paragraph{\cref{eqn:smalldiagonal} converges to 0 a.s.}
We first look at the term
\begin{align*}
\f 1 n
\sum_{\alpha\in \JJ}
\left|\Psi_\alpha\lp
            \sum_{i=1}^r \mathring {\dme}_i \xme^i_\alpha +
              \sum_{j=1}^s \mathring {\eme}_j \vme^j_\alpha
              ;
              \mathring \sigma^2
            \rp
\right|
&\le
    \f{|\JJ|}n \max_{\alpha \in [n]}
        \left|\Psi_\alpha\lp 
              \sum_{i=1}^r \mathring {\dme}_i \xme^i_\alpha +
                \sum_{j=1}^s \mathring {\eme}_j \vme^j_\alpha
                ; \mathring \sigma^2 \rp\right|
    \\
&\le
    \f{2s}{n^{1-1/q}} \sqrt[q]{\f 1 n \sum_{\alpha \in [n]} \left|\Psi_\alpha\lp
              \sum_{i=1}^r \mathring {\dme}_i \xme^i_\alpha +
                \sum_{j=1}^s \mathring {\eme}_j \vme^j_\alpha
              ; \mathring \sigma^2
            \rp
          \right|^q}
    \numberthis\label{eqn:smalldiagonalEasy}
\end{align*}
for any $q > 0$.
Here we used  $|\JJ| \le 2 \rank \Vme \le 2 s$ as noted above.
Now $\left|\Psi_\alpha\lp 
\sum_{i=1}^r \mathring {\dme}_i \xme^i_\alpha +
\sum_{j=1}^s \mathring {\eme}_j \vme^j_\alpha;
\mathring \sigma^2 \rp\right|^q$
is a fixed (independent of $\alpha$ and $n$) polynomially-bounded function of $g^1_\alpha, \ldots, g^{\MM-1}_\alpha$, so by induction hypothesis, 
\[\f 1 n \sum_{\alpha \in [n]} \left|\Psi_\alpha\lp
  \sum_{i=1}^r \mathring {\dme}_i \xme^i_\alpha +
    \sum_{j=1}^s \mathring {\eme}_j \vme^j_\alpha
; \mathring \sigma^2 \rp\right|^q\]
is a.s. uniformly bounded in $n$, so that using a large $q \ge 2$, we see \cref{eqn:smalldiagonalEasy} converges a.s. to 0.

Next, we apply a similar reasoning to the other term and obtain
\begin{align*}
\f 1 n
\sum_{\alpha\in \JJ}
\left|\Psi_\alpha\lp \omega_\alpha; \sigma^2 d_\alpha\rp\right|
&\le
    \f{2s}{n^{1-1/q}} \sqrt[q]{\f 1 n \sum_{\alpha \in [n]}
    \left|\Psi_\alpha\lp \omega_\alpha; \sigma^2 d_\alpha\rp\right|^q
    }
\end{align*}
We in fact already know that
\[\f 1 n \sum_{\alpha \in [n]}
    \left|\Psi_\alpha\lp \omega_\alpha; \sigma^2 d_\alpha\rp\right|^q\]
is a.s. uniformly bounded in $n$
from \cref{eqn:momentUpperBoundDecomposition} in \cref{sec:probA}, so that 
\[
\f 1 n
\sum_{\alpha\in \JJ}
\left|\Psi_\alpha\lp \omega_\alpha; \sigma^2 d_\alpha\rp\right|
\asto 0\]
from which follows the same for \cref{eqn:smalldiagonal}.

\paragraph{\cref{eqn:largediagonal} converges to 0 a.s.}

As mentioned above, to prove this we will use the following smoothness bound of $\Psi_\alpha$, whose proof will be delayed to the end of the section.
Suppose, WLOG, that the polynomially boundedness of $\psi$ presents itself in an inequality $|\psi(x)| \le C \|x\|^p_{p} + C$, for some $p, C > 0$, where $p$ is an integer.
This $p$ will appear explicitly in this smoothness bound below.

\begin{lemma}[Smoothness of $\Psi_\alpha$]\label{lemma:PsiAlphaSmoothnessBound}
Let $w, \Delta w \in \R, \tau^2, \Delta \tau^2 \in \R^{\ge 0}$.
Then
\begin{align*}
&\left| \Psi_\alpha(w+\Delta w ; \tau^2 + \Delta \tau^2) - \Psi_\alpha(w; \tau^2)\right|
\\
&\quad\quad\quad\le
    R
    (|\Delta w| + \Delta \tau^2)
    (1 + \tau^{-2})
        \lp S_\alpha + |w|^p + |\Delta w|^p + \tau^p + (\Delta\tau^2)^{p/2} \rp
    \numberthis
    \label{eqn:PsiSmoothness}
\end{align*}
for some constant $R > 0$, and where
\[S_\alpha \defeq 1 + |g^1_\alpha|^{p} + \cdots + |g^{\MM-1}_\alpha|^p.\]
\end{lemma}

To bound \cref{eqn:largediagonal}, first we expand
\[
\omega_\alpha =
\sum_{i=1}^r \xme_\alpha^i (\mathring {\dme}_i + \hat \epsilon_i)
+ \sum_{j=1}^s \vme_\alpha^j (\mathring {\eme}_j + \check \epsilon_j)
\]
where, by \cref{lemma:omegaExpansion}, $\hat \epsilon \in \R^r, \check \epsilon \in \R^s$ are vectors that go to 0 almost surely with $n$.
Then we apply the smoothness bound \cref{eqn:PsiSmoothness} to get, for each $\alpha \in \JJc$
\begin{align*}
    \left|
    \Psi_\alpha\lp \omega_\alpha; \sigma^2 d_\alpha\rp
    -
    \Psi_\alpha\lp
      \sum_{i=1}^r \mathring {\dme}_i \xme^i_\alpha +
        \sum_{j=1}^s \mathring {\eme}_j \vme^j_\alpha
        ; \mathring \sigma^2 \rp
    \right|
    &\le
        R
        \lp 1 + \min(\sigma^2 d_\alpha, \mathring \sigma^2)^{-1} \rp
        X_\alpha
        Y_\alpha
        \\
    &\le
        R
        \lp 1 + \min(\sigma^2/2, \mathring \sigma^2)^{-1} \rp
        X_\alpha
        Y_\alpha
\end{align*}
using the definition of $\JJc$ that $d_\alpha \ge 1/2, \forall \alpha \in \JJc$.
Here we have defined
\begin{align*}
X_\alpha
    &\defeq
        |\omega_\alpha - 
          \sum_{i=1}^r \mathring {\dme}_i \xme^i_\alpha -
            \sum_{j=1}^s \mathring {\eme}_j \vme^j_\alpha|
        + |\sigma^2 d_\alpha - \mathring \sigma^2|
        \\
    &=
        \left|\sum_{i=1}^r \xme_\alpha^i \hat \epsilon_i
                + \sum_{j=1}^s \vme_\alpha^j \check \epsilon_j
        \right|
        + |\sigma^2 d_\alpha - \mathring \sigma^2|
        \\
Y_\alpha
    &\defeq
        S_\alpha + |\omega_\alpha|^p
        + \left|\sum_{i=1}^r \xme_\alpha^i \hat \epsilon_i
                + \sum_{j=1}^s \vme_\alpha^j \check \epsilon_j
        \right|^p
        + \max(\sigma^2 d_\alpha, \mathring \sigma^2)^{p/2}
        + |\sigma^2 d_\alpha - \mathring \sigma^2|^{p/2}
        .
        \\
\end{align*}
Thus,
\begin{align*}
\cref{eqn:largediagonal}
    &=
        \f 1 n \sum_{\alpha\in \JJc}
        \left|
        \Psi_\alpha\lp \omega_\alpha; \sigma^2 d_\alpha\rp
        -
        \Psi_\alpha\lp
          \sum_{i=1}^r \mathring {\dme}_i \xme^i_\alpha +
            \sum_{j=1}^s \mathring {\eme}_j \vme^j_\alpha;
          \mathring \sigma^2 \rp
        \right|
        \\
    &\le
        R
        \f 1 n
        \lp 1 + \min(\sigma^2/2, \mathring \sigma^2)^{-1} \rp
        \sum_{\alpha \in \JJc}
            X_\alpha
            Y_\alpha
        \\
    &\le
        R
        \lp 1 + \min(\sigma^2/2, \mathring \sigma^2)^{-1} \rp
        \sqrt{\f 1 n \sum_{\alpha \in \JJc}
            X_\alpha^2}
        \sqrt{\f 1 n \sum_{\alpha \in \JJc}
            Y_\alpha^2
            }
        .
\end{align*}
Since $\sigma \asto \mathring \sigma$ and we have assumed $\mathring \sigma > 0$ by \cref{assm:mathringSigmaPositive}, we have $\lp 1 + \min(\sigma^2/2, \mathring \sigma^2)^{-1} \rp$ is almost surely uniformly bounded in $n$.

Therefore, \cref{eqn:largediagonal} can be shown to converge a.s.\ to 0 if we show
\begin{align*}
&\sqrt{\f 1 n \sum_{\alpha \in \JJc} Y_\alpha^2} \quad\text{is a.s.\ uniformly bounded in $n$, and}\\
&\sqrt{\f 1 n \sum_{\alpha \in \JJc} X_\alpha^2} \asto 0
\end{align*}

We prove these two claims in \cref{lemma:YAlphaUnifBounded,lemma:XalphaAsto0} below, which would finish our proof of $\probB \asto 0$, and of our main theorem \cref{thm:NetsorTMasterTheorem} as well.

\begin{lemma}\label{lemma:XalphaAsto0}
$\sqrt{\f 1 n \sum_{\alpha \in \JJc} X_\alpha^2} \asto 0$.
\end{lemma}
\begin{proof}
    Note that
    \begin{align*}
    X_\alpha
        &\le
            \left|\sum_{i=1}^r \xme_\alpha^i \hat \epsilon_i
                    + \sum_{j=1}^s \vme_\alpha^j \check \epsilon_j
            \right|
            + |\mathring \sigma^2 - \sigma^2|
            + |\sigma^2 - \sigma^2 d_\alpha|
            \\
        &\defeq
            P_\alpha + Q_\alpha + R_\alpha
            .
    \end{align*}
    Then by triangle inequality (in $\ell_2$-norm),
    \begin{align*}
    \sqrt{\f 1 n \sum_{\alpha \in \JJc} X_\alpha^2}
        &\le
            \sqrt{\f 1 n \sum_{\alpha \in \JJc} P_\alpha^2}
            + \sqrt{\f 1 n \sum_{\alpha \in \JJc} Q_\alpha^2}
            + \sqrt{\f 1 n \sum_{\alpha \in \JJc} R_\alpha^2}
            .
    \end{align*}
    We now show that each term above converges a.s.\ to 0, which would finish the proof of \cref{lemma:XalphaAsto0}.
    \begin{itemize}
    \item
        Because $\hat \epsilon \asto 0$ and $\check \epsilon \asto 0$, we have, for some constant $C > 0$,
        \begin{align*}
        \f 1 n \sum_{\alpha \in \JJc} P_\alpha^2
            &\le
                C \f 1 n \sum_{\alpha \in \JJc} \lp
                    \sum_{i=1}^r (\xme_\alpha^i \hat \epsilon_i)^2
                    + \sum_{j=1}^s (\vme_\alpha^j \check \epsilon_j)^2
                    \rp
                \\
            &\le
                C \max_{i,j}\{|\hat \epsilon_i|, |\check \epsilon_j|\} \times 
                \f 1 n \sum_{\alpha \in \JJc} \lp
                    \sum_{i=1}^r (\xme_\alpha^i)^2
                    + \sum_{j=1}^s (\vme_\alpha^j)^2
                    \rp
                \\
            &\le
                C \max_{i,j}\{|\hat \epsilon_i|, |\check \epsilon_j|\} \times 
                \f 1 n \sum_{\alpha \in [n]} \lp
                    \sum_{i=1}^r (\xme_\alpha^i)^2
                    + \sum_{j=1}^s (\vme_\alpha^j)^2
                    \rp
                \\
            &\asto
                C \times 0 \times \mathcal{E} = 0
        \end{align*}
        where $\mathcal{E}$ is the Gaussian expectation that $\f 1 n \sum_{\alpha \in [n]} \lp
                    \sum_{i=1}^r (\xme_\alpha^i)^2
                    + \sum_{j=1}^s (\vme_\alpha^j)^2
                    \rp$ converges a.s. to, by inductive hypothesis.
    \item
        The quantity $Q_\alpha$ actually doesn't depend on $\alpha$, so that
        \[\sqrt{\f 1 n \sum_{\alpha \in \JJc} Q_\alpha^2} \le |\mathring \sigma^2 - \sigma^2| \asto 0\]
        by \cref{lemma:sigmaConverges}.
    \item
        Notice $R_\alpha^2 = \sigma^4 (1 - d_\alpha)^2 \le \sigma^4 (1 - d_\alpha)$ because $1 - d_\alpha \in [0, 1/2]$ for $\alpha \in \JJc$.
        Thus,
        \begin{align*}
        \f 1 n \sum_{\alpha \in \JJc} R_\alpha^2
            &\le
                \sigma^4 \f 1 n \sum_{\alpha \in \JJc} 1 - d_\alpha
            \le 
                \sigma^4 \f 1 n \sum_{\alpha \in [n]} 1 - d_\alpha
            =
                \sigma^4 \f 1 n \rank \Vme
        \end{align*}
        by the definition that $d_\alpha = (\PP^\perp_{\Vme})_{\alpha\alpha}$.
        But of course $\rank \Vme \le s$ is bounded relative to $n$.
        So this quantity goes to 0 almost surely) as desired.
    \end{itemize}
\end{proof}

\begin{lemma}\label{lemma:YAlphaUnifBounded}
$\sqrt{\f 1 n \sum_{\alpha \in \JJc} Y_\alpha^2}$ is a.s.\ uniformly bounded in $n$.
\end{lemma}
\begin{proof}
  \newcommand{\XX}{\hat{X}}
  We have
  \begin{align*}
  \sqrt{\f 1 n \sum_{\alpha \in \JJc} Y_\alpha^2}
      &\le
          \sqrt{\f 1 n \sum_{\alpha \in \JJc} S_\alpha^2}
          + \sqrt{\f 1 n \sum_{\alpha \in \JJc} |\omega_\alpha|^{2p}}
          + \sqrt{\f 1 n \sum_{\alpha \in \JJc} \XX_\alpha{}^2}
          + \sqrt{\f 1 n \sum_{\alpha \in \JJc} \max(\sigma^2 d_\alpha, \mathring \sigma^2)^p}
          \\
      &\le
          \sqrt{\f 1 n \sum_{\alpha \in [n]} S_\alpha^2}
          + \sqrt{\f 1 n \sum_{\alpha \in [n]} |\omega_\alpha|^{2p}}
          + \sqrt{\f 1 n \sum_{\alpha \in [n]} \XX_\alpha{}^2}
          + \sqrt{\f 1 n \sum_{\alpha \in [n]} \max(\sigma^2 d_\alpha, \mathring \sigma^2)^p}
  \end{align*}
  where
  \[\XX_\alpha \defeq \left|\sum_{i=1}^r \xme_\alpha^i \hat \epsilon_i
                  + \sum_{j=1}^s \vme_\alpha^j \check \epsilon_j
          \right|^p
          + |\sigma^2 d_\alpha - \mathring \sigma^2|^{p/2}
  \]
  We proceed to show that each of 4 summands above are individually a.s.\ uniformly bounded in $n$.
  \begin{itemize}
  \item
      $S_\alpha^2$ is a polynomially bounded function of $g^1_\alpha, \ldots, g^{\MM-1}_\alpha$, so that by \ref{IH:MomConv}$(\MM-1)$,
      \[\f 1 n \sum_{\alpha \in [n]} S_\alpha^2 \asto C\]
      for some constant $C$, so it is also a.s.\ uniformly bounded in $n$.
  \item
      By \cref{lemma:omegaAlphaUnifomBounded}, we get
      \begin{equation*}
      \f 1 n \sum_{\alpha \in [n]} |\omega_\alpha|^{2p}
      \end{equation*}
      is a.s.\ uniformly bounded in $n$.
  \item
      Using the same reasoning as in the proof of \cref{lemma:XalphaAsto0}, one can easily show
      \[
      \f 1 n \sum_{\alpha \in [n]} \XX_\alpha{}^2 \asto 0
      \]
      so it is also a.s.\ uniformly bounded. 
  \item
      Since $d_\alpha \le 1$, we have $\max(\sigma^2 d_\alpha, \mathring\sigma^2) \le \max(\sigma^2, \mathring \sigma^2)$, which is independent of $\alpha$.
      Therefore,
      \begin{align*}
      \f 1 n \sum_{\alpha \in [n]} \max(\sigma^2 d_\alpha, \mathring \sigma^2)^p
      &\le
          \f 1 n \sum_{\alpha \in [n]} \max(\sigma^2, \mathring \sigma^2)^p
      =
          \max(\sigma^2, \mathring \sigma^2)^{p/2}
      \asto \mathring \sigma^p.
      \end{align*}
      and it is also a.s.\ uniformly bounded in $n$.
  \end{itemize}

\end{proof}
  
Finally, we deliver the promised proof of \cref{lemma:PsiAlphaSmoothnessBound}.

\begin{proof}[Proof of \cref{lemma:PsiAlphaSmoothnessBound}]

By \cref{lemma:stein}, $\Psi_\alpha$ is differentiable in $w$, and
\begin{align}
    \pd_w \Psi_\alpha(w; \tau^2)
    &=
        \inv \tau \EV_{z \sim \Gaus(0, 1)} z \psi(g^1_\alpha, \ldots, g^{\MM - 1}_\alpha, w+ \tau z)
        \label{eqn:PsiAlphaDW}
        \\
    \pd_{\tau^2} \Psi_\alpha(w; \tau^2)
    &=
        \f 1 2 \tau^{-2}
        \EV_{z \sim \Gaus(0, 1)} (z^2-1) \psi(g^1_\alpha, \ldots, g^{\MM - 1}_\alpha, w+ \tau z)
        .
        \label{eqn:PsiAlphaDTau2}
\end{align}

Recall that $|\psi(x)| \le C \|x\|^p_{p} + C$.
We will silently introduce constants $C_1, C_2, \ldots$ depending only on $p$, merging with old constants, typically via \cref{lem:powerbound} or by integrating out some integrands depending only on $p$.
With $z \sim \Gaus(0, 1)$,
\begin{align*}
|\pd_w \Psi_\alpha(w; \tau^2)|
    &\le
        \inv \tau \EV_{z} |z| |\psi(g^1_\alpha, \ldots, g^{\MM - 1}_\alpha, w+ \tau z)|
        \\
    &\le
        \inv \tau C\EV_z |z| \lp 1 + |g^1_\alpha|^{p} + \cdots + |g^{\MM-1}_\alpha|^p + |w + \tau z|^p \rp
        \\
    &\le
        \inv \tau C_1\EV_z |z| \lp 1 + |g^1_\alpha|^{p} + \cdots + |g^{\MM-1}_\alpha|^p + |w|^p + \tau^p |z|^p \rp
        \\
    &\le
        \inv \tau C_2 \lp 1 + |g^1_\alpha|^{p} + \cdots + |g^{\MM-1}_\alpha|^p + |w|^p + \tau^p \rp
        .
\end{align*}
Similarly,
\begin{align*}
|\pd_{\tau^2} \Psi_\alpha(w; \tau^2)|
    &\le
        \f 1 2 \tau^{-2} \EV_z |z^2-1| |\psi(g^1_\alpha, \ldots, g^{\MM - 1}_\alpha, w+ \tau z)|
        \\
    &\le
        \tau^{-2} C_3 \lp 1 + |g^1_\alpha|^{p} + \cdots + |g^{\MM-1}_\alpha|^p + |w|^p + \tau^p \rp
        .
\end{align*}

Therefore, for any $\Delta w \in \R, \Delta \tau^2 \in \R^{\ge 0}$, we have
\begin{align*}
&\phantomeq
    \left| \Psi_\alpha(w+\Delta w; \tau^2 + \Delta \tau^2) - \Psi_\alpha(w; \tau^2)\right|
    \\
&=
    \left|
    \int_0^1 \dd t \lp 
        \Delta w \cdot \pd_w \Psi_\alpha(w + \Delta w t; \tau^2 + \Delta \tau^2 t) +
        \Delta \tau^2 \cdot \pd_{\tau^2} \Psi_\alpha(w + \Delta w t; \tau^2 + \Delta \tau^2 t)
        \rp
    \right|
    \\
&\le
    \int_0^1 \dd t \lp 
        |\Delta w| \cdot |\pd_w \Psi_\alpha(w + \Delta w t; \tau^2 + \Delta \tau^2 t)| +
        |\Delta \tau^2| \cdot |\pd_{\tau^2} \Psi_\alpha(w + \Delta w t; \tau^2 + \Delta \tau^2 t)|
        \rp
    \\
&\le
    (C_2 + C_3)
    (|\Delta w| + |\Delta \tau^2|)
    \\
&\quad
    \int_0^1 \dd t
        ((\tau^2 + \Delta \tau^2 t)^{-1/2} + (\tau^2 + \Delta \tau^2 t)^{-1})
        \times
        \lp S_\alpha + |w+\Delta w t|^p + (\tau^2 + \Delta\tau^2 t)^{p/2} \rp
\end{align*}
where
\[S_\alpha \defeq 1 + |g^1_\alpha|^{p} + \cdots + |g^{\MM-1}_\alpha|^p,\]
which is independent of $t$.
Since $\Delta \tau^2 \ge 0$, $(\tau^2 + \Delta \tau^2 t)^{-1} \le \tau^{-2}$, and we get
\begin{align*}
&\phantomeq
    \left| \Psi_\alpha(w+\Delta w; \tau^2 + \Delta \tau^2) - \Psi_\alpha(w; \tau^2)\right|
    \\
&\le
    C_4
    (|\Delta w| + \Delta \tau^2)
    (\inv \tau + \tau^{-2})
    \int_0^1 \dd t
        \lp S_\alpha + |w+\Delta w t|^p + (\tau^2 + \Delta\tau^2 t)^{p/2} \rp
    \\
&\le
    C_5
    (|\Delta w| + \Delta \tau^2)
    (\inv \tau + \tau^{-2})
    \int_0^1 \dd t
        \lp S_\alpha + |w|^p + |\Delta w|^p t^p + \tau^p + (\Delta\tau^2)^{p/2} t^{p/2} \rp
    \\
&\le
    C_6
    (|\Delta w| + \Delta \tau^2)
    (\inv \tau + \tau^{-2})
        \lp S_\alpha + |w|^p + |\Delta w|^p + \tau^p + (\Delta\tau^2)^{p/2} \rp
\end{align*}
where in the end we have integrated out $t^p$ and $t^{p/2}$.
We finally apply the simplification $\tau^{-1} \le \f 1 2 + \f 1 2 \tau^{-2}$ by AM-GM to get the desired \cref{eqn:PsiSmoothness}.
\end{proof}

\section{Proof of \texorpdfstring{\netsortplus{}}{NetsorT+} Master Theorem Assuming Rank Stability}

\label{sec:netsortplusMasterTheoremProof}

In this section we describe how to augment the proof of \cref{thm:NetsorTMasterTheorem} given in \cref{sec:proofMainTheorem} to yield the proof of \cref{thm:PCNetsorT+MasterTheorem}.
This is very similar to the proof of the \netsorplus{} Master Theorem in \citet{yangTP1}.
The key points to note here are 1) the rank stability assumption \cref{assm:asRankStab} used in \cref{thm:PCNetsorT+MasterTheorem}, and 2) an additional term in \cref{eqn:decompABC} due to fluctuations in the parameter $\bigtheta$.

\subsection{Rank Stability}

By \cref{remk:necessityRankStab}, we see that rank stability assumption is necessary for the parameter-controlled \netsortplus{} Master Theorem.
In the proof of the \netsort{} Master Theorem (\cref{sec:proofMainTheorem}), we had to intricately weave together an induction on rank stability (more generally, \ref{IH:coreSet}) and an induction on moment convergence (\ref{IH:MomConv}).
However, here, to show \cref{thm:PCNetsorT+MasterTheorem}, we just need 1) to induct on \ref{IH:MomConv} and 2)
to invoke \cref{assm:asRankStab} whenever we need to use \cref{lemma:rankStability}, which is when we need to show that pseudo-inverse commutes with almost surely limit, such as in \cref{prop:pseudoinverseLambda}, and when we need to ensure either $\sigma$ is almost surely 0 or is almost surely positive, as in \cref{sec:probB}.

\subsection{Fluctuation of the Parameters}
\label{sec:D}
As in \cref{sec:proofMainTheorem}, we will induct on the G-vars $g^1, \ldots, g^\MM$ in the program to show that
\begin{enumerate}
  \item For any random vector $\bigtheta \in \R^l$ that converges almost surely to a deterministic vector $\mathring{\bigtheta}$ as $n \to \infty$, and for any $\psi(-; -): \R^\MM \times \R^l \to \R$ parameter-controlled at $\mathring{\bigtheta}$,
  \begin{align*}
      \f 1 n \sum_{\alpha=1}^n \psi(g^1_\alpha, \ldots, g^\MM_\alpha; \bigtheta) \asto \EV\psi(Z^{g^1}, \ldots, Z^{g^\MM}; \mathring{\bigtheta}).
  \end{align*}
  for any G-vars $g^1, \ldots, g^\MM$, 
  where $Z^{g^{i}}$ are as defined in \cref{{defn:netsortplusKeyIntuit}}.
  \item Each scalar $\theta$ that is a deterministic function of $g^1, \ldots, g^\MM$ converges a.s.\ to $\mathring \theta$.
\end{enumerate}
The latter trivially follows from former, so we will focus on proving the former in the inductive step.

When we have parameters in nonlinearities, \cref{eqn:decompABC} needs to be modified to contain an additional term $\probD$:

\begin{align*}
    &\phantomeq
        \left|\f 1 n \sum_{\alpha = 1}^n \psi(g^1_\alpha, \ldots, g^{\MM}_\alpha; \bigtheta)
        - \EV \psi(Z^{g^1}, \ldots, Z^{g^\MM}; \mathring{\bigtheta})
        \right|
    \le
        \probD + \probA + \probB + \probC
\end{align*}
where
\begin{align*}
\probD
    &\defeq
        \left|\f 1 n \sum_{\alpha = 1}^n \psi(g^1_\alpha, \ldots, g^{\MM}_\alpha; \bigtheta)
        - \psi(g^1_\alpha, \ldots, g^{\MM}_\alpha; \mathring{\bigtheta})
        \right|
\end{align*}
and $\probA, \probB, \probC$ are as in \cref{eqn:decompABC} but replacing $\psi(-)$ there with $\psi(-; \mathring{\bigtheta})$.
Because $\psi(-; -)$ is parameter-controlled at $\mathring{\bigtheta}$ by assumption, $\psi(-; \mathring{\bigtheta})$ is controlled, and $\probA, \probB, \probC \asto 0$ with the same arguments as before (except using rank stability assumption \cref{assm:asRankStab} where appropriate, instead of \ref{IH:coreSet}).

Now, by the other property of parameter-control, we have
\begin{align*}
\probD
    \le
        \f 1 n \sum_{\alpha = 1}^n 
        \left|\psi(g^1_\alpha, \ldots, g^{\MM}_\alpha; \bigtheta)
        - \psi(g^1_\alpha, \ldots, g^{\MM}_\alpha; \mathring{\bigtheta})
        \right|
    &\le
        \f 1 n \sum_{\alpha = 1}^n 
        f(\bigtheta) \bar \psi(g^1_\alpha, \ldots, g^\MM_\alpha)
        \\
    &=
        f(\bigtheta)
        \f 1 n \sum_{\alpha = 1}^n 
        \bar \psi(g^1_\alpha, \ldots, g^\MM_\alpha)
\end{align*}
for some controlled $\bar \psi: \R^\MM \to \R$ and some $f: \R^l \to \R^{\ge 0} \cup \{\infty\}$ that is continuous at $\mathring{\bigtheta}$ and has $f(\mathring{\bigtheta}) = 0$ (where $\bar \psi$ and $f$ can both depend on $\mathring{\bigtheta}$).
Since $\bigtheta \asto \mathring{\bigtheta}$ by induction hypothesis, we have $f(\bigtheta) \asto 0$.
In addition, by \ref{IH:MomConv}, $\f 1 n \sum_{\alpha = 1}^n 
\bar \psi(g^1_\alpha, \ldots, g^\MM_\alpha)$ converges a.s.\ as well to a finite constant.
Therefore,
\begin{align*}
\probD \asto 0
\end{align*}
as desired.

\subsection{Summary}

The proof of \cref{thm:PCNetsorT+MasterTheorem} would proceed as follows: We induct on \ref{IH:MomConv} with the same setup as \cref{sec:inductiveSetup}, except using \cref{assm:asRankStab} for \cref{prop:pseudoinverseLambda}.
Then we prove the inductive step for \ref{IH:MomConv} as in \cref{sec:inductiveMoments}.
We modify \cref{eqn:decompABC} to add a term $\probD$ as in \cref{sec:D}, which goes to 0 a.s.\ as argued there.
The same arguments for $\probA, \probB, \probC \asto 0$, exhibited in \cref{sec:inductiveMoments} still hold, except that in the proof of $\probB \asto 0$, we apply \cref{assm:asRankStab} (instead of \cref{lemma:rankStability}) to allow us to assume $\mathring \sigma > 0$ and $\sigma > 0$ almost surely.

\section{Proof of \texorpdfstring{\netsortplus{}}{NetsorT+} Master Theorem without Assuming Rank Stability}

\label{sec:proofNetsorTPlusNoRS}

In this section, we prove \cref{thm:PLNetsorT+MasterTheoremNoRS}.

\global\long\def\coreset{\mathcal{M}}%

\global\long\def\vanset{\overline{\mathcal{M}}}%

\begin{defn}
We say a vector $v\in\R^{n}$ has \emph{vanishing moments} if $\frac{1}{n}\sum_{\alpha=1}^{n}v_{\alpha}^{2k}\asto0$ for all integer $k>0$ as $n\to\infty$. We say it has \emph{bounded moments} if there are finite $C_{k}\in\R$ such that $\frac{1}{n}\sum_{\alpha=1}^{n}v_{\alpha}^{2k}\asto C_{k}$ for all $k$.
\end{defn}

Assume each \refNonlinPlus{} instruction only takes G-vars instead of any vector in the program.

\subparagraph*{Proof Plan}

We will inductively rewrite the program and maintain a 1) core set $\coreset$ of G-vars whose elements are orthogonal as vectors and have non-vanishing moments and 2) a set $\vanset$ of G-vars with vanishing moments. In particular, the vectors in $\vanset$ will take the form $Av$ for some matrix $A$ and some $v$ with vanishing moments. Any vector in the original program will always be (mathematically) a linear combination of vectors in $\coreset$ and $\vanset$. Then \ref{IH:MomConv} largely reduces to that of $\coreset$, which has rank stability. 

Every non-initial vector $g$ of $\coreset$ is created by \refMatMulPlus{}, say $g=Ah$ for some matrix $A$ and vector $h$. For any matrix $A$, define $\coreset_{A}^{*}$ to be the collection of all such $h$.

\subsection{Induction Hypotheses}

Let $g^{1},\ldots,g^{M}$ be all of the G-vars in the program, including initial vectors, in order of appearance. We inductively rewrite the program and expand $\coreset$ and $\vanset$, maintaining the following induction hypothesis for each $m\in[M]$.
\begin{description}
\item [IH\label{IH:NoRS}]\!\!\!$(m)$\ \ \ 
  The following hold simultaneously
  \begin{description}
  \item [Rewrite\label{IH:Rewrite}]\!\!\!$(m)$\ \ \ 
    $g^{1},\ldots,g^{m}$ are mathematically linear combinations of vectors in $\coreset$ and $\vanset$ whose coefficients are definable using \refMoment{} applied to $\coreset$ and $\vanset$.
  \item [Moments\label{IH:MomentsNoRS}]\!\!\!$(m)$\ \ \ 
    Let $z^{1},\ldots,z^{k}$ be all vectors in $\coreset$ and $\bar{z}^{1},\ldots,\bar{z}^{\bar{k}}$ be all vectors in $\vanset$. For any pseudo-Lipschitz $\psi:\R^{k+\bar{k}}\to\R$,
  \[
  \frac{1}{n}\sum_{\alpha=1}^{n}\psi(z_{\alpha}^{1},\ldots,z_{\alpha}^{k},\bar{z}_{\alpha}^{1},\ldots,\bar{z}_{\alpha}^{\bar{k}})\asto\EV\psi(Z^{z^{1}},\ldots,Z^{z^{k}},0,\ldots,0).
  \]
  \item [Orthogonal\label{IH:Orthogonal}]\!\!\!$(m)$\ \ \ 
    For all $n$, for any matrix $A$, $\coreset_{A}^{*}$ forms an orthogonal system of vectors where each vector has $\ell_{2}$ norm $\sqrt{n}$
  \item [VanSet\label{IH:Vanset}]\!\!\!$(m)$\ \ \ 
    Every vector of $\vanset$ has vanishing moments
  \end{description}
\end{description}
By induction, we will have \ref{IH:Rewrite}$(M)$ and \ref{IH:MomentsNoRS}$(M)$, which together with \cref{lemma:removeVanishing} imply \cref{thm:PLNetsorT+MasterTheoremNoRS}.

\subsection{Base Case}

WLOG, we may assume the initial vectors have identity covariance by applying Gram-Schmidt. These vectors are added to $\coreset$. Currently, $\vanset$ is empty. \ref{IH:MomentsNoRS} is then automatic by the Strong Law of Large Numbers; the other clauses of the induction hypothesis are obvious.

\subsection{Induction}

Assume \ref{IH:NoRS}$(m-1)$. We seek to show \ref{IH:NoRS}$(m)$.

Suppose $g=g^{m}=Ah$ for some vector $h$. By \ref{IH:Rewrite}$(m-1)$, we have \[h=\phi(z^{1},\ldots,z^{k},\bar{z}^{1},\ldots,\bar{z}^{\bar{k}};\theta_{1},\ldots,\theta_{l})\] where 1) $z^{1},\ldots,z^{k}$ are all of the vectors in $\coreset$, 2) $\bar{z}^{1},\ldots,\bar{z}^{\bar{k}}$ are all of the vectors in $\vanset$, 3) $\theta_{1},\ldots,\theta_{l}$ are scalars created by \refMoment{} from $z^{1},\ldots,z^{k},\bar{z}^{1},\ldots,\bar{z}^{\bar{k}}$ and so converge a.s. to $\mathring{\theta}_{1},\ldots,\mathring{\theta}_{l}$ by \ref{IH:MomentsNoRS}$(m-1)$, and 4) $\phi$ is pseudo-Lipschitz jointly in all of them.

Let 
\[h^{0}\defeq\phi(z^{1},\ldots,z^{k},0,\ldots,0;\theta_{1},\ldots,\theta_{l}) \quad\text{and}\quad \Delta h\defeq h-h^{0}.\] (Note both $h^{0}$ and $\Delta h$ are expressible by \refNonlinPlus{} in terms of $z^{1},\ldots,z^{k},\bar{z}^{1},\ldots,\bar{z}^{\bar{k}};\theta_{1},\ldots,\theta_{l}$). Because $\bar{z}^{1},\ldots,\bar{z}^{\bar{k}}$ have vanishing moments by induction hypothesis and $\phi$ is pseudo-Lipschitz, $\Delta h$ has vanishing moments as well by \cref{lemma:removeVanishing}. We insert $A\Delta h$ into $\vanset$; we will prove that $A\Delta h$ has vanishing moments below in \cref{lemma:ADeltah}.

Now define $\hat{h}^{0}\defeq h^{0}-\Pi h^{0}$ where $\Pi$ is the projection onto the linear span of $\coreset_{A}^{*}$. $\Pi h^{0}$ can be written as a \refNonlinPlus{} like so: $\Pi h^{0}=\sum_{v\in\coreset_{A}^{*}}\frac{v^{\trsp}h^{0}}{n}v$, where we used the induction hypothesis that $\|v\|_{2}=\sqrt{n}$ and $\coreset_{A}^{*}$ is orthogonal. Then $\hat{h}^{0}$ can also be written as a \refNonlinPlus{}.

We proceed by casework on whether $Z^{\hat{h}^{0}}=0$.

1) Suppose $Z^{\hat{h}^{0}}=0$. Then $A\hat{h}^{0}$ has vanishing moments by the same Gaussian conditioning technique as in \cref{sec:proofMainTheorem}, which is possible because we only need to condition on $z^{1},\ldots,z^{k}$, which are all of the G-vars that $\hat{h}^{0}$ depends on, and \ref{IH:Orthogonal}$(m-1)$ implies they have rank stability. Then we add $A\hat{h}^{0}$ to $\vanset$. Because $g=Ah=A(h^{0}+\Delta h)=A(\hat{h}^{0}+\Pi h^{0}+\Delta h)$, we can rewrite all instances of $g$ in the program as the sum of $A\hat{h}^{0}$, $A\Delta h$ (in $\vanset$), and $A\Pi h^{0}=\sum_{v\in\coreset_{A}^{*}}\frac{v^{\trsp}h^{0}}{n}Av$ (a linear combination of $\coreset$).

2) Suppose $Z^{\hat{h}^{0}}\ne0$. Define $h^{1}\defeq h^{0}/\frac{\|\hat{h}^{0}\|^{2}}{n}$ via \refNonlinPlus{} (which is valid since $\frac{\|\hat{h}^{0}\|^{2}}{n}\asto\EV(Z^{\hat{h}^{0}})^{2}\ne0$ by induction hypothesis). We add $Ah^{1}$ to $\coreset$. Because $g=Ah=A(h^{0}+\Delta h)=A(\frac{\|\hat{h}^{0}\|^{2}}{n}h^{1}+\Pi h^{0}+\Delta h)$, we can rewrite all instances of $g$ in the program as the sum of $A\Delta h$ (in $\vanset$), $\frac{\|\hat{h}^{0}\|^{2}}{n}Ah^{1}$, and $A\Pi h^{0}=\sum_{v\in\coreset_{A}^{*}}\frac{v^{\trsp}h^{0}}{n}Av$ (linear combinations of $\coreset$).

With the above updates to $\coreset$ and/or $\vanset$, it's clear that \ref{IH:Rewrite}$(m)$, \ref{IH:Orthogonal}$(m)$ and \ref{IH:Vanset}$(m)$ are true. So it remains to prove \ref{IH:MomentsNoRS}$(m)$.

\subparagraph*{Moments$(m)$}

In the case 1) above (where we did not expand $\coreset$), \ref{IH:MomentsNoRS}$(m)$ follows straightforwardly from \ref{IH:MomentsNoRS}$(m-1)$ and \cref{lemma:removeVanishing}.

In the case 2) above (where we added $Ah^{1}$ to $\coreset$), we need to show
\[
\frac{1}{n}\sum_{\alpha=1}^{n}\psi(Ah_{\alpha}^{1},z_{\alpha}^{1},\ldots,z_{\alpha}^{k},\bar{z}_{\alpha}^{1},\ldots,\bar{z}_{\alpha}^{\bar{k}})\asto\EV\psi(Z^{Ah_{\alpha}^{1}},Z^{z^{1}},\ldots,Z^{z^{k}},0,\ldots,0)
\]
for any pseudo-Lipschitz $\psi$. Since $h^{1}$ is expressible as a \refNonlinPlus{} of purely $z^{1},\ldots,z^{k}$ and some parameters, we can just condition on $z_{\alpha}^{1},\ldots,z_{\alpha}^{k}$ and proceed as in \cref{sec:proofMainTheorem}, after replacing $\bar{z}_{\alpha}^{1},\ldots,\bar{z}_{\alpha}^{\bar{k}}$ by zeros by \cref{lemma:removeVanishing}. (Note this argument wouldn't have worked if we naively put $g=Ah$ in place of $Ah^{1}$ because $h$ depends on $\bar{z}^{1},\ldots,\bar{z}^{\bar{k}}$).

This finishes the induction, barring for the promised \cref{lemma:ADeltah} below.

\begin{lemma}
\label{lemma:ADeltah}$A\Delta h$ has vanishing moments.
\end{lemma}

It's clear that $A\Delta h$ at least has vanishing second moment because $A$ has a uniformly (in $n$) bounded operator norm from standard matrix concentration bounds. We need to work a bit harder to get all moments to vanish.
\begin{proof}
We condition on $z^{1},\ldots,z^{k},\bar{z}^{1},\ldots,\bar{z}^{\bar{k}}$. Suppose
\[
u^{j}=A^{\trsp}v^{j},\quad j=1,\ldots,s,\quad\text{and}\quad\bar{u}^{j}=A^{\trsp}\bar{v}^{j},\quad j=1,\ldots,\bar{s}
\]
are respectively the elements of $\coreset$ and $\vanset$ that are images of MatMul with $A^{\trsp}$. As in the ($\hat{Z}$, $\dot{Z}$)-decomposition in the $n\to\infty$ limit, we can write $A\Delta h$, for finite $n$, as a sum of 1) a Gaussian vector with vanishing 2nd moment and 2) a linear combination of \{$v^{j}\}_{j=1}^{s}$ and $\{\bar{v}^{j}\}_{j=1}^{\bar{s}}$, where the coefficient of $v^{j}$ (resp. $\bar{v}^{j}$) converges to $\sigma_{A}^{2}\EV\partial Z^{\Delta h}/\partial Z^{v^{j}}=0$ (resp. $\sigma_{A}^{2}\EV\partial Z^{\Delta h}/\partial Z^{\bar{v}^{j}}<\infty$). The former has vanishing moments of all order by Gaussianity. Since $v^{j}$ has bounded moments and $\bar{v}^{j}$ has vanishing moments, the latter also has vanishing moments.
\end{proof}
\begin{lemma}
\label{lemma:removeVanishing}Let $z^{1},\ldots,z^{k}$ have bounded moments, $\bar{z}^{1},\ldots,\bar{z}^{\bar{k}}$ have vanishing moments, and $\theta_{1},\ldots,\theta_{l}\in\R$ converge almost surely to deterministic $\mathring{\theta}_{1},\ldots,\mathring{\theta}_{l}\in\R$. Then for any pseudo-Lipschitz $\psi$ and $t>0$,
\[
\frac{1}{n}\sum_{\alpha=1}^{n}|\psi(z_{\alpha}^{1},\ldots,z_{\alpha}^{k},\bar{z}_{\alpha}^{1},\ldots,\bar{z}_{\alpha}^{\bar{k}};\theta_{1},\ldots\theta_{l})-\psi(z_{\alpha}^{1},\ldots,z_{\alpha}^{k},0,\ldots,0;\mathring{\theta}_{1},\ldots,\mathring{\theta}_{l})|^t \asto0.
\]
\end{lemma}
\begin{proof}
  Holder's inequality.
\end{proof}

%% file: main.bbl
\begin{thebibliography}{55}
\providecommand{\natexlab}[1]{#1}
\providecommand{\url}[1]{\texttt{#1}}
\expandafter\ifx\csname urlstyle\endcsname\relax
  \providecommand{\doi}[1]{doi: #1}\else
  \providecommand{\doi}{doi: \begingroup \urlstyle{rm}\Url}\fi

\bibitem[Alemohammad et~al.(2020)Alemohammad, Wang, Balestriero, and
  Baraniuk]{alemohammad2020recurrent}
Sina Alemohammad, Zichao Wang, Randall Balestriero, and Richard Baraniuk.
\newblock The recurrent neural tangent kernel, 2020.

\bibitem[Allen-Zhu et~al.(2018{\natexlab{a}})Allen-Zhu, Li, and
  Liang]{allen-zhu_learning_2018}
Zeyuan Allen-Zhu, Yuanzhi Li, and Yingyu Liang.
\newblock Learning and {Generalization} in {Overparameterized} {Neural}
  {Networks}, {Going} {Beyond} {Two} {Layers}.
\newblock \emph{arXiv:1811.04918 [cs, math, stat]}, November
  2018{\natexlab{a}}.
\newblock URL \url{http://arxiv.org/abs/1811.04918}.

\bibitem[Allen-Zhu et~al.(2018{\natexlab{b}})Allen-Zhu, Li, and
  Song]{allen-zhu_convergence_2018}
Zeyuan Allen-Zhu, Yuanzhi Li, and Zhao Song.
\newblock A {Convergence} {Theory} for {Deep} {Learning} via
  {Over}-{Parameterization}.
\newblock \emph{arXiv:1811.03962 [cs, math, stat]}, November
  2018{\natexlab{b}}.
\newblock URL \url{http://arxiv.org/abs/1811.03962}.

\bibitem[Allen-Zhu et~al.(2018{\natexlab{c}})Allen-Zhu, Li, and
  Song]{allen-zhu_convergence_2018-1}
Zeyuan Allen-Zhu, Yuanzhi Li, and Zhao Song.
\newblock On the {Convergence} {Rate} of {Training} {Recurrent} {Neural}
  {Networks}.
\newblock \emph{arXiv:1810.12065 [cs, math, stat]}, October 2018{\natexlab{c}}.
\newblock URL \url{http://arxiv.org/abs/1810.12065}.

\bibitem[Arora et~al.(2019)Arora, Du, Hu, Li, Salakhutdinov, and
  Wang]{arora_exact_2019}
Sanjeev Arora, Simon~S. Du, Wei Hu, Zhiyuan Li, Ruslan Salakhutdinov, and
  Ruosong Wang.
\newblock On {Exact} {Computation} with an {Infinitely} {Wide} {Neural} {Net}.
\newblock \emph{arXiv:1904.11955 [cs, stat]}, April 2019.
\newblock URL \url{http://arxiv.org/abs/1904.11955}.

\bibitem[Bayati and Montanari(2011)]{bayati_dynamics_2011}
Mohsen Bayati and Andrea Montanari.
\newblock The dynamics of message passing on dense graphs, with applications to
  compressed sensing.
\newblock \emph{IEEE Transactions on Information Theory}, 57\penalty0
  (2):\penalty0 764--785, February 2011.
\newblock ISSN 0018-9448, 1557-9654.
\newblock \doi{10.1109/TIT.2010.2094817}.
\newblock URL \url{http://arxiv.org/abs/1001.3448}.

\bibitem[Chen et~al.(2018)Chen, Pennington, and
  Schoenholz]{chen_dynamical_2018}
Minmin Chen, Jeffrey Pennington, and Samuel Schoenholz.
\newblock Dynamical {Isometry} and a {Mean} {Field} {Theory} of {RNNs}:
  {Gating} {Enables} {Signal} {Propagation} in {Recurrent} {Neural} {Networks}.
\newblock In \emph{Proceedings of the 35th {International} {Conference} on
  {Machine} {Learning}}, volume~80 of \emph{Proceedings of {Machine} {Learning}
  {Research}}, pages 873--882, Stockholmsmässan, Stockholm Sweden, July 2018.
  PMLR.
\newblock URL \url{http://proceedings.mlr.press/v80/chen18i.html}.

\bibitem[Couillet and Debbah(2011)]{couillet_random_2011}
Romain Couillet and Merouane Debbah.
\newblock \emph{Random {Matrix} {Methods} for {Wireless} {Communications}}.
\newblock Cambridge University Press, Cambridge, 2011.
\newblock ISBN 978-0-511-99474-6.
\newblock \doi{10.1017/CBO9780511994746}.
\newblock URL \url{http://ebooks.cambridge.org/ref/id/CBO9780511994746}.

\bibitem[Daniely et~al.(2016)Daniely, Frostig, and Singer]{daniely_toward_2016}
Amit Daniely, Roy Frostig, and Yoram Singer.
\newblock Toward {Deeper} {Understanding} of {Neural} {Networks}: {The} {Power}
  of {Initialization} and a {Dual} {View} on {Expressivity}.
\newblock In D.~D. Lee, M.~Sugiyama, U.~V. Luxburg, I.~Guyon, and R.~Garnett,
  editors, \emph{Advances in {Neural} {Information} {Processing} {Systems} 29},
  pages 2253--2261. Curran Associates, Inc., 2016.
\newblock URL
  \url{http://papers.nips.cc/paper/6427-toward-deeper-understanding-of-neural-networks-the-power-of-initialization-and-a-dual-view-on-expressivity.pdf}.

\bibitem[Donoho et~al.(2009)Donoho, Maleki, and Montanari]{donoho_message_2009}
David~L. Donoho, Arian Maleki, and Andrea Montanari.
\newblock Message {Passing} {Algorithms} for {Compressed} {Sensing}.
\newblock \emph{Proceedings of the National Academy of Sciences}, 106\penalty0
  (45):\penalty0 18914--18919, November 2009.
\newblock ISSN 0027-8424, 1091-6490.
\newblock \doi{10.1073/pnas.0909892106}.
\newblock URL \url{http://arxiv.org/abs/0907.3574}.

\bibitem[Du et~al.(2018)Du, Zhai, Poczos, and Singh]{du_gradient_2018}
Simon~S. Du, Xiyu Zhai, Barnabas Poczos, and Aarti Singh.
\newblock Gradient {Descent} {Provably} {Optimizes} {Over}-parameterized
  {Neural} {Networks}.
\newblock \emph{arXiv:1810.02054 [cs, math, stat]}, October 2018.
\newblock URL \url{http://arxiv.org/abs/1810.02054}.

\bibitem[Du et~al.(2019)Du, Hou, Póczos, Salakhutdinov, Wang, and
  Xu]{du2019graph}
Simon~S. Du, Kangcheng Hou, Barnabás Póczos, Ruslan Salakhutdinov, Ruosong
  Wang, and Keyulu Xu.
\newblock Graph neural tangent kernel: Fusing graph neural networks with graph
  kernels, 2019.

\bibitem[Garriga-Alonso et~al.(2018)Garriga-Alonso, Aitchison, and
  Rasmussen]{garriga-alonso_deep_2018}
Adrià Garriga-Alonso, Laurence Aitchison, and Carl~Edward Rasmussen.
\newblock Deep {Convolutional} {Networks} as shallow {Gaussian} {Processes}.
\newblock \emph{arXiv:1808.05587 [cs, stat]}, August 2018.
\newblock URL \url{http://arxiv.org/abs/1808.05587}.

\bibitem[Hanin(2018)]{hanin_which_2018}
Boris Hanin.
\newblock Which {Neural} {Net} {Architectures} {Give} {Rise} {To} {Exploding}
  and {Vanishing} {Gradients}?
\newblock January 2018.
\newblock URL \url{https://arxiv.org/abs/1801.03744}.

\bibitem[Hanin and Rolnick(2018)]{hanin_how_2018}
Boris Hanin and David Rolnick.
\newblock How to {Start} {Training}: {The} {Effect} of {Initialization} and
  {Architecture}.
\newblock \emph{arXiv:1803.01719 [cs, stat]}, March 2018.
\newblock URL \url{http://arxiv.org/abs/1803.01719}.

\bibitem[Hayou et~al.(2018)Hayou, Doucet, and Rousseau]{hayou_selection_2018}
Soufiane Hayou, Arnaud Doucet, and Judith Rousseau.
\newblock On the {Selection} of {Initialization} and {Activation} {Function}
  for {Deep} {Neural} {Networks}.
\newblock \emph{arXiv:1805.08266 [cs, stat]}, May 2018.
\newblock URL \url{http://arxiv.org/abs/1805.08266}.

\bibitem[Hazan and Jaakkola(2015)]{hazan_steps_2015}
Tamir Hazan and Tommi Jaakkola.
\newblock Steps {Toward} {Deep} {Kernel} {Methods} from {Infinite} {Neural}
  {Networks}.
\newblock \emph{arXiv:1508.05133 [cs]}, August 2015.
\newblock URL \url{http://arxiv.org/abs/1508.05133}.

\bibitem[Hron et~al.(2020)Hron, Bahri, Sohl-Dickstein, and
  Novak]{hron2020infinite}
Jiri Hron, Yasaman Bahri, Jascha Sohl-Dickstein, and Roman Novak.
\newblock Infinite attention: Nngp and ntk for deep attention networks, 2020.

\bibitem[Jacot et~al.(2018)Jacot, Gabriel, and Hongler]{jacot_neural_2018}
Arthur Jacot, Franck Gabriel, and Clément Hongler.
\newblock Neural {Tangent} {Kernel}: {Convergence} and {Generalization} in
  {Neural} {Networks}.
\newblock \emph{arXiv:1806.07572 [cs, math, stat]}, June 2018.
\newblock URL \url{http://arxiv.org/abs/1806.07572}.

\bibitem[Kivelson et~al.(1992)Kivelson, Lee, and Zhang]{kivelson_global_1992}
Steven Kivelson, Dung-Hai Lee, and Shou-Cheng Zhang.
\newblock Global phase diagram in the quantum {Hall} effect.
\newblock \emph{Physical Review B}, 46\penalty0 (4):\penalty0 2223--2238, July
  1992.
\newblock ISSN 0163-1829, 1095-3795.
\newblock \doi{10.1103/PhysRevB.46.2223}.
\newblock URL \url{https://link.aps.org/doi/10.1103/PhysRevB.46.2223}.

\bibitem[Lane et~al.(1955)Lane, Thomas, and Wigner]{lane_giant_1955}
A.~M. Lane, R.~G. Thomas, and E.~P. Wigner.
\newblock Giant {Resonance} {Interpretation} of the {Nucleon}-{Nucleus}
  {Interaction}.
\newblock \emph{Physical Review}, 98\penalty0 (3):\penalty0 693--701, May 1955.
\newblock ISSN 0031-899X.
\newblock \doi{10.1103/PhysRev.98.693}.
\newblock URL \url{https://link.aps.org/doi/10.1103/PhysRev.98.693}.

\bibitem[Le~Roux and Bengio(2007)]{le_roux_continuous_2007}
Nicolas Le~Roux and Yoshua Bengio.
\newblock Continuous neural networks.
\newblock In \emph{Artificial {Intelligence} and {Statistics}}, pages 404--411,
  2007.

\bibitem[Lee et~al.(2018)Lee, Bahri, Novak, Schoenholz, Pennington, and
  Sohl-dickstein]{lee_deep_2018}
Jaehoon Lee, Yasaman Bahri, Roman Novak, Sam Schoenholz, Jeffrey Pennington,
  and Jascha Sohl-dickstein.
\newblock Deep {Neural} {Networks} as {Gaussian} {Processes}.
\newblock In \emph{International {Conference} on {Learning} {Representations}},
  2018.
\newblock URL \url{https://openreview.net/forum?id=B1EA-M-0Z}.

\bibitem[Lee et~al.(2019)Lee, Xiao, Schoenholz, Bahri, Sohl-Dickstein, and
  Pennington]{lee_wide_2019}
Jaehoon Lee, Lechao Xiao, Samuel~S. Schoenholz, Yasaman Bahri, Jascha
  Sohl-Dickstein, and Jeffrey Pennington.
\newblock Wide {Neural} {Networks} of {Any} {Depth} {Evolve} as {Linear}
  {Models} {Under} {Gradient} {Descent}.
\newblock \emph{arXiv:1902.06720 [cs, stat]}, February 2019.
\newblock URL \url{http://arxiv.org/abs/1902.06720}.

\bibitem[Ling et~al.(2019)Ling, He, and Qiu]{ling_spectrum_2019}
Zenan Ling, Xing He, and Robert~C. Qiu.
\newblock Spectrum concentration in deep residual learning: a free probability
  approach.
\newblock \emph{arXiv:1807.11694 [cs, stat]}, February 2019.
\newblock URL \url{http://arxiv.org/abs/1807.11694}.
\newblock arXiv: 1807.11694.

\bibitem[Littwin et~al.(2020)Littwin, Galanti, and
  Wolf]{littwin2020optimization}
Etai Littwin, Tomer Galanti, and Lior Wolf.
\newblock On the optimization dynamics of wide hypernetworks, 2020.

\bibitem[Marčenko and Pastur(1967)]{marcenko_distribution_1967}
V~A Marčenko and L~A Pastur.
\newblock {DISTRIBUTION} {OF} {EIGENVALUES} {FOR} {SOME} {SETS} {OF} {RANDOM}
  {MATRICES}.
\newblock \emph{Mathematics of the USSR-Sbornik}, 1\penalty0 (4):\penalty0
  457--483, April 1967.
\newblock ISSN 0025-5734.
\newblock \doi{10.1070/SM1967v001n04ABEH001994}.
\newblock URL
  \url{http://stacks.iop.org/0025-5734/1/i=4/a=A01?key=crossref.1d0b803ddab02373cb6b0690a61e734a}.

\bibitem[Matthews et~al.(2018)Matthews, Rowland, Hron, Turner, and
  Ghahramani]{matthews_gaussian_2018_arxiv}
Alexander G. de~G. Matthews, Mark Rowland, Jiri Hron, Richard~E. Turner, and
  Zoubin Ghahramani.
\newblock Gaussian {Process} {Behaviour} in {Wide} {Deep} {Neural} {Networks}.
\newblock \emph{arXiv:1804.11271 [cs, stat]}, April 2018.
\newblock URL \url{http://arxiv.org/abs/1804.11271}.

\bibitem[Mezzadri and Snaith(2005)]{mezzadri_recent_2005}
F.~Mezzadri and N.~C. Snaith, editors.
\newblock \emph{Recent {Perspectives} in {Random} {Matrix} {Theory} and
  {Number} {Theory}}.
\newblock Cambridge University Press, 1 edition, June 2005.
\newblock ISBN 978-0-521-62058-1 978-0-511-55049-2.
\newblock \doi{10.1017/CBO9780511550492}.
\newblock URL
  \url{https://www.cambridge.org/core/product/identifier/9780511550492/type/book}.

\bibitem[Mingo and Speicher(2017)]{mingo_free_2017}
J.A. Mingo and R.~Speicher.
\newblock \emph{Free {Probability} and {Random} {Matrices}}.
\newblock Fields {Institute} {Monographs}. Springer New York, 2017.
\newblock ISBN 978-1-4939-6942-5.
\newblock URL \url{https://books.google.com/books?id=d7wpDwAAQBAJ}.

\bibitem[Neal(1995)]{neal_bayesian_1995}
Radford~M Neal.
\newblock \emph{{BAYESIAN} {LEARNING} {FOR} {NEURAL} {NETWORKS}}.
\newblock {PhD} {Thesis}, University of Toronto, 1995.

\bibitem[Novak et~al.(2018)Novak, Xiao, Lee, Bahri, Abolafia, Pennington, and
  Sohl-Dickstein]{novak_bayesian_2018}
Roman Novak, Lechao Xiao, Jaehoon Lee, Yasaman Bahri, Daniel~A Abolafia,
  Jeffrey Pennington, and Jascha Sohl-Dickstein.
\newblock Bayesian {Deep} {Convolutional} {Networks} with {Many} {Channels} are
  {Gaussian} {Processes}.
\newblock \emph{arXiv preprint arXiv:1810.05148}, 2018.

\bibitem[O'Donnell(2014)]{odonnell_analysis_2014}
Ryan O'Donnell.
\newblock \emph{Analysis of boolean functions}.
\newblock Cambridge University Press, New York, NY, 2014.
\newblock ISBN 978-1-107-03832-5.

\bibitem[Pastur(2020)]{pastur2020random}
Leonid Pastur.
\newblock On random matrices arising in deep neural networks. gaussian case,
  2020.

\bibitem[Pennington and Worah(2018)]{pennington_spectrum_2018}
Jeffrey Pennington and Pratik Worah.
\newblock The {Spectrum} of the {Fisher} {Information} {Matrix} of a
  {Single}-{Hidden}-{Layer} {Neural} {Network}.
\newblock In \emph{Advances in {Neural} {Information} {Processing} {Systems}
  31}, page~10, 2018.

\bibitem[Pennington et~al.(2017)Pennington, Schoenholz, and
  Ganguli]{pennington_resurrecting_2017}
Jeffrey Pennington, Samuel Schoenholz, and Surya Ganguli.
\newblock Resurrecting the sigmoid in deep learning through dynamical isometry:
  theory and practice.
\newblock In I.~Guyon, U.~V. Luxburg, S.~Bengio, H.~Wallach, R.~Fergus,
  S.~Vishwanathan, and R.~Garnett, editors, \emph{Advances in {Neural}
  {Information} {Processing} {Systems} 30}, pages 4788--4798. Curran
  Associates, Inc., 2017.
\newblock URL
  \url{http://papers.nips.cc/paper/7064-resurrecting-the-sigmoid-in-deep-learning-through-dynamical-isometry-theory-and-practice.pdf}.

\bibitem[Pennington et~al.(2018)Pennington, Schoenholz, and
  Ganguli]{pennington_emergence_2018}
Jeffrey Pennington, Samuel~S. Schoenholz, and Surya Ganguli.
\newblock The {Emergence} of {Spectral} {Universality} in {Deep} {Networks}.
\newblock \emph{arXiv:1802.09979 [cs, stat]}, February 2018.
\newblock URL \url{http://arxiv.org/abs/1802.09979}.
\newblock arXiv: 1802.09979.

\bibitem[Philipp and Carbonell(2018)]{philipp_nonlinearity_2018}
George Philipp and Jaime~G. Carbonell.
\newblock The {Nonlinearity} {Coefficient} - {Predicting} {Overfitting} in
  {Deep} {Neural} {Networks}.
\newblock \emph{arXiv:1806.00179 [cs, stat]}, May 2018.
\newblock URL \url{http://arxiv.org/abs/1806.00179}.

\bibitem[Poole et~al.(2016)Poole, Lahiri, Raghu, Sohl-Dickstein, and
  Ganguli]{poole_exponential_2016}
Ben Poole, Subhaneil Lahiri, Maithreyi Raghu, Jascha Sohl-Dickstein, and Surya
  Ganguli.
\newblock Exponential expressivity in deep neural networks through transient
  chaos.
\newblock In \emph{Advances {In} {Neural} {Information} {Processing}
  {Systems}}, pages 3360--3368, 2016.

\bibitem[Schoenholz et~al.(2017)Schoenholz, Gilmer, Ganguli, and
  Sohl-Dickstein]{schoenholz_deep_2017}
Samuel~S. Schoenholz, Justin Gilmer, Surya Ganguli, and Jascha Sohl-Dickstein.
\newblock Deep {Information} {Propagation}.
\newblock 2017.
\newblock URL \url{https://openreview.net/pdf?id=H1W1UN9gg}.

\bibitem[Speicher(2009)]{speicher_free_2009}
Roland Speicher.
\newblock Free {Probability} {Theory}.
\newblock \emph{arXiv:0911.0087 [math]}, October 2009.
\newblock URL \url{http://arxiv.org/abs/0911.0087}.

\bibitem[Tao(2012)]{tao_topics_2012}
Terence Tao.
\newblock Topics in random matrix theory.
\newblock \emph{Graduate studies in Mathematics}, 132, 2012.

\bibitem[Tarnowski et~al.(2018)Tarnowski, Warchoł, Jastrzębski, Tabor, and
  Nowak]{tarnowski_dynamical_2018}
Wojciech Tarnowski, Piotr Warchoł, Stanisław Jastrzębski, Jacek Tabor, and
  Maciej~A. Nowak.
\newblock Dynamical {Isometry} is {Achieved} in {Residual} {Networks} in a
  {Universal} {Way} for any {Activation} {Function}.
\newblock \emph{arXiv:1809.08848 [cs, stat]}, September 2018.
\newblock URL \url{http://arxiv.org/abs/1809.08848}.

\bibitem[von Oppen et~al.(2000)von Oppen, Halperin, and
  Stern]{von_oppen_conductivity_2000}
Felix von Oppen, Bertrand~I. Halperin, and Ady Stern.
\newblock Conductivity tensor of striped quantum {Hall} phases.
\newblock \emph{Physical Review Letters}, 84\penalty0 (13):\penalty0
  2937--2940, March 2000.
\newblock ISSN 0031-9007, 1079-7114.
\newblock \doi{10.1103/PhysRevLett.84.2937}.
\newblock URL \url{http://arxiv.org/abs/cond-mat/9910132}.
\newblock arXiv: cond-mat/9910132.

\bibitem[Wigner(1955)]{wigner_characteristic_1955}
Eugene~P. Wigner.
\newblock Characteristic {Vectors} of {Bordered} {Matrices} {With} {Infinite}
  {Dimensions}.
\newblock \emph{The Annals of Mathematics}, 62\penalty0 (3):\penalty0 548,
  November 1955.
\newblock ISSN 0003486X.
\newblock \doi{10.2307/1970079}.
\newblock URL \url{https://www.jstor.org/stable/1970079?origin=crossref}.

\bibitem[Wigner(1958)]{wigner_distribution_1958}
Eugene~P. Wigner.
\newblock On the {Distribution} of the {Roots} of {Certain} {Symmetric}
  {Matrices}.
\newblock \emph{The Annals of Mathematics}, 67\penalty0 (2):\penalty0 325,
  March 1958.
\newblock ISSN 0003486X.
\newblock \doi{10.2307/1970008}.
\newblock URL \url{https://www.jstor.org/stable/1970008?origin=crossref}.

\bibitem[Williams(1997)]{williams_computing_1997}
Christopher K~I Williams.
\newblock Computing with {Infinite} {Networks}.
\newblock In \emph{Advances in neural information processing systems}, page~7,
  1997.

\bibitem[Xiao et~al.(2018)Xiao, Bahri, Sohl-Dickstein, Schoenholz, and
  Pennington]{xiao_dynamical_2018}
Lechao Xiao, Yasaman Bahri, Jascha Sohl-Dickstein, Samuel Schoenholz, and
  Jeffrey Pennington.
\newblock Dynamical {Isometry} and a {Mean} {Field} {Theory} of {CNNs}: {How}
  to {Train} 10,000-{Layer} {Vanilla} {Convolutional} {Neural} {Networks}.
\newblock In \emph{Proceedings of the 35th {International} {Conference} on
  {Machine} {Learning}}, volume~80 of \emph{Proceedings of {Machine} {Learning}
  {Research}}, pages 5393--5402, Stockholmsmässan, Stockholm Sweden, July
  2018. PMLR.
\newblock URL \url{http://proceedings.mlr.press/v80/xiao18a.html}.

\bibitem[Yang(2019{\natexlab{a}})]{yangScalingLimitsWide2019arXiv.org}
Greg Yang.
\newblock Scaling {{Limits}} of {{Wide Neural Networks}} with {{Weight
  Sharing}}: {{Gaussian Process Behavior}}, {{Gradient Independence}}, and
  {{Neural Tangent Kernel Derivation}}.
\newblock \emph{arXiv:1902.04760 [cond-mat, physics:math-ph, stat]}, February
  2019{\natexlab{a}}.

\bibitem[Yang(2019{\natexlab{b}})]{yangTP1}
Greg Yang.
\newblock Tensor programs i: Wide feedforward or recurrent neural networks of
  any architecture are gaussian processes.
\newblock In \emph{Advances in Neural Information Processing Systems}, pages
  9947--9960, 2019{\natexlab{b}}.

\bibitem[Yang(2020)]{yangTP2}
Greg Yang.
\newblock Tensor programs ii: Neural tangent kernel for any architecture, 2020.

\bibitem[Yang and Schoenholz(2018)]{yangVarianceVariation}
Greg Yang and Sam~S. Schoenholz.
\newblock Deep mean field theory: Layerwise variance and width variation as
  methods to control gradient explosion, 2018.
\newblock URL \url{https://openreview.net/forum?id=rJGY8GbR-}.

\bibitem[Yang and Schoenholz(2017)]{yang_mean_2017}
Greg Yang and Samuel~S. Schoenholz.
\newblock Mean {Field} {Residual} {Network}: {On} the {Edge} of {Chaos}.
\newblock In \emph{Advances in neural information processing systems}, 2017.

\bibitem[Yang et~al.(2019)Yang, Pennington, Rao, Sohl-Dickstein, and
  Schoenholz]{yang_mean_2019}
Greg Yang, Jeffrey Pennington, Vinay Rao, Jascha Sohl-Dickstein, and Samuel~S.
  Schoenholz.
\newblock A {Mean} {Field} {Theory} of {Batch} {Normalization}.
\newblock \emph{arXiv:1902.08129 [cond-mat]}, February 2019.
\newblock URL \url{http://arxiv.org/abs/1902.08129}.

\bibitem[Zou et~al.(2018)Zou, Cao, Zhou, and Gu]{zou_stochastic_2018}
Difan Zou, Yuan Cao, Dongruo Zhou, and Quanquan Gu.
\newblock Stochastic {Gradient} {Descent} {Optimizes} {Over}-parameterized
  {Deep} {ReLU} {Networks}.
\newblock \emph{arXiv:1811.08888 [cs, math, stat]}, November 2018.
\newblock URL \url{http://arxiv.org/abs/1811.08888}.

\end{thebibliography}
